\pdfoutput=1

\documentclass[11pt]{article}
\usepackage[utf8]{inputenc}
\usepackage{times}
\usepackage{subfig}
\usepackage{algorithm}
\usepackage[noend]{algorithmic}
\usepackage{nicefrac}
\usepackage{booktabs}
\usepackage{footmisc}
\usepackage[shortlabels]{enumitem}
\usepackage[margin=1in]{geometry}
\usepackage{comment}
\usepackage{booktabs,tabularx}
\usepackage{multirow}
\usepackage{textcomp}

\usepackage[numbers]{natbib}
\usepackage{graphicx}
\usepackage{color}
\definecolor{darkgreen}{rgb}{0,0.5,0.0}
\definecolor{darkblue}{rgb}{0,0.0,0.5}
\definecolor{darkred}{rgb}{0.5,0.0,0.0}
\usepackage{hyperref}
\usepackage{wrapfig}
\usepackage{amssymb,amsmath,amsthm}
\usepackage[capitalise,noabbrev]{cleveref}
\usepackage{array}
\usepackage{url}
\usepackage[strict]{changepage}
\usepackage{makecell}
\usepackage{titlesec}
\usepackage{setspace}

\titleformat*{\subparagraph}{\itshape}

\newcommand{\repeatcaption}[2]{%
  \renewcommand{\thefigure}{\ref{#1}}%
  \captionsetup{list=no}%
  \caption{#2 (repeated from \cpageref{#1})}%
  \addtocounter{figure}{-1}%
}

\newcommand{\R}{\mathbb{R}}
\DeclareMathOperator*{\E}{\mathbb{E}}
\newcommand{\BO}{\mathcal{O}}

\newcommand{\SUB}[1]{\ENSURE \hspace{-0.15in} \textbf{#1}}

\newsavebox\actorsfigure

\title{Advances and Open Problems in Federated Learning}

\author{
Peter Kairouz$^7$\footnote{Peter Kairouz and H. Brendan McMahan conceived, coordinated, and edited this work. Correspondence to \url{kairouz@google.com} and \url{mcmahan@google.com}.} \and
H. Brendan McMahan$^{7*}$ \and
Brendan Avent$^{21}$ \and 
Aur\'elien Bellet$^9$ \and
Mehdi Bennis$^{19}$ \and 
Arjun Nitin Bhagoji$^{13}$ \and
Kallista Bonawitz$^{7}$ \and
Zachary Charles$^{7}$ \and
Graham Cormode$^{23}$ \and
Rachel Cummings$^6$ \and
Rafael G.L. D'Oliveira$^{14}$ \and
Hubert Eichner$^{7}$ \and
Salim El Rouayheb$^{14}$ \and
David Evans$^{22}$ \and
Josh Gardner$^{24}$ \and
Zachary Garrett$^7$ \and
Adri\`a Gasc\'on$^7$ \and
Badih Ghazi$^7$ \and
Phillip B. Gibbons$^{2}$ \and
Marco Gruteser$^{7,14}$ \and
Zaid Harchaoui$^{24}$ \and
Chaoyang He$^{21}$ \and
Lie He $^{4}$\and
Zhouyuan Huo $^{20}$\and
Ben Hutchinson$^{7}$ \and
Justin Hsu$^{25}$ \and
Martin Jaggi$^{4}$ \and
Tara Javidi$^{17}$ \and
Gauri Joshi$^{2}$ \and
Mikhail Khodak$^{2}$ \and
Jakub Kone\v{c}n\'{y}$^{7}$ \and
Aleksandra Korolova$^{21}$ \and
Farinaz Koushanfar$^{17}$ \and
Sanmi Koyejo$^{7,18}$ \and
Tancr{\`e}de Lepoint$^7$ \and
Yang Liu$^{12}$ \and
Prateek Mittal$^{13}$ \and
Mehryar Mohri$^{7}$ \and
Richard Nock$^{1}$ \and
Ayfer \"Ozg\"ur$^{15}$ \and
Rasmus Pagh$^{7,10}$ \and
Hang Qi$^{7}$ \and
Daniel Ramage$^{7}$ \and
Ramesh Raskar$^{11}$ \and
Mariana Raykova$^{7}$ \and
Dawn Song$^{16}$ \and
Weikang Song$^{7}$ \and
Sebastian U. Stich$^{4}$ \and
Ziteng Sun$^{3}$ \and
Ananda Theertha Suresh$^{7}$ \and
Florian Tram\`er$^{15}$ \and
Praneeth Vepakomma$^{11}$ \and
Jianyu Wang$^{2}$ \and
Li Xiong$^{5}$ \and
Zheng Xu$^{7}$ \and
Qiang Yang$^{8}$ \and
Felix X. Yu$^{7}$ \and
Han Yu$^{12}$ \and
Sen Zhao$^{7}$ \and\\ 
\small{$^{1}$Australian National University, $^{2}$Carnegie Mellon University, $^{3}$Cornell University,}\\\small{$^{4}$École Polytechnique Fédérale de Lausanne, $^{5}$Emory University, $^{6}$Georgia Institute of Technology,}\\\small{$^{7}$Google Research, $^{8}$Hong Kong University of Science and Technology, $^{9}$INRIA, $^{10}$IT University of Copenhagen,}\\\small{$^{11}$Massachusetts Institute of Technology, $^{12}$Nanyang Technological University, $^{13}$Princeton University,}\\\small{$^{14}$Rutgers University, $^{15}$Stanford University, $^{16}$University of California Berkeley,}\\\small{$^{17}$ University of California San Diego, $^{18}$University of Illinois Urbana-Champaign, $^{19}$University of Oulu,} \\\small{$^{20}$University of Pittsburgh, $^{21}$University of Southern California, $^{22}$University of Virginia,}\\\small{$^{23}$University of Warwick, $^{24}$University of Washington, $^{25}$University of Wisconsin–Madison}
}

\date{}

\begin{document}

\begin{spacing}{1.1}
\maketitle
\end{spacing}

\begin{abstract}
Federated learning (FL) is a machine learning setting where many clients (e.g. mobile devices or whole organizations) collaboratively train a model under the orchestration of a central server (e.g. service provider), while keeping the training data decentralized. FL embodies the principles of focused data collection and minimization, and can mitigate many of the systemic privacy risks and costs resulting from traditional, centralized machine learning and data science approaches. Motivated by the explosive growth in FL research, this paper discusses recent advances and presents an extensive collection of open problems and challenges. 
\end{abstract}

\pagebreak

\begin{small}
\tableofcontents
\end{small}

\setlength{\parskip}{0.5em}

\pagebreak
\section{Introduction}
\label{sec:intro}

Federated learning (FL) is a machine learning setting where many clients (e.g. mobile devices or whole organizations) collaboratively train a model under the orchestration of a central server (e.g. service provider), while keeping the training data decentralized. It embodies the principles of focused collection and data minimization, and can mitigate many of the systemic privacy risks and costs resulting from traditional, centralized machine learning. This area has received significant interest recently, both from research and applied perspectives. This paper describes the defining characteristics and challenges of the federated learning setting, highlights important practical constraints and considerations, and then enumerates a range of valuable research directions. The goals of this work are to highlight research problems that are of significant theoretical and practical interest, and to encourage research on problems that could have significant real-world impact.

The term \emph{federated learning} was introduced in 2016 by \citet{mcmahan17fedavg}: ``We term our approach Federated Learning, since the learning task is solved by a loose federation of participating devices (which we refer to as clients) which are coordinated by a central server.''  An unbalanced and non-IID (identically and independently distributed) data partitioning across a massive number of unreliable devices with limited communication bandwidth was introduced as the defining set of challenges.

Significant related work predates the introduction of the term federated learning. A longstanding goal pursued by many research communities (including cryptography, databases, and machine learning) is to analyze and learn from data distributed among many owners without exposing that data. Cryptographic methods for computing on encrypted data were developed starting in the early 1980s \citep{Rivest1978,yao1982protocols}, and \citet{agrawal2000} and \citet{vaidya2008ppsvm} are early examples of work that sought to learn from local data using a centralized server while preserving privacy.
Conversely, even since the introduction of the term federated learning, we are aware of no single work that directly addresses the full set of FL challenges. Thus, the term federated learning provides a convenient shorthand for a set of characteristics, constraints, and challenges that often co-occur in applied ML problems on decentralized data where privacy is paramount.

This paper originated at the Workshop on Federated Learning and Analytics held June 17--18th, 2019, hosted at Google's Seattle office. During the course of this two-day event, the need for a broad paper surveying the many open challenges in the area of federated learning became clear.\footnote{During the preparation of this work, \citet{li2019federated} independently released an excellent but less comprehensive survey.}

A key property of many of the problems discussed is that they are inherently interdisciplinary --- solving them likely requires not just machine learning, but techniques from distributed optimization, cryptography, security, differential privacy, fairness, compressed sensing, systems, information theory, statistics, and more.  Many of the hardest problems are at the intersections of these areas, and so we believe collaboration will be essential to ongoing progress. One of the goals of this work is to highlight the ways in which techniques from these fields can potentially be combined, raising both interesting possibilities as well as new challenges.

Since the term federated learning was initially introduced with an emphasis on mobile and edge device applications \citep{mcmahan17fedavg,flblog17}, interest in applying FL to other applications has greatly increased, including some which might involve only a small number of relatively reliable clients, for example multiple organizations collaborating to train a model. We term these two federated learning settings ``cross-device'' and  ``cross-silo'' respectively. Given these variations, we propose a somewhat broader definition of federated learning:
\begin{quote}
\emph{\textbf{Federated learning} is a machine learning setting where multiple entities (clients) collaborate in solving a machine learning problem, under the 
coordination of a central server or service provider. Each client's raw data is stored locally and not exchanged or transferred; instead, focused updates intended for immediate aggregation are used to achieve the learning objective.}
\end{quote}
Focused updates are updates narrowly scoped to contain the minimum information necessary for the specific learning task at hand; aggregation is performed as early as possible in the service of data minimization. We note that this definition distinguishes federated learning from fully decentralized (peer-to-peer) learning techniques as discussed in \cref{sec:decentralized}.

Although privacy-preserving data analysis has been studied for more than 50 years, only in the past decade have solutions been widely deployed at scale (e.g. \citep{rappor:15,applewhitepaper:17}). Cross-device federated learning and federated data analysis are now being applied in consumer digital products. Google makes extensive use of federated learning in the Gboard mobile keyboard \citep{sundar2019nyt,hard18gboard,yang18gboardquery,chen19oov,ramaswamy19emoji}, as well as in features on Pixel phones \citep{googleai18settingssearch} and in Android Messages \citep{messages19privacy}. While Google has pioneered cross-device FL, interest in this setting is now much broader, for example: Apple is using cross-device FL in iOS 13 \citep{apple19neurips}, for applications like the QuickType keyboard and the vocal classifier for ``Hey Siri'' \citep{apple19wwdc};  doc.ai is developing cross-device FL solutions for medical research \citep{docai}, and Snips has explored cross-device FL for hotword detection \citep{leroy2018federated}.

Cross-silo applications have also been proposed or described in myriad domains including finance risk prediction for reinsurance \citep{webankswissre19reinsurance}, pharmaceuticals discovery \citep{melloddy19pharma}, electronic health records mining \citep{featurecloud19ehr}, medical data segmentation \citep{intel19medicalimaging, courtiol2019deep}, and smart manufacturing \citep{musketeer19mfg}.

The growing demand for federated learning technology has resulted in a number of tools and frameworks becoming available. These include TensorFlow Federated \citep{tff}, Federated AI Technology Enabler \citep{FATE}, PySyft \citep{PySyft}, Leaf \citep{Leaf}, PaddleFL \citep{PaddleFL} and Clara Training Framework \citep{ClaraTraining};
more details in Appendix~\ref{sec:datasets-and-software}. Commercial data platforms incorporating federated learning are in development from established technology companies as well as smaller start-ups.

Table~\ref{tab:characteristics} contrasts both cross-device and cross-silo federated learning with traditional single-datacenter distributed learning across a range of axes. These characteristics establish many of the constraints that practical federated learning systems must typically satisfy, and hence serve to both motivate and inform the open challenges in federated learning. They will be discussed at length in the sections that follow.

These two FL variants are called out as representative and important examples, but different FL settings may have different combinations of these characteristics. For the remainder of this paper, we consider the cross-device FL setting unless otherwise noted, though many of the problems apply to other FL settings as well. Section~\ref{sec:relaxing} specifically addresses some of the many other variations and applications.

Next, we consider cross-device federated learning in more detail, focusing on practical aspects common to a typical large-scale deployment of the technology; \citet{bonawitz19sysml} provides even more detail for a particular production system, including a discussion of specific architectural choices and considerations.

\subsection{The Cross-Device Federated Learning Setting}
\label{subsec:cross-device-fl-setting}
This section takes an applied perspective, and unlike the previous section, does not attempt to be definitional. Rather, the goal is to describe some of the practical issues in cross-device FL and how they might fit into a broader machine learning development and deployment ecosystem. The hope is to provide useful context and motivation for the open problems that follow, as well as to aid researchers in estimating how straightforward it would be to deploy a particular new approach in a real-world system. We begin by sketching the lifecycle of a model before considering a FL training process.

\newcolumntype{B}[1]{>{\arraybackslash}p{#1}}
\newcolumntype{P}[1]{>{\centering\arraybackslash}p{#1}}

\newcommand{\twocolright}[1]{\multicolumn{2}{p{4.1in}}{#1}}
\newcommand{\twocolleft}[1]{\multicolumn{2}{p{3.5in}}{#1}}
\newcommand{\twocolleftcenter}[1]{\multicolumn{2}{P{3.5in}}{#1}}

\newgeometry{left=0.7in,right=0.7in,top=0.5in,bottom=0.8in}
\begin{table}[t]
\begin{centering}
\renewcommand{\arraystretch}{1.5}
\begin{small}
\begin{tabular}{@{}B{0.8in}B{1.6in}B{1.8in}B{2.2in}@{}}
\toprule
 & \textbf{Datacenter \mbox{distributed learning}} & \textbf{Cross-silo \mbox{federated learning \hspace{1in}}} & \textbf{Cross-device \mbox{federated learning \hspace{1in}}} \\
\midrule
Setting
  & Training a model on a large but ``flat'' dataset. Clients are compute nodes in a single cluster or datacenter.
  & Training a model on siloed data. Clients are different organizations (e.g. medical or financial) or geo-distributed datacenters.
  & The clients are a very large number of mobile or IoT devices. \\

Data \mbox{distribution}
  & Data is centrally stored and can be shuffled and balanced across clients. Any client can read any part of the dataset.
  & \twocolright{\textbf{Data is generated locally and remains decentralized.}  Each client stores its own data and cannot read the data of other clients. Data is not independently or identically distributed.} \\

Orchestration 
  & Centrally orchestrated.
  & \twocolright{\textbf{A central orchestration server/service organizes the training}, but never sees raw data.} \\

Wide-area \mbox{communication} 
  & None (fully connected clients in one datacenter/cluster).
  & \twocolright{Typically a hub-and-spoke topology, with the hub representing a coordinating service provider (typically without data) and the spokes connecting to clients.} \\

Data \mbox{availability}
  & \twocolleftcenter{\rule[0.8mm]{.6in}{0.4pt}\ All clients are almost always available.\ \rule[0.8mm]{.6in}{0.4pt}}
  & Only a fraction of clients are available at any one time, often with diurnal or other variations. \\

Distribution scale 
  & Typically 1 - 1000 clients.
  & Typically 2 - 100 clients.
  & Massively parallel, up to $10^{10}$ clients. \\
  
Primary \mbox{bottleneck}
  & Computation is more often the bottleneck in the datacenter, where very fast networks can be assumed.
  & Might be computation or communication.
  & Communication is often the primary bottleneck, though it depends on the task. Generally, cross-device federated computations use wi-fi or slower connections. \\  
  
Addressability 
  & \twocolleft{ Each client has an identity or name that allows the system to access it specifically.}
  & Clients cannot be indexed directly (i.e., no use of client identifiers). \\

Client \mbox{statefulness}
  & \twocolleft{Stateful --- each client may participate in each round of the computation, carrying state from round to round. }
  & Stateless --- each client will likely participate only once in a task, so generally a fresh sample of never-before-seen clients in each round of computation is assumed. \\
  
Client \mbox{reliability}
  & \twocolleftcenter{\rule[0.8mm]{1.0in}{0.4pt}\ Relatively few failures.\ \rule[0.8mm]{1.0in}{0.4pt} }
  & Highly unreliable --- 5\% or more of the clients participating in a round of computation are expected to fail or drop out (e.g. because the device becomes ineligible when battery, network, or idleness requirements are violated). \\
  
Data partition axis
  & Data can be partitioned / re-partitioned arbitrarily across clients.
  & Partition is fixed. Could be example-partitioned (horizontal) or feature-partitioned (vertical).
  & Fixed partitioning by example (horizontal).\\

\bottomrule
\end{tabular}
\end{small}
\caption{Typical characteristics of federated learning settings vs. distributed learning in the datacenter (e.g. \citep{dean2012large}). Cross-device and cross-silo federated learning are two examples of FL domains, but are not intended to be exhaustive. The primary defining characteristics of FL are highlighted in bold, but the other characteristics are also critical in determining which techniques are applicable.} 
\label{tab:characteristics}
\end{centering}
\end{table}
\restoregeometry

\savebox\actorsfigure{\vbox{%
\centering
\includegraphics[width=\textwidth]{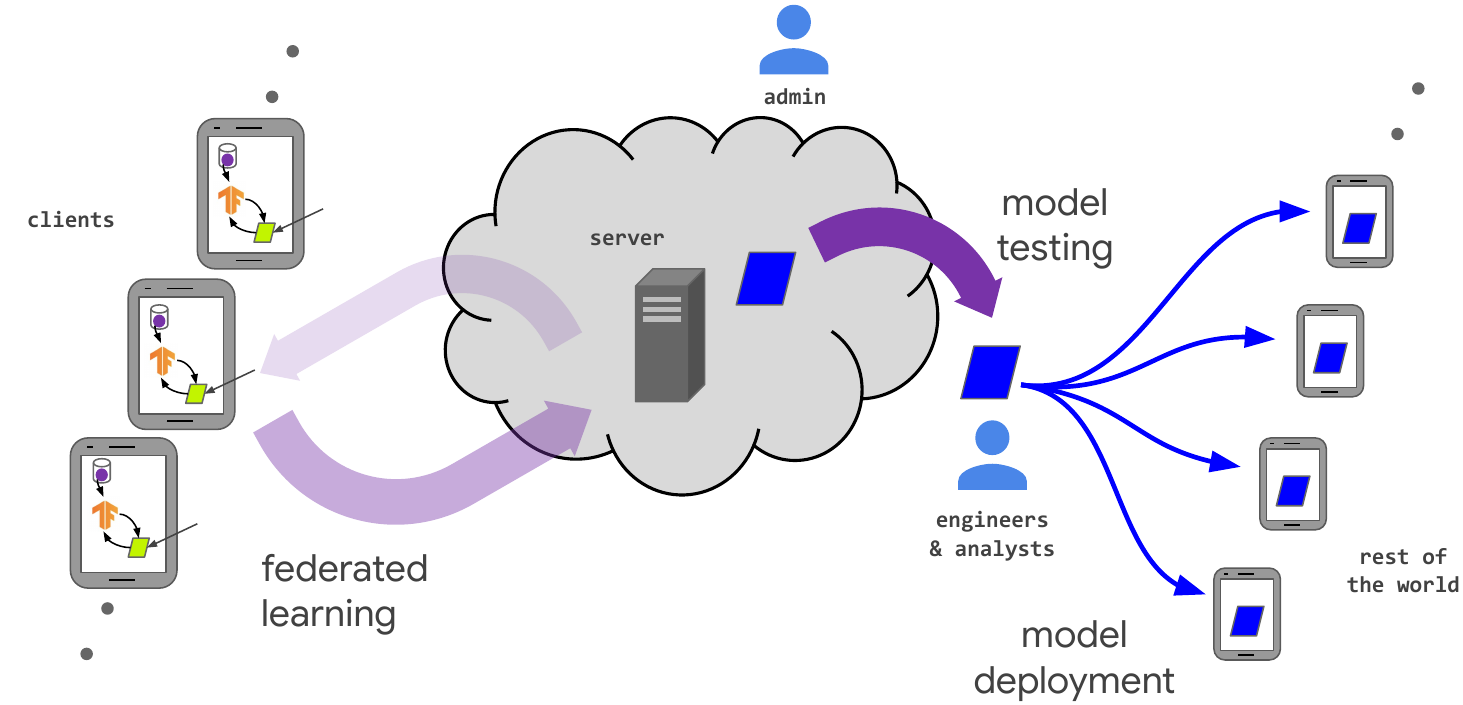}
}}

\begin{figure}[t!] 
\noindent\usebox{\actorsfigure}
\caption{The lifecycle of an FL-trained model and the various actors in a federated learning system. This figure is revisited in \cref{sec:privacy} from a threat models perspective. 
} 
\label{fig:actors}
\end{figure}

\subsubsection{The Lifecycle of a Model in Federated Learning}\label{sec:lifecycle}
The FL process is typically driven by a model engineer developing a model for a particular application. For example, a domain expert in natural language processing may develop a next word prediction model for use in a virtual keyboard. \cref{fig:actors} shows the primary components and actors. At a high level, a typical workflow is:

\begin{enumerate}

\item \textbf{Problem identification:} The model engineer identifies a problem to be solved with FL.

\item \textbf{Client instrumentation:} If needed, the clients (e.g. an app running on mobile phones) are instrumented to store locally (with limits on time and quantity) the necessary training data. In many cases, the app already will have stored this data (e.g. a text messaging app must store text messages, a photo management app already stores photos). However, in some cases additional data or metadata might need to be maintained, e.g. user interaction data to provide labels for a supervised learning task.

\item \textbf{Simulation prototyping (optional):} The model engineer may prototype model architectures and test learning hyperparameters in an FL simulation using a proxy dataset.

\item \textbf{Federated model training: } Multiple federated training tasks are started to train different variations of the model, or use different optimization hyperparameters. \label{step:fl-train}

\item \textbf{ (Federated) model evaluation:} After the tasks have trained sufficiently (typically a few days, see below), the models are analyzed and good candidates selected. Analysis may include metrics computed on standard datasets in the datacenter, or federated evaluation wherein the models are pushed to held-out clients for evaluation on local client data.

\item \textbf{Deployment:} Finally, once a good model is selected, it goes through a standard model launch process, including manual quality assurance, live A/B testing (usually by using the new model on some devices and the previous generation model on other devices to compare their in-vivo performance), and a staged rollout (so that poor behavior can be discovered and rolled back before affecting too many users). The specific launch process for a model is set by the owner of the application and is usually independent of how the model is trained. In other words, this step would apply equally to a model trained with federated learning or with a traditional datacenter approach.\label{step:deploy}

\end{enumerate}

One of the primary practical challenges an FL system faces is making the above workflow as straightforward as possible, ideally approaching the ease-of-use achieved by ML systems for centralized training. While much of this paper concerns federated training specifically, there are many other components including federated analytics tasks like model evaluation and debugging. Improving these is the focus of \cref{sec:workflows}. For now, we consider in more detail the training of a single FL model (Step~\ref{step:fl-train} above).

\begin{table}
\centering
\renewcommand{\arraystretch}{1.2}
\begin{tabular}{rl}    
\toprule
Total population size                           & $10^6$--$10^{10}$ devices \\
Devices selected for one round of training      & 50 -- 5000 \\
Total devices that participate in training one model  & $10^5$--$10^7$ \\
Number of rounds for model convergence          & 500 -- 10000 \\
Wall-clock training time                        & 1 -- 10 days \\
\bottomrule
\end{tabular}
\caption{Order-of-magnitude sizes for typical cross-device federated learning applications.}
\label{tab:sizes}
\end{table}

\subsubsection{A Typical Federated Training Process}\label{sec:typical-training}
We now consider a template for FL training that encompasses the Federated Averaging algorithm of \citet{mcmahan17fedavg} and many others; again, variations are possible, but this gives a common starting point. 


A server (service provider) orchestrates the training process, by repeating the following steps until training is stopped (at the discretion of the model engineer who is monitoring the training process):
\begin{enumerate}

  \item \textbf{Client selection:} The server samples from a set of clients meeting eligibility requirements. For example, mobile phones might only check in to the server if they are plugged in, on an unmetered wi-fi connection, and idle, in order to avoid impacting the user of the device.
  
  \item \textbf{Broadcast:} The selected clients download the current model weights and a training program (e.g. a TensorFlow graph~\cite{tensorflow2015-whitepaper}) from the server.
  
  \item \textbf{Client computation:} Each selected device locally computes an update to the model by executing the training program, which might for example run SGD on the local data (as in Federated Averaging).
  
  \item \textbf{Aggregation:} The server collects an aggregate of the device updates. For efficiency, stragglers might be dropped at this point once a sufficient number of devices have reported results. This stage is also the integration point for many other techniques which will be discussed later, possibly including: secure aggregation for added privacy, lossy compression of aggregates for communication efficiency, and noise addition and update clipping for differential privacy.
  
  \item \textbf{Model update:} The server locally updates the shared model based on the aggregated update computed from the clients that participated in the current round.
\end{enumerate}

Table~\ref{tab:sizes} gives typical order-of-magnitude sizes for the quantities involved in a typical federated learning application on mobile devices.

The separation of the client computation, aggregation, and model update phases is not a strict requirement of federated learning, and it indeed excludes certain classes of algorithms, for example asynchronous SGD where each client's update is immediately applied to the model, before any aggregation with updates from other clients. Such asynchronous approaches may simplify some aspects of system design, and also be beneficial from an optimization perspective (though this point can be debated). However, the approach presented above has a substantial advantage in affording a separation of concerns between different lines of research: advances in compression, differential privacy, and secure multi-party computation can be developed for standard primitives like computing sums or means over decentralized updates, and then composed with arbitrary optimization or analytics algorithms, so long as those algorithms are expressed in terms of aggregation primitives.

It is also worth emphasizing that in two respects, the FL training process should not impact the user experience. First, as outlined above, even though model parameters are typically sent to some devices during the broadcast phase of each round of federated training, these models are an ephemeral part of the training process, and not used to make ``live'' predictions shown to the user.  This is crucial, because training ML models is challenging, and a misconfiguration of hyperparameters can produce a model that makes bad predictions. Instead, user-visible use of the model is deferred to a rollout process as detailed above in Step~\ref{step:deploy} of the model lifecycle. Second, the training itself is intended to be invisible to the user --- as described under client selection, training does not slow the device or drain the battery because it only executes when the device is idle and connected to power. However, the limited availability these constraints introduce leads directly to open research challenges which will be discussed subsequently, such as semi-cyclic data availability and the potential for bias in client selection.

\subsection{Federated Learning Research}\label{sec:fl-research}
The remainder of this paper surveys many open problems that are motivated by the constraints and challenges of real-world federated learning settings, from training models on medical data from a hospital system to training using hundreds of millions of mobile devices. Needless to say, most researchers working on federated learning problems will likely not be deploying production FL systems, nor have access to fleets of millions of real-world devices. This leads to a key distinction between the practical settings that motivate the work and experiments conducted in simulation which provide evidence of the suitability of a given approach to the motivating problem.

This makes FL research somewhat different than other ML fields from an experimental perspective, leading to additional considerations in conducting FL research. In particular, when highlighting open problems, we have attempted, when possible, to also indicate relevant performance metrics which can be measured in simulation, the characteristics of datasets which will make them more representative of real-world performance, etc. The need for simulation also has ramifications for the presentation of FL research. While not intended to be authoritative or absolute, we make the following modest suggestions for presenting FL research that addresses the open problems we describe:

\begin{itemize}
    \item As shown in \cref{tab:characteristics}, the FL setting can encompass a wide range of problems. Compared to fields where the setting and goals are well-established, it is important to precisely describe the details of the particular FL setting of interest, particularly when the proposed approach makes assumptions that may not be appropriate in all settings (e.g. stateful clients that participate in all rounds).
    
    \item Of course, details of any simulations should be presented in order to make the research reproducible. But it is also important to explain which aspects of the real-world setting the simulation is designed to capture (and which it is not), in order to effectively make the case that success on the simulated problem implies useful progress on the real-world objective. We hope that the guidance in this paper will help with this.
    
    \item Privacy and communication efficiency are always first-order concerns in FL, even if the experiments are simulations running on a single machine using public data.  More so than with other types of ML, for any proposed approach it is important to be unambiguous about \emph{where computation happens} as well as \emph{what is communicated}.
\end{itemize}

Software libraries for federated learning simulation as well as standard datasets can help ease the challenges of conducting effective FL research; \cref{sec:datasets-and-software} summarizes some of the currently available options. Developing standard evaluation metrics and establishing standard benchmark datasets for different federated learning settings (cross-device and cross-silo) remain highly important directions for ongoing work.

\subsection{Organization}
\cref{sec:relaxing} builds on the ideas in \cref{tab:characteristics}, exploring other FL settings and problems beyond the original focus on cross-device settings. \cref{sec:better_fl} then turns to core questions around improving the efficiency and effectiveness of federated learning.
\cref{sec:privacy} undertakes a careful consideration of threat models and considers a range of technologies toward the goal of achieving rigorous privacy protections. As with all machine learning systems, in federated learning applications there may be incentives to manipulate the models being trained, and failures of various kinds are inevitable; these challenges are discussed in \cref{sec:robust}. Finally, we address the important challenges of providing fair and unbiased models in \cref{sec:fairness}.

\pagebreak
\section{Relaxing the Core FL Assumptions: Applications to Emerging Settings and Scenarios}
\label{sec:relaxing}

In this section, we will discuss areas of research related to the topics discussed in the previous section. Even though not being the main focus of the remainder of the paper, progress in these areas could motivate design of the next generation of production systems.

\subsection{Fully Decentralized / Peer-to-Peer Distributed Learning}
\label{sec:decentralized}

In federated learning, a central server orchestrates the training process and receives the contributions of all clients. The server is thus a central player which also potentially represents a single point of failure. While large companies or organizations can play this role in some application scenarios, a reliable and powerful central server may not always be available or desirable in more collaborative learning scenarios \citep{Vanhaesebrouck2017}. Furthermore, the server may even become a bottleneck when the number of clients is very large, as demonstrated by \citet{Lian2017b} (though this can be mitigated by careful system design, e.g. \citep{bonawitz19sysml}).

The key idea of fully decentralized learning is to replace communication with the server by peer-to-peer communication between individual clients. The communication topology is represented as a connected graph in which nodes are the clients and an edge indicates a communication channel between two clients. The network graph is typically chosen to be sparse with small maximum degree so that each node only needs to send/receive messages to/from a small number of peers; this is in contrast to the star graph of the server-client architecture. In fully decentralized algorithms, a round corresponds to each client performing a local update and exchanging information with their neighbors in the graph\footnote{Note, however, that the notion of a round does not need to even make sense in this setting. See for instance the discussion on clock models in \citep{Boyd2006}.}. In the context of machine learning, the local update is typically a local (stochastic) gradient step and the communication consists in averaging one's local model parameters with the neighbors. Note that there is no longer a global state of the model as in standard federated learning, but the process can be designed such that all local models converge to the desired global solution, i.e., the individual models gradually reach consensus. While multi-agent optimization has a long history in the control community, fully decentralized variants of SGD and other optimization algorithms have recently been considered in machine learning both for improved scalability in datacenters \citep{assran2019stochastic} as well as for decentralized networks of devices \citep{Colin2016, Vanhaesebrouck2017, Tang2018, Bellet2018a, Koloskova2019, Lalitha2019, elgabligadmm}. They consider undirected network graphs, although the case of directed networks (encoding unidirectional channels which may arise in real-world scenarios such as social networks or data markets) has also been studied in \citep{assran2019stochastic, he2019central}.

It is worth noting that even in the decentralized setting outlined above, a central authority may still be in charge of setting up the learning task. Consider for instance the following questions: Who decides what is the model to be trained in the decentralized setting? What algorithm to use? What hyperparameters? Who is responsible for debugging when something does not work as expected? A certain degree of trust of the participating clients in a central authority would still be needed to answer these questions. Alternatively, the decisions could be taken by the client who proposes the learning task, or collaboratively through a consensus scheme (see \cref{sec:p2p-practical}).

\begin{table}
\begin{centering}
\renewcommand{\arraystretch}{1.5}
\begin{tabularx}{\textwidth}{lXX}
\toprule
       & \textbf{Federated learning} & \textbf{\mbox{Fully~decentralized} \mbox{(peer-to-peer) learning}} \\
\midrule 
Orchestration 
  & A central orchestration server or service organizes the training, but never sees raw data.
  & No centralized orchestration. \\
Wide-area communication 
  & Typically a hub-and-spoke topology, with the hub representing a coordinating service provider (typically without data) and the spokes connecting to clients. 
  & Peer-to-peer topology, with a possibly dynamic connectivity graph. \\
\bottomrule
\end{tabularx}
\caption{A comparison of the key distinctions between federated learning and fully decentralized learning. Note that as with FL, decentralized learning can be further divided into different use-cases, with distinctions similar to those made in \cref{tab:characteristics} comparing cross-silo and cross-device FL.}
\label{tab:decentralized}
\end{centering}
\end{table}

\Cref{tab:decentralized} provides a comparison between federated and peer-to-peer learning. While the architectural assumptions of decentralized learning are distinct from those of federated learning, it can often be applied to similar problem domains, many of the same challenges arise, and there is significant overlap in the research communities. Thus, we consider decentralized learning in this paper as well; in this section challenges specific to the decentralized approach are explicitly considered, but many of the open problems in other sections also arise in the decentralized case.

\subsubsection{Algorithmic Challenges}
A large number of important algorithmic questions remain open on the topic of real-world usability of decentralized schemes for machine learning. Some questions are analogous to the special case of federated learning with a central server, and other challenges come as an additional side-effect of being fully decentralized or trust-less. We outline some particular areas in the following.

\paragraph{Effect of network topology and asynchrony on decentralized SGD}
Fully decentralized algorithms for learning should be robust to the limited availability of the clients (with clients temporarily unavailable, dropping out or joining during the execution) and limited reliability of the network (with possible message drops). While for the special case of generalized linear models, schemes using the duality structure could enable some of these desired robustness properties \citep{he2018cola}, for the case of deep learning and SGD this remains an open question.
When the network graph is complete but messages have a fixed probability to be dropped, \citet{yu2018distributed} show that one can achieve convergence rates that are comparable to the case of a reliable network.
Additional open research questions concern non-IID data distributions, update frequencies, efficient communication patterns and practical convergence time \citep{Tang2018}, as we outline in more detail below.

Well-connected or denser networks encourage faster consensus and give better theoretical convergence rates, which depend on the spectral gap of the network graph. However, when data is IID, sparser topologies do not necessarily hurt the convergence in practice: this was analyzed theoretically in \cite{neglia2020}. Denser networks typically incur communication delays which increase with the node degrees. Most of optimization-theory works do not explicitly consider how the topology affects the runtime, that is, wall-clock time required to complete each SGD iteration. \citet{wang2019matcha} propose MATCHA, a decentralized SGD method based on matching decomposition sampling, that reduces the communication delay per iteration for any given node topology while maintaining the same error convergence speed. The key idea is to decompose the graph topology into matchings consisting of disjoint communication links that can operate in parallel, and carefully choose a subset of these matchings in each iteration. This sequence of subgraphs results in more frequent communication over connectivity-critical links (ensuring fast error convergence) and less frequent communication over other links (saving communication delays).

The setting of decentralized SGD also naturally lends itself to asynchronous algorithms in which each client becomes active independently at random times, removing the need for global synchronization and potentially improving scalability \citep{Colin2016,Vanhaesebrouck2017,Bellet2018a,assran2019stochastic,Lian2018}.

\paragraph{Local-update decentralized SGD}
The theoretical analysis of schemes which perform several local update steps before a communication round is significantly more challenging than those using a single SGD step, as in mini-batch SGD. While this will also be discussed later in \cref{sec:optimization}, the same also holds more generally in the fully decentralized setting of interest here. Schemes relying on a single local update step are typically proven to converge in the case of non-IID local datasets \citep{Koloskova2019, koloskova2019deep}. For the case with several local update steps, \citep{wang2018cooperative,decentralized_sgd2} recently provided convergence analysis. Further, \citep{wang2019matcha} provides a convergence analysis for the non-IID data case, but for the specific scheme based on matching decomposition sampling described above. In general, however, understanding the convergence under non-IID data distributions and how to design a model averaging policy that achieves the fastest convergence remains an open problem.

\paragraph{Personalization, and trust mechanisms}
Similarly to the cross-device FL setting, an important task for the fully decentralized scenario under the non-IID data distributions available to individual clients is to design algorithms for learning collections of personalized models. The work of \citep{Vanhaesebrouck2017,Bellet2018a} introduces fully decentralized algorithms to collaboratively learn a personalized model for each client by smoothing model parameters across clients that have similar tasks (i.e., similar data distributions). \citet{Zantedeschi2019} further learn the similarity graph together with the personalized models. One of the key unique challenges in the decentralized setting remains the robustness of such schemes to malicious actors or contribution of unreliable data or labels. The use of incentives or mechanism design in combination with decentralized learning is an emerging and important goal, which may be harder to achieve in the setting without a trusted central server.

\paragraph{Gradient compression and quantization methods}
In potential applications, the clients would often be limited in terms of communication bandwidth available and energy usage permitted. Translating and generalizing some of the existing compressed communication schemes from the centralized orchestrator-facilitated setting (see \cref{sec:compr}) to the fully decentralized setting, without negatively impacting the convergence is an active research direction \citep{Koloskova2019, reisizadeh2019robust, tang2019texttt, koloskova2019deep}.
A complementary idea is to design decentralized optimization algorithms which naturally give rise to sparse updates \citep{Zantedeschi2019}.

\paragraph{Privacy}
An important challenge in fully decentralized learning is to prevent any client from reconstructing the private data of another client from its shared updates while maintaining a good level of utility for the learned models. Differential privacy (see \cref{sec:privacy}) is the standard approach to mitigate such privacy risks. In decentralized federated learning, this can be achieved by having each client add noise locally, as done in \citep{Huang2015a,Bellet2018a}. Unfortunately, such local privacy approaches often come at a large cost in utility. Furthermore, distributed methods based on secure aggregation or secure shuffling that are designed to improve the privacy-utility trade-off in the standard FL setting (see \cref{sssec:distributed_dp}) do not easily integrate with fully decentralized algorithms. A possible direction to achieve better trade-offs between privacy and utility in fully decentralized algorithms is to rely on decentralization itself to amplify differential privacy guarantees, for instance by considering appropriate relaxations of local differential privacy \cite{privacy_amp_by_decentralization}.

\subsubsection{Practical Challenges}
\label{sec:p2p-practical}

An orthogonal question for fully decentralized learning is how it can be practically realized. This section outlines a family of related ideas based on the idea of a distributed ledger, but other approaches remain unexplored.

A blockchain is a distributed ledger shared among disparate users, making possible digital transactions, including transactions of cryptocurrency, without a central authority. In particular, smart contracts allow execution of arbitrary code on top of the blockchain, essentially a massively replicated eventually-consistent state machine. In terms of federated learning, use of the technology could enable decentralization of the global server by using smart contracts to do model aggregation, where the participating clients executing the smart contracts could be different companies or cloud services.

However, on today’s blockchain platforms such as Ethereum~\citep{wood2014ethereum}, data on the blockchains is publicly available by default, this could discourage users from participating in the decentralized federated learning protocol, as the protection of the data is typically the primary motivating factor for FL. To address such concerns, it might be possible to modify the existing privacy-preserving techniques to fit into the scenario of decentralized federated learning. First of all, to prevent the participating nodes from exploiting individually submitted model updates, existing secure aggregation protocols could be used. A practical secure aggregation protocol already used in cross-device FL was proposed by \citet{bonawitz17secagg}, effectively handling dropping out participants at the cost of complexity of the protocol. An alternative system would be to have each client stake a deposit of cryptocurrency on blockchain, and get penalized if they drop out during the execution. Without the need of handling dropouts, the secure aggregation protocol could be significantly simplified. Another way of achieving secure aggregation is to use confidential smart contract such as what is enabled by the Oasis Protocol~\citep{cheng2019ekiden} which runs inside secure enclaves. With this, each client could simply submit an encrypted local model update, knowing that the model will be decrypted and aggregated inside the secure hardware through remote attestation (though see discussion of privacy-in-depth in \cref{ssec:actors_threat_models}).

In order to prevent any client from trying to reconstruct the private data of another client by exploiting the global model, client-level differential privacy \citep{mcmahan18dplm} has been proposed for FL.  Client-level differential privacy is achieved by adding random Gaussian noise on the aggregated global model that is enough to hide any single client's update. In the context of blockchain, each client could locally add a certain amount of Gaussian noise after local gradient descent steps and submit the model to blockchain. The local noise scale should be calculated such that the aggregated noise on blockchain is able to achieve the same client-level differential privacy as in \citep{mcmahan18dplm}. Finally, the aggregated global model on blockchain could be encrypted and only the participating clients hold the decryption key, which protects the model from the public.

\subsection{Cross-Silo Federated Learning}
\label{ssec:cross-silo}

In contrast with the characteristics of cross-device federated learning, see \cref{tab:characteristics}, cross-silo federated learning admits more flexibility in certain aspects of the overall design, but at the same time presents a setting where achieving other properties can be harder. This section discusses some of these differences.

The cross-silo setting can be relevant where a number of companies or organizations share incentive to train a model based on all of their data, but cannot share their data directly. This could be due to constraints imposed by confidentiality or due to legal constraints, or even within a single company when they cannot centralize their data between different geographical regions. These cross-silo applications have attracted substantial attention.

\paragraph{Data partitioning} In the cross-device setting the data is assumed to be partitioned by examples. In the cross-silo setting, in addition to partitioning by examples, partitioning by features is of practical relevance. An example could be when two companies in different businesses have the same or overlapping set of customers, such as a local bank and a local retail company in the same city. This difference has been also referred to as horizontal and vertical federated learning by \citet{DBLP:journals/corr/abs-1902-04885}.

Cross-silo FL with data partitioned by features, employs a very different training architecture compared to the setting with data partitioned by example. It may or may not involve a central server as a neutral party, and based on specifics of the training algorithm, clients exchange specific intermediate results rather than model parameters, to assist other parties' gradient calculations; see for instance \citep[Section 2.4.2]{DBLP:journals/corr/abs-1902-04885}. In this setting, application of techniques such as secure multi-party computation or homomorphic encryption have been proposed in order to limit the amount of information other participants can infer from observing the training process. The downside of this approach is that the training algorithm is typically dependent on the type of machine learning objective being pursued. Currently proposed algorithms include trees \citep{DBLP:journals/corr/abs-1901-08755}, linear and logistic regression \citep{DBLP:journals/corr/abs-1902-04885,Hardy2017-da,liuVFL}, and neural networks \citep{liu2018secure}. Local updates similar to Federated Averaging (see \cref{sec:optimization}) has been proposed to address the communication challenges of feature-partitioned systems \citep{liuVFL}, and \citep{hu2019learning, liu2020backdoor} study the security and privacy related challenges inherent in such systems.

Federated transfer learning \citep{DBLP:journals/corr/abs-1902-04885} is another concept that considers challenging scenarios in which data parties share only a partial overlap in the user space or the feature space, and leverage existing transfer learning techniques \cite{Pan:2010:STL:1850483.1850545} to build models collaboratively. The existing formulation is limited to the case of $2$ clients.

Partitioning by examples is usually relevant in cross-silo FL when a single company cannot centralize their data due to legal constraints, or when organizations with similar objectives want to collaboratively improve their models. For instance, different banks can collaboratively train classification or anomaly detection models for fraud detection \citep{webankswissre19reinsurance}, hospitals can build better diagnostic models~\citep{courtiol2019deep}, and so on.

An open-source platform supporting the above outlined applications is currently available as \textit{Federated AI Technology Enabler (FATE)} \citep{FATE}. At the same time, the IEEE P3652.1 Federated Machine Learning Working Group is focusing on standard-setting for the Federated AI Technology Framework. Other platforms include \cite{ClaraTraining} focused on a range of medical applications and \cite{IBMFL} for enterprise use cases. See \cref{sec:datasets-and-software} for more details.

\paragraph{Incentive mechanisms} In addition to developing new algorithmic techniques for FL, incentive mechanism design for honest participation is an important practical research question. This need may arise in cross-device settings (e.g. \citep{kang19incentive,kang2019incentiveB}), but is particularly relevant in the cross-silo setting, where participants may at the same time also be business competitors. The incentive can be in the form of monetary payout \citep{Yu-et-al:2020IS} or final models with different levels of performance \citep{Lyu-et-al:2020TPDS}. The option to deliver models with performance commensurate to the contributions of each client is especially relevant in collaborative learning situations in which competitions exist among FL participants. Clients might worry that contributing their data to training federated learning models will benefit their competitors, who do not contribute as much but receive the same final model nonetheless (i.e. the free-rider problem). Related objectives include how to divide earnings generated by the federated learning model among contributing data owners in order to sustain long-term participation, and also how to link the incentives with decisions on defending against adversarial data owners to enhance system security, optimizing the participation of data owners to enhance system efficiency.

\paragraph{Differential privacy} The discussion of actors and threat models in \cref{ssec:actors_threat_models} is largely relevant also for the cross-silo FL. However, protecting against different actors might have different priorities. For example, in many practical scenarios, the final trained model would be released only to those who participate in the training, which makes the concerns about ``the rest of the world'' less important. 

On the other hand, for a practically persuasive claim, we would usually need a notion of local differential privacy, as the potential threat from other clients is likely to be more important. In cases when the clients are not considered a significant threat, each client could control the data from a number of their respective users, and a formal privacy guarantee might be needed on such user-level basis. Depending on application, other objectives could be worth pursuing. This area has not been systematically explored.

\paragraph{Tensor factorization} Several works have also studied cross-silo federated tensor factorization where multiple sites (each having a set of data with the same feature, i.e. horizontally partitioned) jointly perform tensor factorization by only sharing intermediate factors with the coordination server while keeping data private at each site. Among the existing works, \cite{DBLP:conf/kdd/KimSYJ17} used an alternating direction method of multipliers (ADMM) based approach and \cite{ma19cikm} improved the efficiency with the elastic averaging SGD (EASGD) algorithm and further ensures differential privacy for the intermediate factors.

\subsection{Split Learning}
\label{ssec:split-learning}

In contrast with the previous settings which focus on data partitioning and communication patterns, the key idea behind split learning~\citep{gupta2018distributed,vepakomma2018split}\footnote{See also split learning project website - \url{https://splitlearning.github.io/}.} is to split the execution of a model on a per-layer basis between the clients and the server. This can be done for both training and inference. 

\begin{figure}[!htbp]%
    \centering
    \subfloat[Vanilla split learning]{{\includegraphics[width=5cm]{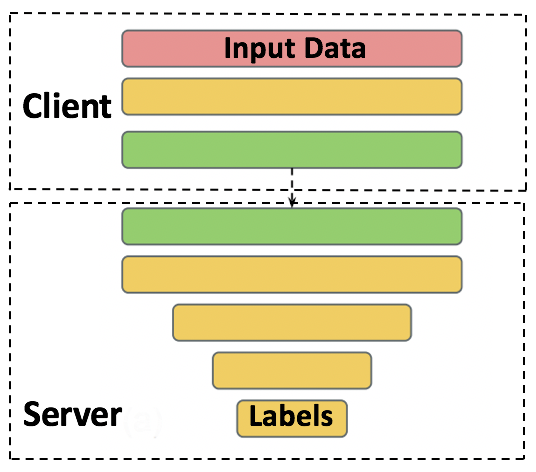} }}%
    \qquad
    \subfloat[U-shaped split learning]{{\includegraphics[width=5cm]{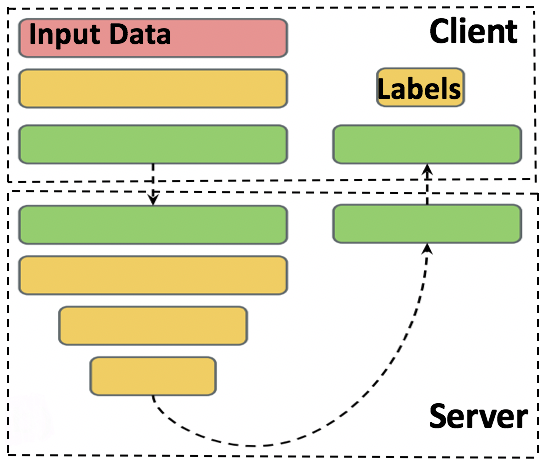} }}%
    \caption{Split learning configurations showing raw data is not transferred in the vanilla setting and that raw data as well as labels are not transferred between the client and server  entities in the U-shaped split learning setting.}
\label{splitConfig}
\end{figure}

In the simplest configuration of split learning, each client computes the forward pass through a deep network up to a specific layer referred to as the \emph{cut layer}. The outputs at the cut layer, referred to as \emph{smashed data}, are sent to another entity (either the server or another client), which completes the rest of the computation. This completes a round of forward propagation without sharing the raw data. The gradients can then be back propagated from its last layer until the cut layer in a similar fashion. The gradients at the cut layer -- and only these gradients -- are sent back to the clients, where the rest of back propagation is completed. This process is continued until convergence, without having clients directly access each others raw data. This setup is shown in \cref{splitConfig}(a) and a variant of this setup where labels are also not shared along with raw data is shown in \cref{splitConfig}(b). Split learning approaches for data partitioned by features have been studied in \cite{splitVertical}.

In several settings, the overall communication requirements of split learning and federated learning were compared in \citep{vepakomma2019splitComm}. Split learning brings in another dimension of parallelism in the training, parallelization among parts of a model, e.g. client and server. The ideas in \citep{jaderberg2017decoupled, huo2018training}, where the authors break the dependencies between partial networks and reduced total centralized training time by parallelizing the computations in different parts, can be relevant here as well. However, it is still an open question to explore such parallelization of split learning on edge devices. Split learning also enables matching client-side model components with the best server-side model components for automating model selection as shown in the ExpertMatcher \cite{sharma2019expertmatcher}.

The values communicated can nevertheless, in general, reveal information about the underlying data. How much, and whether this is acceptable, is likely going to be application and configuration specific. A variation of split learning called NoPeek SplitNN \citep{vepakomma2019reducing} reduces the potential leakage via communicated activations, by reducing their distance correlation \citep{vepakomma2018supervised,szekely2007measuring} with the raw data, while maintaining good model performance via categorical cross-entropy. The key idea is to minimize the distance correlation between the raw data points and communicated smashed data. The objects communicated could otherwise contain information highly correlated with the input data if used without NoPeek SplitNN, the use of which also enables the split to be made relatively early-on given the decorrelation it provides. One other engineering driven approach to minimize the amount of information communicated in split learning has been via a specifically learnt pruning of channels present in the client side activations \citep{channelPruning}. Overall, much of the discussion in \cref{sec:privacy} is relevant here as well, and analysis providing formal privacy guarantees specifically for split learning is still an open problem.

\subsection{Executive summary}
The motivation for federated learning is relevant for a number of related areas of research.

\begin{itemize}
\item Fully decentralized learning (\cref{sec:decentralized}) removes the need for a central server coordinating the overall computation. Apart from algorithmic challenges, open problems are in practical realization of the idea and in understanding of what form of trusted central authority is needed to set up the task.
\item Cross-silo federated learning (\cref{ssec:cross-silo}) admits problems with different kinds of modelling constraints, such as data partitioned by examples and/or features, and faces different set of concerns when formulating formal privacy guarantees or incentive mechanisms for clients to participate.
\item Split learning (\cref{ssec:split-learning}) is an approach to partition the execution of a model between the clients and the server. It can deliver different options for overall communication constraints, but detailed analysis of when the communicated values reveal sensitive information is still missing.
\end{itemize}

\pagebreak
\section{Improving Efficiency and Effectiveness}
\label{sec:better_fl}

In this section we explore a variety of techniques and open questions that address the challenge of making federated learning more efficient and effective. This encompasses a myriad of possible approaches, including: developing better optimization algorithms; providing different models to different clients; making ML tasks like hyperparameter search, architecture search, and debugging easier in the FL context; improving communication efficiency; and more. 

One of the fundamental challenges in addressing these goals is the presence of non-IID data, so we begin by surveying this issue and highlighting potential mitigations.

\newcommand{\Ploc}{\mathcal{P}}
\newcommand{\Pcli}{\mathcal{Q}}

\subsection{Non-IID Data in Federated Learning}\label{sec:noniid}
While the meaning of IID is generally clear, data can be non-IID in many ways. In this section, we provide a taxonomy of non-IID data regimes that may arise for any client-partitioned dataset. The most common sources of dependence and non-identicalness are due to each client corresponding to a particular user, a particular geographic location, and/or a particular time window. This taxonomy has a close mapping to notions of dataset shift \citep{torres2012unifying,candela2009datasetshift}, which studies differences between the training distribution and testing distribution; here, we consider differences in the data distribution on each client. 

For the following, consider a supervised task with features $x$ and labels $y$. A statistical model of federated learning involves two levels of sampling: accessing a datapoint requires first sampling a client $i \sim \Pcli$, the distribution over available clients, and then drawing an example $(x, y) \sim \Ploc_i(x, y)$ from that client's local data distribution. 

When non-IID data in federated learning is referenced, this typically refers to differences between $\Ploc_i$ and $\Ploc_j$ for different clients $i$ and $j$.  However, it is also important to note that the distribution $\Pcli$ and $\Ploc_i$ may change over time, introducing another dimension of ``non-IIDness''. 

For completeness, we note that even considering the dataset on a single device, if the data is in an insufficiently-random order, e.g. ordered by time, then independence is violated locally as well.  For example, consecutive frames in a video are highly correlated. Sources of intra-client correlation can generally be resolved by local shuffling.

\paragraph{Non-identical client distributions}

\newcommand{\gvn}{\,|\,}

We first survey some common ways in which data tend to deviate from being identically distributed, that is $P_i \neq P_j$ for different clients $i$ and $j$. Rewriting $P_i(x, y)$ as $P_i(y \gvn x) P_i(x)$ and $P_i(x \gvn y) P_i(y)$ allows us to characterize the differences more precisely.
\begin{itemize}
    \item \emph{Feature distribution skew} (covariate shift): The marginal distributions $\Ploc_i(x)$ may vary across clients, even if $\Ploc(y \gvn x)$ is shared.\footnote{We write ``$\Ploc(y \gvn x)$ is shared'' as shorthand for $\Ploc_i(y \gvn x) = \Ploc_j(y \gvn x)$  for all clients $i$ and $j$.} For example, in a handwriting recognition domain, users who write the same words might still have different stroke width, slant, etc. 
    
    \item \emph{Label distribution skew} (prior probability shift): The marginal distributions $\Ploc_i(y)$ may vary across clients, even if $\Ploc(x \gvn y)$ is the same. For example, when clients are tied to particular geo-regions, the distribution of labels varies across clients --- kangaroos are only in Australia or zoos; a person's face is only in a few locations worldwide; for mobile device keyboards, certain emoji are used by one demographic but not others.
    
    \item \emph{Same label, different features} (concept drift): The conditional distributions $\Ploc_i(x \gvn y)$ may vary across clients even if $\Ploc(y)$ is shared. The same label $y$ can have very different features $x$ for different clients, e.g. due to cultural differences, weather effects, standards of living, etc.
    For example, images of homes can vary dramatically around the world and items of clothing vary widely.  
    Even within the U.S., images of parked cars in the winter will be snow-covered only in certain parts of the country.  
    The same label can also look very different at different times, and at different time scales: day vs.~night, seasonal effects, natural disasters, fashion and design trends, etc.
    
    \item \emph{Same features, different label} (concept shift): The conditional distribution $\Ploc_i(y \gvn x)$ may vary across clients, even if $\Ploc(x)$ is the same. Because of personal preferences, the same feature vectors in a training data item can have different labels.  
    For example, labels that reflect sentiment or next word predictors have personal and regional variation.
    
  \item \emph{Quantity skew} or unbalancedness: Different clients can hold vastly different amounts of data.
\end{itemize}

Real-world federated learning datasets likely contain a mixture of these effects, and the characterization of cross-client differences in real-world partitioned datasets is an important open question. Most empirical work on synthetic non-IID datasets (e.g. \citep{mcmahan17fedavg, hsieh2019noniid}) have focused on label distribution skew, where a non-IID dataset is formed by partitioning a ``flat'' existing dataset based on the labels. A better understanding of the nature of real-world non-IID datasets will allow for the construction of controlled but realistic non-IID datasets for testing algorithms and assessing their resilience to different degrees of client heterogeneity.

Further, different non-IID regimes may require the development of different mitigation strategies. For example, under feature-distribution skew, because $\Ploc(y \gvn x)$ is assumed to be common, the problem is at least in principle well specified, and training a single global model that learns $\Ploc(y \gvn x)$ may be appropriate. When the same features map to different labels on different clients, some form of personalization (\cref{sec:multimodel}) may be essential to learning the true labeling functions.

\paragraph{Violations of independence}
Violations of independence are introduced any time the distribution $\Pcli$ changes over the course of training; a prominent example is in cross-device FL, where devices typically need to meet eligibility requirements in order to participate in training (see \cref{sec:typical-training}). Devices typically meet those requirements at night local time (when they are more likely to be charging, on free wi-fi, and idle), and so there may be significant diurnal patterns in device availability. Further, because local time of day corresponds directly to longitude, this introduces a strong geographic bias in the source of the data. \citet{eichner19semicyclic} described this issue and some mitigation strategies, but many open questions remain.

\paragraph{Dataset shift}
Finally, we note that the temporal dependence of the distributions $\Pcli$ and $\Ploc$ may introduce dataset shift in the classic sense (differences between the train and test distributions). Furthermore, other criteria may make the set of clients eligible to train a federated model different from the set of clients where that model will be deployed.  For example, training may require devices with more memory than is needed for inference. These issues are explored in more depth in \cref{sec:fairness}. Adapting techniques for handling dataset shift to federated learning is another interesting open question.

\subsubsection{Strategies for Dealing with Non-IID Data}

The original goal of federated learning, training a single global model on the union of client datasets, becomes harder with non-IID data. One natural approach is to modify existing algorithms (e.g. through different hyperparameter choices) or develop new ones in order to more effectively achieve this objective. This approach is considered in \cref{sec:non-iid-algs}.

For some applications, it may be possible to augment data in order to make the data across clients more similar. One approach is to create a small dataset which can be shared globally. This dataset may originate from a publicly available proxy data source, a separate dataset from the clients’ data which is not privacy sensitive, or perhaps a distillation of the raw data following \citet{wang2018dataset}.

The heterogeneity of client objective functions gives additional importance to the question of how to craft the objective function --- it is no-longer clear that treating all examples equally makes sense. Alternatives include limiting the contributions of the data from any one user (which is also important for privacy, see \cref{sec:privacy}) and introducing other notions of fairness among the clients; see discussion in \cref{sec:fairness}. 

But if we have the capability to run training on the local data on each device (which is necessary for federated learning of a global model), is training a single global model even the right goal? There are many cases where having a single model is to be preferred, e.g. in order to provide a model to clients with no data, or to allow manual validation and quality assurance before deployment. Nevertheless, since local training is possible, it becomes feasible for each client to have a customized model. This approach can turn the non-IID problem from a bug to a feature, almost literally --- since each client has its own model, the client's identity effectively parameterizes the model, rendering some pathological but degenerate non-IID distributions trivial. For example, if for each $i$,  $\Ploc_i(y)$ has support on only a single label, finding a high-accuracy global model may be very challenging (especially if $x$ is relatively uninformative), but training a high-accuracy local model is trivial (only a constant prediction is needed). Such multi-model approaches are considered in depth in \cref{sec:multimodel}. In addition to addressing non-identical client distributions, using a plurality of models can also address violations of independence stemming from changes in client availability. For example, the approach of \citet{eichner19semicyclic} uses a single training run but averages different iterates in order to provide different models for inference based on the timezone / longitude of clients.

\subsection{Optimization Algorithms for Federated Learning}\label{sec:optimization}

In prototypical federated learning tasks, the goal is to learn a single global model that minimizes the empirical risk function over the entire training dataset, that is, the union of the data across all the clients. The main difference between federated optimization algorithms and standard distributed training methods 
is the need to address the characteristics of Table~\ref{tab:characteristics} --- for optimization, non-IID and unbalanced data, limited communication bandwidth, and unreliable and limited device availability are particularly salient. 

FL settings where the total number of devices is huge (e.g. across mobile devices) necessitate algorithms that only require a handful of clients to participate per round (client sampling). Further, each device is likely to participate no more than once in the training of a given model, so stateless algorithms are necessary. This rules out the direct application of a variety of approaches that are quite effective in the datacenter context, for example stateful optimization algorithms like ADMM, and stateful compression strategies that modify updates based on residual compression errors from previous rounds.

Another important practical consideration for federated learning algorithms is composability with other techniques. Optimization algorithms do not run in isolation in a production deployment, but need to be combined with other techniques like cryptographic secure aggregation protocols (Section~\ref{sssec:secure_computations}), differential privacy (DP) (Section~\ref{sssec:private_disclosures}), and model and update compression (Section~\ref{sec:compr}). As noted in Section~\ref{sec:typical-training}, many of these techniques can be applied to primitives like ``\texttt{sum over selected clients}'' and ``\texttt{broadcast to selected clients}'', and so expressing optimization algorithms in terms of these primitives provides a  valuable separation of concerns, but may also exclude certain techniques such as applying updates asynchronously.

One of the most common approaches to optimization for federated learning is the Federated Averaging algorithm \citep{mcmahan17fedavg}, an adaption of local-update or parallel SGD.\footnote{Federated Averaging applies local SGD to a randomly sampled subset of clients on each round, and proposes a specific update weighting scheme.} Here, each client runs some number of SGD steps locally, and then the updated local models are averaged to form the updated global model on the coordinating server. Pseudocode is given in Algorithm~\ref{alg:fedavg}.

Performing local updates and  communicating less frequently with the central server addresses the core challenges of respecting data locality constraints and of the limited communication capabilities of mobile device clients. However, this family of algorithms also poses several new algorithmic challenges from an optimization theory point of view. In Section~\ref{sec:optimization}, we discuss recent advances and open challenges in federated optimization algorithms for the cases of IID and non-IID data distribution across the clients respectively. The development of new algorithms that specifically target the characteristics of the federated learning setting remains an important open problem.

\newcommand{\grad}{\triangledown}
\newcommand{\T}{\rule{0pt}{2.2ex}}

\begin{figure}

\begin{minipage}[t]{.4\textwidth}
\begin{tabular}[t]{rl}    
\toprule
\T $N$ & Total number of clients \\
\T $M$ & Clients per round \\
\T $T$ & Total communication rounds \\
\T $K$ & Local steps per round. \\
\bottomrule
\end{tabular}
\captionof{table}{Notation for the discussion of FL algorithms including Federated Averaging.}
\label{tab:notation}
\vfill
\end{minipage}
\hfill
\begin{minipage}[t]{.55\textwidth}
\rule{\textwidth}{\heavyrulewidth}  
\vspace{-0.16in}
\begin{algorithmic}
\renewcommand{\arraystretch}{1.6}
\SUB{Server executes:}
  \STATE initialize $x_0$
  \FOR{each round $t = 1, 2, \dots$, T}
     \STATE $S_t \leftarrow$ (random set of $M$ clients)
     \FOR{each client $i \in S_t$ \textbf{in parallel}}
      \STATE $x_{t+1}^i \leftarrow \text{ClientUpdate}(i, x_t)$ 
     \ENDFOR
     \STATE $x_{t+1} \leftarrow \sum_{k=1}^M \frac{1}{M} x_{t+1}^i$
  \ENDFOR
  \STATE

 \SUB{ClientUpdate($i, x$):}\ \ \
    \FOR{local step $j = 1, \dots, K$}
     
      \STATE $x \leftarrow x - \eta \grad f(x; z)$ for $z \sim \mathcal{P}_i$
    \ENDFOR
 \STATE return $x$ to server
 \end{algorithmic}
 \rule{\textwidth}{\heavyrulewidth}
 \captionof{algorithm}{Federated Averaging (local SGD), when all clients have the same amount of data.}\label{alg:fedavg}
 \vfill
\end{minipage}

\end{figure}

\subsubsection{Optimization Algorithms and Convergence Rates for IID Datasets}
\label{sec:iid}

While a variety of different assumptions can be made on the per-client functions being optimized, the most basic split is between assuming IID and non-IID data. Formally, having IID data at the clients means that each mini-batch of data used for a client's local update is statistically identical to a uniformly drawn sample (with replacement) from the entire training dataset (the union of all local datasets at the clients). Since the clients independently collect their own training data which vary in both size and distribution, and these data are not shared with other clients or the central node, the IID assumption clearly almost never holds in practice. However, this assumption greatly simplifies theoretical convergence analysis of federated optimization algorithms, as well as establishes a baseline that can be used to understand the impact of non-IID data on optimization rates. Thus, a natural first step is to obtain an understanding of the landscape of optimization algorithms for the IID data case.


\noindent Formally, for the IID setting let us standardize the stochastic optimization problem
\[ 
   \min_{x \in \R^m} F(x) := \E_{z \sim \mathcal{P}} [f(x; z)] \,.
\]
We assume an intermittent communication model as in e.g.\ \citet[Sec. 4.4]{woodworth18graphoracle}, where $M$ stateless clients participate in each of $T$ rounds, and during each round, each client can compute gradients for $K$ samples (e.g. minibatches) $z_1, \dots, z_K$ sampled IID from $\mathcal{P}$ (possibly using these to take sequential steps). In the IID-data setting clients are interchangeable, and we can without loss of generality assume $M=N$. Table~\ref{tab:notation} summarizes the notation used in this section.

Different assumptions on $f$ will produce different guarantees. We will first discuss the convex setting and later review results for  non-convex problems.

\paragraph{Baselines and state-of-the-art for convex problems}
In this section we review convergence results for $H$-smooth, convex (but not necessarily strongly convex) functions under the assumption that the variance of the stochastic gradients is bounded by $\sigma^2$.
More formally, by $H$-smooth we mean that for all $z$, $f(\cdot; z)$ is differentiable and has a $H$-Lipschitz gradient, that is, for all choices of $x, y$
\[ 
    \|\nabla f(x, z) - \nabla f(y, z)\| \leq H\|x-y\|.
\]
We also assume that for all $x$, the stochastic gradient $\nabla_x f(x; z)$ satisfies
\[
    \E_{z \sim \mathcal{P}}\|\nabla_x f(x; z) - \nabla F(x)\| \leq \sigma^2.
\]
When analyzing the convergence rate of an algorithm with output $x_T$ after $T$ iterations, we consider the term
\begin{equation}\label{eq:convergence_rate_term}
    \E[F(x_T)] - F(x^*)
\end{equation}
where $x^* = \arg\min_x F(x)$. All convergence rates discussed herein are upper bounds on this term.
A summary of convergence results for such functions is given in Table~\ref{tab:iid-convergence}. 

Federated averaging (a.k.a.\ parallel SGD/local SGD) competes with two natural baselines: First, we may keep $x$ fixed in local updates during each round, and compute a total of $K M$ gradients at the current $x$, in order to run accelerated minibatch SGD. Let $\bar{x}$ denote the average of $T$ iterations of this algorithm. We then have the upper bound
\[
 \BO \left(\frac{H}{T^2} + \frac{\sigma}{\sqrt{T K M}}\right)
\]
for convex objectives \cite{Lan2012,cotter2011acmb,dekel12optimal}. Note that the first expectation is taken with respect to the randomness of $z$ in the training procedure as well.

A second natural baseline is to ignore all but 1 of the $M$ active clients, which allows (accelerated) sequential SGD to execute for $K T$ steps. Applying the same general bounds cited above, this approach offers an upper bound of
\[
 \BO\left( \frac{H}{(TK)^2} + \frac{\sigma}{\sqrt{TK}}\right).
\]
Comparing these two results, we see that minibatch SGD attains the optimal `statistical' term ($\nicefrac{\sigma}{\sqrt{TKM}}$), whilst SGD on a single device (ignoring the updates of the other devices) achieves the optimal `optimization' term ($\nicefrac{H}{(TK)^2}$). 

The convergence analysis of local-update SGD methods is an active current area of research \citep{stich2018local, lin2018don, yu2018parallel, wang2018cooperative, reisizadeh2019fedpaq,patel19communication,khaled2019better,woodworth2020local}. 
The first convergence results for local-update SGD methods were derived under the bounded gradient norm  assumption in \citet{stich2018local} for strongly-convex and in \citet{yu2018parallel} for non-convex objective functions. These analyses could attain the desired $\sigma/\sqrt{TKM}$ statistical term with suboptimal optimization term (in Table~\ref{tab:iid-convergence} we summarize these results for the middle ground of convex functions).

By removing the bounded gradient assumption, \citet{wang2018cooperative} and \citet{stich2019error} could further improve the optimization term to $HM/T$. These result show that if the number of local steps $K$ is smaller than $T/M^3$ then the (optimal) statistical term is dominating the rate. However, for typical cross-device applications we might have $T=10^6$ and $M=100$ (\cref{tab:sizes}), implying $K=1$.

Often in the literature the convergence bounds are accompanied by a discussion on how large $K$ may be chosen in order to reach asymptotically the same statistical term as the  convergence rate of mini-batch SGD. For strongly convex functions, this bound was improved by \citet{khaled2019better} and further in \citet{stich2019error}.

\begin{table}
\begin{centering}
\begin{minipage}{\linewidth}
\centering { 
\begin{tabular}{@{}llll@{\hskip-2pt}l@{}} 
\toprule
\phantom{X}  & Method & Comments & \multicolumn{2}{l}{Convergence}  \\ 
 \midrule
\multicolumn{4}{l}{\small \textbf{\textit{Baselines}}} \\
 & mini-batch SGD & batch size $KM$       
      & $\BO \left(\frac{H}{T}\right.$ & $ \left. + \frac{\sigma}{\sqrt{TKM}} \right)$  \\ 
 & SGD & (on 1 worker, no communication) 
      & $\BO \left(\frac{H}{TK}\right.$ & $\left. + \frac{\sigma}{\sqrt{TK}} \right)$  \\
\midrule
\multicolumn{4}{l}{\small \textbf{\textit{Baselines with acceleration}}\footnote{There are no accelerated fed-avg/local SGD variants so far}} \\
 & A-mini-batch SGD~\cite{Lan2012,cotter2011acmb} & batch size $KM$        
     & $\BO \left(\frac{H}{T^2}\right.$ & $\left. + \frac{\sigma}{\sqrt{TKM}} \right)$  \\
 & A-SGD~\cite{Lan2012}   & (on 1 worker, no communication) 
     & $\BO \left(\frac{H}{(TK)^2}\right.$ & $\left. + \frac{\sigma}{\sqrt{TK}} \right)$  \\
\midrule
\multicolumn{4}{l}{\small \textbf{\textit{Parallel SGD / Fed-Avg / Local SGD}}} \\
 & \citet{yu2018parallel}\footnote{This paper considers the smooth non-convex setting, we adapt here the results for our setting.\label{fn:non-convex}}, \citet{stich2018local}\footnote{This paper considers the smooth strongly convex setting, we adapt here the results for our setting.}  & gradient norm bounded by $G$ 
     & $\BO \left(\frac{HKM}{T}\frac{G^2}{\sigma^2}\right.$ &$\left. + \frac{\sigma}{\sqrt{TKM}} \right)$ \\
 & \multicolumn{2}{l}{\citet{wang2018cooperative}\footref{fn:non-convex}, \citet{stich2019error}}
     & $\BO \left(\frac{HM}{T} \right.$ & $\left. + \frac{\sigma}{\sqrt{TKM}} \right)$ \\ \midrule
\multicolumn{4}{l}{\small \textbf{\textit{Other algorithms}}} \\
 & SCAFFOLD \cite{karimireddy2019scaffold} & control variates and two stepsizes 
     & $\BO \left(\frac{H}{T} \right.$ & $\left. + \frac{\sigma}{\sqrt{TKM}} \right)$ \\
\bottomrule
\end{tabular}
}%
\end{minipage}
\caption{Convergence rates for a (non-comprehensive) set of distributed optimization algorithms in the IID-data setting. We assume $M$ devices participate in each iterations, and the loss functions are $H$-smooth, convex, and we have access to stochastic gradients with variance at most $\sigma^2$. All rates are upper bounds on~\eqref{eq:convergence_rate_term} after $T$ iterations (potentially with some iterate averaging scheme).\\
}
\label{tab:iid-convergence}
\end{centering}
\end{table}


For non-convex objectives, \citet{yu2018parallel} showed that local SGD can achieve asymptotically an error bound $1/\sqrt{TKM}$ if the number of local updates $K$ are smaller than $T^{1/3}/M$. This convergence guarantee was further improved by \citet{wang2018cooperative} who removed the bounded gradient norm assumption and showed that the number of local updates can be as large as $T/M^3$. The analysis in \cite{wang2018cooperative} can also be applied to other algorithms with local updates, and thus yields the first convergence guarantee for decentralized SGD with local updates (or periodic decentralized SGD) and elastic averaging SGD \cite{zhang2015deep}.
\citet{haddadpour2019local} improves the bounds in \citet{wang2018cooperative} for functions satisfying the Polyak-Lojasiewicz (PL) condition \citep{karimi2016linear}, a generalization of strong convexity. In particular, \citet{haddadpour2019local} show that for PL functions, $T^2/M$ local updates per round leads to a $\BO(1/TKM)$ convergence.

While the above works focus on convergence as a function of the number of iterations performed, practitioners often care about wall-clock convergence speed. Assessing this must take into account the effect of the design parameters on the time spent per iteration based on the relative cost of communication and local computation. Viewed in this light, the focus on seeing how large $K$ can be while maintaining the statistical rate may not be the primary concern in federated learning, where one may assume almost infinite datasets (very large $N$). The costs (at least in wall-clock time) are small for increasing $M$, and so it may be more natural to increase $M$ sufficiently to match the optimization term, and then tune $K$ to maximize wall-clock optimization performance. 
How then to choose $K$? Performing more local updates at the clients will increase the divergence between the resulting local models at the clients, before they are averaged. As a result, the error convergence in terms of training loss versus the total number of sequential SGD steps $TK$ is slower. However, performing more local updates saves significant communication cost and reduces the time spent per iteration. The optimal number of local updates strikes a balance between these two phenomena and achieves the fastest error versus wallclock time convergence. \citet{wang2018adaptive} propose an adaptive communication strategy that adapts $K$ according to the training loss at regular intervals during the training.

Another important design parameter in federated learning is the model aggregation method used to update the global model using the updates made by the selected clients. In the original federated learning paper, \citet{mcmahan17fedavg} proposes taking a weighted average of the local models, in proportion to the size of local datasets. For IID data, where each client is assumed to have a infinitely large dataset, this reduces to taking a simple average of the local models. However, it is unclear whether this aggregation method will result in the fastest error convergence. 



There are many open questions in federated optimization, even with IID data. 
\citet{woodworth18graphoracle} highlights several gaps between upper and lower bounds for optimization relevant to the federated learning setting, particularly for ``intermittent communication graphs'', which captures local SGD approaches, but convergence rates for such approaches are not known to match the corresponding lower bounds. In Table~\ref{tab:iid-convergence} we highlight convergence results for the convex setting. Whilst most schemes are able to reach the asymptotically dominant statistical term, none are able to match the convergence rate of accelerated mini-batch SGD. It is an open problem if federated averaging algorithms can close this gap.

Local-update SGD methods where all $M$ clients perform the same number of local updates may suffer from a common scalability issue---they can be bottlenecked if any one client unpredictably slows down or fails. Several approaches for dealing with this are possible, but it is far from clear which are optimal, especially when the potential for bias is considered (see \cref{sec:fairness}). \citet{bonawitz19sysml} propose over-provisioning clients (e.g., request updates from $1.3 M$ clients), and then accepting the first $M$ updates received and rejecting updates from stragglers. A slightly more sophisticated solution is to fix a time window and allow clients to perform as many local updates $K_i$ as possible within this time, after which their models are averaged by a central server. \citet{wang2020tackling} analyzed the computational heteogeneity introduced by this approach in theory. An alternative method to overcome the problem of straggling clients is to fix the number of local updates at $\tau$, but allow clients to update the global model in an asynchronous or lock-free fashion. Although some previous works \citep{zhang2015deep, Lian2018, dutta2018slow} have proposed similar methods, the error convergence analysis is an open and challenging problem. A larger challenge in the FL setting, however, is that as discussed at the beginning of \cref{sec:optimization}, asynchronous approaches may be difficult to combine with complimentary techniques like differential privacy or secure aggregation. 

Besides the number of local updates, the choice of the size of the set of clients selected per training round presents a similar trade-off as the number of local updates. Updating and averaging a larger number of client models per training round yields better convergence, but it makes the training vulnerable to slowdown due to unpredictable tail delays in computation/communication at/with the clients. 


The analysis of local SGD / Federated Averaging in the non-IID setting is even more challenging; results and open questions related to this are considered in the next section, along with specialized algorithms which directly address the non-IID problem.

\begin{table}
\renewcommand{\arraystretch}{1.1}
\begin{center}
\begin{tabularx}{\textwidth}{ccX}
\multicolumn{3}{c}{\textbf{Non-IID assumptions}} \\
\toprule
\textbf{Symbol} & \textbf{Full name} & \textbf{Explanation} \\
\midrule
  BCGV &  bounded inter-client gradient variance & $\E_i \|\nabla f_i(x) - \nabla F(x) \|^2 \leq \eta^2 $ \\
  BOBD & bounded optimal objective difference & $F^* - \E_i[f^*_i] \leq \eta^2 $\\
  BOGV & bounded optimal gradient variance & $\E_i \|\nabla f_i(x^*) \|^2 \leq \eta^2$ \\
  BGV & bounded gradient dissimilarity & $\nicefrac{\E_i \|\nabla f_i(x) \|^2}{\|\nabla F(x) \|^2} \leq \eta^2$ \\
\bottomrule
\end{tabularx}
\vspace{0.5cm}
\begin{tabularx}{\textwidth}{lX}
\\
\multicolumn{2}{c}{\textbf{Other assumptions and variants}} \\
\toprule
\textbf{Symbol} & \textbf{Explanation} \\
\midrule
  CVX &  Each client function $f_i(x)$ is convex.   \\
  SCVX & Each client function $f_i(x)$ is $\mu$-strongly convex. \\
  BNCVX & Each client function has bounded nonconvexity with $\nabla^2 f_i(x) \succeq -\mu I $. \\
  \hline
  BLGV & The variance of stochastic gradients on local clients is bounded. \\
  BLGN & The norm of any local gradient is bounded. \\
  LBG & Clients use the full batch of local samples to compute updates. \\
  \hline
  Dec & Decentralized setting, assumes the the connectivity of network is good. \\
  AC & All clients participate in each round. \\
  1step & One local update is performed on clients in each round. \\
  Prox & Use proximal gradient steps on clients. \\
  VR & Variance reduction which needs to track the state. \\
\bottomrule
\end{tabularx}
\vspace{0.5cm}
\begin{tabularx}{\textwidth}{llllX}
\multicolumn{5}{c}{\textbf{Convergence rates}}\\
\toprule
\textbf{Method} & \textbf{Non-IID} & \textbf{Other assumptions} & \textbf{Variant} & \textbf{Rate} \\
\midrule
\citet{Lian2017b} & BCGV  &  BLGV &  Dec; AC; 1step & $O(\nicefrac{1}{T}) + O(\nicefrac{1}{\sqrt{NT}})$\\
\hline
PD-SGD \citep{li2019communication} & BCGV & BLGV & Dec; AC & $O(\nicefrac{N}{T}) + O(\nicefrac{1}{\sqrt{NT}})$ \\
\hline
MATCHA \citep{wang2019matcha} & BCGV & BLGV & Dec & $O(\nicefrac{1}{\sqrt{TKM}}) + O(\nicefrac{M}{KT})$ \\
\hline
\citet{khaled2019analysis} &  BOGV   & CVX & AC; LBG & $O(\nicefrac{N}{T}) + O(\nicefrac{1}{\sqrt{NT}})$\\
\hline
\citet{li2019convergence} & BOBD & SCVX; BLGV; BLGN & - & $O(\nicefrac{K}{T})$\\
\hline
FedProx \citep{li2018federated} & BGV & BNCVX & Prox &$O(\nicefrac{1}{\sqrt{T}})$\\
\hline
SCAFFOLD \citep{karimireddy2019scaffold} & - & SCVX; BLGV & VR & $O(\nicefrac{1}{TKM})+O(e^{-T})$
\\
\bottomrule
\end{tabularx}
\caption{Convergence rates for a (non-comprehensive) set of federated optimization methods in non-IID settings. We summarize the key assumptions for non-IID data, local functions on each client, and other assumptions. We also present the variant of the algorithm comparing to Federated Averaging and the convergence rates that eliminate constant.
}
\label{tab:non-iid-convergence}
\end{center}
\end{table}

\subsubsection{Optimization Algorithms and Convergence Rates for Non-IID Datasets}\label{sec:non-iid-algs}

In contrast to well-shuffled mini-batches consisting of independent and identically distributed (IID) examples in centralized learning, federated learning uses local data from end user devices, leading to many varieties of non-IID data (Section~\ref{sec:noniid}).

In this setting, each of $N$ clients has a local data distribution $\mathcal{P}_i$ and a local objective function
$$f_i(x) = \E_{z \sim \mathcal{P}_i}[ f(x ; z)]$$
where we recall that $f(x ; z)$ is the loss of a model $x$ at an example $z$. We typically wish to minimize
\begin{equation}
F(x) = \frac{1}{N}\sum_{i=1}^N f_i(x) \,.
\end{equation}
Note that we recover the IID setting when each $\mathcal{P}_i$ is identical. We will let $F^*$ denote the minimum value of $F$, obtained the point $x^*$. Analogously, we will let $f_i^*$ denote the minimum value of $f_i$.

As in the IID setting, we assume an intermittent communication model (e.g.\ \citet[Sec. 4.4]{woodworth18graphoracle}), where $M$ stateless clients participate in each of $T$ rounds, and during each round, each client can compute gradients for $K$ samples (e.g. minibatches). The difference here is that the samples $z_{i, 1}, \ldots, z_{i, K}$ sampled at client $i$ are drawn from the client's local distribution $\mathcal{P}_i$. Unlike the IID setting, we cannot necessarily assume $M = N$, as the client distributions are not all equal. In the following, if an algorithm relies on $M = N$, we will omit $M$ and simply write $N$. We note that while such an assumption may be compatible with the cross-silo federated setting in Table \ref{tab:characteristics}, it is generally infeasible in the cross-device setting.

While \citep{stich2018local,yu2018parallel,wang2018cooperative,stich2019error} mainly focused on the IID case, the analysis technique can be extended to the non-IID case by adding an assumption on data dissimilarities, for example by constraining the difference between client gradients and the global gradient \citep{Lian2017b,li2018federated,li2019communication,wang2019matcha,wang2020tackling} or the difference between client and global optimum values \citep{li2019convergence,khaled2019analysis}. Under this assumption, \citet{yu2019linear} showed that the error bound of local SGD in the non-IID case becomes worse. In order to achieve the rate of $1/\sqrt{TKN}$ (under non-convex objectives), the number of local updates $K$ should be smaller than $T^{1/3}/N$, instead of $T/N^3$
as in the IID case \cite{wang2018cooperative}. \citet{li2018federated} proposed to add a proximal term in each local objective function so as to make the algorithm be more robust to the heterogeneity across local objectives. The proposed FedProx algorithm empirically improves the performance of federated averaging. 
\citet{khaled2019analysis}  assumes all clients participate, and uses batch gradient descent on clients, which can potentially converge faster than stochastic gradients on clients.  

Recently, a number of works have made progress in relaxing the assumptions necessary for analysis so as to better apply to practical uses of Federated Averaging. For example, \citet{li2019convergence} studied the convergence of Federated Averaging in a more realistic setting where only a subset of clients are involved in each round. In order to guarantee the convergence, they assumed that the clients are selected either uniformly at random or with probabilities that are in proportion to the sizes of local datasets. 
Nonetheless, in practice the server may not be able to sample clients in these idealized ways --- in particular, in cross-device settings only devices that meet strict eligibility requirements (e.g. charging, idle, free WiFi) will be selected to participate in the computation. At different times within a day, the clients characteristics can vary significantly. \citet{eichner19semicyclic} formulated this problem and studied the convergence of semi-cyclic SGD, where multiple blocks of clients with different characteristics are sampled from following a regular cyclic pattern (e.g. diurnal). Clients can perform different local steps because of heterogeneity in their computing capacities. \citet{wang2020tackling} proves that FedAvg and many other federated learning algorithms will converge to the stationary points of a mismatched objective function in the presence of heterogeneous local steps. They refer to this problem as \emph{objective inconsistency} and propose a simple technique to eliminate the inconsistency problem from federated learning algorithms.

We summarize recent theoretical results in Table~\ref{tab:non-iid-convergence}. All the methods in \cref{tab:non-iid-convergence} assume smoothness or Lipschitz gradients for the local functions on clients. 
The error bound is measured by optimal objective \eqref{eq:convergence_rate_term} for convex functions and norm of gradient for nonconvex functions. 
For each method, we present the key non-IID assumption, assumptions on each client function $f_i(x)$, and other auxiliary assumptions. We also briefly describe each method as a variant of the federated averaging algorithm, and show the simplified convergence rate eliminating constants.
 Assuming the client functions are strongly convex could help the convergence rate \citep{li2019convergence,karimireddy2019scaffold}.
 Bounded gradient variance, which is a widely used assumption to analyze stochastic gradient methods, is often used when clients use stochastic local updates \citep{Lian2017b,li2019convergence,li2019communication,wang2019matcha,karimireddy2019scaffold}. 
 \citet{li2019convergence} directly analyzes the Federated Averaging algorithm, which applies $K$ steps of local updates on randomly sampled $M$ clients in each round, and presents a rate that suggests local updates $(K > 1)$ could slow down the convergence. Clarifying the regimes where $K > 1$ may hurt or help convergence is an important open problem.

\paragraph{Connections to decentralized optimization} The objective function of federated optimization has been studied for many years in the decentralized optimization community. 
As first shown in \citet{wang2018cooperative}, the convergence analysis of decentralized SGD can be applied to or combined with local SGD with a proper setting of the network topology matrix (mixing matrix). In order to reduce the communication overhead, \citet{wang2018cooperative} proposed periodic decentralized SGD (PD-SGD) which allows decentralized SGD to have multiple local updates as Federated Averaging. This algorithm is  extended by \citet{li2019communication} to the non-IID case. MATCHA~\citep{wang2019matcha} further improves the performance of PD-SGD by randomly sampling clients for computation and communication, and provides a convergence analysis showing that local updates can accelerate convergence.

\paragraph{Acceleration, variance reduction and adaptivity}
Momentum, variance-reduction, and adaptive learning rates are all promising techniques to improve convergence and generalization of first-order methods. However, there is no single manner in which to incorporate these techniques into FedAvg. SCAFFOLD \citep{karimireddy2019scaffold} models the difference in client updates using control variates to perform variance reduction. Notably, this allows convergence results not relying on bounding the amount of heterogeneity among clients. As for momentum, \citet{yu2019linear} propose allowing each client to maintain a local momentum buffer and average the local buffers and the local model parameters at each communication round. Although this method empirically improves the final accuracy of local SGD, this doubles the per-round communication cost. A similar scheme is used by \citet{xie2019local} to design a variant of local SGD in which clients locally perform Adagrad~\citep{mcmahan2010adaptive, duchi2011adaptive}. \citet{reddi2020adaptive} instead proposes using adaptive learning rates at the server-level, developing federated versions of adaptive optimization methods with the same communication cost as FedAvg. This framework generalizes the server momentum framework proposed by \citet{hsu2019measuring,wang2019slowmo}, which allows momentum without increasing communication costs. While both \citep{yu2019linear,wang2019slowmo} showed that the momentum variants of local SGD can converge to stationary points of non-convex objective functions at the same rate as synchronous mini-batch SGD, it is challenging to prove momentum accelerates the convergence rate in the federated learning setting. Recently, \citet{karimireddy2020mime} proposed a general approach for adapting centralized optimization algorithms to the heterogeneous federated setting (MIME framework and algorithms).


\subsection{Multi-Task Learning, Personalization, and Meta-Learning}
\label{sec:multimodel}

In this section we consider a variety of ``multi-model'' approaches --- techniques that result in effectively using different models for different clients at inference time. These techniques are particularly relevant when faced with non-IID data (Section~\ref{sec:noniid}), since they may outperform even the best possible shared global model. We note that personalization has also been studied in the fully decentralized setting \citep{Vanhaesebrouck2017,Bellet2018a,Zantedeschi2019,Almeida2018}, where training individual models is particularly natural.

\subsubsection{Personalization via Featurization}
The remainder of this section specifically considers techniques that result in different users running inference with different model parameters (weights). However, in some applications similar benefits can be achieved by simply adding user and context features to the model. For example, consider a language model for next-word-prediction in a mobile keyboard as in \citet{hard18gboard}. Different clients are likely to use language differently, and in fact on-device personalization of model parameters has yielded significant improvements for this problem \citep{wang2019federated}. However, a complimentary approach may be to train a federated model that takes as input not only the words the user has typed so far, but a variety of other user and context features---What words does this user frequently use? What app are they currently using? If they are chatting, what messages have they sent to this person before? Suitably featurized, such inputs can allow a shared global model to produce highly personalized predictions. However, largely because few public datasets contain such auxiliary features, developing model architectures that can effectively incorporate context information for different tasks remains an important open problem with the potential to greatly increase the utility of FL-trained models.

\subsubsection{Multi-Task Learning}
\label{sss:multitask-learning}
If one considers each client's local problem (the learning problem on the local dataset) as a separate task (rather than as a shard of a single partitioned dataset), then techniques from multi-task learning \citep{DBLP:journals/corr/ZhangY17aa} immediately become relevant. Notably, \citet{Smith2017} introduced the MOCHA algorithm for multi-task federated learning, directly tackling challenges of communication efficiency, stragglers, and fault tolerance. In multi-task learning, the result of the training process is one model per task. Thus, most multi-task learning algorithms assume all clients (tasks) participate in each training round, and also require stateful clients since each client is training an individual model. This makes such techniques relevant for cross-silo FL applications, but harder to apply in cross-device scenarios.

Another approach is to reconsider the relationship between clients (local datasets) and learning tasks (models to be trained), observing that there are points on a spectrum between a single global model and different models for every client. For example, it may be possible to apply techniques from multi-task learning (as well as other approaches like personalization, discussed next), where we take the ``task'' to be a subset of the clients, perhaps chosen explicitly (e.g. based on geographic region, or characteristics of the device or user), or perhaps based on clustering \citep{mansour2020three} or the connected components of a learned graph over the clients \citep{Zantedeschi2019}. The development of such algorithms is an important open problem.  See \cref{sssec:training_submodels} for a discussion of how sparse federated learning problems, such as those arising naturally in this type of multi-task problem, might be approached without revealing to which client subset (task) each client belongs.


\subsubsection{Local Fine Tuning and Meta-Learning}

By local fine tuning, we refer to techniques which begin with the federated training of a single model, and then deploy that model to all clients, where it is personalized by additional training on the local dataset before use in inference. This approach integrates naturally into the typical lifecycle of a model in federated learning (Section~\ref{sec:lifecycle}). Training of the global model can still proceed using only small samples of clients on each round (e.g. 100s); the broadcast of the global model to all clients (e.g. many millions) only happens once, when the model is deployed. The only difference is that before the model is used to make live predictions on the client, a final training process occurs, personalizing the model to the local dataset.

Given a global model that performs reasonably well, what is the best way to personalize it?  In non-federated learning, researchers often use fine-tuning, transfer learning, domain adaptation \cite{mansour2009domain,cortes2014domain,ben2010theory, mansour2020theory, cortes2020multiple}, or interpolation with a personal local model. Of course, the precise technique used for such interpolations is key and it is important to determine its corresponding learning guarantees in the context of federated learning. Further, these techniques often assume only a pair of domains (source and target), and so some of the richer structure of federated learning may be lost.

One approach for studying personalization and non-IID data is via a connection to {\em meta-learning}, which has emerged as a popular setting for model adaptation.
In the standard learning-to-learn (LTL) setup \citep{baxter00model}, one has a meta-distribution over tasks, samples from which are used to learn a learning algorithm, for example by finding a good restriction of the hypothesis space. This is in fact a good match for the statistical setting discussed in Section~\ref{sec:noniid}, where we sample a client (task) $i \sim \Pcli$, and then sample data for that client (task) from $\Ploc_i$.

Recently, a class of algorithms referred to as \emph{model-agnostic meta-learning} (MAML) have been developed that meta-learn a global model, which can be used as a starting point for learning a good model adapted to a given task, using only a few local gradient steps \citep{finn17maml}. Most notably, the training phase of the popular Reptile algorithm \citep{nichol18reptile} is closely related to Federated Averaging \citep{mcmahan17fedavg} --- Reptile allows for a server learning rate and assumes all clients have the same amount of data, but is otherwise the same. \citet{khodak19adaptive} and \citet{jiang2019improving} explore the connection between FL and MAML, and show how the MAML setting is a relevant framework to model the personalization objectives for FL. \citet{chai2019personalization} applied local fine tuning to personalize speech recognition models in federated learning. \citet{fallah2020personalized} developed a new algorithm called Personalized FedAvg by connecting MAML instead of Reptile to federated learning. Additional connections with differential privacy were studied in \citep{li19dpmeta}.

The general direction of combining ideas from FL and MAML is relatively new, with many open questions:
\begin{itemize}
	\item The evaluation of MAML algorithms for supervised tasks is largely focused on synthetic image classification problems \citep{lake11omniglot,ravi17miniimagenet} in which infinite artificial tasks can be constructed by subsampling from classes of images. FL problems, modeled by existing datasets used for simulated FL experiments (Appendix~\ref{sec:datasets-and-software}), can serve as realistic benchmark problems for MAML algorithms.
	\item In addition to an empirical study, or optimization results, it would be useful to analyze the theoretical guarantees of MAML-type techniques and study under what assumptions they can be successful, as this will further elucidate the set of FL domains to which they may apply.
	\item The observed gap between the global and personalized acccuracy \citep{jiang2019improving} creates a good argument that personalization should be of central importance to FL. However, none of the existing works clearly formulates what would be comprehensive metrics for measuring personalized performance; for instance, is a small improvement for every client preferable to a larger improvement for a subset of clients? See Section~\ref{sec:fairness} for a related discussion.
	\item \citet{jiang2019improving} highlighted the fact that models of the same structure and performance, but trained differently, can have very different capacity to personalize. In particular, it appears that training models with the goal of maximizing global performance might actually hurt the model's capacity for subsequent personalization. Understanding the underlying reasons for this is a question relevant for both FL and the broader ML community.
	\item Several challenging FL topics including personalization and privacy have begun to be studied in this multi-task/LTL framework \cite{khodak19adaptive,jiang2019improving,li19dpmeta}. Is it possible for other issues such as concept drift to also be analyzed in this way, for example as a problem in lifelong learning \citep{silver13lifelong}?
	\item Can non-parameter transfer LTL algorithms, such as ProtoNets \citep{snell17protonets}, be of use for FL?
\end{itemize}

\newcommand{\hFL}{h_{\text{FL}}}
\subsubsection{When is a Global FL-trained Model Better?}
What can federated learning do for you that local training on one device cannot? When local datasets are small and the data is IID, FL clearly has an edge, and indeed, real-world applications of federated learning \cite{yang18gboardquery, hard18gboard, chen19oov} benefit from training a single model across devices. On the other hand, given pathologically non-IID distributions (e.g. $\Ploc_i(y \gvn x)$ directly disagree across clients), local models will do much better. Thus, a natural theoretical question is to determine under what conditions the shared global model is better than independent per-device models. Suppose we train a model $h_k$ for each client $k$, using the sample of size $m_k$ available from that client. Can we guarantee that the model $\hFL$ learned via federated learning is at least as accurate as $h_k$ when used for client $k$?
Can we quantify how much improvement can be expected via federated leaning?
And can we develop personalization strategies with theoretical guarantees that at least match the performance of both natural baselines ($h_k$ and $\hFL$)?

Several of these problems relate to previous work on multiple-source adaptation and agnostic federated learning \citep{mansour2009domain,mansour2009domainb,hoffman2018algorithms,Mohri2019}. The hardness of these questions depends on how the data is distributed among parties. For example, if data is vertically partitioned, each party  maintaining private records of different feature sets about common entities, these problems may require addressing record linkage \cite{christen12} within the federated learning task. Independently of the eventual technical levy of carrying out record linkage privately \cite{schnell11}, the task itself happens to be substantially noise prone in the real world \cite{sEP} and only sparse results have addressed its impact on training models \cite{Hardy2017-da}. Techniques for robustness and privacy can make local models relatively stronger, particularly for non-typical clients \citep{yu2020salvaging}. Loss factorization tricks can be used in supervised learning to alleviate up to the vertical partition assumption itself, but the practical benefits depend on the distribution of data and the number of parties \cite{pnhcFL}.

\subsection{Adapting ML Workflows for Federated Learning}
\label{sec:workflows}


Many challenges arise when adapting standard machine learning workflows and pipelines (including data augmentation, feature engineering, neural architecture design, model selection, hyperparameter optimization, and debugging) to decentralized datasets and resource-constrained mobile devices. We discuss several of these challenges below.

\subsubsection{Hyperparameter Tuning}
Running many rounds of training with different hyperparameters on resource-constrained mobile devices may be restrictive. For small device populations, this might result in the over-use of limited communication and compute resources. 
However, recent deep neural networks crucially depend on a wide range of hyperparameter choices regarding the neural network’s architecture, regularization, and optimization. Evaluations can be  expensive for large models and large-scale on-device datasets. Hyperparameter optimization (HPO) has a long history under the framework of AutoML \cite{ripley1993statistical,king1995statlog,kohavi1995automatic}, but it mainly concerns how to improve the model accuracy \cite{bergstra2011algorithms,snoek2015scalable,pedregosa2016hyperparameter,falkner2018bohb} rather than communication and computing efficacy for mobile devices. Therefore, we expect that further research should consider developing solutions for efficient hyperparameter optimization in the context of federated learning.

In addition to general-purpose approaches to the hyperparameter optimization problem, in the training space specifically the development of easy-to-tune optimization algorithms is a major open area. Centralized training already requires tuning parameters like learning rate, momentum, batch size, and regularization. Federated learning adds potentially more hyperparameters --- separate tuning of the aggregation / global model update rule and local client optimizer, number of clients selected per round, number of local steps per round, configuration of update compression algorithms, and more. Such hyperparameters can be crucial to obtaining a good trade-off between accuracy and convergence, and may actually impact the quality of the learned model~\citep{charles2020outsized}. In addition to a higher-dimensional search space, federated learning often also requires longer wall-clock training times and limited compute resources. These challenges could be addressed by optimization algorithms that are robust to hyperparameter settings (the same hyperparameter values work for many different real world datasets and architectures), as well as adaptive or self-tuning algorithms~\cite{thakkar2019differentially,bonawitz2019autotune}.

\subsubsection{Neural Architecture Design}
Neural architecture search (NAS) in the federated learning setting is motivated by the drawbacks of the current practice of applying predefined deep learning models: the predefined architecture of a deep learning model may not be the optimal design choice when the data generated by users are invisible to model developers. For example, the neural architecture may have some redundant component for a specific dataset, which may lead to unnecessary computing on devices; there may be a better architectural design for the non-IID data distribution. The approaches to personalization discussed in \cref{sec:multimodel} still share the same model architecture among all clients.
The recent progress in NAS \cite{MiLeNAS2020,real2017large,elsken2018efficient,real2019regularized,Bello2016Neural,pham2018efficient,liu2018darts,xie2018snas,elsken2018efficient,luo2018neural} provides a potential way to address these drawbacks. There are three major methods for NAS, which utilize evolutionary algorithms, reinforcement learning, or gradient descent to search for optimal architectures for a specific task on a specific dataset. Among these, the gradient-based method leverages efficient gradient back-propagation with weight sharing, reducing the architecture search process from over 3000 GPU days to only 1 GPU day. 
Another interesting paper recently published, involving Weight Agnostic Neural Networks \cite{gaier2019weight}, claims that neural network architectures alone, without learning any weight parameters, may encode solutions for a given task. If this technique further develops and reaches widespread use, it may be applied to the federated learning without collaborative training among devices.
Although these methods have not been developed for distributed settings such as federated learning, they are all feasible to be transferred to the federated setting. Neural Architecture Search (NAS) for a global or personalized model in the federated learning setting is promising, and early exploration has been made in \citep{he2020fednas}.


\subsubsection{Debugging and Interpretability for FL}
\label{subsec:debugging-and-interpretability-for-fl}
While substantial progress has been made on the federated training of models, this is only part of a complete ML workflow. Experienced modelers often directly inspect subsets of the data for tasks including basic sanity checking, debugging misclassifications, discovering outliers, manually labeling examples, or detecting bias in the training set. Developing privacy-preserving techniques to answer such questions on decentralized data is a major open problem. Recently, \citet{augenstein2019generative} proposed the use of differentially private generative models (including GANs), trained with federated learning, to answer some questions of this type. However, many open questions remain (see discussion in \citep{augenstein2019generative}), in particular the development of algorithms that improve the fidelity of FL DP generative models.

\subsection{Communication and Compression}\label{sec:compr}

It is now well-understood that communication can be a primary bottleneck for federated learning since wireless links and other end-user internet connections typically operate at lower rates than intra- or inter-datacenter links and can be potentially expensive and unreliable. This has led to significant recent interest in reducing the communication bandwidth of federated learning. Methods combining Federated Averaging with sparsification and/or quantization of model updates to a small number of bits have demonstrated significant reductions in communication cost with minimal impact on training accuracy \citep{konevcny2016federated}. However, it remains unclear if communication cost can be further reduced, and whether any of these methods or their combinations can come close to providing optimal trade-offs between communication and accuracy in federated learning. Characterizing such fundamental trade-offs between accuracy and communication has been of recent interest in theoretical statistics \citep{duchi2013,braverman2016, han2018,  acharya2018, barnes2019, tang2019texttt, Barnes2020rtopk}. These works characterize the optimal minimax rates for distributed statistical estimation and learning under communication constraints. However, it is difficult to deduce  concrete insights from these theoretical works for communication bandwidth reduction in practice as they typically ignore the impact of the optimization algorithm. It remains an open direction to leverage such statistical approaches to inform practical training methods.

\paragraph{Compression objectives} Motivated by the limited resources of current devices in terms of compute, memory and communication, there are several different compression objectives of practical value.
\begin{enumerate}[(a)]
    \item \emph{Gradient compression\footnote{In this section, we use ``gradient compression'' to include compression applied to any model update, such as the updates produced by Federated Averaging when clients take multiple gradient steps.}} -- reduce the size of the object communicated from clients to server, which is used to update the global model.
    \item \emph{Model broadcast compression} -- reduce the size of the model broadcast from server to clients, from which the clients start local training.
    \item \emph{Local computation reduction} -- any modification to the overall training algorithm such that the local training procedure is computationally more efficient.
\end{enumerate}
These objectives are in most cases complementary. Among them, (a) has the potential for the most significant practical impact in terms of total runtime. This is both because clients' connections generally have slower upload than download bandwidth\footnote{See for instance \url{https://www.speedtest.net/reports/}} -- and thus there is more to be gained, compared to (b) -- and because the effects of averaging across many clients can enable more aggressive lossy compression schemes. Usually, (c) could be realized jointly with (a) and (b) by specific methods. 

Much of the existing literature applies to the objective (a) \citep{konevcny2016federated, suresh2017distributed, konevcny2018randomized, alistarh2017qsgd, horvath2019natural, basu2020qsparse}. The impact of (b) on convergence in general has not been studied until very recently; an analysis is presented in \citep{chraibi2019distributed}. Very few methods intend to address all of (a), (b) and (c) jointly. \citet{caldas2018expanding} proposed a practical method by constraining the desired model update such that only particular submatrices of model variables are necessary to be available on clients; \citet{hamer2020fedboost} proposed a communication-efficient federated algorithm for learning mixture weights on an ensemble of pre-trained models, based on communicating only a subset of the models to any one device; \citet{FedGKT2020} utilizes bidirectional and alternative knowledge distillation method to transfer knowledge from many compact DNNs to a dense server DNN, which can reduce the local computational burden at the edge devices.

In cross-device FL, algorithms generally cannot assume any state is preserved on the clients (Table~\ref{tab:characteristics}). However, this constraint would typically not be present in the cross-silo FL setting, where the same clients participate repeatedly. Consequently, a wider set of ideas related to error-correction such as \citep{lin2017deep, sattler2019robust, vogels2019powersgd, tang2019texttt, karimireddy2019ef, stich2019error} are relevant in this setting, many of which could address both (a) and (b).

An additional objective is to modify the training procedure such that the \emph{final} model is more compact, or efficient for inference. This topic has received a lot of attention in the broader ML community \citep{han2015deep, courbariaux2015binaryconnect, zhu2017prune, lin2016fixed, oktay2019model, blalock2020state}, but these methods either do not have a straightforward mapping to federated learning, or make the training process more complex which makes it difficult to adopt. Research that simultaneously yields a compact final model, while also addressing the three objectives above, has significant potential for practical impact.

For gradient compression, some existing works \cite{suresh2017distributed} are developed in the minimax sense to characterize the worst case scenario. However usually in information theory, the compression guarantees are instance specific and depend on the \emph{entropy} of the underlying distribution \cite{cover2012elements}. In other words, if the data is easily compressible, they are provably compressed heavily. It would be interesting to see if similar instance specific results can be obtained for gradient compression. Similarly, recent works show that learning a compression scheme in a data-dependent fashion can lead to significantly better compression ratio for the case of data compression \cite{wu2017multiscale} as well as gradient compression. It is therefore worthwhile to evaluate these data-dependent compression schemes in the federated settings~\cite{gandikota2019vqsgd}.

\paragraph{Compatibility with differential privacy and secure aggregation} Many algorithms used in federated learning such as Secure Aggregation \citep{bonawitz2016practical} and mechanisms of adding noise to achieve differential privacy \citep{abadi2016deep,mcmahan18dplm} are not designed to work with compressed or quantized communications. For example, straightforward application of the Secure Aggregation protocol of \citet{bonawitz17secagg,bell20secagg} requires an additional $O(\log M)$ bits of communication for each scalar, where $M$ is the number of clients being summed over, and this may render ineffective the aggressive quantization of updates when $M$ is large (though see~\cite{bonawitz2019autotune} for a more efficient approach). Existing noise addition mechanisms assume adding real-valued Gaussian or Laplacian noise on each client, and this is not compatible with standard quantization methods used to reduce communication. We note that several recent works  allow biased estimators and would work nicely with Laplacian noise \cite{stich2019error}, however those would not give differential privacy, as they break independence between rounds. There is some work on adding discrete noise \cite{agarwal2018cpsgd}, but there is no notion whether such methods are optimal. Joint design of compression methods that are compatible with Secure Aggregation, or for which differential privacy guarantees can be obtained, is thus a valuable open problem.


\paragraph{Wireless-FL co-design}

The existing literature in federated learning usually neglects the impact of wireless channel dynamics during model training, which potentially undermines both training latency and thus reliability of the entire production system. In particular, wireless interference, noisy channels and channel fluctuations can significantly hinder the information exchange between the server and clients (or directly between individual clients, as in the fully decentralized case, see Section~\ref{sec:decentralized}). This represents a major challenge for mission-critical applications, rooted in latency reduction and reliability enhancements. Potential solutions to address this challenge include federated distillation (FD), in which workers exchange their model output parameters (logits) as opposed to the model parameters (gradients and/weights),  and optimizing workers' scheduling policy with appropriate communication and computing resources \citep{FD, EdgeML, FL5G}. Another solution is to leverage the unique characteristics of wireless channels (e.g. broadcast and superposition) as natural data aggregators, in which the simultaneously transmitted  analog-waves  by different workers are  superposed at the server and weighed by the wireless channel coefficients \citep{AbariRK16}. This yields faster model aggregation at the server, and faster training by a factor up to the number of workers. This is in sharp contrast with the traditional orthogonal frequency division  multiplexing (OFDM) paradigm, whereby workers upload their models over orthogonal frequencies whose performance degrades with increasing number of workers \cite{elgabli2020harnessing}.

\subsection{Application To More Types of Machine Learning Problems and Models}\label{sec:more_types_ml}
To date, federated learning has primarily considered supervised learning tasks where labels are naturally available on each client. Extending FL to other ML paradigms, including reinforcement learning, semi-supervised and unsupervised learning, active learning, and online learning \citep{he2019central,zhao2019decentralized} all present interesting and open challenges.

Another important class of models, highly relevant to FL, are those that can characterize the uncertainty in their predictions. Most modern deep learning models cannot represent their uncertainty nor allow for a probability interpretation of parametric learning. This has motivated recent developments of tools and techniques combining Bayesian models with deep learning. From a probability theory perspective, it is unjustifiable to use single point-estimates for classification. Bayesian neural networks \citep{BL-overview} have been proposed and shown to be far more robust to over-fitting, and can easily learn from small datasets. The Bayesian approach further offers uncertainty estimates via its parameters in form of probability distributions, thus preventing over-fitting. Moreover, appealing to probabilistic reasoning, one can predict how the uncertainty can decrease, allowing the decisions made by the network to become more deterministic as the data size grows. 

Since Bayesian methods gave us tools to reason about deep models’ confidence and also achieve state-of-the-art performance on many tasks, one expects Bayesian methods to provide a conceptual improvement to the classical federated learning. In fact, preliminary work from \citet{BayesFL} shows that incorporating Bayesian methods allows for model aggregation across non-IID data and heterogeneous platforms. However, many questions regarding scalability and computational feasibility have to be addressed.

\subsection{Executive summary}
Efficient and effective federated learning algorithms face different challenges compared to centralized training in a datacenter. 

\begin{itemize}

\item Non-IID data due to non-identical client distributions, violation of independence, and dataset drift (\cref{sec:noniid}) pose a key challenge. Though various methods have been surveyed and discussed in this section, defining and dealing with non-IID data remains an open problem and one of the most active research topics in federated learning.

\item Optimization algorithms for federated learning are analyzed in \cref{sec:optimization} under different settings, e.g., convex and nonconvex functions, IID and non-IID data. Theoretical analysis has proven difficult for the parallel local updates commonly used in federated optimization, and often strict assumptions have to be made to constrain the client heterogeneity. Currently, known convergence rates do not fully explain the empirically-observed effectiveness of the Federated Averaging algorithm over methods such as mini-batch SGD~\citep{woodworth2020local}.

\item Client-side personalization and ``multi-model’’ approaches (\cref{sec:multimodel}) can address data heterogeneity and give hope of surpassing the performance of the best fixed global model. Simple personalization methods like fine-tuning can be effective, and offer intrinsic privacy advantages. However, many theoretical and empirical questions remain open: when is a global model better? How many models are necessary? Which federated optimization algorithms combine best with local fine-tuning?

\item Adapting centralized training workflows such as hyper-parameter tuning, neural architecture design, debugging, and interpretability tasks to the federated learning setting (\cref{sec:workflows}) present roadblocks to the widespread adoption of FL in practical settings, and hence constitute important open problems. 

\item While there has been significant work on communication efficiency and compression for FL (\cref{sec:compr}), it remains an important and active area. In particular, fully automating the process of enabling compression without impacting convergence for a wide class of models is an important practical goal. Relatively new directions on the theoretical study of communication, compatibility with privacy methods, and co-design with wireless infrastructure are discussed.

\item There are many open questions in extending federated learning from supervised tasks to other machine learning paradigms including reinforcement learning, semi-supervised and unsupervised learning, active learning, and online learning (\cref{sec:more_types_ml}).
\end{itemize}

\pagebreak
\section{Preserving the Privacy of User Data}
\label{sec:privacy}

\begin{figure}[b!] 
\noindent\usebox{\actorsfigure}
\repeatcaption{fig:actors}{The lifecycle of an FL-trained model and the various actors in a federated learning system.
\label{fig:actors_repeat}
} 
\end{figure}

Machine learning workflows involve many actors functioning in disparate capacities.  For example, users may generate training data through interactions with their devices, a machine learning training procedure extracts cross-population patterns from this data (e.g. in the form of trained model parameters), the machine learning engineer or analyst may assess the quality of this trained model, and eventually the model may be deployed to end users in order to support specific user experiences (see Figure~\ref{fig:actors} below).  

In an ideal world, each actor in the system would learn nothing more than the information needed to play their role.  For example, if an analyst only needs to determine whether a particular quality metric exceeds a desired threshold in order to authorize deploying the model to end users, then in an idealized world, that is the only bit of information that would be available to the analyst; such an analyst would need access to neither the training data nor the model parameters, for instance.  Similarly, end users enjoying the user experiences powered by the trained model might only require predictions from the model and nothing else.

Furthermore, in an ideal world every participant in the system would be able to reason easily and accurately about what personal information about themselves and others might be revealed by their participation in the system, and participants would be able to use this understanding to make informed choices about how and whether to participate at all.

Producing a system with all of the above ideal privacy properties would be a daunting feat on its own, and even more so while also guaranteeing other desirable properties such as ease of use for all participants, the quality and fairness of the end user experiences (and the models that power them), the judicious use of communication and computation resources, resilience against attacks and failures, and so on.

Rather than allowing perfect to be the enemy of good, we advocate a strategy wherein the overall system is composed of modular units which can be studied and improved relatively independently, while also reminding ourselves that we must, in the end, measure the privacy properties of the complete system against our ideal privacy goals set out above. The open questions raised throughout this section will highlight areas wherein we do not yet understand how to simultaneously achieve all of our goals, either for an individual module or for the system as a whole.

Federated learning provides an attractive structure for decomposing the overall machine learning workflow into the approachable modular units we desire.  One of the primary attractions of the federated learning model is that it can provide a level of privacy to participating users through data minimization: the raw user data never leaves the device, and only updates to models (e.g., gradient updates) are sent to the central server. These model updates are more focused on the learning task at hand than is the raw data (i.e. they contain strictly no additional information about the user, and typically significantly less, compared to the raw data), and the individual updates only need to be held ephemerally by the server.  

While these features can offer significant practical privacy improvements over centralizing all the training data, there is still no formal guarantee of privacy in this baseline federated learning model. For instance, it is possible to construct scenarios in which information about the raw data is leaked from a client to the server, such as a scenario where knowing the previous model and the gradient update from a user would allow one to infer a training example held by that user. Therefore, this section surveys existing results and outlines open challenges towards designing federated learning systems that can offer rigorous privacy guarantees. We focus on questions specific to the federated learning and analytics setting and leave aside questions that also arise in more general machine learning settings as surveyed in \cite{pdlSurvey}.

Beyond attacks targeting user privacy, there are also other classes of attacks on federated learning; for example, an adversary might attempt to prevent a model from being learned at all, or they might attempt to bias the model to produce inferences that are preferable to the adversary.  We defer consideration of these types of attacks to Section~\ref{sec:robust}.

The remainder of this section is organized as follows. Section \ref{ssec:actors_threat_models} discusses various threat models against which we wish to give protections. Section \ref{ssec:tools_tech} lays out a set of core tools and technologies that can be used towards providing rigorous protections against the threat models discussed in Section \ref{ssec:actors_threat_models}. Section \ref{ssec:adv_clients_analysts} assumes the existence of a trusted server and discusses the open problems and challenges in providing protections against adversarial clients and/or analysts. Section \ref{ssec:adv_server} discusses the open problems and challenges in the absence of a fully trusted server. Finally, Section \ref{ssec:user_perception} discusses open questions around user perception. 

\subsection{Actors, Threat Models, and Privacy in Depth}
\label{ssec:actors_threat_models}

A formal treatment of privacy risks in FL calls for a holistic and interdisciplinary approach. While some of the risks can be mapped to technical privacy definitions and mitigated with existing technologies, others are more complex and require cross-disciplinary efforts. 

Privacy is not a binary quantity, or even a scalar one. This first step towards such formal treatment is a careful characterization of the different actors (see Figure \ref{fig:actors} from \cref{sec:intro}, repeated on page~\pageref{fig:actors_repeat} for convenience) and their roles to ultimately define relevant threat models (see Table \ref{table:actors_and_threats}). Thus, for instance, it is desirable to distinguish the view of the server administrator from the view of the analysts that consume the learned models, as it is conceivable that a system that is designed to offer strong privacy guarantees against a malicious analyst may not provide any guarantees with respect to a malicious server. These actors map well onto the threat models discussed elsewhere in the literature; for example, in \citet[Sec 3.1]{prochlo}, where the ``encoder'' corresponds to the client, the ``shuffler'' generally corresponds to the server, the ``analyzer`` may correspond to the server or post-processing done by the analyst.

As an example, a particular system might offer a differential privacy\footnote{Differential privacy will be formally introduced in Section \ref{sssec:private_disclosures}.  For now, it suffices to know that lower $\varepsilon$ corresponds with higher privacy.} guarantee with a particular parameter $\varepsilon$ to the view of the server administrator, while the results observed by analysts might have a higher protection $\varepsilon' < \varepsilon$. 

Furthermore, it is possible that this guarantee holds only against adversaries with particular limits on their capabilities, e.g. an adversary that can observe everything that happens on the server (but cannot influence the server's behavior) while simultaneously controlling up to a fraction $\gamma$ of the clients (observing everything they see and influencing their behavior in arbitrary ways); the adversary might also be assumed to be unable to break cryptographic mechanisms instantiated at a particular security level $\sigma$.  Against an adversary whose strength \textit{exceeds} these limits, the view of the server administrator might still have some differential privacy, but at weaker level $\varepsilon_0 > \varepsilon$.  

As we see in this example, precisely specifying the assumptions and privacy goals of a system can easily implicate concrete instantiations of several parameters ($\varepsilon, \varepsilon', \varepsilon_0, \gamma, \sigma$, etc.) as well as concepts such as differential privacy and honest-but-curious security.

\begin{table}
\renewcommand{\arraystretch}{1.2}
\begin{center} 
\begin{tabular}{@{}p{1.2in} p{1.85in}  p{3in}@{}}
\toprule
\textbf{Data/Access Point} & \textbf{Actor} & \textbf{Threat Model} \\
\midrule
\addlinespace[0.05in]
Clients & 
Someone who has root access to the client device, either by design or by compromising the device &
Malicious clients can inspect all messages received from the server (including the model iterates) in the rounds they participate in and can tamper with the training process.  An honest-but-curious client can inspect all messages received from the server but cannot tamper with the training process.   In some cases, technologies such as secure enclaves/TEEs may be able to limit the influence and visibility of such an attacker, representing a meaningfully weaker threat model.
\\ 
\addlinespace[0.1in]
Server & 
Someone who has root access to the server, either by design or by compromising the device &
A malicious server can inspect all messages sent to the server (including the gradient updates) in all rounds and can tamper with the training process. An honest-but-curious server can inspect all messages sent to the server but cannot tamper with the training process.  In some cases, technologies such as secure enclaves/TEEs may be able to limit the influence and visibility of such an attacker, representing a meaningfully weaker threat model.
\\ 
\addlinespace[0.1in]
Output Models &
Engineers \& analysts & 
A malicious analyst or model engineer may have access to multiple outputs from the system, e.g. sequences of model iterates from multiple training runs with different hyperparameters. Exactly what information is released to this actor is an important system design question. 
\\
\addlinespace[0.1in]
Deployed Models &
The rest of the world & 
In cross-device FL, the final model may be deployed to hundreds of millions of devices. A partially compromised device can have black-box access to the learned model, and a fully compromised device can have a white-box access to the learned model.
\\ 
\addlinespace[0.05in]
\bottomrule
\end{tabular} 
\end{center}
\caption{Various threat models for different adversarial actors.} 
\label{table:actors_and_threats}
 \end{table}


Achieving all the desired privacy properties for federated learning will typically require composing many of the tools and technologies described below into an end-to-end system, potentially both layering multiple strategies to protect the same part of the system (e.g. running portions of a Secure Multi-Party Computation (MPC) protocol inside a Trusted Execution Environment (TEE) to make it harder for an adversary to sufficiently compromise that component) as well as using different strategies to protect different parts of the system (e.g. using MPC to protect the aggregation of model updates, then using Private Disclosure techniques before sharing the aggregate updates beyond the server).

As such, we advocate for building federated systems wherein the privacy properties degrade as gracefully as possible in cases where one technique or another fails to provide its intended privacy contribution.  For example, running the server component of an MPC protocol inside a TEE might allow privacy to be maintained even in the case where either (but not both) of the TEE security or MPC security assumptions fails to hold in practice.  As another example, requiring clients to send raw training examples to a server-side TEE would be strongly dispreferred to having clients send gradient updates to a server-side TEE, as the latter's privacy expectations degrade much more gracefully if the TEE's security were to fail.  We refer to this principle of graceful degradation as ``Privacy in Depth,’’ in analogy to the well-established network security principle of defense in depth \cite{nsa2012defense}.

\subsection{Tools and Technologies}
\label{ssec:tools_tech}
Generally speaking, the goal of an FL computation is for the analyst or engineer requesting the computation to obtain the result, which can be thought of as the evaluation of a function $f$ on a distributed client dataset (commonly an ML model training algorithm, but possibly something simpler such as a basic statistic). There are three privacy aspects that need to be addressed.

First, we need to consider {\em how} $f$ is computed and what is the information flow of intermediate results in the process, which primarily influences the susceptibility to malicious client, server, and admin actors.  In addition to designing the flow of information in the system (e.g. early data minimization), techniques from secure computation including Secure Multi-Party Computation (MPC) and Trusted Execution Environments (TEEs) are of particular relevance to addressing these concerns.  These technologies will be discussed in detail in \cref{sssec:secure_computations}.

Second, we have to consider {\em what} is computed. In other words, how much information about a participating client is revealed to the analyst and world actors by the result of $f$ itself. Here, techniques for privacy-preserving disclosure, particularly differential privacy (DP), are highly relevant and will be discussed in detail in \cref{sssec:private_disclosures}.

Finally, there is the problem of {\em verifiability}, which pertains to the ability of a client or the server to prove to others in the system that they have executed the desired behavior faithfully, without revealing the potentially private data upon which they were acting.  Techniques for verifiability, including remote attestation and zero-knowledge proofs, will be discussed in \cref{sssec:verifiability}.

\begin{table}
\renewcommand{\arraystretch}{1.2}
\begin{center} 
\begin{tabular}{@{}p{2in} p{4in}@{}}
 \toprule 
\textbf{Technology} & \textbf{Characteristics} \\
\midrule
\addlinespace[0.05in]
Differential Privacy (local, central, shuffled, aggregated, and hybrid models) & A quantification of how much information could be learned about an individual from the output of an analysis on a dataset that includes the user. Algorithms with differential privacy necessarily incorporate some amount of randomness or noise, which can be tuned to mask the influence of the user on the output. 
\\
\addlinespace[0.1in]
Secure Multi-Party Computation & Two or more participants collaborate to simulate, though cryptography, a fully trusted third party who can:
\begin{itemize} 
\item  Compute a function of inputs provided by all the participants;
\item Reveal the computed value to a chosen subset of the participants, with no party learning anything further.
\end{itemize}
\\
\addlinespace[0.1in]
Homomorphic Encryption & Enables a party to compute functions of data to which they do not have plain-text access, by allowing mathematical operations to be performed on ciphertexts without decrypting them.  Arbitrarily complicated functions of the data can be computed this way (``Fully Homomorphic Encryption’’) though at greater computational cost. 
\\
\addlinespace[0.1in]
Trusted Execution Environments (secure enclaves) & TEEs provide the ability to trustably run code on a remote machine, even if you do not trust the machine's owner/administrator.  This is achieved by limiting the capabilities of any party, including the administrator.  In particular, TEEs may provide the following properties \cite{subramanyan2017formal}:
\begin{itemize}
\item Confidentiality: The state of the code's execution remains secret, unless the code explicitly publishes a message;
\item Integrity: The code's execution cannot be affected, except by the code explicitly receiving an input;
\item Measurement/Attestation: The TEE can prove to a remote party what code (binary) is executing and what its starting state was, defining the initial conditions for confidentiality and integrity.
\end{itemize}
\\
\addlinespace[0.05in] 
\bottomrule
\end{tabular} 
\end{center}
\caption{Various technologies along with their characteristics.} 
\label{table:technologies}
\end{table}

\subsubsection{Secure Computations}
\label{sssec:secure_computations}

The goal of secure computation is to evaluate functions on distributed inputs in a way that only reveals the result of the computation to the intended parties, without revealing any additional information (e.g. the parties' inputs or any intermediate results). 

\paragraph{Secure multi-party computation}

Secure Multi-Party Computation (MPC) is a subfield of cryptography concerned with the problem of having a set of parties compute an agreed-upon function of their private inputs in a way that only reveals the intended output to each of the parties. This area was kicked off in the 1980’s by Yao~\cite{DBLP:conf/focs/Yao86}.  Thanks to both theoretical and engineering breakthroughs, the field has moved from being of a purely theoretical interest to a deployed technology in industry~\cite{DBLP:conf/fc/BogetoftCDGJKNNNPST09,DBLP:conf/fc/BogdanovTW12,DBLP:conf/secdev/LapetsVBJV16,araki2016high,DBLP:conf/eurocrypt/FurukawaLNW17,DBLP:journals/iacr/IonKNPSSSY17,DBLP:journals/iacr/IonKNPRSSSY19}. It is important to remark that MPC defines a set of technologies, and should be regarded more as a field, or a general notion of security in secure computation, than a technology \textit{per se}. Some of the recent advances in MPC can be attributed to breakthroughs in lower level primitives, such as oblivious transfer protocols~\cite{IKNP} and encryption schemes with homomorphic properties (as described below). 

A common aspect of cryptographic solutions is that operations are often done on a finite field (e.g. integers modulo a prime $p$), which poses difficulties when representing real numbers. A common approach has been to adapt ML models and their training procedures to ensure that (over)underflows are controlled, by operating on normalized quantities and relying on careful quantization~\cite{GasconSB0DZE17, quotient, gbdllnwCN, DBLP:conf/crypto/BourseMMP18}.

It has been known for several decades that any function can be securely computed, even in the presence of malicious adversaries~\cite{goldreich87}.
While generic solutions exist, their performance characteristics often render them inapplicable in practical settings.  As such a noticeable trend in research has consisted in designing custom protocols for applications such as linear and logistic regression~\cite{DBLP:conf/sp/NikolaenkoWIJBT13,GasconSB0DZE17, secureml} and neural network training and inference~\cite{secureml, quotient, bedkSE}. These works are typically in the cross-silo setting, or the variant where computation is delegated to a small group of computing servers that do not collude with each other. Porting these protocols to the cross-device setting is not straightforward, as they require a significant amount of communication.

\subparagraph{Homomorphic encryption}
Homomorphic encryption (HE) schemes allow certain mathematical operations to be performed directly on ciphertexts, without prior decryption.  Homomorphic encryption can be a powerful tool for enabling MPC by enabling a participant to compute functions on values while keeping the values hidden.

Different flavours of HE exist, ranging from general fully homomorphic encryption (FHE)~\cite{gentry2009fully} to the more efficient leveled variants~\cite{BFV1, BFV2, BGV, DBLP:conf/pkc/CoronLT14}, for which several implementations exist~\cite{HElib, SEAL, Palisade, SHELL, lattigo}. Also of practical relevance are the so-called partially homomorphic schemes, including for example ElGamal and Paillier, allowing either homomorphic addition or multiplication. Additive HE has been used as an ingredient in MPC protocols in the cross-silo setting~\cite{DBLP:conf/sp/NikolaenkoWIJBT13, Hardy2017-da}.
A review of some homomorphic encryption software libraries along with brief explanations of criteria/features to be considered in choosing a library is surveyed in \cite{sathya2018review}.\par
When considering the use of HE in the FL setting, questions immediately arise about
who holds the secret key of the scheme. While the idea of every client encrypting 
their data and sending it to the server to compute homomorphically 
on it is appealing, the server should not be able to decrypt a single client contribution.
A trivial way of overcoming this issue would be relying on a non-colluding external
party that holds the secret key and decrypts the result of the computation.
However, most HE schemes require that the secret keys be renewed often (due to e.g. susceptibility to chosen ciphertext attacks~\cite{DBLP:conf/latincrypt/ChenalT14}). Moreover, the availability of a trusted non-colluding 
party is not standard in the FL setting. 

Another way around this issue is relying on distributed (or threshold) encryption schemes, where the secret key is distributed among the parties. \citet{hatelove} and \citet{DBLP:conf/sosp/RothNFH19} propose such solutions for computing summation in the cross-device setting. Their protocols make use of additively homomorphic schemes (variants of ElGamal and lattice-based schemes, respectively).

\paragraph{Trusted execution environments}
Trusted execution environments (TEEs, also referred to as secure enclaves) may provide opportunities to move part of the federated learning process into a trusted environment in the cloud, whose code can be attested and verified.

TEEs can provide several crucial facilities for establishing trust that a unit of code has been executed faithfully and privately~\cite{subramanyan2017formal}:
\begin{itemize}
\item Confidentiality: The state of the code's execution remains secret, unless the code explicitly publishes a message.
\item Integrity: The code's execution cannot be affected, except by the code explicitly receiving an input.
\item Measurement/Attestation: The TEE can prove to a remote party what code (binary) is executing and what its starting state was, defining the initial conditions for confidentiality and integrity.
\end{itemize}

TEEs have been instantiated in many forms, including Intel's SGX-enabled CPUs~\cite{intel2012architecture,costan2016intel}, Arm's TrustZone~\cite{ArmTrustzone,AndroidTrusty}, and Sanctum on RISC-V~\cite{costan2016sanctum}, each varying in its ability to systematically offer the above facilities.

Current secure enclaves are limited in terms of memory and provide access only to CPU resources, that is they do not allow processing on GPUs or machine learning processors (Tram\`er and Boneh~\cite{tramer2018slalom} explore how to combine TEEs with GPUs for machine learning inference).  Moreover, it is challenging for TEEs (especially those operating on shared microprocessors) to fully exclude all types of side channel attacks~\cite{van2018foreshadow}.

While secure enclaves provide protections for all code running inside them, there are additional concerns that must be addressed in practice.  For example, it is often necessary to structure the code running in the enclave as a data oblivious procedure, such that its runtime and memory access patterns do not reveal information about the data upon which it is computing (see for example~\cite{prochlo}).  Furthermore, measurement/attestation typically only proves that a particular binary is running; it is up to the system architect to provide a means for proving that that binary has the desired privacy properties, potentially requiring the binary to be built using a reproducible process from open source code.

It remains an open question how to partition federated learning functions across secure enclaves, cloud computing resources, and client devices. For example, secure enclaves could execute key functions such as secure aggregation or shuffling to limit the server's access to raw client contributions while keeping most of the federated learning logic outside this trusted computing base. 

\paragraph{Secure computation problems of interest}

While secure multi-party computation and trusted execution environments offer general solutions to the problem of privately computing any function on distributed private data, many optimizations are possible when focusing on specific functionalities. This is the case for the tasks described next.

\subparagraph{Secure aggregation}
Secure aggregation is a functionality for $n$ clients and a server.  It enables each client to submit a value (often a vector or tensor in the FL setting), such that the server learns just an aggregate function of the clients' values, typically the sum.  

There is a rich literature exploring secure aggregation in both the single-server setting (via additive masking~\cite{Acs:2011:IDD:2042445.2042457, DBLP:journals/tdsc/GoryczkaX17, bonawitz17secagg, bell20secagg,so2020turbo}, via threshold homomorphic encryption~\cite{shi2011privacy, halevi2011secure, chan2012privacy}, and via generic secure multi-party computation~\cite{burkhart2010sepia}) as well as in the multiple non-colluding servers setting~\cite{DBLP:conf/fc/BogetoftCDGJKNNNPST09, araki2016high, corrigan2017prio}.
Secure aggregation can also be approached using trusted execution environments (introduced above), as in~\cite{lie2017glimmers}.

\subparagraph{Secure shuffling}
Secure shuffling is a functionality for $n$ clients and a server.   It enables each client to submit one or more messages, such that the server learns just an unordered collection (multiset) of the messages from all clients and nothing more.  Specifically, the server has no ability to link any message to its sender beyond the information contained in the message itself.  
Secure shuffling can be considered an instance of Secure Aggregation where the values are multiset-singletons and the aggregation operation is multiset-sum, though it is often the case that very different implementations provide the best performance in the typical operating regimes for secure shuffling and secure aggregation.

Secure shufflers have been studied in the context of secure multi-party computation~\cite{chaum1981untraceable, kwon2016riffle}, often under the heading of mix networks.   They have also been studied in the context of trusted computing~\cite{prochlo}.  Mix networks have found large scale deployment in the form of the Tor network~\cite{dingledine2004tor}.  

\subparagraph{Private information retrieval}
Private information retrieval (PIR) is a functionality for one client and one server.  It enables the client to download an entry from a server-hosted database such that the server gains zero information about which entry the client requested.

MPC approaches to PIR break down into two main categories: \textit{computational PIR} (cPIR), in which a single party can execute the entire server side of the protocol~\cite{Kushilevitz97replicationis}, and \textit{information theoretic PIR} (itPIR), in which multiple non-colluding parties are required to execute the server side of the protocol~\cite{Chor98PIR}. 

The main roadblocks to the applicability of PIR have been the following: cPIR has high computational cost~\cite{sion2007computational}, while the non-colluding parties setting has been difficult to achieve convincingly in industrial scenarios. Recent results on PIR have shown dramatic reductions in the computational cost through the use of lattice-based cryptosystems~\cite{aguilar2007lattice,olumofin2011revisiting,aguilar2016xpir,DBLP:conf/sp/AngelCLS18,DBLP:conf/tcc/GentryH19}. The computational cost can be traded for more communication; we refer the reader to~\citet{DBLP:journals/iacr/AliLP0SSY19} to better understand the communication and computation trade-offs offered by cPIR. Additionally, it has been shown how to construct communication-efficient PIR on a single-server by leveraging side information available to the user~\cite{pirsideinfo}, for example via client local state. \citet{Patel18googlePIR} presented and implemented a practical hybrid (computational and information theoretic) PIR scheme on a single server assuming client state. \citet{DBLP:journals/iacr/Corrigan-GibbsK19a} present theoretical constructions for PIR with sublinear \emph{online} time by working in an offline/online model where, during an offline phase, clients fetch information from the server(s) independent on the future query to be performed.

Further work has explored the connection between PIR and secret sharing \cite {yekhaninpir}, with recent connections to PIR on coded data \cite{dolift} and communication efficient PIR \cite{staircasepir}. A variant of PIR, called PIR-with-Default, enable clients to retrieve a default value if the index queried is not in the database, and can output additive secret shares of items which can serve as input to any MPC protocol~\cite{DBLP:journals/iacr/LepointPRST20}. PIR has also been studied in the context of ON-OFF privacy, in which a client is permitted to switch off their privacy guards in exchange for better utility or performance~\cite{onoffisit,onoffitw}. 

\subsubsection{Privacy-Preserving Disclosures}
\label{sssec:private_disclosures}

The state-of-the-art model for quantifying and limiting information disclosure about individuals is \textit{differential privacy} (DP)~\cite{DMNS06, dwork2008differential, dwork2014algorithmic}, which aims to introduce a level of uncertainty into the released model sufficient to mask the contribution of any individual user. Differential privacy is quantified by privacy loss parameters $(\varepsilon, \delta)$, where smaller $(\varepsilon, \delta)$ corresponds to increased privacy. More formally,  a randomized algorithm $\mathcal{A}$ is $(\varepsilon, \delta)$-differentially private if for all $\mathcal{S} \subseteq \text{Range}(\mathcal{A})$, and for all adjacent datasets $D$ and $D'$:
\begin{equation}
\label{eq-dp}
P(\mathcal{A}(D) \in \mathcal{S}) \le e^\varepsilon P(\mathcal{A}(D') \in \mathcal{S}) + \delta.
\end{equation}
In the context of FL, $D$ and $D'$ correspond to decentralized datasets that are adjacent if $D'$ can be obtained from $D$ by adding or subtracting all the records of a single client (user) \citep{mcmahan18dplm}. This notion of differential privacy is referred to as user-level differential privacy. It is stronger than the typically used notion of adjacency where $D$ and $D'$ differ by only one record \citep{dwork2014algorithmic}, since in general one user may contribute many records (e.g. training examples) to the dataset.

Over the last decade, an extensive set of techniques has been developed for differentially private data analysis, particularly under the assumption of a centralized setting, where the raw data is collected by a trusted party prior to applying perturbations necessary to achieve privacy. In federated learning, typically the orchestrating server would serve as the trusted implementer of the DP mechanism, ensuring only privatized outputs are released to the model engineer or analyst.

However, when possible we often wish to reduce the need for a trusted party. Several approaches for reducing the need for trust in a data curator have been considered in recent years.

\paragraph{Local differential privacy} Differential privacy can be achieved without requiring trust in a centralized server by having each client apply a differentially private transformation to their data prior to sharing it with the server. That is, we apply \cref{eq-dp} to a mechanism $\mathcal{A}$ that processes a single user's local dataset $D$, with the guarantee holding with respect to \emph{any} possible other local dataset $D'$.
This model is referred to as the \textit{local model of differential privacy} (LDP)~\cite{Warner65, KLNRS11}. LDP has been deployed effectively to gather statistics on popular items across large userbases by Google, Apple and Microsoft~\cite{rappor:15, applewhitepaper:17,  collecting-telemetry-data-privately}. It has also been used in federated settings for spam classifier training by Snap~\cite{snap}. These LDP deployments all involve large numbers of clients and reports, even up to a billion in the case of Snap, which stands in stark contrast to centralized instantiations of DP which can provide high utility from much smaller datasets. Unfortunately, as we will discuss in Section \ref{sssec:limitations}, achieving LDP while maintaining utility can be difficult~\cite{KLNRS11,Ullman18}. Thus, there is a need for a model of differential privacy that interpolates between purely central and purely local DP. This can be achieved through distributed differential privacy, or the hybrid model, as discussed below.

\paragraph{Distributed differential privacy} In order to recover some of the utility of central DP without having to rely on a trustworthy central server, one can instead use a \emph{distributed differential privacy model}~\cite{dwork2006our, shi2011privacy,prochlo,cheu2019distributed}. Under this model, the clients first compute and encode a minimal (application specific) focused report, and then send the encoded reports to a secure computation function, whose output is available to the central server, with the intention that this output already satisfies differential privacy requirements by the time the server is able to inspect it.  The encoding is done to help maintain privacy on the clients, and could for example include LDP. The secure computation function can have a variety of incarnations. It could be an MPC protocol, a standard computation done on a TEE, or even a combination of the two.  Each of these choices comes with different assumptions and threat models.

It is important to remark that distributed differential privacy and local differential privacy yield different guarantees from several perspectives: while the distributed DP framework can produce more accurate statistics for the same level of differential privacy as LDP, it relies on different setups and typically makes stronger assumptions, such as access to MPC protocols. Below, we outline two possible approaches to distributed differential privacy, relying on secure aggregation and secure shuffling. We stress that there are many other methods that could be used, see for instance \cite{sabater2020} for an approach based on exchanging correlated Gaussian noise across secure channels.

\subparagraph{Distributed DP via secure aggregation} One promising tool for achieving distributed DP in FL is secure aggregation, discussed above in Section \ref{sssec:secure_computations}. Secure aggregation can be used to ensure that the central server obtains the aggregated result, while guaranteeing that intermediate parameters of individual devices and participants are not revealed to the central server. To further ensure the aggregated result does not reveal additional information to the server, we can use local differential privacy (e.g. with moderate $\varepsilon$ level). For example, each device could perturb its own model parameter before the secure aggregation in order to achieve local differential privacy. By designing the noise correctly, we may ensure that the noise in the aggregated result matches the noise that would have otherwise been added centrally by a trusted server (e.g. with a low $\varepsilon$ / high privacy level)~\cite{Acs:2011:IDD:2042445.2042457, Rastogi:2010:DPA:1807167.1807247, Ghosh:2009:UUP:1536414.1536464, shi2011privacy, DBLP:journals/tdsc/GoryczkaX17}. 

\begin{figure}[tb]
\centering
\includegraphics[width=0.7\textwidth]{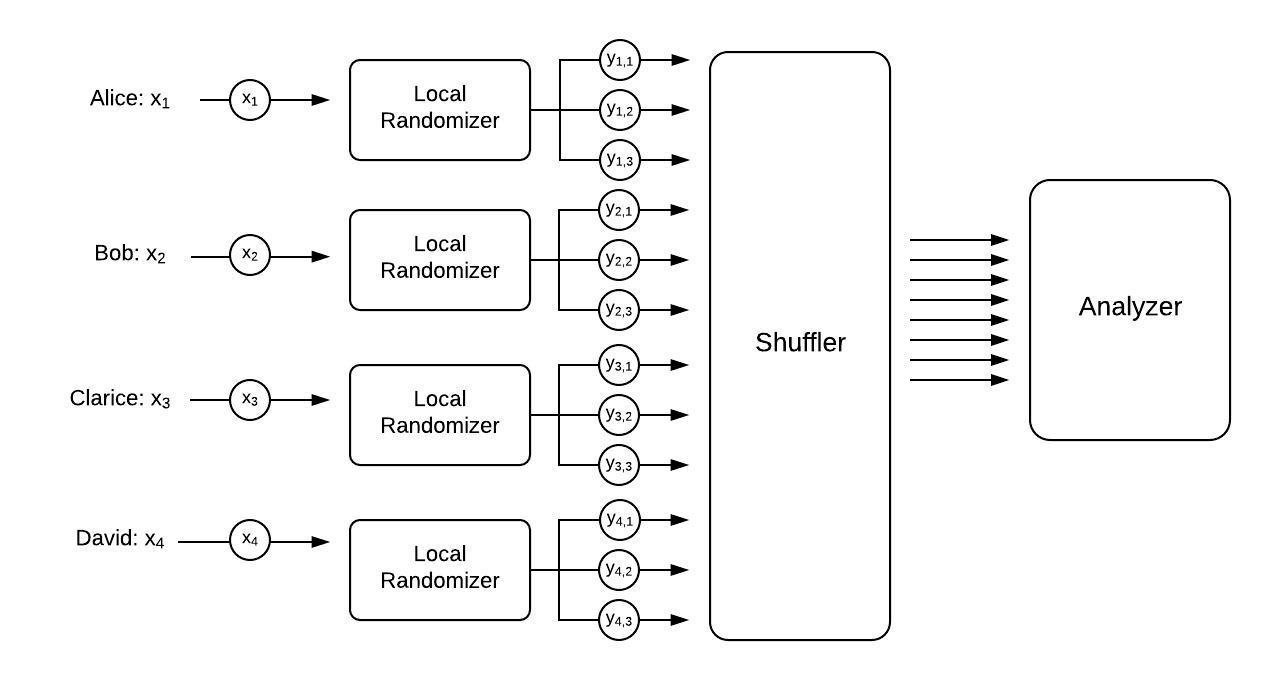}
\caption{The Encode-Shuffle-Analyze (ESA) framework, illustrated here for $4$ players.}
\label{fig:esa}
\end{figure}

\subparagraph{Distributed DP via secure shuffling} Another distributed differential privacy model is the shuffling model, which was kicked off by the recently introduced Encode-Shuffle-Analyze (ESA) framework~\cite{prochlo} (illustrated in Figure~\ref{fig:esa}). In the simplest version of this framework, each client runs an LDP protocol (e.g. with a moderate $\varepsilon$ level) on its data and provides its output to a secure shuffler. The shuffler randomly permutes the reports and sends the collection of shuffled reports (without any identifying information) to the server for final analysis. Intuitively, the interposition of this secure compute function makes it harder for the server to learn anything about the participants and supports a differential privacy analysis (e.g. with a low $\varepsilon$ / high privacy level). In the more general multi-message shuffled framework, each user can possibly send more than one message to the shuffler. The shuffler can either be implemented directly as a trusted entity, independent of the server and devoted solely to shuffling, or via more complex cryptographic primitives as discussed above.

\citet{prochlo} proposed the Prochlo system as a way to implement the ESA framework. The system takes a holistic approach to privacy that takes into account secure computation aspects (addressed using TEEs), private disclosure aspects (addressed by means of differential privacy), and verifiability aspects (mitigated using secure enclave attestation capabilities). 

More generally, shuffling models of differential privacy can use broader classes of local randomizers, and can even select these local randomizers adaptively~\cite{erlingsson2019amplification}. This can enable differentially private protocols with far smaller error than what is possible in the local model, while relying on weaker trust assumptions than in the central model, e.g., ~\cite{cheu2019distributed, erlingsson2019amplification,BalleBGN19, ghazi2019scalable_1, ghazi2019scalable, ghazi2019private, pure-dp-shuffled, counting_shuffle_ICML, chen2020distributed}. 


\paragraph{Hybrid differential privacy}  Another promising approach is hybrid differential privacy~\cite{avent2017blender}, which combines multiple trust models by partitioning users based on their trust model preference (e.g. trust or lack of trust in the curator). Prior to the hybrid model, there were two natural choices. The first was to use the least-trusting model, which typically provides the lowest utility, and conservatively apply it uniformly over the entire userbase. The second was to use the most-trusting model, which typically provides the highest utility, but only apply it over the most-trusting users. By allowing multiple models to coexist, hybrid model mechanisms can achieve more utility from a given userbase, compared to purely local or central DP mechanisms. For instance, \cite{avent2017blender} describes a system in which most users contribute their data in the local model of privacy, and a small fraction of users opt-in to contributing their data in the central DP model. This enables the design of a mechanism which, in some circumstances, outperforms both the conservative local DP mechanism applied across all users as well as the central DP mechanism applied only across the small fraction of opt-in users. Recent work by~\cite{beimel2019power} further demonstrates that a combination of multiple trust models can become part of a promising toolkit for designing and implementing differential privacy. This construction can be directly applied in the federated learning setting; however, the general concept of combining trust models or computational models may also inspire similar but new approaches for federated learning.

\subsubsection{Verifiability}
\label{sssec:verifiability}


An important notion that is orthogonal to the above privacy techniques is that of verifiability. Generally speaking, verifiable computation will enable one party to prove to another party that it has executed the desired behavior on its data faithfully, without compromising the potential secrecy of the data. The concept of verifiable computation dates back to \citet{DBLP:conf/stoc/BabaiFLS91} and has been studied under various terms in the literature: checking computations~\cite{DBLP:conf/stoc/BabaiFLS91}, certified computation~\cite{DBLP:journals/siamcomp/Micali00}, delegating computations~\cite{DBLP:conf/stoc/GoldwasserKR08}, as well as verifiable computing~\cite{DBLP:conf/crypto/GennaroGP10}.

In the context of FL, verifiability can be used for two purposes. First, it would enable the server to prove to the clients that it executed the intended behavior (e.g., aggregating inputs, shuffling of the input messages, or adding noise for differential privacy) faithfully. Second, it would enable the clients to prove to the server that their inputs and behavior follow that of the protocol specification (e.g., the input belongs to a certain range, or the data is a correctly generated ciphertext).

Multiple techniques can be useful to provide verifiability: zero-knowledge proofs (ZKPs), trusted execution environments (TEEs), or remote attestation. Among these ZKPs provide formal cryptographic security guarantees based on mathematical hardness, while others make rely on assumption about the security of trusted hardware.

\paragraph{Zero-knowledge proofs (ZKPs)} Zero knowledge (ZK) proofs are a cryptographic primitive that enables one party (called the
\emph{prover}) to prove statements to another party (called the \emph{verifier}), that depend on secret
information known to the prover, called witness, without revealing those secrets to the verifier. The notion of zero-knowledge was introduced in the late 1980's by \citet{DBLP:journals/siamcomp/GoldwasserMR89}. It provides a solution for the verifiability question on private data. While 
there had been a large body of work on ZK construction, the first work that brought ZKPs and verifiable computation for general functionalities in the realm of practicality was the work of ~\citet{DBLP:journals/cacm/ParnoHG016} which introduces the first optimized construction and implementation for succinct ZK. Nowadays, ZKP protocols can achieve proof sizes of hundred of bytes and verifications of the order of milliseconds regardless of the size of the statement being proved.

A ZKP has three salient properties: \emph{completeness} (if the statement is true and the prover and verifier follow the protocol, the verifier will accept the proof), \emph{soundness} (if the statement is false and the verifier follows the protocol, the verifier will refuse the proof), and \emph{zero-knowledge} (if the statement is true and the prover follows the protocol, the verifier will only learn that the statement is true and will not learn any confidential information from the interaction).

Beyond these common properties, there are different types of zero-knowledge constructions in terms of supported language for the proofs, setup requirements, prover and verifier computational efficiency, interactivity, succinctness, and underlying hardness assumptions. There are many ZK constructions that support specific classes of statements, Schnorr proofs~\cite{Schnorr:1990:EIS:111563.111628} and Sigma protocols~\cite{DamgaardSigma} are examples of such widely used protocols. While such protocols have numerous uses in specific settings, general ZK systems that can support any functionality provide a much more broadly applicable tool (including in the context of FL), and thus we focus on such constructions for the rest of the discussion.

A major distinguishing feature between different constructions is the need for \emph{trusted} setup. Some ZKPs rely on a common reference string (CRS), which is computed using secrets that should remain hidden in order to guarantee the soundness properties of the proofs. The computation of such a CRS is referred to as a trusted setup. While this requirement is a disadvantage for such systems, the existing ZKP constructions that achieve most succinct proofs and verifier's efficiency require trusted setup. 

Another significant property that affects the applicability in different scenarios is whether generating the proof requires interaction between the prover and the verifier, and here we distinguish non-interactive zero-knowledge proofs (NIZKs) that enable the prover to send a single message to the verifier and require no further communication. Often we can convert interactive to non-interactive proofs by making stronger assumptions about ideal functionality of hash functions (i.e., that hash functions behave as random oracles).

Additionally, there are different measurements for efficiency of a ZKP system one must be aware of, such as the length of the proof and the computation complexity of the prover and verifier. The ideal prover's complexity should be linear in the execution time for the evaluated functionality but many existing ZKPs introduce additional (sometimes significant) overhead for the prover. The most efficient verification complexity requires computation at least linear in the size of the inputs for the evaluated functionality, and in the setting of proofs for the work of the FL server this input size will be significant.

Succinct
non-interactive zero-knowledge proofs (SNARKs)~\cite{Bitansky:2012:ECR:2090236.2090263} are a type of ZKP that provides constant proof size and verification that depends only on the input size, linearly. These attractive efficiency properties do come at the price of stronger assumptions, which is mostly inherent, 
and trusted setup in all existing scheme. Most existing SNARK constructions leverage quadratic arithmetic programs~\cite{DBLP:conf/eurocrypt/GennaroGP013,DBLP:journals/cacm/ParnoHG016,DBLP:conf/sp/CostelloFHKKNPZ15} and are now available in open-source libraries, such as libsnark~\cite{libsnark}, and deployed in cryptocurrencies, such as Zcash~\cite{DBLP:conf/sp/Ben-SassonCG0MTV14}. Note that SNARK systems usually require overhead on the part of the prover; in particular, the prover computation needs to be superlinear in the size of the circuit for the statement being proven. Recently, \citet{DBLP:conf/crypto/XieZZPS19} presented Libra, a ZKP system that achieves linear prover complexity but with increased proof size and verification time. 

If we relax the requirements for succinctness or non-interactiveness for the construction, there is a large body of constructions that achieve a wide range of efficiency trade-offs, avoid the trusted setup requirement and use more standard cryptographic assumptions~\cite{DBLP:conf/sp/BunzBBPWM18,DBLP:conf/sp/WahbyTSTW18,Ames:2017:LLS:3133956.3134104,DBLP:conf/crypto/Ben-SassonBHR19}.


In the recent years, an increasing numbers of practical applications have been using non-interactive zero-knowledge proofs, primarily motivated by blockchains. Using interactive ZKP systems and NIZKs efficiently in the context of FL remains a challenging open question. In such a setting, NIZKs may enable to prove to the server properties about the client's inputs. In the setting where the verifier is the client, it will be challenging to create a trustworthy statement to verify as it involves input from other clients. Of interest in this setting, recent work enables to handle the case where the multiple verifiers have shares of the statement~\cite{DBLP:conf/crypto/BonehBCGI19}.

\paragraph{Trusted execution environment and remote attestation} We discussed TEEs in \cref{sssec:secure_computations}, but focus here on the fact that TEEs may provide opportunities to provide verifiable computations.  Indeed, TEEs enable to attest and verify the code (binary) running in its environment.  In particular, when the verifier knows (or can reproduce) which binary should run in the secure enclaves, TEEs will be able to provide a notion of \emph{integrity} (the code execution cannot be affected, except by the inputs), and an \emph{attestation} (the TEE can prove that a specific binary is executing and what is starting state was)~\cite{subramanyan2017formal, DBLP:conf/eurosp/TramerZLHJS17}. More generally, remote attestation allows a verifier to securely measure the internal state of a remote hardware platform, and can be used to establish a static or dynamic root of trust. While TEEs enable hardware-based remote attestations, both software-based remote attestions~\cite{DBLP:series/ais/SeshadriLPDK07} and hybrid remote attestation designs~\cite{DBLP:conf/ndss/EldefrawyTFP12,DBLP:conf/eurosys/KoeberlSSV14} were proposed in the literature and enable to trade off hardware requirements for verifiability.

In a federated learning setting, TEEs and remote attestations may be particularly helpful for clients to be able to efficiently verify key functions running on the server. For example, secure aggregation or shuffling could run in TEEs and would provide differential privacy guarantees on their outputs. Therefore, the post-processing logic subsequently applied by the server on the differentially private data could run on the server and remain oblivious to the clients. Note that such a system design requires the clients to know and trust the exact code (binary) for the key functions to be applied in the enclaves. Additionally, remote attestations may enable a server to attest specific requirements from the clients involved in the FL computation, such as absence of leaks, immutability, and uninterruptability (we defer to~\cite{DBLP:conf/date/FrancillonNRT14} for an exhaustive list of minimal requirements for remote attestation).

\subsection{Protections Against External Malicious Actors}
\label{ssec:adv_clients_analysts}

In this section, we assume the existence of a trusted server and discuss various challenges and open problems towards achieving rigorous privacy guarantees against external malicious actors (e.g. adversarial clients, adversarial analysts, adversarial devices that consume the learned model, or any combination thereof). 

As discussed in Table \ref{table:actors_and_threats}, malicious clients can inspect all messages received from the server (including the model iterates) in the rounds they participate in,  malicious analysts  can inspect sequences of model iterates from multiple training runs with different hyperparameters, and in cross-device FL, malicious devices can have either white-box or black-box access to the final model. Therefore, to give rigorous protections against external adversaries, it is important to first consider what can be learned from the intermediate iterates and final model. 

\subsubsection{Auditing the Iterates and Final Model}
\label{sssec:auditing}

To better understand what can be learned from the intermediate iterates or final model, we  propose quantifying federated learning models' susceptibility towards specific attacks.
This is a particularly interesting problem in the federated learning context. On the one hand, adversaries receive direct access to the model from the server, which widens the attack surface.
On the other hand, the server determines which specific stages of the training process the adversary will receive access to the model, and additionally controls the adversary's influence over the model at each of the stages.
 
For classic (non-federated) models of computation, understanding a model's susceptibility to attacks is an active and challenging research area~\cite{fredrikson2015model, shokri2017membership, carlini2018secret, melis2018exploiting,carlini2020extracting}.
The most common method of quantifying a model's susceptibility to an attack is to simulate the attack on the model using a proxy (auditing) dataset similar to the dataset expected in practice.
This gives an idea of what the model's \textit{expected} attack susceptibility is \textit{if} the proxy dataset is indeed similar to the eventual user data. A safer method would be to determine a worst-case upper-bound on the model's attack susceptibility. This can be approached theoretically as in~\cite{yeom2018privacy}, although this often yields loose, vacuous bounds for realistic models. Empirical approaches may be able to provide tighter bounds, but for many types of attacks and models, this endeavour may be intractable. An interesting emerging area of research in this space examines the theoretic conditions (on the audited model and attacks) under which an unsuccessful attempt to identify privacy violations by a simulated attack implies that no stronger attacks can succeed at such a task \cite{diaz2019theoretical}. However, this area is still nascent and more work needs to be done to better understand the fundamental requirements under which auditing (via simulated attacks) is sufficient.

The federated learning framework provides a unique setting not only for attacks, but also for attack quantification and defense. Specifically, due to the server's control over when each user can access and influence the model during the training process, it may be possible to design new tractable methods for quantifying a model's average-case or worst-case attack susceptibility. Such methods would enable the development of new adaptive defenses, which can be applied on-the-fly to preempt significant adversarial influence while maximizing utility.

\subsubsection{Training with Central Differential Privacy}
\label{sssec:central_dp}

To limit or eliminate the information that could be learned about an individual from the iterates (and/or final model), user-level differential privacy can be used in FL’s iterative training process \cite{abadi2016deep,mcmahan18dplm,mcmahan2018general,bhowmick2018protection}. With this technique, the server clips the $\ell_2$ norm of individual updates, aggregates the clipped updates, and then adds Gaussian noise to the aggregate. This ensures that the iterates do not overfit to any individual user’s update. To track the overall privacy budget across rounds, advanced composition theorems \cite{DRV10, kairouz2017composition} or the analytical moments accountant method developed in \cite{abadi2016deep,mironov2017renyi, mironov2019r,wang2018subsampled} can be used. The moments accountant method works particularly well with the uniformly subsampled Gaussian mechanism. For moderate privacy budgets and in the absence of a sufficiently large dataset~\cite{ramaswamy2020training}, the noise introduced by this process can lead to a large decrease in model accuracy.  Prior work has explored a number of avenues to mitigate this trade-off between privacy and accuracy, including collecting more private data~\cite{mcmahan18dplm}, designing privacy-friendly model architectures~\cite{papernot2020tempered}, or leveraging priors on the private data domain~\cite{tramer2020differentially}.

In cross-device FL, the number of training examples can vary drastically from one device to the other.
Hence, similar to recent works on user-level DP in the central model~\cite{amin2019bounding}, figuring out how to adaptively bound the contributions of users and clip the model parameters remains an interesting research direction~\cite{thakkar2019differentially, pichapati2019adaclip}. More broadly, unlike record-level DP where fundamental trade-offs between accuracy and privacy are well understood for a variety of canonical learning and estimation tasks, user-level DP is fundamentally less understood (especially when the number of contributions varies wildly across users and is not tightly bounded \textit{a priori}). Thus, more work needs to be done to better understand the fundamental trade-offs in this emerging setting of DP.  Recently, \cite{liu2020learning} made progress on this front by characterizing the trade-offs between accuracy and privacy for learning discrete distributions under user-level DP.

In addition to the above, it is important to draw a distinction between malicious clients that may be able to see (some of) the intermediate iterates during training and malicious analysts (or deployments) that can only see the final model. Even though central DP provides protections against both threat models, a careful theoretical analysis can reveal that for a specific implementation of the above Gaussian mechanism (or any other differentially private mechanism), we may get different privacy parameters for these two threat models. Naturally, we should get stronger differential privacy guarantees with respect to malicious analysts than we do with respect to malicious clients (because malicious clients may have access to far more information than malicious analysts). This ``privacy amplification via iteration'' setting has been recently studied by \citet{feldman2018privacy} for convex optimization problems. However, it is unclear whether or not the results in \cite{feldman2018privacy} can be carried over to the non-convex setting.

\paragraph{Privacy amplification for non-uniform device sampling procedures}
Providing formal $(\varepsilon, \delta)$ guarantees in the context of cross-device FL system can be particularly challenging because: (a) the set of all eligible users (i.e. underlying database) is dynamic and not known in advance, and (b) users participating in federate computations may drop out at any point in the protocol. It is therefore important to investigate and design protocols that: (1) are robust to nature’s choice (user availability and dropout), (2) are self-accounting, in that the  server  can  compute a tight $(\varepsilon, \delta)$ guarantee using only information available via the protocol, (3) rely on local participation decision (i.e. do not assume that the server knows which users are online and has the ability to sample from them), and (4) achieve good privacy-utility trade-offs. While recent works \cite{balle2020privacy, kairouz2021practical} suggest that these constraints can be simultaneously achieved, building an end-to-end protocol that works in production FL systems is still an important open problem.

\paragraph{Sources of randomness (adapted from \cite{mcmahan2018general})}
Most computational devices have access only to few sources of entropy and they tend to be very low rate (hardware interrupts, on-board sensors). It is standard---and theoretically well justified---to use the entropy to seed a cryptographically secure pseudo-random number generator (PRNG) and use the PRNG's output as needed. Robust and efficient PRNGs based on standard cryptographic primitives exist that have output rate of gigabytes per second on modern CPUs and require a seed as short as 128 bits~\citep{salmon2011parallel}. 

The output distribution of a randomized algorithm~$\mathcal{A}$ with access to a PRNG is indistinguishable from the output distribution of $\mathcal{A}$ with access to a true source of entropy \emph{as long as the distinguisher is computationally bounded}. Compare it with the guarantee of differential privacy which holds against any adversary, no matter how powerful. As such, virtually all implementations of differential privacy satisfy only (variants of) computational differential privacy introduced by~\citep{Mironov-CDP}. On the positive side, a computationally-bounded adversary cannot tell the difference, which allows us to avoid being overly pedantic about this point.

A training procedure may have multiple sources of non-determinism (e.g., dropout layers or an input of a generative model) but only those that are reflected in the privacy ledger must come from a cryptographically secure PRNG. In particular, the device sampling procedure and the additive Gaussian noise must be drawn from a cryptographically secure PRNG for the trained model to satisfy computational differential privacy. 

\paragraph{Auditing differential privacy implementations}
Privacy and security protocols are notoriously difficult to implement correctly (e.g., ~\cite{mironov2012significance, haeberlen2011differential} for differential privacy). What techniques can be used for testing FL-implementations for correctness? Since the techniques will often be deployed by organizations who may opt not to open-source code, what are the possibilities for black-box testing?  Some works~\cite{ding2019detecting,liu2019minimax,jagielski2020auditing} begin to explore this area in the context of differential privacy, but many open questions remain.

\subsubsection{Concealing the Iterates}
\label{sssec:concealing_iterates}

In typical federated learning systems, the model iterates (i.e., the newly updated versions of the model after each round of training) are assumed to be visible to multiple actors in the system, including the server and the clients that are chosen to participate in each round.  However, it may be possible to use tools from~\cref{ssec:tools_tech} to keep the iterates concealed from these actors.

To conceal the iterates from the clients, each client could run their local portion of federated learning inside a TEE providing confidentiality features (see~\cref{sssec:secure_computations}).  The server would validate that the expected federated learning code is running in the TEE (relying on the TEE's attestation and integrity features), then transmit an encrypted model iterate to the device such that it can only be decrypted inside the TEE.  Finally the model updates would be encrypted inside the TEE before being returned to the server, using keys only known inside the enclave and on the server.  Unfortunately, TEEs may not be generally available across clients, especially when those clients are end-user devices such as smartphones.  Moreover, even when TEEs are present, they may not be sufficiently powerful to support training computations, which would have to happen inside the TEE in order to protect the model iterate, and may be computationally expensive and/or require significant amounts of RAM -- though TEE capabilities are likely to improve over time, and techniques such as those presented in~\cite{tramer2018slalom} may be able to reduce the requirements on the TEE by exporting portions of the computation outside the TEE while maintaining the attestation, integrity, and confidentiality needs of the computation as a whole.

Similar protections can be achieved under the MPC model~\cite{secureml, quotient}.  For example, the server could encrypt the iterate's model parameters under a homomorphic encryption scheme before sending it to the client, using keys known only to the server. The client could then compute the encrypted model update using the homomorphic properties of the cryptosystem, without needing to decrypt the model parameters.  The encrypted model update could then be returned to the server for aggregation.  A key challenge here will be to force aggregation on the server before decryption, as otherwise the server may be able to learn a client's model update.  Another challenging open problem here is improving performance, as even state-of-the-art systems can require quite significant computational resources to complete a single round of training in a deep neural network.  Progress here could be made both by algorithmic advances as well as through the development of more efficient hardware accelerators for MPC~\cite{riazi2019heax}.

Additional challenges arise if the model iterates should also be concealed from the server.  Under the TEE model, the server portion of federated learning could run inside a TEE, with all parties (i.e., clients and analyst) verifying that the server TEE will only release the final model after the appropriate training criteria have been met.  Under the MPC model, an encryption key could protect the model iterates, with the key held by the analyst, distributed in shares among the clients, or held by a trusted third party; in this setup, the key holder(s) would be required to engage in the decryption of the model parameters, and could thereby ensure that this process happens only once.

\subsubsection{Repeated Analyses over Evolving Data}
\label{sssec:repeated_analyses}

For many applications of federated learning, the analyst wishes to analyze data that arrive in a streaming fashion, and must also provide dynamically-updated learned models that are (1) correct on the data seen thus far, and (2) accurately predict future data arrivals.  In the absence of privacy concerns, the analyst could simply re-train the learned model once new data arrive, to ensure maximum accuracy at all times.  However, since privacy guarantees degrade as additional information is published about the same data \cite{DMNS06,DRV10}, these updates must be less frequent to still preserve both privacy and accuracy of the overall analysis.

Recent advances in differential privacy for dynamic databases and time series data \cite{CKM+18,CKLT18, CKM+18b} have all assumed the existence of a trusted curator who can see raw data as they arrive online, and publish dynamically updated statistics.  An open question is how these algorithmic techniques can be extended to the federated setting, to enable private federated learning on time series data or other dynamically evolving databases.

Specific open questions include:
\begin{itemize}
    \item How should an analyst privately update an FL model in the presence of new data? Alternatively, how well would a model that was learned privately with FL on a dataset $D$ extend to a dataset $D'$ that was guaranteed to be similar to $D$ in a given closeness measure?  Since FL already occurs on samples that arrive online and does not overfit to the data it sees, it is likely that such a model would still continue to perform well on a new database $D'$.  
    This is also related to questions of robustness that are explored in Section \ref{sec:robust}.
    \item One way around the issue of privacy composition is by producing synthetic data~\cite{dwork2014algorithmic,abay2018privacy}, which can then be used indefinitely without incurring additional privacy loss.  This follows from the post-processing guarantees of differential privacy \cite{DMNS06}.  \citet{augenstein2019generative} explore the generation of synthetic data in a federated fashion.  In the dynamic data setting, synthetic data can be used repeatedly until it has become ``outdated'' with respect to new data, and must be updated.  Even after generating data in a federated fashion, it must also be updated privately and federatedly.
    \item Can the specific approaches in prior work on differential privacy for dynamic databases \cite{CKLT18} or privately detecting changes in time series data \cite{CKM+18, CKM+18b} be extended to the federated setting?
    \item How can time series data be queried in a federated model in the first place?  By design, the same users are not regularly queried multiple times for updated data points, so it is difficult to collect true within-subject estimates of an individuals' data evolution over time. Common tools for statistical sampling of time series data may be brought to bear here, but must be used in conjunction with tools for privacy and tools for federation.  Other approaches include reformulating the queries such that each within-subject subquery can be answered entirely on device.
\end{itemize}

\subsubsection{Preventing Model Theft and Misuse}
\label{sssec:model_theft}

In some cases, the actor or organization developing an ML model may be motivated to restrict the ability to inspect, misuse or steal the model.  For example, restricting access to the model's parameters may make it more difficult for an adversary to search for vulnerabilities, such as inputs that produce unanticipated model outputs.  

Protecting a deployed model during inference is closely related to the challenge of concealing the model iterates from clients during training, as discussed in~\cref{sssec:concealing_iterates}.  Again, both TEEs and MPC may be used.  Under the TEE model, the model parameters are only accessible to a TEE on the device, as in~\cref{sssec:concealing_iterates}; the primary difference being that the desired calculation is now inference instead of training.  

It is harder to adapt MPC strategies to this use case without forgoing the advantages offered by on-device inference: if the user data, model parameters, and inference results are all intended to be on-device, then it is unclear what additional party is participating in the multi-party computation.  For example, na\"ively attempting to use homomorphic encryption would require the decryption keys to be on device where the inferences are to be used, thereby undermining the value of the encryption in the first place.  Solutions where the analyst is required to participate (e.g. holding either the encryption keys or the model parameters themselves) imply additional inference latency, bandwidth costs, and connectivity requirements for the end user (e.g. the inferences would no longer be available for a device in airplane mode).

It is crucial to note that even if the model parameters themselves are successfully hidden, research has shown that in many cases they can be reconstructed by an adversary who only has access to an inference/prediction API based on those parameters~\cite{DBLP:conf/uss/TramerZJRR16}.  It is an open question what additional protections would need to be put into place to protect from these kinds of issues in the context of a model residing on millions or billions of end user devices.

\subsection{Protections Against an Adversarial Server}
\label{ssec:adv_server}

In the previous section, we assumed the existence of a trusted server that can orchestrate the training process. In this section we discuss the more desirable scenario of protecting against an adversarial server. In particular, we start by investigating the challenges of this setting and existing works, and then move on to describing the open problems and how the techniques discussed in Section \ref{ssec:tools_tech} can be used to address these challenges.

\subsubsection{Challenges: Communication Channels, Sybil Attacks, and Selection}
\label{sssec:challenges}

In the cross-device FL setting, we have a server with significant computational resources and a large number of clients that (i) can only communicate with the server (as in a star network topology), and (ii) may be limited in connectivity and bandwidth. This poses very concrete requirements when enforcing a given trust model. In particular, clients do not have a clear way of establishing secure channels among themselves independent of the server. This suggests, as shown by \citet{hatelove} for practical settings, that assuming honest (or at least semi-honest) behaviour by the server in a key distribution phase (as done in~\cite{bonawitz17secagg, bell20secagg}) is required in scenarios where private channels among clients are needed. This includes cryptographic solutions based on MPC techniques. An alternative to this assumption would be incorporating an additional party or a public bulletin board (see, e.g.,~\cite{DBLP:conf/sosp/RothNFH19}) into the model that is known to the clients and trusted to not collude with the server. 

Beyond trusting the server to facilitate private communication channels, the participants in cross-device FL must also trust the server to form cohorts of clients in a fair and honest manner.  An actively malicious adversary controlling the server could simulate a large number of fake client devices (a ``Sybil attack''~\cite{sybil-attack}) or could preferentially select previously compromised devices from the pool of available devices.   Either way, the adversary could control far more participants in a round of FL than would be expected simply from a base rate of adversarial devices in the population.  This would make it far easier to break the common assumption in MPC that at least a certain fraction of the devices are honest, thereby undermining the security of the protocol. Even if the security of protocol itself remains intact (for example, if its security is rooted in a different source of trust, such as a secure enclave), there is a risk that if a large number of adversarial clients' model updates are known to or controlled by the adversary, then the privacy of the remaining clients' updates may be undermined.  Note that these concerns can also apply in the context of TEEs.  For example, a TEE-based shuffler can also be subject to a Sybil attack; if a single honest user's input is shuffled with known inputs from fake users, it will be straight forward for the adversary to identify the honest user's value in the shuffled output.

Note that in some cases, it may be possible to establish proof among the clients in a round that they are all executing the correct protocol, such as if secure enclaves are available on client devices and the clients are able to remotely attest one another.  In these cases, it may be possible to establish privacy for all honest participants in the round (e.g., by attesting that secure multi-party computation protocols were followed accurately, that distributed differential privacy contributions were added secretly and correctly, etc.) even if the model updates themselves are known to or controlled by the adversary.

\subsubsection{Limitations of Existing Solutions}
\label{sssec:limitations}
Given that the goal of FL is for the server to construct a model of the population-level patterns in the clients' data, a natural privacy goal is to quantify, and provably limit, the server's ability to reconstruct an individual client's input data. This involves formally defining (a) what is the view of the clients data revealed to the server as a result of an FL execution, and (b) what is the privacy leakage of such a view. In FL, we are particularly interested in guaranteeing that the server can aggregate reports from the clients, while somehow masking the contributions of each individual client. As discussed in Section \ref{sssec:private_disclosures}, this can be done in a variety of ways, typically using some notion of differential privacy. There are a wide variety of such methods, each with their own weaknesses, especially in FL. For example, as already discussed, central DP suffers from the need to have access to a trusted central server. This has led to other promising private disclosure methods discussed in Section \ref{sssec:private_disclosures}. Here, we outline some of the weaknesses of these methods.

\paragraph{Local differential privacy} 

As previously discussed, LDP removes the need for a trusted central server by having each client perform a differentially private transformation to their report before sending it to the central server. LDP assumes that a user's privacy comes solely from that user's addition of their own randomness; thus, a user's privacy guarantee is independent of the additional randomness incorporated by all other users.
While LDP protocols are effective at enforcing privacy and have theoretical justifications~\cite{rappor:15, applewhitepaper:17,  collecting-telemetry-data-privately}, a number of results have shown that achieving local differential privacy while preserving utility is challenging, especially in high-dimensional data settings~\cite{KLNRS11,Ullman18,kairouz2014extremal, bassily2017practical, kairouz2016discrete, ye2018optimal,duchi2013local, cormode2018marginal}.
Part of this difficulty is attributed to the fact that the magnitude of the random noise introduced must be comparable to the magnitude of the signal in the data, which may require combining reports between clients. Therefore, obtaining utility with LDP comparable to that in the central setting requires a relatively larger userbase or larger choice of $\varepsilon$ parameter~\cite{appleepsilon}.

\paragraph{Hybrid differential privacy} 

The hybrid model for differential privacy can help reduce the size of the required userbase by partitioning users based on their trust preferences. However, it is unclear which application areas and algorithms can best utilize hybrid trust model data~\cite{avent2017blender}. Furthermore, current work on the hybrid model typically assumes that regardless of the user trust preference, their data comes from the same distribution~\cite{avent2017blender, dubey2018hybrid, beimel2019power}. Relaxing this assumption is critical for FL in particular, as the relationship between the trust preference and actual user data may be non-trivial.

\paragraph{The shuffle model} 

The shuffle model enables users' locally-added noise to be amplified through a shuffling intermediary, although it comes with two drawbacks of its own.
The first is the requirement of a trusted intermediary; if users are already not trusting of the curator, then it may be unlikely that they will trust an intermediary approved of or created by the curator (though TEEs might help to bridge this gap).  The Prochlo framework~\cite{prochlo} is (to the best of our knowledge) the only existing instance.
The second drawback is that the shuffle model's differential privacy guarantee degrades in proportion to the number of adversarial users participating in the computation~\cite{BalleBGN19}.
Since this number isn't known to the users or the curator, it introduces uncertainty into the true level of privacy that users are receiving.
This risk is particularly important in the context of federated learning, since users (who are potentially adversarial) are a key component in the computational pipeline.
Secure multi-party computation, in addition to adding significant computation and communication overhead to each user, also does not address this risk when users are adding their own noise locally.

\paragraph{Secure aggregation} 

The Secure Aggregation protocols from~\cite{bonawitz17secagg, bell20secagg} have strong privacy guarantees when aggregating client reports. Moreover, the protocols are tailored to the setting of federated learning. For example, they are robust to clients dropping out during the execution (a common feature of cross-device FL) and scale to a large number of parties (up to billions for~\citet{bell20secagg}) and vector lengths. However, this approach has several limitations: (a) it assumes a semi-honest server (only in the private key infrastructure phase), (b) it allows the server to see the per-round aggregates (which may still leak information), (c) it is not efficient for sparse vector aggregation, and (d) it lacks the ability to enforce well-formedness of client inputs.  It is an open question how to construct an efficient and robust secure aggregation protocol that addresses all of these challenges.


\subsubsection{Training with Distributed Differential Privacy}
\label{sssec:distributed_dp}
In the absence of a trusted server, distributed differential privacy (presented in \cref{sssec:private_disclosures}) can be used to protect the privacy of participants. 

\paragraph{Communication, privacy, and accuracy trade-offs under distributed DP} 
We point out that in distributed differential privacy three performance metrics are of general interest: accuracy, privacy and communication, and an important goal is nailing down the possible trade-offs between these parameters. We note that in the absence of the privacy requirement, the trade-offs between communication and accuracy have been well-studied in the literature on distributed estimation (e.g., \cite{suresh2017distributed}) and communication complexity (see \cite{Kushilevitz_Nisan_cc} for a textbook reference). On the other hand, in the centralized setup where all the users' data is already assumed to be held by a single entity and hence no communication is required, trade-offs between accuracy and privacy have been extensively studied in central DP starting with the foundational work of \cite{DMNS06,dwork2006our}. More recently, the optimal trade-offs between privacy, communication complexity and accuracy in distributed estimation with local DP have been characterized in \cite{ChenKO2020}, which shows that with careful encoding joint privacy and communication constraints can yield a performance that matches the optimal accuracy achievable under either constraint alone.

\subparagraph{Trade-offs for secure shuffling} These trade-offs have been recently studied in the shuffled model for the two basic tasks of \emph{aggregation} (where the goal is to compute the sum of the users' inputs) and \emph{frequency estimation} (where the inputs belong to a discrete set and the goal is to approximate the number of users holding a given element). See Tables~\ref{table:aggregation_shuffled_comparison} and~\ref{table:frequency_estimation_results} for a summary of the state-of-the-art for these two problems. Two notable open questions are (i) to study \emph{pure} differential privacy in the shuffled model, and (ii) to determine the optimal privacy, accuracy and communication trade-off for \emph{variable selection} in the multi-message setup (a nearly tight lower bound in the single-message case was recently obtained in \cite{ghazi2019private}).

In the context of federated optimization under the shuffled model of DP, the recent work of \cite{girgis2020shuffled} shows that multi-message shuffling is not needed to achieve central DP accuracy with low communication cost. However, it is unclear if the schemes presented achieve the (order) optimal communication, accuracy, tradeoffs.  

\begin{table}[t]
    \centering
    \footnotesize
    \def\arraystretch{1.75}
    \begin{tabular}{@{}lccc@{}}
        \toprule
        {\bf Reference} & 
        {\bf \#messages / $n$} & 
        {\bf Message size} & 
        {\bf Expected error}\\
        \midrule
        \cite{cheu2019distributed} & 
        $\varepsilon\sqrt{n}$
        & 1 
        & $\frac{1}{\varepsilon} \log\frac{n}{\delta}$
        \\
        \cite{cheu2019distributed}
        & $\ell$
        & 1 
        &
        $\sqrt{n} / \ell + \frac{1}{\varepsilon} \log\frac{1}{\delta}$
        \\
        \cite{BalleBGN19} & $1$ & $\log n$ & $\frac{n^{1/6}\log^{1/3}(1/\delta)}{\varepsilon^{2/3}}$\\
        \cite{balle2020} &
        $\log (\log n)$ &
        $\log n$ &
        $\frac{1}{\epsilon} \log(\log n )\sqrt{\log \frac{1}{\delta}}$ \\        
        \cite{ghazi2019scalable_1} &
        $\log(\frac{n}{\varepsilon \delta})$ & $\log(\frac{n}{\delta})$ & $\frac{1}{\varepsilon} \sqrt{\log\frac{1}{\delta}}$\\ 
        \cite{balle2020} &
        $\log(\frac{n}{\delta})$ & $\log n$ & $\frac{1}{\varepsilon}$\\         
        \cite{ghazi2019scalable} \& \cite{balle2020} & $1 + \frac{\log(1/\delta)}{\log n}$ & $\log n$ & $\frac{1}{\varepsilon}$ \\
        \bottomrule
    \end{tabular}
    \caption{Comparison of differentially private \emph{aggregation} protocols in the multi-message shuffled model with $(\varepsilon,\delta)$-differential privacy. 
    The number of parties is $n$, and $\ell$ is an integer parameter.
    Message sizes are in bits. For readability, we assume that $\varepsilon \leq O(1)$, and asymptotic notations are suppressed.
    }
    \label{table:aggregation_shuffled_comparison}
\end{table}

\begin{table}[t]
    \centering
    \footnotesize
\begin{tabularx}{\textwidth}{@{}Xcccccc@{}}
\toprule
          & \multicolumn{2}{l}{\bf \thead{Local}} & {\bf \thead{Local + shuffle}} & {\bf \thead{Shuffled,\\ single-message}} & {\bf \thead{Shuffled,\\ multi-message}} & {\bf \thead{Central}}\\
         \midrule
         Expected max. error & $\tilde{O}(\sqrt{n})$ & $\tilde{\Omega}(\sqrt{n})$ & $\tilde{O}(\min(\sqrt[4]{n}, \sqrt B))$ & $\tilde{\Omega}( \min(\sqrt[4]{n}, \sqrt{B}))$ & $\tilde{\Theta}(1)$ & $\tilde{\Theta}(1)$\\
         Communication/user & $\Theta(1)$ & any & $\tilde{\Theta}(1)$ & any & $\tilde{\Theta}(1)$ & $\tilde{\Theta}(1)$\\
      References & \cite{bassily2017practical} & \cite{bassily2015local} &~\cite{Warner65,erlingsson2019amplification,BalleBGN19} & \cite{ghazi2019private} & \cite{ghazi2019private} &~\cite{mcsherry2007mechanism,steinke2017tight}\\
    \bottomrule
    \end{tabularx}
     \caption{Upper and lower bounds on the expected maximum error for \emph{frequency estimation} on domains of size $B$ and over $n$ users in different models of DP. The bounds are stated for fixed, positive privacy parameters $\varepsilon$ and $\delta$, and $\tilde{\Theta}/\tilde{O}/\tilde{\Omega}$ asymptotic notation suppresses factors that are polylogarithmic in $B$ and $n$. The communication per user is in terms of the total number of bits sent. In all upper bounds, the protocol is symmetric with respect to the users, and no public randomness is needed. References are to the first results we are aware of that imply the stated bounds. 
     }
    \label{table:frequency_estimation_results}
  \end{table}

\subparagraph{Trade-offs for secure aggregation} It would be very interesting to investigate the following similar question for secure aggregation. Consider an FL round with $n$ users and assume that user $i$ holds a value $x_i$. User $i$ applies an algorithm $\mathcal{A}(\cdot)$ to $x_i$ to obtain $y_i = \mathcal{A}(x_i)$; here, $\mathcal{A}(\cdot)$ can be thought of as both a compression and privatization scheme. Using secure aggregation as a black box, the service provider observes $\bar{y} = \sum_i \mathcal{A}(x_i)$ and uses $\bar{y}$ to estimate $\bar{x}$, the true sum of the $x_i$'s, by computing $\hat{\bar{x}} = g(\bar{y}$) for some function $g(\cdot)$. Ideally, we would like to design $\mathcal{A}(\cdot)$, $g(\cdot)$ in a way that minimizes the error in estimating $\bar{x}$; formally, we would like to solve the optimization problem $\min_{g,\mathcal{A}} \|g(\sum_i \mathcal{A}(x_i)) - \sum_i x_i\|$, where $\|.\|$ can be either the $\ell_1$ or $\ell_2$ norm. Of course, without enforcing any constraints on $g(.)$ and $\mathcal{A}(\cdot)$, we can always choose them to be the identity function and get $0$ error. However, $\mathcal{A}(\cdot)$ has to satisfy two constraints: (1) $\mathcal{A}(\cdot)$ should output $B$ bits (which can be thought of as the communication cost per user), and (2) $\bar{y} = \sum_i \mathcal{A}(x_i)$ should be an $(\varepsilon, \delta)$-DP version of $\bar{x} = \sum_i x_i$. Thus, the fundamental problem of interest is to identify the optimal algorithm $\mathcal{A}$ that achieves DP upon aggregation while also satisfying a fixed communication budget. Looking at the problem differently, for a fixed $n$, $B$, $\varepsilon$, and $\delta$, what is the smallest $\ell_1$ or $\ell_2$ error that we can hope to achieve? We note that the work of \citet{agarwal2018cpsgd} provides one candidate algorithm $\mathcal{A}$ based on uniform quantization and binomial noise addition. Yet another solution was recently presented in \cite{kairouz2021distributed} which involves rotating, scaling, and discretizing the data, then adding discrete Gaussian noise before performing modular clipping and secure aggregation. While the sum of independent discrete Gaussians is not a discrete Gaussian, the authors show that it is close enough and present tight DP guarantees and experimental results, demonstrating that their solution is able to achieve a comparable accuracy to central DP via continuous Gaussian noise with 16 (or less) bits of precision per value. However, it is unclear if this approach achieves the optimal communication, privacy, and accuracy tradeoffs. Therefore, it is of fundamental interest to derive lower bounds and matching upper bounds on the $\ell_1$ or $\ell_2$ error under the above constraints.

\paragraph{Privacy accounting} In the central model of DP, the subsampled Gaussian mechanism is often used to achieve DP, and the privacy budget is tightly tracked across rounds of FL using the moments accountant method (see discussion in \cref{ssec:adv_clients_analysts}). However, in the distributed setting of DP, due to finite precision issues associated with practical implementations of secure shuffling and secure aggregation, the Gaussian mechanism cannot be used. Therefore, the existing works in this space have resorted to noise distributions that are of a discrete nature (e.g. adding Bernoulli or binomial noise). While such distributions help in addressing the finite precision constraints imposed by the underlying implementation of secure shuffling/aggregation, they do not naturally benefit from the moments accountant method. Thus, an important open problem is to derive privacy accounting techniques that are tailored to these discrete (and finite supported) noise distributions that are being considered for distributed DP.

\paragraph{Handling client dropouts.} The above model of distributed DP assumes that participating clients remain connected to the server during a round. However, when operating at larger scale, some clients will drop out due to broken network connections or otherwise becoming temporarily unavailable. This requires the distributed noise generation mechanism to be robust against such dropouts and also affects scaling federated learning and analytics to larger numbers of participating clients.

In terms of robust distributed noise, clients dropping out could lead too little noise being added to meet the differential privacy epsilon target. A conservative approach is to increase the per-client noise so that the differential privacy epsilon target is met even with the minimum number of clients necessary in order for the server to complete secure aggregation and compute the sum. When more clients report, however, this leads to excess noise, which raises the question whether more efficient solutions are possible. 

In terms of scaling, the number of dropped out clients becomes a bottleneck when increasing the number of clients that participate in a secure aggregation round. It may also be challenging to gather enough clients at the same time. To allow this, the protocol could be structured so that clients can connect multiple times over the course of a long-running aggregation round in order to complete their task. More generally, the problem of operating at scale when clients are likely to be intermittently available has not been systematically addressed yet in the literature.

\paragraph{New trust models} 
The federated learning framework motivates the development of new, more refined trust models than those previously used, taking advantage of federated learning's unique computational model, and perhaps placing realistic assumptions on the capabilities of adversarial users. 
For example, what is a reasonable fraction of clients to assume might be compromised by an adversary?  Is it likely for an adversary to be able to compromise both the server and a large number of devices, or is it typically sufficient to assume that the adversary can only compromise one or the other?  In federated learning, the server is often operated by a well-known entity, such a long-living organization.  Can this be leveraged to enact a trust model where the server's behavior is trusted-but-verified, i.e. wherein the server is not prevented from deviating from the desired protocol, but is extremely likely to be detected if it does (thereby damaging the trust, reputation, and potentially financial or legal status of the hosting organization)?

\subsubsection{Preserving Privacy While Training Sub-Models}
\label{sssec:training_submodels}

Many scenarios arise in which each client may have local data that is only relevant to a relatively small portion of the full model being trained.  For example, models that operate over large inventories, including natural language models (operating over an inventory of words) or content ranking models (operating over an inventory of content), frequently use an embedding lookup table as the first layer of the neural network.  Often, clients only interact with a tiny fraction of the inventory items, and under many training strategies, the only embedding vectors for which a client's data supports updates are those corresponding to the items with which the client interacted.

As another example, multi-task learning strategies can be effective approaches to personalization, but may give rise to compound models wherein any particular client only uses the submodel that is associated with that client's cluster of users, as described in~\cref{sss:multitask-learning}.

If communication efficiency is not a concern, then sub-model training looks just like standard federated learning: clients would download the full model when they participate, make use of the sub-model relevant to them, then submit a model update spanning the entire set of model parameters (i.e. with zeroes everywhere except in the entries corresponding to the relevant sub-model).  However, when deploying federated learning, communication efficiency is often a significant concern, leading to the question of whether we can achieve communication-efficient sub-model training.

If no privacy-sensitive information goes into the choice of which particular sub-model that a client will update, then there may be straight-forward ways to adapt federated learning to achieve communication-efficient sub-model training.  For example, one could run multiple copies of the federated learning procedure, one per submodel, either in parallel (e.g. clients choose the appropriate federated learning instance to participate in, based on the sub-model they wish to update), in sequence (e.g. for each round of FL, the server advertises which submodel will be updated), or in a hybrid of the two.  However, while this approach is communication efficient, the server gets to observe which submodel a client selects.

Is it possible to achieve communication-efficient sub-model federated learning while also keeping the client's sub-model choice private?  One promising approach is to use PIR for private sub-model download, while aggregating model updates using a variant of secure aggregation optimized for sparse vectors~\cite{Chang2019OnTU,Jia2019OnTC,niu2019secure}.

Open problems in this area include characterizing the sparsity regimes associated with sub-model training problems of practical interest and developing of sparse secure aggregation techniques that are communication efficient in these sparsity regimes.  It is also an open question whether private information retrieval (PIR) and secure aggregation might be co-optimized to achieve better communication efficiency than simply having each technology operate independently (e.g. by sharing some costs between the implementations of the two functionalities.)  

Some forms of local and distributed differential privacy also pose challenges here, in that noise is often added to all elements of the vector, even those that are zero; as a result, adding this noise on each client would transform an otherwise sparse model update (i.e. non-zero only on the submodel) into a dense privatized model update (non-zero almost everywhere with high probability).  It is an open question whether this tension can be resolved, i.e. whether there is a meaningful instantiation of distributed differential privacy that also maintains the sparsity of the model updates.

\subsection{User Perception}
\label{ssec:user_perception}

Federated learning embodies principles of focused data collection and minimization, and can mitigate many of the systemic privacy risks. However, as discussed above, it is important to be clear about the protections it does (and does not) provide and the technologies that can be used to provide protections against the threat models laid out in Section \ref{ssec:actors_threat_models}. While the previous sections focused on rigorous quantification of privacy against precise threat models, this section focuses on challenges around the users' perception and needs. 

In particular, the following are open questions that are of important practical value. Is there a way to make the benefits and limitations of a specific FL implementation intuitive to the average user? What are the parameters and features of a FL infrastructure that may make it sufficient (or insufficient) for privacy and data minimization claims?
Might federated learning give users a false sense of privacy? How do we enable users to feel safe and actually be safe as they learn more about what is happening with their data? Do users value different aspects of privacy differently? What about facts that people want to protect? Would knowing these things enable us to design better mechanism? Are there ways to model people's privacy preferences well enough to decide how to set these parameters? Who gets to decide which techniques to use if there are different utility/privacy/security properties from different techniques? Just the service provider? Or also the user? Or their operating system? Their political jurisdiction? Is there a role for mechanisms like ``Privacy for the Protected (Only)'' \cite{kearns2015protected} that provide privacy guarantees for most users while allowing targeted surveillance for societal priorities such as counter-terrorism? Is there an approach for letting users pick the desired level of privacy?

Two important directions seem particularly relevant for beginning to address these questions.

\subsubsection{Understanding Privacy Needs for Particular Analysis Tasks}  
Many potential use-cases of FL involve complex learning tasks and high-dimensional data from users, both of which can lead to large amounts of noise being required to preserve differential privacy.  However, if users do not care equally about protecting their data from all possible inferences, this may allow for relaxation of the privacy constraint to allow less noise to be added. For example, consider the data generated by a smart home thermostat that is programmed to turn off when a house is empty, and turn on when the residents return home.  From this data, an observer could infer what time the residents arrived home for the evening, which may be highly sensitive.  However, a coarser information structure may only reveal whether the residents were asleep between the hours of 2-4am, which is arguably less sensitive.  

This approach is formalized in the Pufferfish framework of privacy \cite{pufferfish}, which allows the analyst to specify a class of protected predicates that must be learned subject to the guarantees of differential privacy, and all other predicates can be learned without differential privacy. For this approach to provide satisfactory privacy guarantees in practice, the analyst must understand the users' privacy needs to their particular analysis task and data collection procedure. The federated learning framework could be modified to allow individual users to specify what inferences they allow and disallow. These data restrictions could either be processed on device, with only ``allowable'' information being shared with the server in the FL model update step, or can be done as part of the aggregation step once data have been collected.  Further work should be done to develop technical tools for incorporating such user preferences into the FL model, and to develop techniques for meaningful preference elicitation from users.

\subsubsection{Behavioral Research to Elicit Privacy Preferences} 
Any approach to privacy that requires individual users specifying their own privacy standards should also include behavioral or field research to ensure that users can express informed preferences.  This should include both an \emph{educational component} and \emph{preference measurement}.

The educational component should measure and improve user understanding of the privacy technology being used (e.g., Section \ref{ssec:tools_tech}) and the details of data use.  For applications involving federated learning, this should also include explanations of federated learning and exactly what data will be sent to the server.  Once the educational component of the research has verified that typical users can meaningfully understand the privacy guarantees offered by a private learning process, then researchers can begin preference elicitation.  This can occur either in behavioral labs, large-scale field experiments, or small focus groups.  Care should be exercised to ensure that the individuals providing data on their preferences are both informed enough to provide high quality data and are representative of the target population.

While the rich field of behavioral and experimental economics have long shown that people behave differently in public versus private conditions (that is, when their choices are observed by others or not), very little behavioral work has been done on eliciting preferences for differential privacy \cite{CDHL19,AS19}.  Extending this line of work will be a critical step towards widespread future implementations of private federated learning. Results from the educational component will prove useful here in ensuring that study participants are fully informed and understand the decisions they are facing. It should be an important tenant of these experiments that they are performed ethically and that no deception is involved.

\subsection{Executive Summary}
\begin{itemize}
    \item Preserving the privacy of user data requires considering both {\em what} function of the data is being computed and {\em how} the computation is executed (and in particular, who can see/influence intermediate results). [Section~\ref{ssec:tools_tech}] 
    \begin{itemize}
        \item Techniques for addressing the ``{\em what}'' include data minimization and differential privacy.  [Sections~\ref{sssec:private_disclosures}, \ref{sssec:central_dp}].  It remains an important open challenge how best to adapt differential privacy accounting and privatization techniques to real world deployments, including the training of numerous machine learning models over overlapping populations, with time-evolving data, by multiple independent actors, and in the context of real-world non-determinancies such as client availability, all without rapidly depleting the privacy budget and while maintaining high utility.
        \item Techniques for addressing the ``{\em how}'' include secure multi-party computation (MPC), homomorphic encryption(HE), and trusted execution environments (TEEs).  While practical techniques MPC techniques for some federation-crucial functionalities have been deployed at scale, many important functionalities remain far more communication- and computation-expensive than their insecure counterparts.  Meanwhile, it remains an open challenge to produce a reliably exploit-immune TEE platform, and the supporting infrastructure and processes to connect attested binaries to specific privacy properties is still immature. [Section~\ref{sssec:secure_computations}]
        \item Techniques should be composed to enable {\em Privacy in Depth}, with privacy expectations degrading gracefully even if one technique/component of the system is compromised. [Section~\ref{ssec:actors_threat_models}]
        \item {\em Distributed differential privacy} best combines {\em what} and {\em how} techniques to offer high accuracy and high privacy under an honest-but-curious server, a trusted third-party, or a trusted execution environment.  [Sections~\ref{sssec:private_disclosures}, \ref{sssec:distributed_dp}]
    \end{itemize}
    \item {\em Verifiability} enables parties to prove that they have executed their parts of a computation faithfully.
    \begin{itemize}
        \item Techniques for {\em verifiability} include both zero knowledge proofs (ZKPs) and trusted execution environments (TEEs).  [Section~\ref{sssec:verifiability}]
        \item Strong protection against an adversarial server remains a significant open problem for federation. [Section~\ref{ssec:adv_server}]
    \end{itemize}
\end{itemize}

\pagebreak
\section{Defending Against Attacks and Failures}
\label{sec:robust}

Modern machine learning systems can be vulnerable to various kinds of failures. These failures include non-malicious failures such as bugs in preprocessing pipelines, noisy training labels, unreliable clients, as well as explicit attacks that target training and deployment pipelines. Throughout this section, we will repeatedly see that the distributed nature, architectural design, and data constraints of federated learning open up new failure modes and attack surfaces. Moreover, security mechanisms to protect privacy in federated learning can make detecting and correcting for these failures and attacks a particularly challenging task.

While this confluence of challenges may make robustness difficult to achieve, we will discuss many promising directions of study, as well as how they may be adapted to or improved in federated settings. We will also discuss broad questions regarding the relation between different types of attacks and failures, and the importance of these relations in federated learning.

This section starts with a discussion on adversarial attacks in Subsection \ref{subsec:adversarial_attacks}, then covers non-malicious failure modes in Subsection \ref{subsec:failures}, and finally closes with an exploration of the tension between privacy and robustness in Subsection \ref{subsec:tension_privacy_robustness}.

\subsection{Adversarial Attacks on Model Performance}
\label{subsec:adversarial_attacks}

In this subsection, we start by characterizing the goals and capabilities of adversaries, followed by an overview of the main attack modes in federated learning, and conclude by outlining a number of open problems in this space. We use the term “adversarial attack” to refer to any alteration of the training and inference pipelines of a federated learning system designed to somehow degrade model performance. Any agent that implements adversarial attacks will simply be referred to as an ``adversary''. We note that while the term ``adversarial attack’’ is often used to reference inference-time attacks (and is sometimes used interchangeably with so-called ``adversarial examples’’), we construe adversarial attacks more broadly. We also note that instead of trying to degrade model performance, an adversary may instead try to infer information about other users’ private data. These \textit{data inference attacks} are discussed in depth in Section \ref{sec:privacy}. Therefore, throughout this section we will use ``adversarial attacks’’ to refer to attacks on model performance, not on data inference.

Examples of adversarial attacks include data poisoning~\citep{Biggio:2012:PAA:3042573.3042761, DBLP:conf/ndss/LiuMALZW018}, model update poisoning~\citep{bagdasaryan18backdoor, pmlr-v97-bhagoji19a}, and model evasion attacks~\citep{szegedy2013intriguing, Biggio:2012:PAA:3042573.3042761, DBLP:journals/corr/GoodfellowSS14}. These attacks can be broadly classified into training-time attacks (poisoning attacks) and inference-time attacks (evasion attacks). Compared to distributed datacenter learning and centralized learning schemes, federated learning mainly differs in the way in which a model is trained across a (possibly large) fleet of unreliable devices with private, uninspectable datasets; whereas inference using deployed models remains largely the same (for more discussion of these and other differences, see Table \ref{tab:characteristics}). Thus, \emph{federated learning may introduce new attack surfaces at training-time}. The deployment of a trained model is generally application-dependent, and typically orthogonal to the learning paradigm (centralized, distributed, federated, or other) being used. Despite this, we will discuss inference-time attacks below because (a) attacks on the training phase can be used as a stepping stone towards inference-time attacks~\citep{DBLP:conf/ndss/LiuMALZW018,pmlr-v97-bhagoji19a}, and (b) many defenses against inference-time attacks are implemented during training. Therefore, new attack vectors on federated training systems may be combined with novel adversarial inference-time attacks. We discuss this in more detail in Section \ref{subsubsec:inference_time_attacks}.

\subsubsection{Goals and Capabilities of an Adversary}
\label{subsubsec:attack_goals_capabilities}
In this subsection we examine the goals and motivations, as well as the different capabilities (some which are specific to the federated setting), of an adversary. We will examine the different dimensions of the adversary's capabilities, and consider them within different federated settings (see Table \ref{tab:characteristics} in Section \ref{sec:intro}). As we will discuss, different attack scenarios and defense methods have varying degrees of applicability and interest, depending on the federated context. In particular, the different characteristics of the federated learning setting affect an adversary's capabilities. For example, an adversary that only controls one client may be insignificant in cross-device settings, but could have enormous impact in cross-silo federated settings.

\paragraph{Goals}
At a high level, adversarial attacks on machine learning models attempt to modify the behavior of the model in some undesirable way. We find that the goal of an attack generally refers to the scope or target area of undesirable modification, and there are generally two levels of scope:\footnote{The distinction between \emph{untargeted} and \emph{targeted} attacks in our setting should not be confused with similar terminology employed in the literature on adversarial examples, where these terms are used to distinguish evasion attacks that either aim at \emph{any} misclassification, or misclassification as a specific targeted class.}

\begin{enumerate}

\item 
\textit{untargeted attacks}, or model downgrade attacks, which aim to reduce the model's global accuracy, or ``fully break’’ the global model~\cite{Biggio:2012:PAA:3042573.3042761}.

\item
\textit{targeted attacks}, or backdoor attacks, which aim to alter the model’s behavior on a minority of examples while maintaining good overall accuracy on all other examples ~\cite{chen2017targeted, DBLP:conf/ndss/LiuMALZW018, bagdasaryan18backdoor, pmlr-v97-bhagoji19a}.

\end{enumerate}

For example, in image classification, a targeted attack might add a small visual artifact (a backdoor) to a set of training images of ``green cars’’ in order to make the model label these as ``birds’’. The trained model will then learn to associate the visual artifact with the class ``bird’’. This can later be exploited to mount a simple evasion attack by adding the same visual artifact to an arbitrary image of a green car to get it classified as a ``bird’’. Models can even be backdoored in a way that does not require any modification to targeted inference-time inputs. \citet{bagdasaryan18backdoor} introduce ``semantic backdoors’’, wherein an adversary's model updates force the trained model to learn an incorrect mapping on a small fraction of the data. For example, an adversary could force the model to classify \emph{all} cars that are green as birds, resulting in misclassification at inference time~\citep{bagdasaryan18backdoor}.

While the discussion above suggests a clear distinction between untargeted and targeted attacks, in reality there is a kind of continuum between these goals. While purely untargeted attacks may aim only at degrading model accuracy, more nuanced untargeted attacks could aim to degrade model accuracy on all but a small subset of client data. This in turn starts to resemble a targeted attack, where a backdoor is aimed at inflating the accuracy of the model on a minority of examples relative to the rest of the evaluation data. Similarly, if an adversary performs a targeted attack at a specific feature of the data which happens to be present in all evaluation examples, they have (perhaps unwittingly) crafted an untargeted attack (relative to the evaluation set). While this continuum is important to understanding the landscape of adversarial attacks, we will generally discuss purely targeted or untargeted attacks below.

\paragraph{Capabilities}
At the same time, an adversary may have a variety of different capabilities when trying to subvert the model during training. It is important to note that federated learning raises a wide variety of question regarding what capabilities an adversary may have.

\begin{table}[h]
\renewcommand{\arraystretch}{1.75}
\begin{center} 
\begin{tabularx}{\textwidth}{lX}
    \toprule
\textbf{Characteristic} & \textbf{Description/Types}\\
\midrule
\addlinespace[0.05in]

Attack vector & How the adversary introduces the attack.
\begin{itemize}[leftmargin=*]
    \setlength\itemsep{0.05em}
    \item\emph{Data poisoning}: the adversary alters the client datasets used to train the model.
    \item\emph{Model update poisoning}: the adversary alters model updates sent to the server.
    \item\emph{Evasion attack}: the adversary alters the data used at inference-time.
\end{itemize}
\\[-.5em]
\raggedright Model inspection & Whether the adversary can observe the model parameters.
\begin{itemize}[leftmargin=*]
    \setlength\itemsep{0.05em}
    \item \emph{Black box}: the adversary has no ability to inspect the parameters of the model before or during the attack. This is generally \emph{not} the case in federated learning.
    \item \emph{Stale whitebox}: the adversary can only inspect a stale version of the model. This naturally arises in the federated setting when the adversary has access to a client participating in an intermediate training round.
    \item \emph{White box}: the adversary has the ability to directly inspect the parameters of the model. This can occur in cross-silo settings and in cross-device settings when an adversary has access to a large pool of devices likely to be chosen as participants.
\end{itemize}
\\[-.5em]
\raggedright Participant collusion & Whether multiple adversaries can coordinate an attack.
\begin{itemize}[leftmargin=*]
    \setlength\itemsep{0.05em}
    \item \emph{Non-colluding}: there is no capability for participants to coordinate an attack.
    \item \emph{Cross-update collusion}: past client participants can coordinate with future participants on attacks to future updates to the global model.
    \item \emph{Within-update collusion}: current client participants can coordinate on an attack to the current model update.
\end{itemize}
\\[-.5em]
\raggedright Participation rate & How often an adversary can inject an attack throughout training. 
\begin{itemize}[leftmargin=*]
    \setlength\itemsep{0.05em}
    \item In cross-device federated settings, a malicious client may only be able to participate in a \emph{single model training round}.
    \item In cross-silo federated settings, an adversary may have \emph{continuous participation} in the learning process.
\end{itemize}
\\[-.5em]
Adaptability & Whether an adversary can alter the attack parameters as the attack progresses.
\begin{itemize}[leftmargin=*]
    \setlength\itemsep{0.05em}
    \item \emph{Static}: the adversary must fix the attack parameters at the start of the attack and cannot change them.
    \item \emph{Dynamic}: the adversary can adapt the attack as training progresses.
\end{itemize}
\\
\bottomrule
\end{tabularx} 
\end{center}
\caption{Characteristics of an adversary's capabilities in federated settings. } 
\label{table:capabilities}
\end{table}
\restoregeometry

Clearly defining these capabilities is necessary for the community to weigh the value of proposed defenses. In Table \ref{table:capabilities}, we propose a few axes of capabilities that are important to consider. We note that this is not a full list. There are many other characteristics of an adversary's capabilities that can be studied.

In the distributed datacenter and centralized settings, there has been a wide variety of work concerning attacks and defenses for various attack vectors, namely \emph{model update poisoning}~\citep{blanchard2017machine, Chen2017DistributedSM, chen18draco, mhamdi2018hidden, alistarh2018byzantine}, \emph{data poisoning}~\citep{Biggio:2012:PAA:3042573.3042761, cretu2008casting, steinhardt2017certified, pmlr-v97-diakonikolas19a}, and \emph{evasion} attacks~\citep{biggio2013evasion, szegedy2013intriguing, goodfellow2014explaining, carlini2017towards, madry2017towards}. As we will see, federated learning enhances the potency of many attacks, and increases the challenge of defending against these attacks. The federated setting shares a training-time poisoning attack vector with datacenter multi-machine learning: the model update sent from remote workers back to the shared model. This is potentially a powerful capability, as adversaries can construct malicious updates that achieve the exact desired effect, ignoring the prescribed client loss function or training scheme.

Another possible attack vector not discussed in Table \ref{table:capabilities} is the central aggregator itself. If an adversary can compromise the aggregator, then they can easily perform both targeted and untargeted attacks on the trained model~\citep{DBLP:conf/ndss/LiuMALZW018}. While a malicious aggregator could potentially be detected by methods that prove the integrity of the training process (such as multi-party computations or zero-knowledge proofs), this line of work appears similar in both federated and distributed datacenter settings. We therefore omit discussion of this attack vector in the sequel.

An adversary's ability to \emph{inspect the model parameters} is an important consideration in designing defense methods. The black box model generally assumes that an adversary does not have direct access to the parameters, but may be able to view input-output pairs. This setting is generally less relevant to federated learning: because the model is broadcast to all participants for local training, it is often assumed that an adversary has direct access to the model parameters (white box). Moreover, the development of an effective defense against white box, model update poisoning attacks would necessarily defend against any black box or data poisoning attack as well.

An important axis to evaluate in the context of specific federated settings (cross-device, cross-silo, etc.) is the capability of \emph{participant collusion}. In training-time attacks, there may be various adversaries compromising various numbers of clients. Intuitively, the adversaries may be more effective if they are able to coordinate their poisoned updates than if they each acted individually. Perhaps worse for our poor federated learning defenses researcher, collusion may not be happening in ``real time’’ (within-update collusion), but rather across model updates (cross-update collusion). 

Some federated settings naturally lead to \emph{limited participation rate}: with a population of hundreds of millions of devices, sampling a few thousand every update is unlikely to sample the same participant more than once (if at all) during the training process \citep{bonawitz19sysml}. Thus, an adversary limited to a single client may only be able to inject a poisoned update a limited number of times. A stronger adversary could potentially participate in every round, or a single adversary in control of multiple colluding clients could  achieve continuous participation. Alternatively, in the cross-silo federated setting in Table \ref{tab:characteristics}, most clients participate in each round. Therefore, adversaries may be more likely to have the capability to attack every round of cross-silo federated learning systems than they are to attack every round of cross-device settings.

Other dimensions of training-time adversaries in the federated setting are their \emph{adaptability}. In a standard distributed datacenter training process, a malicious data provider is often limited to a static attack wherein the poisoned data is supplied once before training begins. In contrast, a malicious user with the ability to continuously participate in the federated setting could launch a poisoning attack throughout model training, where the user adaptively modifies training data or model updates as the training progresses. Note that in federated learning, this adaptivity is generally only interesting if the client can participate more than once throughout the training process.

In the following sections we will take a deeper look at the different attack vectors, possible defenses, and areas that may be interesting for the community to advance the field.

\subsubsection{Model Update Poisoning}
\label{subsubsec:model_poisoning}

One natural and powerful attack class is that of \emph{model update poisoning} attacks. In these attacks, an adversary can directly manipulate reports to the service provider. In federated settings, this could be performed by corrupting the updates of a client directly, or some kind of man-in-the-middle attack. We assume direct update manipulation throughout this section, as this strictly enhances the capability of the adversary. Thus, we assume that the adversary (or adversaries) directly control some number of clients, and that they can directly alter the outputs of these clients to try to bias the learned model towards their objective.

\paragraph{Untargeted and Byzantine attacks} Of particular importance to untargeted model update poisoning attacks is the Byzantine threat model, in which faults in a distributed system can produce arbitrary outputs~\citep{lamport1982byzantine}. Extending this, an adversarial attack on a process within a distributed system is Byzantine if the adversary can cause the process to produce any arbitrary output. Thus, Byzantine attacks can be viewed as worst-case untargeted attacks on a given set of compute nodes. Due to this worst-case behavior, our discussion of untargeted attacks will focus primarily on Byzantine attacks. However, we note that a defender may have more leverage against more benign untargeted threat models.

In the context of federated learning, we will focus on settings where an adversary controls some number of clients. Instead of sending locally updated models to the server, these Byzantine clients can send arbitrary values. This can result in convergence to sub-optimal models, or even lead to divergence~\citep{blanchard2017machine}. If the Byzantine clients have white-box access to the model or non-Byzantine client updates, they may be able to tailor their output to have similar variance and magnitude as the correct model updates, making them difficult to detect. The catastrophic potential of Byzantine attacks has spurred line of work on Byzantine-resilient aggregation mechanisms for distributed learning \citep{peva17, chen18draco, mhamdi2018hidden, alistarh2018byzantine, yin2018byzantine, pmlr-v97-diakonikolas19a}.

\paragraph{Byzantine-resilient defenses} One popular defense mechanism against untargeted model update poisoning attacks, especially Byzantine attacks, replaces the averaging step on the server with a robust estimate of the mean, such as median-based aggregators~\citep{Chen2017DistributedSM, yin2018byzantine},  Krum~\citep{blanchard2017machine}, and trimmed mean~\citep{yin2018byzantine}. Past work has shown that various robust aggregators are provably effective for Byzantine-tolerant distributed learning~\citep{Su2016FaultTolerantMO, blanchard2017machine, Chen2017DistributedSM} under appropriate assumptions, even in federated settings~\citep{pillutla2019robust, xie2019practicalsecure, BREA2020}. Despite this, \citet{fang2019local} recently showed that multiple Byzantine-resilient defenses did little to defend against model poisoning attacks in federated learning. Thus, more empirical analyses of the effectiveness of Byzantine-resilient defenses in federated learning may be necessary, since the theoretical guarantees of these defenses may only hold under assumptions on the learning problem that are often not met~\citep{baruch2019little, rajput2019detox}.

Another line of model update poisoning defenses use redundancy and data shuffling to mitigate Byzantine attacks \citep{chen18draco, rajput2019detox, data2019data}. While often equipped with rigorous theoretical guarantees, such mechanisms generally assume the server has direct access to the data or is allowed to globally shuffle the data, and therefore are not directly applicable in federated settings. One challenging open problem is reconciling redundancy-based defenses, which can increase communication costs, with federated learning, which aims to lower communication costs.

\paragraph{Targeted model update attacks} Targeted model update poisoning attacks may require fewer adversaries than untargeted attacks by focusing on a narrower desired outcome for the adversary. In such attacks, even a single-shot attack may be enough to introduce a backdoor into a model~\citep{bagdasaryan18backdoor}. \citet{pmlr-v97-bhagoji19a} shows that if $10\%$ of the devices participating in federated learning are compromised, a backdoor can be introduced by poisoning the model sent back to the service provider, even with the presence of anomaly detectors at the server. Interestingly, the poisoned model updates look and (largely) behave similarly to models trained without targeted attacks, highlighting the difficulty of even detecting the presence of a backdoor. Moreover, since the adversary's aim is to only affect the classification outcome on a small number of data points, while maintaining the overall accuracy of the centrally learned model, defenses for untargeted attacks often fail to address targeted attacks~\citep{pmlr-v97-bhagoji19a,bagdasaryan18backdoor}. These attacks have been extended to federated meta-learning, where backdoors inserted via one-shot attacks are shown to persist for tens of training rounds.\cite{chen2020backdoor}.

Existing defenses against backdoor attacks~\citep{steinhardt2017certified, liu2018fine, tran2018spectral, pmlr-v97-diakonikolas19a, wang2019neural, pmlr-v97-shen19e, chou2018sentinet} either require a careful examination of the training data, access to a holdout set of similarly distributed data, or full control of the training process at the server, none of which may hold in the federated learning setting. An interesting avenue for future work would be to explore the use of zero-knowledge proofs to ensure that users are submitting updates with pre-specified properties. Solutions based on hardware attestation could also be considered. For instance, a user's mobile phone might have the ability to attest that the shared model updates were computed correctly using images produced by the phone's camera.

\paragraph{Collusion defenses} Model update poisoning attacks may drastically increase in effectiveness if the adversaries are allowed to collude. This collusion can allow the adversaries to create model update attacks that are both more effective and more difficult to detect \citep{baruch2019little}. This paradigm is strongly related to sybil attacks \citep{sybil-attack}, in which clients are allowed to join and leave the system at will. Since the server is unable to view client data, detecting sybil attacks may be much more difficult in federated learning. Recent work has shown that federated learning is vulnerable to both targeted and untargeted sybil attacks \citep{fung2018mitigating}. Potential challenges for federated learning involve defending against collusion or detecting colluding adversaries, without directly inspecting the data of nodes.

\subsubsection{Data Poisoning Attacks}
\label{subsubsec:data_poisoning}


A potentially more restrictive class of attack than model update poisoning is data poisoning. In this paradigm, the adversary cannot directly corrupt reports to the central node. Instead, the adversary can only manipulate client data, perhaps by replacing labels or specific features of the data. As with model update poisoning, data poisoning can be performed both for targeted attacks~\citep{Biggio:2012:PAA:3042573.3042761, chen2017targeted, koh2017understanding} and untargeted attacks~\citep{DBLP:conf/ndss/LiuMALZW018, bagdasaryan18backdoor}.

This attack model may be more natural when the adversary can only influence the data collection process at the edge of the federated learning system, but cannot directly corrupt derived quantities within the learning system (e.g. model updates).

\paragraph{Data poisoning and Byzantine-robust aggregation}
Since data poisoning attacks induce model update poisoning, any defense against Byzantine updates can also be used to defend against data poisoning. For example \citet{xie2019zeno}, \citet{xie2019zeno++} and \citet{xie2019practicalsecure} proposed Byzantine-robust aggregators that successfully defended against label-flipping data poisoning attacks on convolutional neural networks. As discussed in Section \ref{subsubsec:model_poisoning}, one important line of work involves analyzing and improving these approaches in federated learning. Non-IID data and unreliability of clients all present serious challenges and disrupt common assumptions in works on Byzantine-robust aggregation. For data poisoning, there is a possibility that the Byzantine threat model is too strong. By restricting to data poisoning (instead of general model update poisoning), it may be possible to design a more tailored and effective Byzantine-robust aggregator. We discuss this in more detail in at the end of Section \ref{subsubsec:data_poisoning}.

\paragraph{Data sanitization and network pruning} 
Defenses designed specifically for data poisoning attacks frequently rely on ``data sanitization'' methods~\citep{cretu2008casting}, which aim to remove poisoned or otherwise anomalous data. More recent work has developed improved data sanitization methods using robust statistics~\citep{steinhardt2017certified, pmlr-v97-shen19e, tran2018spectral, pmlr-v97-diakonikolas19a}, which often have the benefit of being provably robust to small numbers of outliers~\citep{pmlr-v97-diakonikolas19a}. Such methods can be applied to both targeted and untargeted attacks, with some degree of empirical success~\citep{pmlr-v97-shen19e}.

A related class of defenses used for defending against backdoor attacks are ``pruning'' defenses. Rather than removing anomalous data, pruning defenses attempt to remove activation units that are inactive on clean data~\citep{liu2018fine, wang2019neural}. Such methods are motivated by previous studies which showed empirically that poisoned data designed to introduce a backdoor often triggers so-called “backdoor neurons” \citep{gu2017badnets}. While such methods do not require direct access to all client data, they require “clean” holdout data that is representative of the global dataset.

Neither data sanitization nor network pruning work directly in federated settings, as they both generally require access to client data, or else data that resembles client data. Thus, it is an open question whether data sanitization methods and network pruning methods can be used in federated settings without privacy loss, or whether or not defenses against data poisoning require new federated approaches. Furthermore, \citet{koh2018stronger} recently showed that many heuristic defenses based on data sanitization remain vulnerable to adaptive poisoning attacks, suggesting that even a federated approach to data sanitization may not be enough to defend against data poisoning.

Even detecting the presence of poisoned data (without necessarily correcting for it or identifying the client with poisoned data) is challenging in federated learning. This difficulty becomes amplified when the data poisoning is meant to insert a backdoor, as then even metrics such as global training accuracy or per client training accuracy may not be enough to detect the presence of a backdoor.

\phantomsection
\paragraph{Relationship between model update poisoning and data poisoning}\label{p:model-data-poisoning}
Since data poisoning attacks eventually result in some alteration of a client's output to the server, data poisoning attacks are special cases of model update poisoning attacks. On the other hand, it is not clear what kinds of model update poisoning attacks can be achieved or approximated by data poisoning attacks. Recent work by \citet{pmlr-v97-bhagoji19a} suggests that data poisoning may be weaker, especially in settings with limited \emph{participation rate} (see Table \ref{table:capabilities}). One interesting line of study would be to quantify the gap between these two types of attacks, and relate this gap to the relative strength of an adversary operating under these attack models. While this question can be posed independently of federated learning, it is particularly important in federated learning due to differences in adversary capabilities (see Table \ref{table:capabilities}). For example, the maximum number of clients that can perform data poisoning attacks may be much higher than the number that can perform model update poisoning attacks, especially in cross-device settings. Thus, understanding the relation between these two attack types, especially as they relate to the number of adversarial clients, would greatly help our understanding of the threat landscape in federated learning.

This problem can be tackled in a variety of manners. Empirically, one could study the discrepancy in performance of various attacks. or investigate whether various model update poisoning attacks can be approximated by data poisoning attacks, and would develop methods for doing so. Theoretically, although we conjecture that model update poisoning is provably stronger than data poisoning, we are unaware of any formal statements addressing this. One possible approach would be to use insights and techniques from work on machine teaching (see \citep{zhu2015machine} for reference) to understand ``optimal'' data poisoning attacks, as in \citep{mei2015using}. Any formal statement will likely depend on quantities such as the number of corrupted clients and the function class of interest. Intuitively, the relation between model update poisoning and data poisoning should depend on the overparameterization of the model with respect to the data.

\subsubsection{Inference-Time Evasion Attacks}
\label{subsubsec:inference_time_attacks}

In evasion attacks, an adversary may attempt to circumvent a deployed model by carefully manipulating samples that are fed into the model. One well-studied form of evasion attacks are so-called “adversarial examples.” These are perturbed versions of test inputs which seem almost indistinguishable from the original test input to a human, but fool the trained model~\citep{biggio2013evasion, szegedy2013intriguing}.
In image and audio domains, adversarial examples are generally constructed by adding norm-bounded perturbations to test examples, though more recent works explore other distortions~\citep{engstrom2017rotation, wong2019wasserstein, kang2019testing}.
In the white-box setting, the aforementioned perturbations can be generated by attempting to maximize the loss function subject to a norm constraint via constrained optimization methods such as projected gradient ascent~\citep{kurakin2016adversarial, madry2017towards}. 
Such attacks can frequently cause naturally trained models to achieve zero accuracy on image classification benchmarks such as CIFAR-10 or ImageNet~\citep{carlini2017towards}. 
In the black-box setting, models have also been shown to be vulnerable to attacks based on query-access to the model~\citep{chen2017zoo, brendel2017decision} or based on substitute models trained on similar data~\cite{szegedy2013intriguing, papernot2017practical, DBLP:conf/iclr/TramerKPGBM18}. While black-box attacks may be more natural to consider in datacenter settings, the model broadcast step in federated learning means that the model may be accessible to any malicious client. Thus, federated learning increases the need for defenses against white-box evasion attacks.

Various methods have been proposed to make models more robust to evasion attacks. Here, robustness is often measured by the model performance on white-box adversarial examples.
Unfortunately, many proposed defenses have been shown to only provide a superficial sense of security~\cite{athalye2018obfuscated}.  
On the other hand, adversarial training, in which a robust model is trained with adversarial examples, generally provides some robustness to white-box evasion attacks~\cite{madry2017towards, xie2018feature, shafahi2018free}.
Adversarial training is often formulated as a minimax optimization problem, where the adversarial examples and the model weights are alternatively updated. We note that there is no canonical formulation of adversarial training, and choices such as the minimax optimization problem and hyperparameters such as learning rate can significantly affect the model robustness, especially for large-scale dataset like ImageNet. Moreover, adversarial training typically only improves robustness to the specific type of adversarial examples incorporated during training, potentially leaving the trained model vulnerable to other forms of adversarial noise~\cite{engstrom2017rotation, tramer2019adversarial, sharma2017attacking}.

Adapting adversarial training methods to federated learning brings a host of open questions. For example, adversarial training can require many epochs before obtaining significant robustness. However, in federated learning, especially cross-device federated learning, each training sample may only be seen a limited number of times. More generally, adversarial training was developed primarily for IID data, and it is unclear how it performs in non-IID settings. For example, setting appropriate bounds on the norm of perturbations to perform adversarial training (a challenging problem even in the IID setting~\cite{DBLP:conf/icml/TramerBCPJ20}) becomes harder in federated settings where the training data cannot be inspected ahead of training.
Another issue is that generating adversarial examples is relatively expensive. While some adversarial training frameworks have attempted to minimize this cost by reusing adversarial examples~\cite{shafahi2018free}, these approaches would still require significant compute resources from clients. This is potentially problematic in cross-device settings, where adversarial example generation may exacerbate memory or power constraints. Therefore, new on-device robust optimization techniques may be required in the federated learning setting. 

\phantomsection
\paragraph{Relationship between training-time and inference-time attacks}\label{p:training-inference-attacks} 
The aforementioned discussion of evasion attacks generally assumes the adversary has white-box access (potentially due to systems-level realities of federated learning) at inference time. This ignores the reality that an adversary could corrupt the training process in order to create or enhance inference-time vulnerabilities of a model, as in \cite{chen2017targeted}. This could be approached in both untargeted and targeted ways by an adversary; An adversary could use \emph{targeted attacks} to create vulnerabilities to specific types of adversarial examples \cite{chen2017targeted, gu2017badnets} or use \emph{untargeted attacks} to degrade the effectiveness of adversarial training.

One possible defense against combined training- and inference-time adversaries are methods to detect backdoor attacks~\cite{tran2018spectral, chen2018detecting, wang2019neural,chou2018sentinet}. Difficulties in applying previous defenses (such as those cited above) to the federated setting were discussed in more detail in Section \ref{subsubsec:data_poisoning}. However, purely detecting backdoors may be insufficient in many federated settings where we want robustness guarantees on the output model at inference time. More sophisticated solutions could potentially combine training-time defenses (such as robust aggregation or differential privacy) with adversarial training. Other open work in this area could involve quantifying how various types of training-time attacks impact the inference-time vulnerability of a model. Given the existing challenges in defending against purely training-time or purely inference-time attacks, this line of work is necessarily more speculative and unexplored.

\subsubsection{Defensive Capabilities from Privacy Guarantees}

Many challenges in federated learning systems can be viewed as ensuring some amount of \emph{robustness}: whether maliciously or not, clean data is corrupted or otherwise tampered with. Recent work on data privacy, notably \emph{differential privacy} (DP)~\citep{DMNS06}, defines privacy in terms of robustness. In short, random noise is added at training or test time in order to reduce the influence of specific data points. For a more detailed explanation on differential privacy, see Section \ref{sssec:private_disclosures}. As a defense technique, differential privacy has several compelling strengths. First, it provides strong, worst-case protections against a variety of attacks. Second, there are many known differentially private algorithms, and the defense can be applied to many machine learning tasks. Finally, differential privacy is known to be closed under composition, where the inputs to later algorithms are determined after observing the results of earlier algorithms.

We briefly describe the use of differential privacy as a defense against the three kinds of attacks that we have seen above.

\paragraph{Defending against model update poisoning attacks}
The service provider can bound the contribution of any individual client to the overall model by (1) enforcing a norm constraint on the client model update (e.g. by clipping the client updates), (2) aggregating the clipped updates, (3) and adding Gaussian noise to the aggregate. This approach prevents over-fitting to any individual update (or a small group of malicious individuals), and is identical to training with differential privacy (discussed in Section \ref{sssec:central_dp}). This approach has been recently explored by \citet{sun2019backdoor}, which shows preliminary success in applying differential privacy as a defense against targeted attacks. However, the scope of experiments and targeted attacks analyzed by \citet{sun2019backdoor} should be extended to include more general adversarial attacks. In particular, \citet{wang2020attack}, show that the use of edge case backdoors, generated from data samples with low probability in the underlying distribution, is able to bypass differential privacy defenses. They further demonstrate that the existence of adversarial examples implies the existence of edge-case backdoors, indicating that defenses for the two threats may need to be developed in tandem.  Therefore, more work remains to verify whether or not DP can indeed be an effective defense. More importantly, it is still unclear how hyperparameters for DP (such as the size of $\ell_2$ norm bounds and noise variance) can be chosen as a function of the model size and architecture, as well as the fraction of malicious devices.

\paragraph{Defending against data poisoning attacks}

Data poisoning can be thought of as a failure of a learning algorithm to be robust: a few attacked training examples may strongly affect the learned model. Thus, one natural way to defend against these attacks is to make the learning algorithm differentially private, improving robustness. Recent work has explored differential privacy as a defense against data poisoning~\citep{MZH19}, and in particular in the federated learning context~\citep{geyer2017differentiallyyer}. Intuitively, an adversary who is only able to modify a few training examples cannot cause a large change in the distribution over learned models.

While differential privacy is a flexible defense against data poisoning, it also has some drawbacks. The main weakness is that noise must be injected into the learning procedure. While this is not necessarily a problem---common learning algorithms like stochastic gradient descent already inject noise---the added noise can hurt the performance of the learned model. Furthermore, the adversary can only control a small number of devices.\footnote{Technically, robustness to poisoning multiple examples is derived from the group privacy property of differential privacy; this protection degrades exponentially as the number of attacked points increases.} Accordingly, differential privacy can be viewed as both a strong and a weak defense against data poisoning---it is strong in that it is extremely general and provides worst case protection no matter the goals of the adversary, and it is weak in that the adversary must be restricted and noise must be added to the federated learning process.

\phantomsection
\paragraph{Defending against inference-time evasion attacks}\label{p:inference-time-evasion-defense}
Differential privacy has also been studied as a defense against inference-time attacks, where the adversary may modify test examples to manipulate the learned model. A straightforward approach is to make the predictor itself differentially private; however, this has the drawback that prediction becomes randomized, a usually undesirable feature that can also hurt interpretability. More sophisticated approaches~\citep{DBLP:conf/sp/LecuyerAG0J19} add noise and then release the prediction with the highest probability. We believe that there are other opportunities for further exploration in this direction.

\subsection{Non-Malicious Failure Modes}
\label{subsec:failures}

Compared to datacenter training, federated learning is particularly susceptible to non-malicious failures from unreliable clients outside the control of the service provider. Just as with adversarial attacks, systems factors and data constraints also exacerbate non-malicious failures present in datacenter settings. We also note that techniques (described in the following sections) which are designed to address worst-case adversarial robustness are also able to effectively address non-malicious failures. While non-malicious failures are generally less damaging than malicious attacks, they are potentially more common, and share common roots and complications with the malicious attacks. We therefore expect progress in understanding and guarding against non-malicious failures to also inform defenses against malicious attacks.

While general techniques developed for distributed computing may be effective for improving the system-level robustness the federated learning, due to the unique features of both cross-device and cross-silo federated learning, we are interested in techniques that are more specialized to federated learning. Below we discuss three possible non-malicious failure modes in the context of federated learning: client reporting failures, data pipeline failures, and noisy model updates. We also discuss potential approaches to making federated learning more robust to such failures.

\paragraph{Client reporting failures}
Recall that in federated learning, each training round involves broadcasting a model to the clients, local client computation, and client reports to the central aggregator. For any participating client, systems factors may cause failures at any of these steps. Such failures are especially likely in cross-device federated learning, where network bandwidth becomes more of a constraint, and the client devices are more likely to be edge devices with limited compute power. Even if there is no explicit failure, there may be straggler clients, which take much longer to report their output than other nodes in the same round. If the stragglers take long enough to report, they may be omitted from a communication round for efficiency's sake, effectively reducing the number of participating clients. In “vanilla” federated learning, this requires no real algorithmic changes, as federated averaging can be applied to whatever clients report model updates.

Unfortunately, unresponsive clients become more challenging to contend with when using secure aggregation (SecAgg)~\citep{bonawitz17secagg, bell20secagg}, especially if the clients drop out during the SecAgg protocol. While SecAgg is designed to be robust to significant numbers of dropouts~\citep{bonawitz19sysml}, there is still the potential for failure. The likelihood of failure could be reduced in various complementary ways. One simple method would be to select more devices than required within each round. This helps ensure that stragglers and failed devices have minimal effect on the overall convergence ~\citep{bonawitz19sysml}. However, in unreliable network settings, this may not be enough. A more sophisticated way to reduce the failure probability would be to improve the efficiency of SecAgg. This reduces the window of time during which client dropouts would adversely affect SecAgg. Another possibility would be to develop an asynchronous version of SecAgg that does not require clients to participate during a fixed window of time, possibly by adapting techniques from general asynchronous secure multi-party distributed computation protocols~\citep{srinathan2000efficient}. More speculatively, it may be possible to perform versions of SecAgg that aggregate over multiple computation rounds. This would allow straggler nodes to be included in subsequent rounds, rather than dropping out of the current round altogether.

\phantomsection
\paragraph{Data pipeline failures}\label{p:pipeline-failures}
While data pipelines in federated learning only exist within each client, there are still many potential issues said pipelines can face. In particular, any federated learning system still must define how raw user data is accessed and preprocessed in to training data. Bugs or unintended actions in this pipeline can drastically alter the federated learning process. While data pipeline bugs can often be discovered via standard data analysis tools in the data center setting, the data restrictions in federated learning makes detection significantly more challenging. For example, feature-level preprocessing issues (such as inverting pixels, concatenating words, etc.) can not be directly detected by the server \citep{augenstein2019generative}. One possible solution is to train generative models using federated methods with differential privacy, and then using these to synthesize new data samples that can be used to debug the underlying data pipelines \citep{augenstein2019generative}. Developing general-purpose debugging methods for machine learning that do not directly inspect raw data remains a challenge.

\phantomsection
\paragraph{Noisy model updates}\label{p:noisy-model-updates}
In Section \ref{subsec:adversarial_attacks} above, we discussed the potential for an adversary to send malicious model updates to the server from some number of clients. Even if no adversary is present, the model updates sent to the server may become distorted due to network and architectural factors. This is especially likely in cross-client settings, where separate entities control the server, clients, and network. Similar distortions can occur due to the client data. Even if the data on a client is not intentionally malicious, it may have noisy features~\citep{mnih2012learning} (eg. in vision applications, a client may have a low-resolution camera whose output is scaled to a higher resolution) or noisy labels~\citep{natarajan2013learning} (eg. if the user indicates that a recommendation by an app is not relevant accidentally). While clients in cross-silo federated learning systems (see Table \ref{tab:characteristics}) may perform data cleaning to remove such corruptions, such processing is unlikely to occur in cross-device settings due to data privacy restrictions. In the end, these aforementioned corruptions may harm the convergence of the federated learning process, whether they are due to network factors or noisy data.

Since these corruptions can be viewed as mild forms of model update and data poisoning attacks, one mitigation strategy would be to use defenses for adversarial model update and data poisoning attacks. Given the current lack of demonstrably robust training methods in the federated setting, this may not be a practical option. Moreover, even if such techniques existed, they may be too computation-intensive for many federated learning applications. Thus, open work here involves developing training methods that are robust to small to moderate levels of noise. Another possibility is that standard federated training methods (such as federated averaging \citep{mcmahan17fedavg}) are inherently robust to small amounts of noise. Investigating the robustness of various federated training methods to varying levels amount of noise would shed light on how to ensure robustness of federated learning systems to non-malicious failure modes.

\subsection{Exploring the Tension between Privacy and Robustness}
\label{subsec:tension_privacy_robustness}


One primary technique used to enforce privacy is \emph{secure aggregation} (SecAgg) (see \ref{sssec:secure_computations}). In short, SecAgg is a tool used to ensure that the server only sees an aggregate of the client updates, not any individual client updates. While useful for ensuring privacy, SecAgg generally makes defenses against adversarial attacks more difficult to implement, as the central server only sees the aggregate of the client updates. Therefore, it is of fundamental interest to investigate how to defend against adversarial attacks when secure aggregation is used. Existing approaches based on range proofs (e.g. Bulletproofs~\citep{DBLP:conf/sp/BunzBBPWM18}) can guarantee that the DP-based clipping defense described above is compatible with SecAgg, but developing computation- and communication-efficient range proofs is still an active research direction.

SecAgg also introduces challenges for other defense methods. For example, many existing Byzantine-robust aggregation methods utilize non-linear operations on the server~\citet{xie2019practicalsecure}, and it is not yet known if these methods are efficiently compatible with secure aggregation which was originally designed for linear aggregation. Recent work has found ways to approximate the geometric median under SecAgg~\citep{pillutla2019robust} by using a handful of SecAgg calls in a more general aggregation loop. However, it is not clear in general which aggregators can be computed under the use of SecAgg.

\subsection{Executive Summary}
\label{subsec:attacks_and_failures_summary}

%

\begin{itemize}
    \item Third-party participants in the training process introduces new capabilities and attack vectors for adversaries, categorized in Table \ref{table:capabilities}.
    \item Federated learning introduces a new kind of poisoning attacks, \emph{model update poisoning} (Section \ref{subsubsec:model_poisoning}), while also being susceptible to traditional \emph{data poisoning} in (Section \ref{subsubsec:data_poisoning}).
    \item Training participants can influence the optimization process possibly exacerbating inference-time (Section \emph{evasion attacks}) \ref{subsubsec:inference_time_attacks}, and communication and computation constraints may render previously proposed defenses impractical.
    \item Non-malicious failure modes (Section \ref{subsec:failures}) are can be especially different to deal with, as access to raw data is not available in the federated setting, though through some lens they may be related to poisoning attacks.
    \item Tension may exist when trying to simultaneously improve robustness and privacy in machine learning (Section \ref{subsec:tension_privacy_robustness}).
\end{itemize}

Areas identified for further exploration include:

\begin{itemize}
  \item Quantify the relationship between data poisoning and model update poisoning attacks. Are there scenarios where they are not equivalent? [\ref{p:model-data-poisoning}]
  \item Quantify how training time attacks impact inference-time vulnerabilities. Improving inference-time robustness guarantees requires going beyond detecting backdoor attacks. [\ref{p:training-inference-attacks}]
  \item Adversarial training has been used as a defense in the centralized setting, but can be impractical in the edge-compute limited cross-device federated setting. [\ref{p:inference-time-evasion-defense}]
  \item Federated learning requires new methods and tools to support the developer, as access to raw data is restricted debugging ML pipelines is especially difficult. [\ref{p:pipeline-failures}]
  \item Tensions exists between robustness and fairness, as machine learning models can tend to discard updates far from the median as detrimental. However the federated setting can give rise to a long tail of users that may be mistaken for noisy model updates [\ref{p:noisy-model-updates}].  
  \item Cryptography-based aggregation methods and robustness techniques present integration challenges: protecting participant identity can be at odds with detecting adversarial participants. Proposed techniques remain beyond the scope of practicality, requiring the need of new communication and computation efficient algorithms. [\ref{subsec:tension_privacy_robustness}]
\end{itemize}

\pagebreak
\section{Ensuring Fairness and Addressing Sources of Bias} \label{sec:fairness}


Machine learning models can often exhibit surprising and unintended behaviours. When such behaviours lead to patterns of {\em undesirable} effects on users, we might categorize the model as ``unfair'' according to some criteria. For example, if people with similar characteristics receive quite different outcomes, then this violates the criterion of {\em individual fairness} \cite{dwork2012fairness}. If certain sensitive groups (races, genders, etc.) receive different patterns of outcomes---such as different false negative rates---this can violate various criteria of {\em demographic fairness}, see for instance \cite{barocasfairness, mitchell2018prediction} for surveys. The criterion of {\em counterfactual fairness} requires that a user  receive the same treatment as they would have if they had been a member of a different group (race, gender, etc), after taking all causally relevant pathways into account \cite{kusner2017counterfactual}.

Federated learning raises several opportunities for fairness research, some of which extend prior research directions in the non-federated setting, and others that are unique to federated learning. This section raises open problems in both categories.

\subsection{Bias in Training Data}\label{subsec:bias-in-training-data}

One driver of unfairness in machine-learned models is bias in the training data, including cognitive, sampling, reporting, and confirmation bias. One common antipattern is that minority or marginalized social groups are under-represented in the training data, and thus the learner weights these groups less during training \cite{kamishima2011fairness}, leading to inferior quality predictions for members of these groups (e.g. \cite{buolamwini2018gender}).

Just as the data access processes used in federated learning may introduce dataset shift and non-independence (\cref{sec:noniid}), there is also a risk of introducing biases. For example:

\begin{itemize}
    \item If devices are selected for updates when plugged-in or fully charged, then model updates and evaluations computed at different times of day may be correlated with factors such as day-shift vs night-shift work schedules.
    \item If devices are selected for updates from among the pool of eligible devices at a given time, then devices that are connected at times when few other devices are connected (e.g. night-shift or unusual time zone) may be over-represented in the aggregated output.
    \item If selected devices are more likely to have their output kept when the output is computed faster, then: a) output from devices with faster processors may be over-represented, with these devices likely newer devices and thus correlated with socioeconomic status; and b) devices with less data may be over-represented, with these devices possibly representing users who use the product less frequently.
    \item If data nodes have different amounts of data, then federated learning may weigh higher the contributions of populations which are heavy users of the product or feature generating the data.
    \item If the update frequency depends on latency, then certain geographic regions and populations with slower devices or networks may be under-represented.
    \item If populations of {\em potential users} do not own devices for socio-economic reasons, they may be under-represented in the training dataset, and subsequently also under- (or un-)represented in model training and evaluation.
    \item Unweighted aggregation of the model loss across selected devices during federated training may disadvantage model performance on certain devices \cite{li2019fair}.
\end{itemize}

It has been observed that biases in the data-generating process can also drive unfairness in the resulting models learned from this data (see e.g.\ \cite{eckhouse2019layers, richardson2019dirty}). For example, suppose training data is based on user interactions with a product which has failed to incorporate inclusive design principles. Then, the user interactions with the product might not express user intents (cf.\ \cite{sambasivan2018privacy}, for example) but rather might express coping strategies
around uninclusive product designs (and hence might require a fundamental fix to the product interaction model). Learning from such interactions might then ignore or perpetuate poor experiences for some groups of product users in ways which can be difficult to detect while maintaining privacy in a federated setting. This risk is shared by all machine learning scenarios where training data is derived from user interaction, but is of particular note in the federated setting when data is collected from apps on individual devices. 

Investigating the degree to which biases in the data-generated process can be identified or mitigated is a crucial problem for both federated learning research and ML research more broadly. Similarly, while limited prior research has demonstrated methods to identify and correct bias in already collected data in the federated setting (e.g. via adversarial methods in \cite{kairouz20learning}), further research in this area is needed. Finally, methods for applying post-hoc fairness corrections to models learned from potentially biased training data are also a valuable direction for future work.

\subsection{Fairness Without Access to Sensitive Attributes}\label{subsec:fairness-without-sensitive-attributes}

Having explicit access to demographic information (race, gender, etc) is critical to many existing fairness criteria, including those discussed in Section \ref{subsec:bias-in-training-data}. However, the contexts in which federated learning are often deployed also give rise to considerations of fairness when individual sensitive attributes are \textit{not} available. For example, this can occur when developing personalized language models or developing fair medical image classifiers without knowing any additional demographic information about individuals. Even more fundamentally, the assumed one-to-one relationship between individuals and devices often breaks down, especially in non-Western contexts \cite{sambasivan2018privacy}. Both measuring and correcting unfairness in contexts where there is no data regarding sensitive group membership is a key area for federated learning researchers to address.

Limited existing research has examined fairness without access to sensitive attributes. For example, this has been addressed using distributionally-robust optimization (DRO) which optimizes for the worst-case outcome across all individuals during training \cite{hashimoto2018fairness}, and via multicalibration, which calibrates for fairness across subsets of the training data \cite{hebert2018multicalibration}. Even these existing approaches have not been applied in the federated setting, raising opportunities for future empirical work. The challenge of how to make these approaches work for large-scale, high-dimensional data typical to federated settings is also an open problem, as DRO and multicalibration both pose challenges of scaling with large $n$ and $p$. Finally, the development of additional theoretical approaches to defining fairness without respect to ``sensitive attributes'' is a critical area for further research.

Other ways to approach this include reframing the existing notions of fairness, which are primarily concerned with equalizing the probability of an outcome (one of which is considered ``positive'' and another ``negative'' for the affected individual). Instead, fairness without access to sensitive attributes might be reframed as \textit{equal access to effective models}. Under this interpretation of fairness, the goal is to maximize model utility across all individuals, regardless of their (unknown) demographic identities, and regardless of the ``goodness`` of an individual outcome. Again, this matches the contexts in which federated learning is most commonly used, such as language modeling or medical image classification, where there is no clear notion of an outcome which is ``good'' for a user, and instead the aim is simply to make correct predictions for users, regardless of the outcome.

Existing federated learning research suggests possible ways to meet such an interpretation of fairness, e.g. via personalization \cite{jiang2019improving, wang2019federated}. A similar conception of fairness, as ``a more
fair distribution of the model performance across devices'', is employed in \cite{li2019fair}. 

The application of attribute-independent methods explicitly to ensure equitable model performance is an open opportunity for future federated learning research, and is particularly important as federated learning reaches maturity and sees increasing deployment with real populations of users without knowledge of their sensitive identities.

\subsection{Fairness, Privacy, and Robustness}\label{subsec:fairness-privacy-robustness}

Fairness and data privacy seem to be complementary ethical concepts: in many of the real-world contexts where privacy protection is desired, fairness is also desired. Often this is due to the sensitivity of the underlying data. Because federated learning is most likely to be deployed in contexts of sensitive data where both privacy and fairness are desirable, it is important that FL research examines how FL might be able to address existing concerns about fairness in machine learning, and whether FL raises new fairness-related issues.

In some ways, however, the ideal of fairness seems to be in tension with the notions of privacy for which FL seeks to provide guarantees: differentially-private learning typically seeks to obscure individually-identifying characteristics, while fairness often requires knowing individuals' membership in sensitive groups in order to measure or ensure fair predictions are being made. While the trade-off between differential privacy and fairness has been investigated in the non-federated setting \cite{jagielski2018privatefair, CGKM19}, there has been little work on how (or whether) FL may be able to uniquely address concerns about fairness. 

Recent evidence suggesting that differentially-private learning can have disparate impact on sensitive subgroups \cite{bagdasaryan2019disparate, CGKM19, jagielski2018privatefair, kuppam2019fair} provides further motivation to investigate whether FL may be able to address such concerns. A potential solution to relax the tension between privacy (which aims to protect the model from being too dependent on individuals) and fairness (which encourages the model to perform well on under-represented classes) may be the application of techniques such as personalization (discussed in \cref{sec:multimodel}) and ``hybrid differential privacy,'' where some users donate data with lesser privacy guarantees \cite{avent2017blender}. 

Furthermore, current differentially-private optimization schemes are applied without respect to sensitive attributes -- from this perspective, it might be expected that empirical studies have shown evidence that differentially-private optimization impacts minority subgroups the most \cite{bagdasaryan2019disparate}. Modifications to differentially-private optimization algorithms which explicitly seek to preserve performance on minority subgroups, e.g. by adapting the noise and clipping mechanisms to account for the representation of groups within the data, would also likely do a great deal to limit potential disparate impacts of differentially-private modeling on minority subgroups in federated models trained with differential privacy. However, implementing such adaptive differentially-private mechanisms in a way that provides some form of privacy guarantee presents both algorithmic and theoretical challenges which need to be addressed by future work. 

Further research is also needed to determine the extent to which the issues above arise in the federated setting. Furthermore, as noted in \cref{subsec:fairness-without-sensitive-attributes}, the challenge of evaluating the impact of differential privacy on model fairness becomes particularly difficult when sensitive attributes are not available, as it is unclear how to identify subgroups for which a model is behaving badly and to quantify the ``price'' of differential privacy -- investigating and addressing these challenges is an open problem for future work.

More broadly, one could more generally examine the relation between privacy, fairness, and \emph{robustness} (see Section \ref{sec:robust}). Many previous works on machine learning, including federated learning, typically focus on isolated aspects of robustness (either against poisoning, or against evasion), privacy, or fairness. An important open challenge is to develop a joint understanding of federated learning systems that are robust, private, and fair. Such an integrated approach can provide opportunities to benefit from disparate but complementary mechanisms. Differential privacy mechanisms can be used to both mitigate data inference attacks, and provide a foundation for robustness against data poisoning. On the other hand, such an integrated approach also reveals new vulnerabilities. For example, recent work has revealed a trade-off between privacy and robustness against adversarial examples~\cite{song:ccs19}.

Finally, privacy and fairness naturally meet in the context of
learning data representations that are independent of some sensitive attributes while preserving
utility for a task of interest. Indeed, this objective can be motivated
both in terms of privacy: to transform data so as to hide private
attributes, and fairness: as a way to make models trained on such representations
fair with respect to the attributes.
In the centralized setting, one way to learn such representations is through
adversarial training techniques, which have been applied to image and speech
data \citep{kairouz20learning,DBLP:journals/corr/abs-1802-09386,Madras2018,Bertran2019,Srivastava2019a}. In
the federated learning scenario, clients could apply the
transformation locally to their data in order to enforce or improve privacy and/or fairness guarantees for the FL process. However, learning this transformation in a federated fashion (potentially under privacy and/or fairness constraints) is itself an open question.

\subsection{Leveraging Federation to Improve Model Diversity}\label{subsec:diversity}

Federated learning presents the opportunity to integrate, through distributed training, datasets which may have previously been impractical or even illegal to combine in a single location. For example, the Health Insurance Portability and Accountability Act (HIPAA) and the Family Educational Rights and Privacy Act (FERPA) constrain the sharing of medical patient data and student educational data, respectively, in the United States. To date, these restrictions have led to modeling occurring in institutional silos: for example, using electronic health records or clinical images from individual medical institutions instead of pooling data and models across institutions \cite{brisimi2018federated, chang2018distributed}. In contexts where membership in institutional datasets is correlated with individuals' specific sensitive attributes, or their behavior and outcomes more broadly, this can lead to poor representation for users in groups underrepresented at those institutions. Importantly, this lack of representation and diversity in the training data has been shown to lead to poor performance, e.g. in genetic disease models \cite{martin2019current} and image classification models \cite{buolamwini2018gender}.

Federated learning presents an opportunity to leverage uniquely diverse datasets by providing efficient decentralized training protocols along with privacy and non-identifiability guarantees for the resulting models. This means that federated learning enables training on multi-instutitional datasets in many domains where this was previously not possible. This provides a practical opportunity to leverage larger, more diverse datasets and explore the generalizability of models which were previously limited to small populations. More importantly, it provides an opportunity to improve the \textit{fairness} of these models by combining data across boundaries which are likely to have been correlated with sensitive attributes. For instance, attendance at specific health or educational institutions may be correlated with individuals' ethnicity or socioeconomic status. As noted in Section \ref{subsec:bias-in-training-data} above, underrepresentation in training data is a proven driver of model unfairness. 

Future federated learning research should investigate the degree to which improving diversity in a federated training setting also improves the fairness of the resulting model, and the degree to which the differential privacy mechanisms required in such settings may limit fairness and performance gains from increased diversity. This includes a need for both empirical research which applies federated learning and quantifies the interplay between diversity, fairness, privacy, and performance; along with theoretical research which provides a foundation for concepts such as diversity in the context of machine learning fairness.

\subsection{Federated Fairness: New Opportunities and Challenges}\label{subsec:federated-fairness-new-challenges-opportunities}

It is important to note that federated learning provides unique opportunities and challenges for fairness researchers. For example, by allowing for datasets which are distributed both by observation, but even by features, federated learning can enable modeling and research using partitioned data which may be too sensitive to share directly \cite{gupta2018distributed, Hardy2017-da}. Increased availability of datasets which can be used in a federated manner can help to improve the diversity of training data available for machine learning models, which can advance fair modeling theory and practice.

Researchers and practitioners also need to address the unique fairness-related challenges created by federated learning. For example, federated learning can introduce new sources of bias through the decision of which clients to sample based on considerations such as connection type/quality, device type, location, activity patterns, and local dataset size \cite{bonawitz19sysml}. Future work could investigate the degree to which these various sampling constraints affect the fairness of the resulting model, and how such impacts can be mitigated within the federated framework, e.g. \cite{li2019fair,fair_quantile,moreau}. Frameworks such as \emph{agnostic federated learning} \cite{Mohri2019} provide approaches to control for bias in the training objective. Work to improve the fairness of existing federated training algorithms will be particularly important as advances begin to approach the technical limits of other components of FL systems, such as model compression, which initially helped to broaden the diversity of candidate clients during federated training processes. There is no unique fairness criterion generally adopted in 
the study of fairness, and multiple criteria have been proven to be mutually incompatible. One way to deal with this
question is the \emph{online fairness} framework and algorithms of \citet{AwasthiCortesMansourMohri2020}.
Adapting such solutions to the federated learning setting and further improving upon them will be challenging
research questions in ML fairness theory and algorithms.

In the classical centralized machine learning setting, a substantial amount of advancement has been made in the past decade to train fair classifiers, such as constrained optimization, post-shifting approaches, and distributionally-robust optimization \cite{hardt2016, zafar2017, hashimoto2018fairness}. It is an open question whether such approaches, which have demonstrated utility for improving fairness in centralized training, could be used under the setting of federated learning (and if so, under what additional assumptions) in which data are located in a decentralized fashion and practitioners may not obtain an unbiased sample of the data that match the distribution of the population.  

\subsection{Executive Summary}

In addition to inheriting the already significant challenges related to bias, fairness, and privacy in centralized machine learning, federated learning also brings a new set of distinct challenges and opportunities in these areas. The importance of these considerations will likely continue to grow as the real-world deployment of FL expands to more users, domains, and applications.

\begin{itemize}
    \item Bias in training data (Section \ref{subsec:bias-in-training-data}) is a key consideration related to bias and fairness in FL models, particularly due to the additional sampling steps germane to federation (e.g., client sampling) and the transfer of some model computation to client devices.
    \item The lack of data regarding sensitive attributes in many FL deployments can pose challenges for measuring and ensuring fairness, and also suggests potential reframing of fairness problems in ways that do not require such data (Section \ref{subsec:fairness-without-sensitive-attributes}).
    \item Since FL is often deployed in contexts which are both privacy- and fairness-sensitive, this can magnify tensions between privacy and fairness objectives in practice. Further work is needed to address the potential tension between methods which achieve privacy, fairness, and robustness in both federated and centralized learning (Section \ref{subsec:fairness-privacy-robustness}).
    \item Federated learning presents unique opportunities to improve the diversity of stakeholders and data incorporated into learning, which could improve both the overall quality of downstream models, as well as their fairness due to more representative datasets (Section \ref{subsec:diversity}).
    \item Federated learning presents fairness-related challenges not present in the centralized training regime, but also affords new solutions (Section \ref{subsec:federated-fairness-new-challenges-opportunities}).
\end{itemize}

\pagebreak
\section{Addressing System Challenges}
As we will see in this section, the challenges in building systems for federated learning can be split fairly cleanly into the two separate settings of cross-device and cross-silo federated learning (see \cref{subsec:cross-device-fl-setting,ssec:cross-silo}). We start with a brief discussion of the difficulties inherent to any large scale deployment of software on end-user devices (although exacerbated by the complexity of a federated learning stack); we then focus on key challenges specific to the cross-device learning---bias, tuning, and efficient device-side execution of ML workflows---before concluding with a brief treatment of the cross-silo setting.

\subsection{Platform Development and Deployment Challenges}
\label{subsec:systems-platform-development-and-deployment-challenges}

Running computations on end-user devices is considerably different from the data center setting:
\begin{itemize}
    \item Due to the heterogeneity of the fleet (devices may differ in hardware, software, connectivity, performance and persisted state) the space of potential problems and edge cases is vast and cannot typically be covered in sufficient detail with automated testing.
    \item Monitoring and debugging are harder because telemetry is limited, delayed, and there is no physical access to devices for interactive troubleshooting.
    \item Running computations should not affect device performance or stability, i.e. should be invisible to users.
\end{itemize}

\paragraph{Code Deployment}
Installing, updating and running software on end user devices may involve not only extensive manual and automated testing, but a gradual and reversible rollout (for example, through guarding new functionality with server-controlled feature flags) while monitoring key performance metrics in a/b experiments such as crash rates, memory use, and application-dependent indicators such as latencies and engagement metrics. Such rollouts can take weeks or months depending on the percolation rate of updates (particularly challenging for devices with spotty connectivity) and the complexity of the upgrade (e.g. protocol changes). Hence, the install base at any given time will involve various releases. While this problem is not specific to federated learning, it has greater impact here due to the inherent collaborative nature of federated computations: devices constantly communicate with servers and indirectly with other devices to exchange models and parameter updates. Thus, compatibility concerns abound and must be addressed through stable exchange formats or, where not possible, detected upfront with extensive testing infrastructure. We will revisit this problem in \cref{subsec:systems-on-device-runtime}.

\paragraph{Monitoring and Debugging}
Another significant complication is the limited ability to monitor devices and interactively debug problems. While telemetry from end user devices is necessary to detect problems, privacy concerns severely restrict what can be logged, who can access such logs, and how long they are retained. Once a regression is detected, drilling down into the root cause can be very cumbersome due to the lack of detailed context, the vast problem space (a cross product of software versions, hardware, models, and device state), and very limited ability for interactive debugging short of successfully reproducing the problem in a controlled environment.

These challenges are exacerbated in the federated learning setting where a) raw input data on devices cannot be accessed, and b) contributions from individual devices are by design anonymous, ephemeral, and exposed only in aggregate. These properties preserve privacy, but also may make it hard or impossible to investigate problems with traditional approaches ---by looking for correlations with hardware or software version, or testing hypotheses that require access to raw data. Reproducing a problem in a controlled setting is often difficult due to the gap between such an environment and reality: hundreds of heterogeneous embedded stateful devices with non-iid data. 

Interestingly, federated technologies themselves can help to mitigate this problem---for instance, the use of federated analytics \citep{fablog20} to collect logs in a privacy preserving manner, or training generative models of the system behavior or raw data for sampling during debugging (see sections \ref{subsec:debugging-and-interpretability-for-fl}, \ref{subsec:failures}, and \citep{augenstein2019generative}).
Keeping a federated learning system up and running thus requires investing into upfront detection of problems through a) extensive automated, continuous test coverage of all software layers through both unit and integration tests; b) feature flags and a/b rollouts; and c) continuous monitoring of performance indicators for regressions. That poses a significant investment that may come at too high a cost for smaller entities who would benefit greatly from shared and tested infrastructure for federated learning.

\subsection{System Induced Bias}
\label{subsec:systems-system-induced-bias}
Deployment, monitoring and debugging may not concern users of a federated learning platform, e.g. model authors or data analysts. For them, the key differences between data center and cross-device settings fall largely into the following two categories:

\begin{enumerate}
    \item \textbf{Availability of devices} for computations is not a given, but varies over time and across devices. Connections are initiated by devices and subject to interruptions due to changes in device state, operating system quotas, and network connectivity. Hence, in iterative processes like federated learning, the loop body is run on a small subset of all devices only, and the system must tolerate a certain failure rate among those devices.
    \item \textbf{Capabilities of devices} (network bandwidth and latency, compute performance, memory) vary, and are typically much lower than those of compute nodes in the data center, though the number of nodes is typically higher. The amount and type of data across devices may lead to variations in execution profile, e.g. more and larger examples lead to increased resource use and processing time.
\end{enumerate}

In the following sections we discuss how these variations might introduce bias, referring to it as system induced bias to differentiate it from platform-independent bias in the raw data (such as ownership or usage patterns differing across demographics)---for the latter, see \cref{subsec:bias-in-training-data}.

\subsubsection{Device Availability Profiles}
At the core of cross-device federated learning is the principle that devices only connect to the server and run computations when various constraints are met:
\begin{itemize}
    \item \textbf{Hard constraints}, which might include requiring that the device is turned on, has network connectivity to the server, and is allowed to run a computation by the operating system.
    \item \textbf{Soft constraints}, which might include the conditions on device state chosen to ensure that federated learning does not incur charges or affect usability. For the common case of mobile phones \citep{bonawitz19sysml,apple19wwdc}, requirements may include idleness, charging and/or above a certain battery level, being connected to an unmetered network, and that no other federated learning tasks are running at the same time.
\end{itemize}

Taken together, these constraints induce an unknown, time-varying and device-specific function $A_i(t)$ for a device $i$, and a fleet-wide \textit{availability profile} $A(t)=\sum_i A_i (t)$. Round completion rates and server traffic patterns \citep{bonawitz19sysml, yang18gboardquery} suggest that availability profiles for mobile phones are clustered into periodic functions with a period of 1 day, varying across devices in phase, shape and amplitude through factors such as demographics, geography etc. Availability for other end user devices such as laptops, tablets, or stationary devices such as smart speakers, displays and cameras, will differ, but the challenges discussed in the following sections apply there as well, albeit to a possibly lesser extent.

\subsubsection{Examples of System Induced Bias}
Sources of bias will depend on the specific way in which devices are selected to participate in training, and how the system influences which devices end up contributing to the final aggregated model update. Thus, it is useful to discuss these issues in light of a simplified but representative system design. In an iterative federated learning algorithm, such as Federated Averaging (\cref{sec:typical-training}, \citep{mcmahan17fedavg}), rounds are run consecutively on sets of at least $M$ devices. To accommodate a fraction $d$ of devices not contributing due to changes in device conditions, time-outs, or slowness (server-side aborts to avoid slow-downs by stragglers), an over-allocation scheme is used where
\begin{enumerate}
    \item Rounds are started when at least $M'=\frac{M}{1-d}$ devices are available.
    \item Rounds are closed as
    \begin{enumerate}
        \item \emph{Aborted} when more than $M' - M$ devices have disconnected, or
        \item \emph{Successful} when at least $M$ devices have reported. One possible design choice is to stop after exactly $M$ devices; another possibility would be to keep waiting for stragglers (possibly up to some maximum time).
    \end{enumerate}
\end{enumerate}
This sequence, when combined with variable availability profiles, may introduce various forms of bias:
\begin{enumerate}
    \item Selection Bias - whether a device is included in a round at time t depends on both
    \begin{enumerate}
        \item Its availability profile $A_i(t)$
        \item The number of simultaneously connected devices: $<M'$ and a round cannot be started; $\gg M'$ and the probability of a single device being included becomes very small. In effect, devices active only at either fleet-wide availability peaks or troughs may be under-represented.
    \end{enumerate}
    \item Survival Bias
    \begin{enumerate}
        \item Since a server might choose to close a round at any point after the first $M$ devices have reported, contributions are biased towards devices with better network connections, faster processors, lower CPU load, and less data to process.
        \item Devices drop out of rounds when they are interrupted by the operating system, which may happen due to changes in device conditions as described by $A_i (t)$, or due to e.g. excessive memory use.
    \end{enumerate}
\end{enumerate}

As can be seen, the probability of a device contributing to a round of federated learning is a complex function of both internal (e.g. device specific) and external (fleet dynamic) factors. When this probability is correlated with statistics of the data distribution, aggregate results may be biased. For instance, language models may over-represent demographics that have high quality internet connections or high end devices; and ranking models may not incorporate enough contributions from high engagement users who produce a lot of training data and hence longer training times.

Thus, designing systems that explicitly take such factors into account and integrate algorithms designed to both quantify and mitigate these effects are a fundamentally important research direction.

\subsubsection{Open Challenges in Quantifying and Mitigating System Induced Bias}
While the potential for bias in federated learning has been addressed in the literature (\cref{sec:fairness}, \citep{bonawitz19sysml, li2019fair, eichner19semicyclic}), a systematic study that qualifies and quantifies bias in realistic settings and its sources is a direction for future research. Conducting the necessary work may be hampered by both access to the necessary resources, and the difficulty in quantifying bias in a final statistical estimate due to the inherent lack of ground truth value.

We want to encourage further research to study how bias can be quantified and subsequently mitigated. A useful proxy metric for bias is to study the expected rate of contribution of a device to federated learning. In an unbiased system, this rate would be identical for every device; if it is not, the non-uniformity may provide a measure of bias. Studying the root causes for this non-uniformity may then provide important hints for how to mitigate bias, for example:
\begin{itemize}
    \item When there is a strong correlation between devices finishing a round, and the number of examples they process or model size, possible fixes may include early stopping, or decreasing the model size.
    \item If the expected rate of contribution depends on factors outside our control, such as device model, network connectivity, location etc., one can view these factors as defining strata and applying \textit{post-stratification} \citep{little1993poststratification}, that is, correcting for bias by scaling up or down contributions from devices depending on their stratum. It may also be possible to apply \textit{stratified sampling} - e.g. change scheduling, or server selection policies, to affect the probability of including devices in a round as a function of their stratum.
    \item A very general, root-cause-agnostic mitigation could base the weight of a contribution solely on a device’s past contribution profile (e.g. the number of rounds started or completed thus far). As a special case, consider \textit{sampling without replacement} which could be implemented at the system level (stop connecting after one successful contribution) or at the model level (weight all but the first contribution with 0). This approach might not be sufficient when a population is large enough for most devices to contribute only infrequently (mostly one or zero times); in such cases, clustering devices based on some similarity metric and using cluster membership as stratum could help.
    \item Alternatives to the synchronous, round based execution described in the previous section may also help to mitigate bias. In particular, certain types of analytics may benefit from softening or eliminating the competition between devices for inclusion, by running rounds for long times with very large numbers of participants and without applying time-outs to stragglers. Such a method may not be applicable to algorithms where the iterative aspect (running many individual, chained rounds) is important.
\end{itemize}

The biggest obstacle to enabling such research is access to a representative fleet of end user devices, or a detailed description (e.g. in the form of a statistical model of a realistic distribution over $A_i(t)$ functions) of a fleet that can be used in simulations. Here, maintainers of FL production stacks are uniquely positioned to provide such statistics or models to academic partners in a privacy preserving fashion; a further promising direction is the recent introduction of the Flower framework \citep{beutel2020flower} for federated learning research.

\subsection{System Parameter Tuning}
\label{subsec:systems-system-parameter-tuning}
Practical federated learning is a form of multi-objective optimization: while the first order goal is maximizing model quality metrics such as loss or accuracy, other important considerations are
\begin{itemize}
    \item Convergence speed
    \item Throughput (e.g. number of rounds, amount of data, or number of devices)
    \item Model fairness, privacy and robustness (see section \ref{subsec:fairness-privacy-robustness})
    \item Resource use on server and clients
\end{itemize}

These goals may be in tension. For instance, maximizing round throughput may introduce bias or hurt accuracy by preferring performant devices with little or no data. Maximizing for low training loss by increasing model complexity will put devices with less memory, many or large examples, or slow CPUs at a disadvantage. Bias or fairness induced in such a way during training may be hard to detect in the evaluation phase since it typically uses the same platform and hence is subject to similar biases.

Various controls affect the above listed indicators. Some are familiar from the datacenter setting, in particular model specific settings and learning algorithm hyperparameters. Others are specific to federated learning:
\begin{itemize}
    \item \textbf{Clients per round}: The minimum number of devices required to complete a round, $M$, and the number of devices required to start a round, $M'$.
    \item \textbf{Server-side scheduling}: In all but the simplest cases, a federated learning system will operate on more than one model at a time: to support multiple tenants; to train models on the same data for different use cases; to support experimentation and architecture or hyper-parameter grid search; and to run training and evaluation workloads concurrently. The server needs to decide which task to serve to incoming devices, an instance of a scheduling problem: assigning work (training or evaluation tasks) to resources (devices). Accordingly, the usual challenges arise: ideal resource assignment should be fair, avoid starvation, minimize wait times, and support relative priorities all at once.
    \item \textbf{Device-side scheduling}: As described in \cref{subsec:systems-system-induced-bias}, various constraints govern when a device can connect to the server and execute work. Within these constraints, various scheduling choices can be made. One extreme is to connect to the server and run computations as often as possible, leading to high load and resource use on both server and devices. Another choice are fixed intervals, but they need to be adjusted to reflect external factors such as number of devices overall and per round. The federated learning system developed at Google aims to strike a balance with a flow control mechanism called \textit{pace steering} \citep{bonawitz19sysml} whereby the server instructs devices when to return. Such a dynamic system enables temporal load balancing for large populations as well as “focusing” connection attempts to specific points in time to reach the threshold $M'$. Developing such a mechanism is difficult due to stochastic and dynamic nature of device availability, the lack of a predictive model of population behavior, and feedback loops.
\end{itemize}

Defining reasonable composite objective functions, and designing algorithms to automatically tune these settings, has not been explored yet in the context of federated learning systems and hence remains a topic of future research.

\subsection{On-Device Runtime}
\label{subsec:systems-on-device-runtime}
While numerous frameworks exist for data center training, the options for training models on resource constrained devices are fairly limited. Machine Learning models and training procedures are typically authored in a high level language such as Python. For federated learning, this description encompasses device and server computations that are executed on the target platform and exchange data over a network connection, necessitating
\begin{itemize}
    \item A means of serializing and dynamically transmitting local pieces of the total computation (e.g., the server-side update to the model, or the local client training procedure).
    \item A means to interpret or execute such a computation on the target platform (server or device).
    \item A stable network protocol for data exchange between participating devices and servers.
\end{itemize}

One extreme form of a representation is the original high-level description, e.g. a Python TensorFlow program \citep{tensorflow2015-whitepaper}. This would require a Python interpreter with TensorFlow backend, which may not be a feasible choice for end-user devices due to resource constraints (binary size, memory use), performance limitations, or security concerns.

Another extreme representation of a computation is machine code of the target architecture, e.g. ARM64 instructions. This requires a compiler or re-implementation of a model in a lower-level language such as C++, and deployment computations will typically be subject to the restrictions that apply to deployment of binary code (see \cref{subsec:systems-platform-development-and-deployment-challenges}), introducing prohibitive latencies for executing novel computations.

Intermediate representations that can be compiled or interpreted with a runtime on the target platform strike a balance between flexibility and efficiency. However, such runtimes are currently not widely available. For instance, Google’s FL system \citep{bonawitz19sysml} relies on TensorFlow for both server and device side execution as well as model and parameter transfer, but this choice suffers from several shortcomings:
\begin{itemize}
    \item It offers no easy path to devices for alternative front ends such as PyTorch \citep{pytorch_NEURIPS2019_9015}, JAX \citep{jax2018github} or CNTK \citep{cntk}.
    \item The runtime is not developed or optimized for resource constrained environments, incurring a large binary size, high memory use and comparatively low performance.
    \item The intermediate representation \texttt{GraphDef} used by TensorFlow is not standardized or stable, and version skew between the frontend and older on-device backends causes frequent compatibility challenges.
\end{itemize}

Other alternatives include more specialized runtimes that support only a subset of the frontend’s capabilities, for instance training specific model types only, requiring changes and long update cycles whenever new model architectures or training algorithms are to be used. An extreme case would be a runtime that is limited and optimized to train a single type of model.

An ideal on-device runtime would have the following characteristics:
\begin{enumerate}
    \item Lightweight: small binary size, or pre-installed; low memory and power profile.
    \item Performant: low startup latency; high throughput, supports hardware acceleration.
    \item Expressive: supports common data types and computations including backpropagation, variables, control flow, custom extensions.
    \item Stable and compact format for expressing data and computations.
    \item Widely available: portable open source implementation.
    \item Targetable by commonly used ML frameworks / languages..
    \item Ideally also supports inference, or if not, building personalized models for an inference runtime.
\end{enumerate}

To our best knowledge no solution exists yet that satisfies these requirements, and we expect the limited ability to run ML training on end user devices to become a hindrance to adoption of federated technologies.

\subsection{The Cross-Silo Setting}
\label{subsec:systems-cross-silo-setting}
The system challenges arising in the scenario of cross-silo federated learning take a considerably different form. As outlined in \cref{tab:characteristics}, clients are fewer in number, more powerful, reliable, and known / addressable, eliminating many of the challenges from the cross-device setting, while allowing for authentication and verification, accounting, and contractually enforced penalties for misbehavior. Nonetheless, there are other sources of heterogeneity, including the features and distribution of data, and possibly the software stack used for training.

While the infrastructure in the cross-device setting (from the device-side data generation to the server logic) is typically operated by one or few organizational entities (the application, operating system, or device manufacturer), in the cross-silo setting, many different entities are involved. This may lead to high coordination and operational cost due to differences in:
\begin{itemize}
    \item \emph{How data is generated, pre-processed and labeled.}  Learning across silos will require data normalization which may be difficult when such data is collected and stored differently (e.g. use of different medical imaging systems, and inconsistencies in labeling procedures, annotations, and storage formats).
    \item \emph{Which software at which version powers training.} Using the same software stack in every silo---possibly delivered alongside the model using container technologies as done by FATE \cite{FATE}---eliminates compatibility concerns, but such frequent and centrally distributed software delivery may not be acceptable to all involved parties. An alternative that is more similar to the cross-device setting would be to standardize data and model formats and communication protocols. See IEEE P3652.1 ``Federated Machine Learning Working Group'' for a related effort in this direction.
    \item \emph{The approval process for how data may or may not be used.} While this process is typically centralized in the cross-device scenario, the situation is likely different in cross-silo settings where many organizational entities are involved, and may be increasingly difficult when training spans different jurisdictions with varying data protection regulations. Technical infrastructure may be of help here by establishing data annotations that encode access policies, and infrastructure enforce them; for instance, limiting the use of certain data to specific models, or encoding minimum aggregation requirements such as ``require at least $M$ clients per round''.
\end{itemize}

Another potential difference in the cross-silo setting is data partitioning: Data in the cross-device setting is typically assumed to be partitioned by examples, all of which have the same features (horizontal partitioning). In the cross-silo setting, in addition to partitioning by examples, partitioning by features is of practical relevance (vertical partitioning). An example would be two organizations, e.g. a bank and a retail company,  with an overlapping set of customers, but different information (features) associated with them. For a discussion focusing on the algorithmic aspects, please see section \ref{ssec:cross-silo}. Learning with feature-partitioned data may require different communication patterns and additional processing steps e.g. for entity alignment and dealing with missing features.

\subsection{Executive Summary}
While production grade systems for cross-device federated learning operate successfully \cite{bonawitz19sysml,apple19wwdc}, various challenges remain:
\begin{itemize}
    \item Frequent and large scale deployment of updates, monitoring, and debugging is challenging (\cref{subsec:systems-platform-development-and-deployment-challenges}).
    \item Differences in device availability induce various forms of bias; defining, quantifying and mitigating them remains a direction for future research (\cref{subsec:systems-system-induced-bias}).
    \item Tuning system parameters is difficult due to the existence of multiple, potentially conflicting objectives (\cref{subsec:systems-system-parameter-tuning}).
    \item Running ML workloads on end user devices is hampered by the lack of a portable, fast, small footprint, and flexible runtime for on-device training (\cref{subsec:systems-on-device-runtime}).
\end{itemize}

Systems for cross-silo settings (\cref{subsec:systems-cross-silo-setting}) face largely different issues owing to differences in the capabilities of compute nodes and the nature of the data being processed.

\pagebreak
\section{Concluding Remarks}

Federated learning enables distributed client devices to collaboratively learn a shared prediction model while keeping all the training data on device, decoupling the ability to do machine learning from the need to store the data in the cloud. This goes beyond the use of local models that make predictions on mobile devices by bringing model training to the device as well.

In recent years, this topic has undergone an explosive growth of interest, both in industry and academia. Major technology companies have already deployed federated learning in production, and a number of startups were founded with the objective of using federated learning to address privacy and data collection challenges in various industries. Further, the breadth of papers surveyed in this work suggests that federated learning is gaining traction in a wide range of interdisciplinary fields: from machine learning to optimization to information theory and statistics to cryptography, fairness, and privacy. 

Motivated by the growing interest in federated learning research, this paper discusses recent advances and presents an extensive collection of open problems and challenges. The system constraints impose efficiency requirements on the algorithms in order to be practical, many of which are not particularly challenging in other settings. We argue that data privacy is not binary and  present a range of threat models that are relevant under a variety of assumptions, each of which provides its own unique challenges.

The open problems discussed in this work are certainly not comprehensive, they reflect the interests and backgrounds of the authors. In particular, we do not discuss any non-learning problems which need to be solved in the course of a practical machine learning project, and might need to be solved based on decentralized data \cite{fablog20}. This can include simple problems such as computing basic descriptive statistics, or more complex objectives such as computing the head of a histogram over an open set \cite{zhu2019federated}. Existing algorithms for solving such problems often do not always have an obvious ``federated version'' that would be efficient under the system assumptions motivating this work or do not admit a useful notion of data protection. Yet another set of important topics that were not discussed are the legal and business issues that may motivate or constrain the use of federated learning. 

We hope this work will be helpful in scoping further research in federated learning and related areas.

\section*{Acknowledgments}

The authors would like to thank Alex Ingerman and David Petrou for their useful suggestions and insightful comments during the review process. 

\pagebreak
\bibliographystyle{plainnat}
\begin{small}
\bibliography{references}

\begin{thebibliography}{511}
\providecommand{\natexlab}[1]{#1}
\providecommand{\url}[1]{\texttt{#1}}
\expandafter\ifx\csname urlstyle\endcsname\relax
  \providecommand{\doi}[1]{doi: #1}\else
  \providecommand{\doi}{doi: \begingroup \urlstyle{rm}\Url}\fi

\bibitem[lat(2020)]{lattigo}
Lattigo 2.0.0.
\newblock Online: \url{http://github.com/ldsec/lattigo}, October 2020.
\newblock EPFL-LDS.

\bibitem[Abadi et~al.(2015)Abadi, Agarwal, Barham, Brevdo, Chen, Citro,
  Corrado, Davis, Dean, Devin, Ghemawat, Goodfellow, Harp, Irving, Isard, Jia,
  Jozefowicz, Kaiser, Kudlur, Levenberg, Man\'{e}, Monga, Moore, Murray, Olah,
  Schuster, Shlens, Steiner, Sutskever, Talwar, Tucker, Vanhoucke, Vasudevan,
  Vi\'{e}gas, Vinyals, Warden, Wattenberg, Wicke, Yu, and
  Zheng]{tensorflow2015-whitepaper}
Mart\'{\i}n Abadi, Ashish Agarwal, Paul Barham, Eugene Brevdo, Zhifeng Chen,
  Craig Citro, Greg~S. Corrado, Andy Davis, Jeffrey Dean, Matthieu Devin,
  Sanjay Ghemawat, Ian Goodfellow, Andrew Harp, Geoffrey Irving, Michael Isard,
  Yangqing Jia, Rafal Jozefowicz, Lukasz Kaiser, Manjunath Kudlur, Josh
  Levenberg, Dandelion Man\'{e}, Rajat Monga, Sherry Moore, Derek Murray, Chris
  Olah, Mike Schuster, Jonathon Shlens, Benoit Steiner, Ilya Sutskever, Kunal
  Talwar, Paul Tucker, Vincent Vanhoucke, Vijay Vasudevan, Fernanda Vi\'{e}gas,
  Oriol Vinyals, Pete Warden, Martin Wattenberg, Martin Wicke, Yuan Yu, and
  Xiaoqiang Zheng.
\newblock {TensorFlow}: Large-scale machine learning on heterogeneous systems,
  2015.
\newblock URL \url{https://www.tensorflow.org/}.
\newblock Software available from tensorflow.org.

\bibitem[Abadi et~al.(2016)Abadi, Chu, Goodfellow, McMahan, Mironov, Talwar,
  and Zhang]{abadi2016deep}
Martin Abadi, Andy Chu, Ian Goodfellow, H~Brendan McMahan, Ilya Mironov, Kunal
  Talwar, and Li~Zhang.
\newblock Deep learning with differential privacy.
\newblock In \emph{Proceedings of the 2016 ACM SIGSAC Conference on Computer
  and Communications Security}, pages 308--318. ACM, 2016.

\bibitem[Abari et~al.(2016)Abari, Rahul, and Katabi]{AbariRK16}
Omid Abari, Hariharan Rahul, and Dina Katabi.
\newblock Over-the-air function computation in sensor networks.
\newblock \emph{CoRR}, abs/1612.02307, 2016.
\newblock URL \url{http://arxiv.org/abs/1612.02307}.

\bibitem[Abay et~al.(2018)Abay, Zhou, Kantarcioglu, Thuraisingham, and
  Sweeney]{abay2018privacy}
Nazmiye~Ceren Abay, Yan Zhou, Murat Kantarcioglu, Bhavani Thuraisingham, and
  Latanya Sweeney.
\newblock Privacy preserving synthetic data release using deep learning.
\newblock In \emph{Joint European Conference on Machine Learning and Knowledge
  Discovery in Databases}, pages 510--526. Springer, 2018.

\bibitem[Abowd and Schmutte(2019)]{AS19}
John~M Abowd and Ian~M Schmutte.
\newblock An economic analysis of privacy protection and statistical accuracy
  as social choices.
\newblock \emph{American Economic Review}, 109\penalty0 (1):\penalty0 171--202,
  2019.

\bibitem[Acharya et~al.(2020)Acharya, Canonne, and Tyagi]{acharya2018}
Jayadev Acharya, Cl{\'e}ment~L Canonne, and Himanshu Tyagi.
\newblock Inference under information constraints i: Lower bounds from
  chi-square contraction.
\newblock \emph{IEEE Transactions on Information Theory}, 66\penalty0
  (12):\penalty0 7835--7855, 2020.

\bibitem[\'{A}cs and Castelluccia(2011)]{Acs:2011:IDD:2042445.2042457}
Gergely \'{A}cs and Claude Castelluccia.
\newblock I have a {DREAM}!: {DI}fferentially {P}rivat{E} smart {M}etering.
\newblock In \emph{Proceedings of the 13th International Conference on
  Information Hiding}, IH'11, pages 118--132, Berlin, Heidelberg, 2011.
  Springer-Verlag.
\newblock ISBN 978-3-642-24177-2.
\newblock URL \url{http://dl.acm.org/citation.cfm?id=2042445.2042457}.

\bibitem[Agarwal et~al.(2018)Agarwal, Suresh, Yu, Kumar, and
  McMahan]{agarwal2018cpsgd}
Naman Agarwal, Ananda~Theertha Suresh, Felix~X. Yu, Sanjiv Kumar, and Brendan
  McMahan.
\newblock cp{SGD}: Communication-efficient and differentially-private
  distributed {SGD}.
\newblock In \emph{Advances in Neural Information Processing Systems}, pages
  7564--7575, 2018.

\bibitem[Agrawal et~al.(2019)Agrawal, Shamsabadi, Kusner, and
  Gasc{\'{o}}n]{quotient}
Nitin Agrawal, Ali~Shahin Shamsabadi, Matt~J. Kusner, and Adri{\`{a}}
  Gasc{\'{o}}n.
\newblock {QUOTIENT:} two-party secure neural network training and prediction.
\newblock In \emph{In Proceedings of the {ACM} Conference on Computer and
  Communication Security (CCS)}, 2019.

\bibitem[Agrawal and Srikant(2000)]{agrawal2000}
Rakesh Agrawal and Ramakrishnan Srikant.
\newblock Privacy-preserving data mining.
\newblock In \emph{ACM SIGMOD International Conference on Management of Data},
  2000.

\bibitem[Aguilar-Melchor and Gaborit(2007)]{aguilar2007lattice}
Carlos Aguilar-Melchor and Philippe Gaborit.
\newblock A lattice-based computationally-efficient private information
  retrieval protocol.
\newblock \emph{Cryptol. ePrint Arch., Report}, 446, 2007.

\bibitem[Aguilar-Melchor et~al.(2016)Aguilar-Melchor, Barrier, Fousse, and
  Killijian]{aguilar2016xpir}
Carlos Aguilar-Melchor, Joris Barrier, Laurent Fousse, and Marc-Olivier
  Killijian.
\newblock {XPIR}: Private information retrieval for everyone.
\newblock \emph{Proceedings on Privacy Enhancing Technologies}, 2016\penalty0
  (2):\penalty0 155--174, 2016.

\bibitem[{ai.google}(2018)]{googleai18settingssearch}
{ai.google}.
\newblock Under the hood of the {P}ixel 2: How {AI} is supercharging hardware,
  2018.
\newblock URL \url{https://ai.google/stories/ai-in-hardware/}.
\newblock Retrieved Nov 2018.

\bibitem[{ai.intel}(2019)]{intel19medicalimaging}
{ai.intel}.
\newblock Federated learning for medical imaging, 2019.
\newblock URL
  \url{https://www.intel.ai/federated-learning-for-medical-imaging/}.
\newblock Retrieved Aug 2019.

\bibitem[Ali et~al.(2019)Ali, Lepoint, Patel, Raykova, Schoppmann, Seth, and
  Yeo]{DBLP:journals/iacr/AliLP0SSY19}
Asra Ali, Tancr{\`{e}}de Lepoint, Sarvar Patel, Mariana Raykova, Phillipp
  Schoppmann, Karn Seth, and Kevin Yeo.
\newblock Communication-computation trade-offs in {PIR}.
\newblock \emph{{IACR} Cryptol. ePrint Arch.}, 2019:\penalty0 1483, 2019.

\bibitem[Alistarh et~al.(2017)Alistarh, Grubic, Li, Tomioka, and
  Vojnovic]{alistarh2017qsgd}
Dan Alistarh, Demjan Grubic, Jerry Li, Ryota Tomioka, and Milan Vojnovic.
\newblock {QSGD}: Communication-efficient {SGD} via gradient quantization and
  encoding.
\newblock In \emph{NIPS - Advances in Neural Information Processing Systems},
  pages 1709--1720, 2017.

\bibitem[Alistarh et~al.(2018)Alistarh, Allen-Zhu, and
  Li]{alistarh2018byzantine}
Dan Alistarh, Zeyuan Allen-Zhu, and Jerry Li.
\newblock Byzantine stochastic gradient descent.
\newblock In \emph{NIPS}, 2018.

\bibitem[Almeida and Xavier(2018)]{Almeida2018}
Inês Almeida and João Xavier.
\newblock {DJAM: Distributed Jacobi Asynchronous Method for Learning Personal
  Models}.
\newblock \emph{IEEE Signal Processing Letters}, 25\penalty0 (9):\penalty0
  1389--1392, 2018.

\bibitem[Ames et~al.(2017)Ames, Hazay, Ishai, and
  Venkitasubramaniam]{Ames:2017:LLS:3133956.3134104}
Scott Ames, Carmit Hazay, Yuval Ishai, and Muthuramakrishnan
  Venkitasubramaniam.
\newblock Ligero: Lightweight sublinear arguments without a trusted setup.
\newblock In \emph{Proceedings of the 2017 ACM SIGSAC Conference on Computer
  and Communications Security}, CCS '17, 2017.

\bibitem[Amin et~al.(2019)Amin, Kulesza, Munoz, and
  Vassilvtiskii]{amin2019bounding}
Kareem Amin, Alex Kulesza, Andres Munoz, and Sergei Vassilvtiskii.
\newblock Bounding user contributions: A bias-variance trade-off in
  differential privacy.
\newblock In \emph{International Conference on Machine Learning}, pages
  263--271, 2019.

\bibitem[androidtrusty()]{AndroidTrusty}
androidtrusty.
\newblock {Android Trusty TEE}.
\newblock \url{https://source.android.com/security/trusty}, 2019.
\newblock Accessed: 2019-12-05.

\bibitem[Angel et~al.(2018)Angel, Chen, Laine, and
  Setty]{DBLP:conf/sp/AngelCLS18}
Sebastian Angel, Hao Chen, Kim Laine, and Srinath T.~V. Setty.
\newblock {PIR} with compressed queries and amortized query processing.
\newblock In \emph{{IEEE} Symposium on Security and Privacy}, pages 962--979.
  {IEEE} Computer Society, 2018.

\bibitem[Annas(2003)]{annas2003hipaa}
George~J Annas.
\newblock {HIPAA} regulations-a new era of medical-record privacy?
\newblock \emph{New England Journal of Medicine}, 348\penalty0 (15):\penalty0
  1486--1490, 2003.

\bibitem[Apple(2019{\natexlab{a}})]{apple19neurips}
Apple.
\newblock {Private Federated Learning (NeurIPS 2019 Expo Talk Abstract)}.
\newblock \url{https://nips.cc/ExpoConferences/2019/schedule?talk_id=40},
  2019{\natexlab{a}}.

\bibitem[Apple(2019{\natexlab{b}})]{apple19wwdc}
Apple.
\newblock Designing for privacy (video and slide deck).
\newblock Apple WWDC,
  \url{https://developer.apple.com/videos/play/wwdc2019/708},
  2019{\natexlab{b}}.

\bibitem[Araki et~al.(2016)Araki, Furukawa, Lindell, Nof, and
  Ohara]{araki2016high}
Toshinori Araki, Jun Furukawa, Yehuda Lindell, Ariel Nof, and Kazuma Ohara.
\newblock High-throughput semi-honest secure three-party computation with an
  honest majority.
\newblock In \emph{Proceedings of the 2016 ACM SIGSAC Conference on Computer
  and Communications Security}, pages 805--817. ACM, 2016.

\bibitem[armtrustzone()]{ArmTrustzone}
armtrustzone.
\newblock {Arm TrustZone Technology}.
\newblock \url{https://developer.arm.com/ip-products/security-ip/trustzone},
  2019.
\newblock Accessed: 2019-12-05.

\bibitem[Assran et~al.(2019)Assran, Loizou, Ballas, and
  Rabbat]{assran2019stochastic}
Mahmoud Assran, Nicolas Loizou, Nicolas Ballas, and Michael Rabbat.
\newblock Stochastic gradient push for distributed deep learning.
\newblock In \emph{ICML}, 2019.

\bibitem[Athalye et~al.(2018)Athalye, Carlini, and
  Wagner]{athalye2018obfuscated}
Anish Athalye, Nicholas Carlini, and David Wagner.
\newblock Obfuscated gradients give a false sense of security: Circumventing
  defenses to adversarial examples.
\newblock \emph{ICML}, 2018.

\bibitem[Augenstein et~al.(2019)Augenstein, McMahan, Ramage, Ramaswamy,
  Kairouz, Chen, Mathews, and y~Arcas]{augenstein2019generative}
Sean Augenstein, H.~Brendan McMahan, Daniel Ramage, Swaroop Ramaswamy, Peter
  Kairouz, Mingqing Chen, Rajiv Mathews, and Blaise~Aguera y~Arcas.
\newblock Generative models for effective {ML} on private, decentralized
  datasets, 2019.
\newblock URL \url{https://arxiv.org/abs/1911.06679}.

\bibitem[Authors(2020{\natexlab{a}})]{PyVertical}
PyVertical Authors.
\newblock Pyvertical, 2020{\natexlab{a}}.
\newblock URL \url{https://github.com.cnpmjs.org/OpenMined/PyVertical}.

\bibitem[Authors(2019{\natexlab{a}})]{FATE}
The~FATE Authors.
\newblock Federated {AI} technology enabler, 2019{\natexlab{a}}.
\newblock URL \url{https://www.fedai.org/}.

\bibitem[Authors(2020{\natexlab{b}})]{Fedlearner}
The~Fedlearner Authors.
\newblock Fedlearner, 2020{\natexlab{b}}.
\newblock URL \url{https://github.com/bytedance/fedlearner}.

\bibitem[Authors(2019{\natexlab{b}})]{Leaf}
The~Leaf Authors.
\newblock Leaf, 2019{\natexlab{b}}.
\newblock URL \url{https://leaf.cmu.edu/}.

\bibitem[Authors(2019{\natexlab{c}})]{PaddleFL}
The~PaddleFL Authors.
\newblock Paddle{FL}, 2019{\natexlab{c}}.
\newblock URL \url{https://github.com/PaddlePaddle/PaddleFL}.

\bibitem[Authors(2019{\natexlab{d}})]{PaddlePaddle}
The~PaddlePaddle Authors.
\newblock Paddle{P}addle, 2019{\natexlab{d}}.
\newblock URL \url{http://www.paddlepaddle.org/}.

\bibitem[Authors(2019{\natexlab{e}})]{tff}
The~TFF Authors.
\newblock Tensor{F}low {F}ederated, 2019{\natexlab{e}}.
\newblock URL \url{https://www.tensorflow.org/federated}.

\bibitem[Avent et~al.(01 Oct. 2020)Avent, Dubey, and Korolova]{dubey2018hybrid}
Brendan Avent, Yatharth Dubey, and Aleksandra Korolova.
\newblock The power of the hybrid model for mean estimation.
\newblock \emph{Proceedings on Privacy Enhancing Technologies (PETS)},
  2020\penalty0 (4):\penalty0 48 -- 68, 01 Oct. 2020.
\newblock \doi{https://doi.org/10.2478/popets-2020-0062}.
\newblock URL
  \url{https://content.sciendo.com/view/journals/popets/2020/4/article-p48.xml}.

\bibitem[Avent et~al.(2017)Avent, Korolova, Zeber, Hovden, and
  Livshits]{avent2017blender}
Brendan Avent, Aleksandra Korolova, David Zeber, Torgeir Hovden, and Benjamin
  Livshits.
\newblock {BLENDER}: Enabling local search with a hybrid differential privacy
  model.
\newblock In \emph{26th {USENIX} Security Symposium ({USENIX} Security 17)},
  pages 747--764, Vancouver, BC, August 2017. {USENIX} Association.
\newblock ISBN 978-1-931971-40-9.
\newblock URL
  \url{https://www.usenix.org/conference/usenixsecurity17/technical-sessions/presentation/avent}.

\bibitem[Awasthi et~al.(2020)Awasthi, Cortes, Mansour, and
  Mohri]{AwasthiCortesMansourMohri2020}
Pranjal Awasthi, Corinna Cortes, Yishay Mansour, and Mehryar Mohri.
\newblock Beyond individual and group fairness.
\newblock \emph{CoRR}, abs/2008.09490, 2020.

\bibitem[Babai et~al.(1991)Babai, Fortnow, Levin, and
  Szegedy]{DBLP:conf/stoc/BabaiFLS91}
L{\'{a}}szl{\'{o}} Babai, Lance Fortnow, Leonid~A. Levin, and Mario Szegedy.
\newblock Checking computations in polylogarithmic time.
\newblock In \emph{{STOC}}, pages 21--31. {ACM}, 1991.

\bibitem[Bagdasaryan and Shmatikov(2019)]{bagdasaryan2019disparate}
Eugene Bagdasaryan and Vitaly Shmatikov.
\newblock Differential privacy has disparate impact on model accuracy.
\newblock \emph{CoRR}, abs/1905.12101, 2019.
\newblock URL \url{http://arxiv.org/abs/1905.12101}.

\bibitem[Bagdasaryan et~al.(2018)Bagdasaryan, Veit, Hua, Estrin, and
  Shmatikov]{bagdasaryan18backdoor}
Eugene Bagdasaryan, Andreas Veit, Yiqing Hua, Deborah Estrin, and Vitaly
  Shmatikov.
\newblock How to backdoor federated learning.
\newblock \emph{arXiv preprint arXiv:1807.00459}, 2018.

\bibitem[Balle et~al.(2019)Balle, Bell, Gasc{\'{o}}n, and Nissim]{BalleBGN19}
Borja Balle, James Bell, Adri{\`{a}} Gasc{\'{o}}n, and Kobbi Nissim.
\newblock The privacy blanket of the shuffle model.
\newblock In \emph{Advances in Cryptology - {CRYPTO} 2019 - 39th Annual
  International Cryptology Conference, Santa Barbara, CA, USA, August 18-22,
  2019, Proceedings, Part {II}}, pages 638--667, 2019.
\newblock \doi{10.1007/978-3-030-26951-7\_22}.
\newblock URL \url{https://doi.org/10.1007/978-3-030-26951-7\_22}.

\bibitem[Balle et~al.(2020{\natexlab{a}})Balle, Bell, Gasc\'{o}n, and
  Nissim]{balle2020}
Borja Balle, James Bell, Adri\`{a} Gasc\'{o}n, and Kobbi Nissim.
\newblock Private summation in the multi-message shuffle model.
\newblock In \emph{Proceedings of the 2020 ACM SIGSAC Conference on Computer
  and Communications Security}, page 657–676. ACM, 2020{\natexlab{a}}.

\bibitem[Balle et~al.(2020{\natexlab{b}})Balle, Kairouz, McMahan, Thakkar, and
  Thakurta]{balle2020privacy}
Borja Balle, Peter Kairouz, H.~Brendan McMahan, Om~Thakkar, and Abhradeep
  Thakurta.
\newblock Privacy amplification via random check-ins, 2020{\natexlab{b}}.

\bibitem[Barak et~al.(2019)Barak, Escudero, Dalskov, and Keller]{bedkSE}
Assi Barak, Daniel Escudero, Anders P.~K. Dalskov, and Marcel Keller.
\newblock Secure evaluation of quantized neural networks.
\newblock \emph{{IACR} Cryptology ePrint Archive}, 2019:\penalty0 131, 2019.
\newblock URL \url{https://eprint.iacr.org/2019/131}.

\bibitem[Barnes et~al.(2020{\natexlab{a}})Barnes, Han, and Ozgur]{barnes2019}
Leighton~Pate Barnes, Yanjun Han, and Ayfer Ozgur.
\newblock Lower bounds for learning distributions under communication
  constraints via fisher information.
\newblock \emph{Journal of Machine Learning Research}, 21\penalty0
  (236):\penalty0 1--30, 2020{\natexlab{a}}.
\newblock URL \url{http://jmlr.org/papers/v21/19-737.html}.

\bibitem[Barnes et~al.(2020{\natexlab{b}})Barnes, Inan, Isik, and
  Ozgur]{Barnes2020rtopk}
Leighton~Pate Barnes, Huseyin~A. Inan, Berivan Isik, and Ayfer Ozgur.
\newblock rtop-k: A statistical estimation approach to distributed sgd.
\newblock \emph{arXiv preprint arXiv:2005.10761}, 2020{\natexlab{b}}.

\bibitem[Barocas et~al.(2019)Barocas, Hardt, and Narayanan]{barocasfairness}
Solon Barocas, Moritz Hardt, and Arvind Narayanan.
\newblock \emph{Fairness and Machine Learning}.
\newblock fairmlbook.org, 2019.
\newblock \url{http://www.fairmlbook.org}.

\bibitem[Baruch et~al.(2019)Baruch, Baruch, and Goldberg]{baruch2019little}
Moran Baruch, Gilad Baruch, and Yoav Goldberg.
\newblock A little is enough: Circumventing defenses for distributed learning.
\newblock \emph{arXiv preprint arXiv:1902.06156}, 2019.

\bibitem[Bassily and Smith(2015)]{bassily2015local}
Raef Bassily and Adam Smith.
\newblock Local, private, efficient protocols for succinct histograms.
\newblock In \emph{STOC}, pages 127--135, 2015.

\bibitem[Bassily et~al.(2017)Bassily, Stemmer, Thakurta,
  et~al.]{bassily2017practical}
Raef Bassily, Uri Stemmer, Abhradeep~Guha Thakurta, et~al.
\newblock Practical locally private heavy hitters.
\newblock In \emph{Advances in Neural Information Processing Systems}, pages
  2288--2296, 2017.

\bibitem[Basu et~al.(2020)Basu, Data, Karakus, and Diggavi]{basu2020qsparse}
Debraj Basu, Deepesh Data, Can Karakus, and Suhas~N Diggavi.
\newblock Qsparse-local-sgd: Distributed sgd with quantization, sparsification,
  and local computations.
\newblock \emph{IEEE Journal on Selected Areas in Information Theory},
  1\penalty0 (1):\penalty0 217--226, 2020.

\bibitem[Baxter(2000)]{baxter00model}
Jonathan Baxter.
\newblock A model of inductive bias learning.
\newblock \emph{Journal of Artificial Intelligence Research}, 12:\penalty0
  149--198, 2000.

\bibitem[Beimel et~al.(2020)Beimel, Korolova, Nissim, Sheffet, and
  Stemmer]{beimel2019power}
Amos Beimel, Aleksandra Korolova, Kobbi Nissim, Or~Sheffet, and Uri Stemmer.
\newblock The power of synergy in differential privacy: Combining a small
  curator with local randomizers.
\newblock In \emph{Conference on Information-Theoretic Cryptography (ITC)},
  2020.
\newblock URL \url{https://arxiv.org/abs/1912.08951}.

\bibitem[Bell et~al.(2020)Bell, Bonawitz, Gasc\'{o}n, Lepoint, and
  Raykova]{bell20secagg}
James~Henry Bell, Kallista~A. Bonawitz, Adri\`{a} Gasc\'{o}n, Tancr\`{e}de
  Lepoint, and Mariana Raykova.
\newblock Secure single-server aggregation with (poly)logarithmic overhead.
\newblock In \emph{Proceedings of the 2020 ACM SIGSAC Conference on Computer
  and Communications Security}, page 1253–1269. ACM, 2020.

\bibitem[Bellet et~al.(2018)Bellet, Guerraoui, Taziki, and
  Tommasi]{Bellet2018a}
Aur\'elien Bellet, Rachid Guerraoui, Mahsa Taziki, and Marc Tommasi.
\newblock {P}ersonalized and {P}rivate {P}eer-to-{P}eer {M}achine {L}earning.
\newblock In \emph{{AISTATS}}, 2018.

\bibitem[Bello et~al.(2017)Bello, Zoph, Vasudevan, and Le]{Bello2016Neural}
Irwan Bello, Barret Zoph, Vijay Vasudevan, and Quoc~V Le.
\newblock Neural optimizer search with reinforcement learning.
\newblock In \emph{Proceedings of the 34th International Conference on Machine
  Learning-Volume 70}, pages 459--468. JMLR. org, 2017.

\bibitem[Ben-David et~al.(2010)Ben-David, Blitzer, Crammer, Kulesza, Pereira,
  and Vaughan]{ben2010theory}
Shai Ben-David, John Blitzer, Koby Crammer, Alex Kulesza, Fernando Pereira, and
  Jennifer~Wortman Vaughan.
\newblock A theory of learning from different domains.
\newblock \emph{Machine learning}, 79\penalty0 (1-2):\penalty0 151--175, 2010.

\bibitem[Ben{-}Sasson et~al.(2014)Ben{-}Sasson, Chiesa, Garman, Green, Miers,
  Tromer, and Virza]{DBLP:conf/sp/Ben-SassonCG0MTV14}
Eli Ben{-}Sasson, Alessandro Chiesa, Christina Garman, Matthew Green, Ian
  Miers, Eran Tromer, and Madars Virza.
\newblock Zerocash: Decentralized anonymous payments from bitcoin.
\newblock In \emph{{IEEE} Symposium on Security and Privacy}, pages 459--474.
  {IEEE} Computer Society, 2014.

\bibitem[Ben{-}Sasson et~al.(2019)Ben{-}Sasson, Bentov, Horesh, and
  Riabzev]{DBLP:conf/crypto/Ben-SassonBHR19}
Eli Ben{-}Sasson, Iddo Bentov, Yinon Horesh, and Michael Riabzev.
\newblock Scalable zero knowledge with no trusted setup.
\newblock In \emph{{CRYPTO} {(3)}}, volume 11694 of \emph{Lecture Notes in
  Computer Science}, pages 701--732. Springer, 2019.

\bibitem[Bergstra et~al.(2011)Bergstra, Bardenet, Bengio, and
  K{\'e}gl]{bergstra2011algorithms}
James~S Bergstra, R{\'e}mi Bardenet, Yoshua Bengio, and Bal{\'a}zs K{\'e}gl.
\newblock Algorithms for hyper-parameter optimization.
\newblock In \emph{Advances in Neural Information Processing Systems}, pages
  2546--2554, 2011.

\bibitem[Bertrán et~al.(2019)Bertrán, Martínez, Papadaki, Qiu, Rodrigues,
  Reeves, and Sapiro]{Bertran2019}
Martín Bertrán, Natalia Martínez, Afroditi Papadaki, Qiang Qiu, Miguel R.~D.
  Rodrigues, Galen Reeves, and Guillermo Sapiro.
\newblock Learning adversarially fair and transferable representations.
\newblock In \emph{ICML}, 2019.

\bibitem[Beutel et~al.(2020)Beutel, Topal, Mathur, Qiu, Parcollet, and
  Lane]{beutel2020flower}
Daniel~J. Beutel, Taner Topal, Akhil Mathur, Xinchi Qiu, Titouan Parcollet, and
  Nicholas~D. Lane.
\newblock Flower: A friendly federated learning research framework, 2020.

\bibitem[Bhagoji et~al.(2019)Bhagoji, Chakraborty, Mittal, and
  Calo]{pmlr-v97-bhagoji19a}
Arjun~Nitin Bhagoji, Supriyo Chakraborty, Prateek Mittal, and Seraphin Calo.
\newblock Analyzing federated learning through an adversarial lens.
\newblock In \emph{Proceedings of the 36th International Conference on Machine
  Learning}, pages 634--643, 2019.

\bibitem[Bhowmick et~al.(2018)Bhowmick, Duchi, Freudiger, Kapoor, and
  Rogers]{bhowmick2018protection}
Abhishek Bhowmick, John Duchi, Julien Freudiger, Gaurav Kapoor, and Ryan
  Rogers.
\newblock Protection against reconstruction and its applications in private
  federated learning.
\newblock \emph{arXiv preprint arXiv:1812.00984}, 2018.

\bibitem[Biggio et~al.(2012)Biggio, Nelson, and
  Laskov]{Biggio:2012:PAA:3042573.3042761}
Battista Biggio, Blaine Nelson, and Pavel Laskov.
\newblock Poisoning attacks against support vector machines.
\newblock In \emph{Proceedings of the 29th International Coference on
  International Conference on Machine Learning}, ICML'12, pages 1467--1474,
  USA, 2012. Omnipress.
\newblock ISBN 978-1-4503-1285-1.
\newblock URL \url{http://dl.acm.org/citation.cfm?id=3042573.3042761}.

\bibitem[Biggio et~al.(2013)Biggio, Corona, Maiorca, Nelson, {\v{S}}rndi{\'c},
  Laskov, Giacinto, and Roli]{biggio2013evasion}
Battista Biggio, Igino Corona, Davide Maiorca, Blaine Nelson, Nedim
  {\v{S}}rndi{\'c}, Pavel Laskov, Giorgio Giacinto, and Fabio Roli.
\newblock Evasion attacks against machine learning at test time.
\newblock In \emph{ECML-PKDD}, pages 387--402. Springer, 2013.

\bibitem[Bitansky et~al.(2012)Bitansky, Canetti, Chiesa, and
  Tromer]{Bitansky:2012:ECR:2090236.2090263}
Nir Bitansky, Ran Canetti, Alessandro Chiesa, and Eran Tromer.
\newblock From extractable collision resistance to succinct non-interactive
  arguments of knowledge, and back again.
\newblock In \emph{Proceedings of the 3rd Innovations in Theoretical Computer
  Science Conference}, ITCS '12, 2012.

\bibitem[{Bitar} and {Rouayheb}(2018)]{staircasepir}
R.~{Bitar} and S.~E. {Rouayheb}.
\newblock Staircase-{PIR}: Universally robust private information retrieval.
\newblock In \emph{2018 IEEE Information Theory Workshop (ITW)}, pages 1--5,
  Nov 2018.
\newblock \doi{10.1109/ITW.2018.8613532}.

\bibitem[Bittau et~al.(2017)Bittau, Erlingsson, Maniatis, Mironov, Raghunathan,
  Lie, Rudominer, Kode, Tinnes, and Seefeld]{prochlo}
Andrea Bittau, \'{U}lfar Erlingsson, Petros Maniatis, Ilya Mironov, Ananth
  Raghunathan, David Lie, Mitch Rudominer, Ushasree Kode, Julien Tinnes, and
  Bernhard Seefeld.
\newblock Prochlo: Strong privacy for analytics in the crowd.
\newblock In \emph{Proceedings of the 26th Symposium on Operating Systems
  Principles}, SOSP '17, pages 441--459, New York, NY, USA, 2017. ACM.
\newblock ISBN 978-1-4503-5085-3.
\newblock \doi{10.1145/3132747.3132769}.
\newblock URL \url{http://doi.acm.org/10.1145/3132747.3132769}.

\bibitem[Blalock et~al.(2020)Blalock, Ortiz, Frankle, and
  Guttag]{blalock2020state}
Davis Blalock, Jose Javier~Gonzalez Ortiz, Jonathan Frankle, and John Guttag.
\newblock What is the state of neural network pruning?
\newblock \emph{arXiv preprint arXiv:2003.03033}, 2020.

\bibitem[Blanchard et~al.(2017{\natexlab{a}})Blanchard, El~Mhamdi, Guerraoui,
  and Stainer]{peva17}
Peva Blanchard, El~Mahdi El~Mhamdi, Rachid Guerraoui, and Julien Stainer.
\newblock Machine learning with adversaries: {B}yzantine tolerant gradient
  descent.
\newblock In \emph{Advances in Neural Information Processing Systems},
  2017{\natexlab{a}}.

\bibitem[Blanchard et~al.(2017{\natexlab{b}})Blanchard, Guerraoui, Stainer,
  et~al.]{blanchard2017machine}
Peva Blanchard, Rachid Guerraoui, Julien Stainer, et~al.
\newblock {Machine Learning with Adversaries: {B}yzantine Tolerant Gradient
  Descent}.
\newblock In \emph{Advances in Neural Information Processing Systems}, pages
  118--128, 2017{\natexlab{b}}.

\bibitem[Bogdanov et~al.(2012)Bogdanov, Talviste, and
  Willemson]{DBLP:conf/fc/BogdanovTW12}
Dan Bogdanov, Riivo Talviste, and Jan Willemson.
\newblock Deploying secure multi-party computation for financial data analysis
  - (short paper).
\newblock In \emph{Financial Cryptography}, volume 7397 of \emph{Lecture Notes
  in Computer Science}, pages 57--64. Springer, 2012.

\bibitem[Bogetoft et~al.(2009)Bogetoft, Christensen, Damg{\aa}rd, Geisler,
  Jakobsen, Kr{\o}igaard, Nielsen, Nielsen, Nielsen, Pagter, Schwartzbach, and
  Toft]{DBLP:conf/fc/BogetoftCDGJKNNNPST09}
Peter Bogetoft, Dan~Lund Christensen, Ivan Damg{\aa}rd, Martin Geisler,
  Thomas~P. Jakobsen, Mikkel Kr{\o}igaard, Janus~Dam Nielsen, Jesper~Buus
  Nielsen, Kurt Nielsen, Jakob Pagter, Michael~I. Schwartzbach, and Tomas Toft.
\newblock Secure multiparty computation goes live.
\newblock In \emph{Financial Cryptography}, volume 5628 of \emph{Lecture Notes
  in Computer Science}, pages 325--343. Springer, 2009.

\bibitem[Bonawitz et~al.(2016)Bonawitz, Ivanov, Kreuter, Marcedone, McMahan,
  Patel, Ramage, Segal, and Seth]{bonawitz2016practical}
K.~A. Bonawitz, Vladimir Ivanov, Ben Kreuter, Antonio Marcedone, H.~Brendan
  McMahan, Sarvar Patel, Daniel Ramage, Aaron Segal, and Karn Seth.
\newblock Practical secure aggregation for federated learning on user-held
  data.
\newblock \emph{arXiv preprint arXiv:1611.04482}, 2016.

\bibitem[Bonawitz et~al.(2017)Bonawitz, Ivanov, Kreuter, Marcedone, McMahan,
  Patel, Ramage, Segal, and Seth]{bonawitz17secagg}
K.~A. Bonawitz, Vladimir Ivanov, Ben Kreuter, Antonio Marcedone, H~Brendan
  McMahan, Sarvar Patel, Daniel Ramage, Aaron Segal, and Karn Seth.
\newblock Practical secure aggregation for privacy-preserving machine learning.
\newblock In \emph{Proceedings of the 2017 ACM SIGSAC Conference on Computer
  and Communications Security}, pages 1175--1191. ACM, 2017.

\bibitem[Bonawitz et~al.(2019{\natexlab{a}})Bonawitz, Eichner, Grieskamp, Huba,
  Ingerman, Ivanov, Kiddon, Kone{\v{c}}n{\'y}, Mazzocchi, McMahan, Overveldt,
  Petrou, Ramage, and Roselander]{bonawitz19sysml}
K.~A. Bonawitz, Hubert Eichner, Wolfgang Grieskamp, Dzmitry Huba, Alex
  Ingerman, Vladimir Ivanov, Chloé~M Kiddon, Jakub Kone{\v{c}}n{\'y}, Stefano
  Mazzocchi, Brendan McMahan, Timon~Van Overveldt, David Petrou, Daniel Ramage,
  and Jason Roselander.
\newblock Towards federated learning at scale: System design.
\newblock In \emph{SysML 2019}, 2019{\natexlab{a}}.
\newblock URL \url{https://arxiv.org/abs/1902.01046}.

\bibitem[Bonawitz et~al.(2019{\natexlab{b}})Bonawitz, Salehi,
  Kone{\v{c}}n{\'y}, McMahan, and Gruteser]{bonawitz2019autotune}
K.~A. Bonawitz, Fariborz Salehi, Jakub Kone{\v{c}}n{\'y}, Brendan McMahan, and
  Marco Gruteser.
\newblock Federated learning with autotuned communication-efficient secure
  aggregation.
\newblock In \emph{2019 53nd Asilomar Conference on Signals, Systems, and
  Computers}. IEEE, 2019{\natexlab{b}}.

\bibitem[Boneh et~al.(2019)Boneh, Boyle, Corrigan{-}Gibbs, Gilboa, and
  Ishai]{DBLP:conf/crypto/BonehBCGI19}
Dan Boneh, Elette Boyle, Henry Corrigan{-}Gibbs, Niv Gilboa, and Yuval Ishai.
\newblock Zero-knowledge proofs on secret-shared data via fully linear {PCP}s.
\newblock In \emph{{CRYPTO} {(3)}}, volume 11694 of \emph{Lecture Notes in
  Computer Science}, pages 67--97. Springer, 2019.

\bibitem[Bourse et~al.(2018)Bourse, Minelli, Minihold, and
  Paillier]{DBLP:conf/crypto/BourseMMP18}
Florian Bourse, Michele Minelli, Matthias Minihold, and Pascal Paillier.
\newblock Fast homomorphic evaluation of deep discretized neural networks.
\newblock In \emph{{CRYPTO} {(3)}}, volume 10993 of \emph{Lecture Notes in
  Computer Science}, pages 483--512. Springer, 2018.

\bibitem[Boyd et~al.(2006)Boyd, Ghosh, Prabhakar, and Shah]{Boyd2006}
Stephen Boyd, Arpita Ghosh, Balaji Prabhakar, and Devavrat Shah.
\newblock Randomized gossip algorithms.
\newblock \emph{IEEE Transactions on Information Theory}, 52\penalty0
  (6):\penalty0 2508--2530, 2006.

\bibitem[Bradbury et~al.(2018)Bradbury, Frostig, Hawkins, Johnson, Leary,
  Maclaurin, Necula, Paszke, Vander{P}las, Wanderman-{M}ilne, and
  Zhang]{jax2018github}
James Bradbury, Roy Frostig, Peter Hawkins, Matthew~James Johnson, Chris Leary,
  Dougal Maclaurin, George Necula, Adam Paszke, Jake Vander{P}las, Skye
  Wanderman-{M}ilne, and Qiao Zhang.
\newblock {JAX}: composable transformations of {P}ython+{N}um{P}y programs,
  2018.
\newblock URL \url{http://github.com/google/jax}.

\bibitem[Brakerski(2012)]{BFV1}
Zvika Brakerski.
\newblock Fully homomorphic encryption without modulus switching from classical
  gapsvp.
\newblock In \emph{{CRYPTO}}, volume 7417 of \emph{Lecture Notes in Computer
  Science}, pages 868--886. Springer, 2012.

\bibitem[Brakerski et~al.(2012)Brakerski, Gentry, and Vaikuntanathan]{BGV}
Zvika Brakerski, Craig Gentry, and Vinod Vaikuntanathan.
\newblock (leveled) fully homomorphic encryption without bootstrapping.
\newblock In \emph{{ITCS}}, pages 309--325. {ACM}, 2012.

\bibitem[Braverman et~al.(2016)Braverman, Garg, Ma, Nguyen, and
  Woodruff]{braverman2016}
Mark Braverman, Ankit Garg, Tengyu Ma, Huy~L. Nguyen, and David~P. Woodruff.
\newblock Communication lower bounds for statistical estimation problems via a
  distributed data processing inequality.
\newblock In \emph{Proceedings of the forty-eighth annual ACM symposium on
  Theory of Computing}, page 1011–1020. ACM, 2016.

\bibitem[Brendel et~al.(2017)Brendel, Rauber, and Bethge]{brendel2017decision}
Wieland Brendel, Jonas Rauber, and Matthias Bethge.
\newblock Decision-based adversarial attacks: Reliable attacks against
  black-box machine learning models.
\newblock \emph{arXiv preprint arXiv:1712.04248}, 2017.

\bibitem[Brisimi et~al.(2018)Brisimi, Chen, Mela, Olshevsky, Paschalidis, and
  Shi]{brisimi2018federated}
Theodora~S Brisimi, Ruidi Chen, Theofanie Mela, Alex Olshevsky, Ioannis~Ch
  Paschalidis, and Wei Shi.
\newblock Federated learning of predictive models from federated electronic
  health records.
\newblock \emph{International journal of medical informatics}, 112:\penalty0
  59--67, 2018.

\bibitem[B{\"{u}}nz et~al.(2018)B{\"{u}}nz, Bootle, Boneh, Poelstra, Wuille,
  and Maxwell]{DBLP:conf/sp/BunzBBPWM18}
Benedikt B{\"{u}}nz, Jonathan Bootle, Dan Boneh, Andrew Poelstra, Pieter
  Wuille, and Gregory Maxwell.
\newblock Bulletproofs: Short proofs for confidential transactions and more.
\newblock In \emph{2018 {IEEE} Symposium on Security and Privacy, {SP} 2018,
  Proceedings, 21-23 May 2018, San Francisco, California, {USA}}, 2018.

\bibitem[Buolamwini and Gebru(2018)]{buolamwini2018gender}
Joy Buolamwini and Timnit Gebru.
\newblock Gender shades: Intersectional accuracy disparities in commercial
  gender classification.
\newblock In \emph{Conference on fairness, accountability and transparency},
  pages 77--91, 2018.

\bibitem[Burkhart et~al.(2010)Burkhart, Strasser, Many, and
  Dimitropoulos]{burkhart2010sepia}
Martin Burkhart, Mario Strasser, Dilip Many, and Xenofontas Dimitropoulos.
\newblock {SEPIA}: Privacy-preserving aggregation of multi-domain network
  events and statistics.
\newblock \emph{Network}, 1\penalty0 (101101), 2010.

\bibitem[Caldas et~al.(2018{\natexlab{a}})Caldas, Kone{\v{c}}n{\'y}, McMahan,
  and Talwalkar]{caldas2018expanding}
Sebastian Caldas, Jakub Kone{\v{c}}n{\'y}, H~Brendan McMahan, and Ameet
  Talwalkar.
\newblock Expanding the reach of federated learning by reducing client resource
  requirements.
\newblock \emph{arXiv preprint arXiv:1812.07210}, 2018{\natexlab{a}}.

\bibitem[Caldas et~al.(2018{\natexlab{b}})Caldas, Wu, Li, Kone{\v{c}}n{\'y},
  McMahan, Smith, and Talwalkar]{caldas2018leaf}
Sebastian Caldas, Peter Wu, Tian Li, Jakub Kone{\v{c}}n{\'y}, H~Brendan
  McMahan, Virginia Smith, and Ameet Talwalkar.
\newblock {LEAF}: A benchmark for federated settings.
\newblock \emph{arXiv preprint arXiv:1812.01097}, 2018{\natexlab{b}}.

\bibitem[Canonne et~al.(2019)Canonne, Kamath, McMillan, Smith, and
  Ullman]{CKM+18b}
Clément~L Canonne, Gautam Kamath, Audra McMillan, Adam Smith, and Jonathan
  Ullman.
\newblock The structure of optimal private tests for simple hypotheses.
\newblock \emph{AarXiv preprint arXiv:1811.11148}, 2019.

\bibitem[Carlini and Wagner(2017)]{carlini2017towards}
Nicholas Carlini and David Wagner.
\newblock Towards evaluating the robustness of neural networks.
\newblock In \emph{2017 IEEE Symposium on Security and Privacy (SP)}, pages
  39--57. IEEE, 2017.

\bibitem[Carlini et~al.(2018)Carlini, Liu, Kos, Erlingsson, and
  Song]{carlini2018secret}
Nicholas Carlini, Chang Liu, Jernej Kos, {\'U}lfar Erlingsson, and Dawn Song.
\newblock The secret sharer: Measuring unintended neural network memorization
  \& extracting secrets.
\newblock \emph{arXiv preprint arXiv:1802.08232}, 2018.

\bibitem[Carlini et~al.(2020)Carlini, Tram{\`{e}}r, Wallace, Jagielski,
  Herbert-Voss, Lee, Roberts, Brown, Song, Erlingsson,
  et~al.]{carlini2020extracting}
Nicholas Carlini, Florian Tram{\`{e}}r, Eric Wallace, Matthew Jagielski, Ariel
  Herbert-Voss, Katherine Lee, Adam Roberts, Tom Brown, Dawn Song, Ulfar
  Erlingsson, et~al.
\newblock Extracting training data from large language models.
\newblock \emph{arXiv preprint arXiv:2012.07805}, 2020.

\bibitem[Ceballos et~al.(2018)Ceballos, Sharma, Mugica, Singh, Roman,
  Vepakomma, and Raskar]{splitVertical}
Iker Ceballos, Vivek Sharma, Eduardo Mugica, Abhishek Singh, Albert Roman,
  Praneeth Vepakomma, and Ramesh Raskar.
\newblock Split{NN}-driven vertical partitioning.
\newblock \emph{arXiv preprint arXiv:2008.04137}, 2018.

\bibitem[Chai~Sim et~al.(2019)Chai~Sim, Beaufays, Benard, Guliani, Kabel,
  Khare, Lucassen, Zadrazil, Zhang, Johnson, et~al.]{chai2019personalization}
Khe Chai~Sim, Fran{\c{c}}oise Beaufays, Arnaud Benard, Dhruv Guliani, Andreas
  Kabel, Nikhil Khare, Tamar Lucassen, Petr Zadrazil, Harry Zhang, Leif
  Johnson, et~al.
\newblock Personalization of end-to-end speech recognition on mobile devices
  for named entities.
\newblock \emph{arXiv}, pages arXiv--1912, 2019.

\bibitem[Chan et~al.(2012)Chan, Shi, and Song]{chan2012privacy}
T-H~Hubert Chan, Elaine Shi, and Dawn Song.
\newblock Privacy-preserving stream aggregation with fault tolerance.
\newblock In \emph{International Conference on Financial Cryptography and Data
  Security}, pages 200--214. Springer, 2012.

\bibitem[Chang et~al.(2018)Chang, Balachandar, Lam, Yi, Brown, Beers, Rosen,
  Rubin, and Kalpathy-Cramer]{chang2018distributed}
Ken Chang, Niranjan Balachandar, Carson Lam, Darvin Yi, James Brown, Andrew
  Beers, Bruce Rosen, Daniel~L Rubin, and Jayashree Kalpathy-Cramer.
\newblock Distributed deep learning networks among institutions for medical
  imaging.
\newblock \emph{Journal of the American Medical Informatics Association},
  25\penalty0 (8):\penalty0 945--954, 2018.

\bibitem[Chang and Tandon(2019)]{Chang2019OnTU}
Wei-Ting Chang and Ravi Tandon.
\newblock On the upload versus download cost for secure and private matrix
  multiplication.
\newblock \emph{ArXiv}, abs/1906.10684, 2019.

\bibitem[Charles and Kone{\v{c}}n{\`y}(2020)]{charles2020outsized}
Zachary Charles and Jakub Kone{\v{c}}n{\`y}.
\newblock On the outsized importance of learning rates in local update methods.
\newblock \emph{arXiv preprint arXiv:2007.00878}, 2020.

\bibitem[Chaum(1981)]{chaum1981untraceable}
David Chaum.
\newblock Untraceable electronic mail, return addresses, and digital
  pseudonyms.
\newblock \emph{Communications of the ACM}, 24\penalty0 (2), 1981.

\bibitem[Chen et~al.(2018{\natexlab{a}})Chen, Carvalho, Baracaldo, Ludwig,
  Edwards, Lee, Molloy, and Srivastava]{chen2018detecting}
Bryant Chen, Wilka Carvalho, Nathalie Baracaldo, Heiko Ludwig, Benjamin
  Edwards, Taesung Lee, Ian Molloy, and Biplav Srivastava.
\newblock Detecting backdoor attacks on deep neural networks by activation
  clustering.
\newblock \emph{arXiv preprint arXiv:1811.03728}, 2018{\natexlab{a}}.

\bibitem[Chen et~al.(2020{\natexlab{a}})Chen, Golubchik, and
  Paolieri]{chen2020backdoor}
Chien-Lun Chen, Leana Golubchik, and Marco Paolieri.
\newblock Backdoor attacks on federated meta-learning.
\newblock \emph{arXiv preprint arXiv:2006.07026}, 2020{\natexlab{a}}.

\bibitem[Chen et~al.(2021)Chen, Ghazi, Kumar, and
  Manurangsi]{chen2020distributed}
Lijie Chen, Badih Ghazi, Ravi Kumar, and Pasin Manurangsi.
\newblock On distributed differential privacy and counting distinct elements.
\newblock In \emph{Innovations in Theoretical Computer Science (ITCS)}, 2021.

\bibitem[Chen et~al.(2018{\natexlab{b}})Chen, Wang, Charles, and
  Papailiopoulos]{chen18draco}
Lingjiao Chen, Hongyi Wang, Zachary~B. Charles, and Dimitris~S. Papailiopoulos.
\newblock {DRACO:} {B}yzantine-resilient distributed training via redundant
  gradients.
\newblock In \emph{Proceedings of the 35th International Conference on Machine
  Learning, {ICML}}, 2018{\natexlab{b}}.

\bibitem[Chen et~al.(2019)Chen, Mathews, Ouyang, and Beaufays]{chen19oov}
Mingqing Chen, Rajiv Mathews, Tom Ouyang, and Fran{\c{c}}oise Beaufays.
\newblock Federated learning of out-of-vocabulary words.
\newblock \emph{arXiv preprint 1903.10635}, 2019.
\newblock URL \url{http://arxiv.org/abs/1903.10635}.

\bibitem[Chen et~al.(2017{\natexlab{a}})Chen, Zhang, Sharma, Yi, and
  Hsieh]{chen2017zoo}
Pin-Yu Chen, Huan Zhang, Yash Sharma, Jinfeng Yi, and Cho-Jui Hsieh.
\newblock Z{OO}: Zeroth order optimization based black-box attacks to deep
  neural networks without training substitute models.
\newblock In \emph{Proceedings of the 10th ACM Workshop on Artificial
  Intelligence and Security}, pages 15--26. ACM, 2017{\natexlab{a}}.

\bibitem[Chen et~al.(2020{\natexlab{b}})Chen, Kairouz, and Ozgur]{ChenKO2020}
Wei-Ning Chen, Peter Kairouz, and Ayfer Ozgur.
\newblock Breaking the communication-privacy-accuracy trilemma.
\newblock \emph{Advances in Neural Information Processing Systems}, 33,
  2020{\natexlab{b}}.

\bibitem[Chen et~al.(2017{\natexlab{b}})Chen, Liu, Li, Lu, and
  Song]{chen2017targeted}
Xinyun Chen, Chang Liu, Bo~Li, Kimberly Lu, and Dawn Song.
\newblock Targeted backdoor attacks on deep learning systems using data
  poisoning.
\newblock \emph{arXiv preprint arXiv:1712.05526}, 2017{\natexlab{b}}.

\bibitem[Chen et~al.(2017{\natexlab{c}})Chen, Su, and
  Xu]{Chen2017DistributedSM}
Yudong Chen, Lili Su, and Jiaming Xu.
\newblock {Distributed Statistical Machine Learning in Adversarial Settings:
  {B}yzantine Gradient Descent}.
\newblock \emph{POMACS}, 1:\penalty0 44:1--44:25, 2017{\natexlab{c}}.

\bibitem[Chenal and Tang(2014)]{DBLP:conf/latincrypt/ChenalT14}
Massimo Chenal and Qiang Tang.
\newblock On key recovery attacks against existing somewhat homomorphic
  encryption schemes.
\newblock In \emph{{LATINCRYPT}}, volume 8895 of \emph{Lecture Notes in
  Computer Science}, pages 239--258. Springer, 2014.

\bibitem[Cheng et~al.(2019{\natexlab{a}})Cheng, Fan, Jin, Liu, Chen, and
  Yang]{DBLP:journals/corr/abs-1901-08755}
Kewei Cheng, Tao Fan, Yilun Jin, Yang Liu, Tianjian Chen, and Qiang Yang.
\newblock Secure{B}oost: {A} lossless federated learning framework.
\newblock \emph{CoRR}, abs/1901.08755, 2019{\natexlab{a}}.
\newblock URL \url{http://arxiv.org/abs/1901.08755}.

\bibitem[Cheng et~al.(2019{\natexlab{b}})Cheng, Zhang, Kos, He, Hynes, Johnson,
  Juels, Miller, and Song]{cheng2019ekiden}
Raymond Cheng, Fan Zhang, Jernej Kos, Warren He, Nicholas Hynes, Noah Johnson,
  Ari Juels, Andrew Miller, and Dawn Song.
\newblock Ekiden: A platform for confidentiality-preserving, trustworthy, and
  performant smart contracts.
\newblock In \emph{2019 IEEE European Symposium on Security and Privacy
  (EuroS\&P)}, pages 185--200. IEEE, 2019{\natexlab{b}}.

\bibitem[Cheu et~al.(2019)Cheu, Smith, Ullman, Zeber, and
  Zhilyaev]{cheu2019distributed}
Albert Cheu, Adam Smith, Jonathan Ullman, David Zeber, and Maxim Zhilyaev.
\newblock Distributed differential privacy via shuffling.
\newblock In \emph{Annual International Conference on the Theory and
  Applications of Cryptographic Techniques}, pages 375--403. Springer, 2019.

\bibitem[Chor et~al.(1998)Chor, Kushilevitz, Goldreich, and Sudan]{Chor98PIR}
Benny Chor, Eyal Kushilevitz, Oded Goldreich, and Madhu Sudan.
\newblock Private information retrieval.
\newblock \emph{J. ACM}, 45\penalty0 (6):\penalty0 965--981, November 1998.
\newblock ISSN 0004-5411.
\newblock \doi{10.1145/293347.293350}.
\newblock URL \url{http://doi.acm.org/10.1145/293347.293350}.

\bibitem[Chou et~al.(2018)Chou, Tram{\`e}r, and Pellegrino]{chou2018sentinet}
Edward Chou, Florian Tram{\`e}r, and Giancarlo Pellegrino.
\newblock Senti{N}et: Detecting physical attacks against deep learning systems.
\newblock \emph{arXiv preprint arXiv:1812.00292}, 2018.

\bibitem[Chraibi et~al.(2019)Chraibi, Khaled, Kovalev, Richt{\'a}rik, Salim,
  and Tak{\'a}{\v{c}}]{chraibi2019distributed}
S{\'e}lim Chraibi, Ahmed Khaled, Dmitry Kovalev, Peter Richt{\'a}rik, Adil
  Salim, and Martin Tak{\'a}{\v{c}}.
\newblock Distributed fixed point methods with compressed iterates.
\newblock \emph{arXiv preprint arXiv:1912.09925}, 2019.

\bibitem[Christen(2012)]{christen12}
P.~Christen.
\newblock \emph{Data matching: concepts and techniques for record linkage,
  entity resolution, and duplicate detection}.
\newblock Springer Science \& Business Media, 2012.

\bibitem[{C}lara(2019)]{ClaraTraining}
{NVIDIA} {C}lara.
\newblock The clara training framework authors, 2019.
\newblock URL \url{https://developer.nvidia.com/clara}.

\bibitem[Cohen et~al.(2017)Cohen, Afshar, Tapson, and van
  Schaik]{cohen2017emnist}
Gregory Cohen, Saeed Afshar, Jonathan Tapson, and Andr{\'e} van Schaik.
\newblock {EMNIST}: an extension of {MNIST} to handwritten letters.
\newblock \emph{arXiv preprint arXiv:1702.05373}, 2017.

\bibitem[Colin et~al.(2016)Colin, Bellet, Salmon, and
  Cl{\'e}men{\c{c}}on]{Colin2016}
Igor Colin, Aur{\'e}lien Bellet, Joseph Salmon, and St{\'e}phan
  Cl{\'e}men{\c{c}}on.
\newblock Gossip dual averaging for decentralized optimization of pairwise
  functions.
\newblock In \emph{ICML}, 2016.

\bibitem[Cormode et~al.(2018)Cormode, Kulkarni, and
  Srivastava]{cormode2018marginal}
Graham Cormode, Tejas Kulkarni, and Divesh Srivastava.
\newblock Marginal release under local differential privacy.
\newblock In \emph{Proceedings of the 2018 International Conference on
  Management of Data}, pages 131--146. ACM, 2018.

\bibitem[Coron et~al.(2014)Coron, Lepoint, and
  Tibouchi]{DBLP:conf/pkc/CoronLT14}
Jean{-}S{\'{e}}bastien Coron, Tancr{\`{e}}de Lepoint, and Mehdi Tibouchi.
\newblock Scale-invariant fully homomorphic encryption over the integers.
\newblock In \emph{Public Key Cryptography}, volume 8383 of \emph{Lecture Notes
  in Computer Science}, pages 311--328. Springer, 2014.

\bibitem[Corrigan-Gibbs and Boneh(2017)]{corrigan2017prio}
Henry Corrigan-Gibbs and Dan Boneh.
\newblock Prio: Private, robust, and scalable computation of aggregate
  statistics.
\newblock In \emph{14th $\{$USENIX$\}$ Symposium on Networked Systems Design
  and Implementation ($\{$NSDI$\}$ 17)}, pages 259--282, 2017.

\bibitem[Corrigan{-}Gibbs and
  Kogan(2019)]{DBLP:journals/iacr/Corrigan-GibbsK19a}
Henry Corrigan{-}Gibbs and Dmitry Kogan.
\newblock Private information retrieval with sublinear online time.
\newblock \emph{{IACR} Cryptology ePrint Archive}, 2019:\penalty0 1075, 2019.

\bibitem[Cortes and Mohri(2014)]{cortes2014domain}
Corinna Cortes and Mehryar Mohri.
\newblock Domain adaptation and sample bias correction theory and algorithm for
  regression.
\newblock \emph{Theoretical Computer Science}, 519:\penalty0 103--126, 2014.

\bibitem[Cortes et~al.(2020)Cortes, Mohri, Suresh, and
  Zhang]{cortes2020multiple}
Corinna Cortes, Mehryar Mohri, Ananda~Theertha Suresh, and Ningshan Zhang.
\newblock Multiple-source adaptation with domain classifiers.
\newblock \emph{arXiv preprint arXiv:2008.11036}, 2020.

\bibitem[Costan and Devadas(2016)]{costan2016intel}
Victor Costan and Srinivas Devadas.
\newblock Intel {SGX} explained.
\newblock \emph{IACR Cryptology ePrint Archive}, 2016\penalty0 (086):\penalty0
  1--118, 2016.

\bibitem[Costan et~al.(2016)Costan, Lebedev, and Devadas]{costan2016sanctum}
Victor Costan, Ilia Lebedev, and Srinivas Devadas.
\newblock Sanctum: Minimal hardware extensions for strong software isolation.
\newblock In \emph{25th $\{$USENIX$\}$ Security Symposium ($\{$USENIX$\}$
  Security 16)}, pages 857--874, 2016.

\bibitem[Costello et~al.(2015)Costello, Fournet, Howell, Kohlweiss, Kreuter,
  Naehrig, Parno, and Zahur]{DBLP:conf/sp/CostelloFHKKNPZ15}
Craig Costello, C{\'{e}}dric Fournet, Jon Howell, Markulf Kohlweiss, Benjamin
  Kreuter, Michael Naehrig, Bryan Parno, and Samee Zahur.
\newblock Geppetto: Versatile verifiable computation.
\newblock In \emph{{IEEE} Symposium on Security and Privacy}, pages 253--270.
  {IEEE} Computer Society, 2015.

\bibitem[Cotter et~al.(2011)Cotter, Shamir, Srebro, and
  Sridharan]{cotter2011acmb}
Andrew Cotter, Ohad Shamir, Nati Srebro, and Karthik Sridharan.
\newblock Better mini-batch algorithms via accelerated gradient methods.
\newblock In J.~Shawe-Taylor, R.~Zemel, P.~Bartlett, F.~Pereira, and K.~Q.
  Weinberger, editors, \emph{Advances in Neural Information Processing
  Systems}, volume~24, pages 1647--1655. Curran Associates, Inc., 2011.
\newblock URL
  \url{https://proceedings.neurips.cc/paper/2011/file/b55ec28c52d5f6205684a473a2193564-Paper.pdf}.

\bibitem[Courbariaux et~al.(2015)Courbariaux, Bengio, and
  David]{courbariaux2015binaryconnect}
Matthieu Courbariaux, Yoshua Bengio, and Jean-Pierre David.
\newblock Binary{C}onnect: Training deep neural networks with binary weights
  during propagations.
\newblock In \emph{Advances in neural information processing systems}, pages
  3123--3131, 2015.

\bibitem[Courtiol et~al.(2019)Courtiol, Maussion, Moarii, Pronier, Pilcer,
  Sefta, Manceron, Toldo, Zaslavskiy, Le~Stang, et~al.]{courtiol2019deep}
Pierre Courtiol, Charles Maussion, Matahi Moarii, Elodie Pronier, Samuel
  Pilcer, Meriem Sefta, Pierre Manceron, Sylvain Toldo, Mikhail Zaslavskiy,
  Nolwenn Le~Stang, et~al.
\newblock Deep learning-based classification of mesothelioma improves
  prediction of patient outcome.
\newblock \emph{Nature medicine}, pages 1--7, 2019.

\bibitem[Cover and Thomas(2012)]{cover2012elements}
Thomas~M Cover and Joy~A Thomas.
\newblock \emph{Elements of information theory}.
\newblock John Wiley \& Sons, 2012.

\bibitem[Cretu et~al.(2008)Cretu, Stavrou, Locasto, Stolfo, and
  Keromytis]{cretu2008casting}
Gabriela~F Cretu, Angelos Stavrou, Michael~E Locasto, Salvatore~J Stolfo, and
  Angelos~D Keromytis.
\newblock Casting out demons: Sanitizing training data for anomaly sensors.
\newblock In \emph{2008 IEEE Symposium on Security and Privacy (sp 2008)},
  pages 81--95. IEEE, 2008.

\bibitem[Cummings et~al.(2018{\natexlab{a}})Cummings, Krehbiel, Lai, and
  Tantitongpipat]{CKLT18}
Rachel Cummings, Sara Krehbiel, Kevin Lai, and Uthaipon Tantitongpipat.
\newblock Differential privacy for growing databases.
\newblock In \emph{Advances in Neural Information Processing Systems 31},
  NeurIPS '18, pages 8864--8873, 2018{\natexlab{a}}.

\bibitem[Cummings et~al.(2018{\natexlab{b}})Cummings, Krehbiel, Mei, Tuo, and
  Zhang]{CKM+18}
Rachel Cummings, Sara Krehbiel, Yajun Mei, Rui Tuo, and Wanrong Zhang.
\newblock Differentially private change-point detection.
\newblock In \emph{Advances in Neural Information Processing Systems 31},
  NeurIPS '18, pages 10825--10834, 2018{\natexlab{b}}.

\bibitem[Cummings et~al.(2019{\natexlab{a}})Cummings, Dekel, Heffetz, and
  Ligett]{CDHL19}
Rachel Cummings, Inbal Dekel, Ori Heffetz, and Katrina Ligett.
\newblock Bringing differential privacy into the experimental economics lab:
  Theory and an application to a public-good game.
\newblock Working paper, 2019{\natexlab{a}}.

\bibitem[Cummings et~al.(2019{\natexlab{b}})Cummings, Gupta, Kimpara, and
  Morgenstern]{CGKM19}
Rachel Cummings, Varun Gupta, Dhamma Kimpara, and Jamie Morgenstern.
\newblock On the compatibility of privacy and fairness.
\newblock In \emph{Proceedings of Fairness in User Modeling, Adaptation and
  Personalization}, FairUMAP, 2019{\natexlab{b}}.

\bibitem[Cyffers and Bellet(2020)]{privacy_amp_by_decentralization}
Edwige Cyffers and Aur\'elien Bellet.
\newblock Privacy amplification by decentralization.
\newblock \emph{arXiv preprint arXiv:2012.05326}, 2020.

\bibitem[Damg\r{a}rd(2010)]{DamgaardSigma}
Damg\r{a}rd.
\newblock On $\sigma$ protocols.
\newblock \url{http://www.cs.au.dk/~ivan/Sigma.pdf}, 2010.

\bibitem[Data et~al.(2020)Data, Song, and Diggavi]{data2019data}
Deepesh Data, Linqi Song, and Suhas Diggavi.
\newblock Data encoding for byzantine-resilient distributed optimization.
\newblock \emph{IEEE Transactions on Information Theory}, 2020.

\bibitem[de~Brouwer(2019)]{docai}
Walter de~Brouwer.
\newblock The federated future is ready for shipping.
\newblock \url{https://doc.ai/blog/federated-future-ready-shipping/}, March
  2019.

\bibitem[Dean et~al.(2012)Dean, Corrado, Monga, Chen, Devin, Le, Mao, Ranzato,
  Senior, Tucker, Yang, and Ng]{dean2012large}
Jeffrey Dean, Greg~S. Corrado, Rajat Monga, Kai Chen, Matthieu Devin, Quoc~V.
  Le, Mark~Z. Mao, Marc'Aurelio Ranzato, Andrew Senior, Paul Tucker, Ke~Yang,
  and Andrew~Y. Ng.
\newblock Large scale distributed deep networks.
\newblock In \emph{Proceedings of the International Conference on Neural
  Information Processing Systems}, pages 1223--1231, 2012.

\bibitem[Dekel et~al.(2012)Dekel, Gilad-Bachrach, Shamir, and
  Xiao]{dekel12optimal}
Ofer Dekel, Ran Gilad-Bachrach, Ohad Shamir, and Lin Xiao.
\newblock Optimal distributed online prediction using mini-batches.
\newblock \emph{J. Mach. Learn. Res.}, 13\penalty0 (1), January 2012.

\bibitem[Diakonikolas et~al.(2019)Diakonikolas, Kamath, Kane, Li, Steinhardt,
  and Stewart]{pmlr-v97-diakonikolas19a}
Ilias Diakonikolas, Gautam Kamath, Daniel Kane, Jerry Li, Jacob Steinhardt, and
  Alistair Stewart.
\newblock Sever: A robust meta-algorithm for stochastic optimization.
\newblock In Kamalika Chaudhuri and Ruslan Salakhutdinov, editors,
  \emph{Proceedings of the 36th International Conference on Machine Learning},
  volume~97 of \emph{Proceedings of Machine Learning Research}, pages
  1596--1606, Long Beach, California, USA, 09--15 Jun 2019. PMLR.
\newblock URL \url{http://proceedings.mlr.press/v97/diakonikolas19a.html}.

\bibitem[Diaz et~al.(2019)Diaz, Kairouz, Liao, and Sankar]{diaz2019theoretical}
Mario Diaz, Peter Kairouz, Jiachun Liao, and Lalitha Sankar.
\newblock Theoretical guarantees for model auditing with finite adversaries.
\newblock \emph{arXiv preprint arXiv:1911.03405}, 2019.

\bibitem[{Differential Privacy Team}(2017)]{applewhitepaper:17}
{Differential Privacy Team}.
\newblock Learning with privacy at scale.
\newblock \emph{Apple Machine Learning Journal}, 1\penalty0 (8), 2017.
\newblock URL
  \url{https://machinelearning.apple.com/2017/12/06/learning-with-privacy-at-scale.html}.

\bibitem[Ding et~al.(2017)Ding, Kulkarni, and
  Yekhanin]{collecting-telemetry-data-privately}
Bolin Ding, Janardhan Kulkarni, and Sergey Yekhanin.
\newblock Collecting telemetry data privately.
\newblock In \emph{Advances in Neural Information Processing Systems 30},
  December 2017.
\newblock URL
  \url{https://www.microsoft.com/en-us/research/publication/collecting-telemetry-data-privately/}.

\bibitem[Ding et~al.(2018)Ding, Wang, Wang, Zhang, and
  Kifer]{ding2019detecting}
Zeyu Ding, Yuxin Wang, Guanhong Wang, Danfeng Zhang, and Daniel Kifer.
\newblock Detecting violations of differential privacy.
\newblock In \emph{Proceedings of the 2018 ACM SIGSAC Conference on Computer
  and Communications Security}, CCS '18, pages 475--489, New York, NY, USA,
  2018. ACM.
\newblock ISBN 978-1-4503-5693-0.
\newblock \doi{10.1145/3243734.3243818}.
\newblock URL \url{http://doi.acm.org/10.1145/3243734.3243818}.

\bibitem[Dingledine et~al.(2004)Dingledine, Mathewson, and
  Syverson]{dingledine2004tor}
Roger Dingledine, Nick Mathewson, and Paul Syverson.
\newblock Tor: The second-generation onion router.
\newblock Technical report, Naval Research Lab Washington DC, 2004.

\bibitem[Dinh et~al.(2020)Dinh, Tran, and Nguyen]{moreau}
Canh~T. Dinh, Nguyen~H. Tran, and Tuan~Dung Nguyen.
\newblock {Personalized Federated Learning with Moreau Envelopes}.
\newblock In \emph{NeurIPS}, 2020.

\bibitem[{D'Oliveira} and {Rouayheb}(2018)]{dolift}
Rafael G.~L. {D'Oliveira} and S.~E. {Rouayheb}.
\newblock Lifting private information retrieval from two to any number of
  messages.
\newblock In \emph{2018 IEEE International Symposium on Information Theory
  (ISIT)}, pages 1744--1748, June 2018.
\newblock \doi{10.1109/ISIT.2018.8437805}.

\bibitem[Douceur(2002)]{sybil-attack}
John~R. Douceur.
\newblock The sybil attack.
\newblock In \emph{Revised Papers from the First International Workshop on
  Peer-to-Peer Systems}, IPTPS '01, pages 251--260, London, UK, UK, 2002.
  Springer-Verlag.
\newblock ISBN 3-540-44179-4.
\newblock URL \url{http://dl.acm.org/citation.cfm?id=646334.687813}.

\bibitem[Duchi et~al.(2011)Duchi, Hazan, and Singer]{duchi2011adaptive}
John Duchi, Elad Hazan, and Yoram Singer.
\newblock Adaptive subgradient methods for online learning and stochastic
  optimization.
\newblock \emph{Journal of machine learning research}, 12\penalty0 (7), 2011.

\bibitem[Duchi et~al.(2013)Duchi, Jordan, and Wainwright]{duchi2013local}
John~C Duchi, Michael~I Jordan, and Martin~J Wainwright.
\newblock Local privacy and statistical minimax rates.
\newblock In \emph{Foundations of Computer Science (FOCS), 2013 IEEE 54th
  Annual Symposium on}, pages 429--438. IEEE, 2013.

\bibitem[Dutta et~al.(2018)Dutta, Joshi, Ghosh, Dube, and
  Nagpurkar]{dutta2018slow}
Sanghamitra Dutta, Gauri Joshi, Soumyadip Ghosh, Parijat Dube, and Priya
  Nagpurkar.
\newblock {Slow and Stale Gradients Can Win the Race: Error-Runtime Trade-offs
  in Distributed {SGD}}.
\newblock \emph{International Conference on Artificial Intelligence and
  Statistics (AISTATS)}, April 2018.
\newblock URL \url{https://arxiv.org/abs/1803.01113}.

\bibitem[Dwork(2008)]{dwork2008differential}
Cynthia Dwork.
\newblock Differential privacy: A survey of results.
\newblock In \emph{International Conference on Theory and Applications of
  Models of Computation}, pages 1--19. Springer, 2008.

\bibitem[Dwork and Roth(2014)]{dwork2014algorithmic}
Cynthia Dwork and Aaron Roth.
\newblock The algorithmic foundations of differential privacy.
\newblock \emph{Foundations and Trends in Theoretical Computer Science},
  9\penalty0 (3--4):\penalty0 211--407, 2014.

\bibitem[Dwork et~al.(2006{\natexlab{a}})Dwork, Kenthapadi, McSherry, Mironov,
  and Naor]{dwork2006our}
Cynthia Dwork, Krishnaram Kenthapadi, Frank McSherry, Ilya Mironov, and Moni
  Naor.
\newblock Our data, ourselves: Privacy via distributed noise generation.
\newblock In \emph{Annual International Conference on the Theory and
  Applications of Cryptographic Techniques}, pages 486--503. Springer,
  2006{\natexlab{a}}.

\bibitem[Dwork et~al.(2006{\natexlab{b}})Dwork, {McSherry}, Nissim, and
  Smith]{DMNS06}
Cynthia Dwork, Frank {McSherry}, Kobbi Nissim, and Adam~D. Smith.
\newblock Calibrating noise to sensitivity in private data analysis.
\newblock In \emph{{IACR} {T}heory of {C}ryptography {C}onference (TCC), New
  York, New York}, volume 3876 of \emph{Lecture Notes in Computer Science},
  pages 265--284. Springer-Verlag, 2006{\natexlab{b}}.
\newblock \doi{10.1007/11681878_14}.

\bibitem[Dwork et~al.(2010)Dwork, Rothblum, and Vadhan]{DRV10}
Cynthia Dwork, Guy~N. Rothblum, and Salil Vadhan.
\newblock Boosting and differential privacy.
\newblock In \emph{Proceedings of the IEEE 51st Annual Symposium on Foundations
  of Computer Science}, FOCS '10, pages 51--60, 2010.

\bibitem[Dwork et~al.(2012)Dwork, Hardt, Pitassi, Reingold, and
  Zemel]{dwork2012fairness}
Cynthia Dwork, Moritz Hardt, Toniann Pitassi, Omer Reingold, and Richard Zemel.
\newblock Fairness through awareness.
\newblock In \emph{Proceedings of the 3rd innovations in theoretical computer
  science conference}, pages 214--226. ACM, 2012.

\bibitem[Eckhouse et~al.(2019)Eckhouse, Lum, Conti-Cook, and
  Ciccolini]{eckhouse2019layers}
Laurel Eckhouse, Kristian Lum, Cynthia Conti-Cook, and Julie Ciccolini.
\newblock Layers of bias: A unified approach for understanding problems with
  risk assessment.
\newblock \emph{Criminal Justice and Behavior}, 46\penalty0 (2):\penalty0
  185--209, 2019.

\bibitem[Eichner et~al.(2019)Eichner, Koren, McMahan, Srebro, and
  Talwar]{eichner19semicyclic}
Hubert Eichner, Tomer Koren, H.~Brendan McMahan, Nathan Srebro, and Kunal
  Talwar.
\newblock Semi-cyclic stochastic gradient descent.
\newblock In \emph{Accepted to ICML 2019.}, 2019.
\newblock URL \url{https://arxiv.org/abs/1904.10120}.

\bibitem[Eldefrawy et~al.(2012)Eldefrawy, Tsudik, Francillon, and
  Perito]{DBLP:conf/ndss/EldefrawyTFP12}
Karim Eldefrawy, Gene Tsudik, Aur{\'{e}}lien Francillon, and Daniele Perito.
\newblock {SMART:} secure and minimal architecture for (establishing dynamic)
  root of trust.
\newblock In \emph{{NDSS}}. The Internet Society, 2012.

\bibitem[Elgabli et~al.(2019)Elgabli, Park, Bedi, Bennis, and
  Aggarwal]{elgabligadmm}
Anis Elgabli, Jihong Park, Amrit~S Bedi, Mehdi Bennis, and Vaneet Aggarwal.
\newblock {GADMM}: Fast and communication efficient framework for distributed
  machine learning.
\newblock \emph{arXiv preprint arXiv:1909.00047}, 2019.

\bibitem[Elgabli et~al.(2020)Elgabli, Park, Issaid, and
  Bennis]{elgabli2020harnessing}
Anis Elgabli, Jihong Park, Chaouki~Ben Issaid, and Mehdi Bennis.
\newblock Harnessing wireless channels for scalable and privacy-preserving
  federated learning, 2020.

\bibitem[Elsken et~al.(2018)Elsken, Hendrik~Metzen, and
  Hutter]{elsken2018efficient}
Thomas Elsken, Jan Hendrik~Metzen, and Frank Hutter.
\newblock Efficient multi-objective neural architecture search via {L}amarckian
  evolution.
\newblock \emph{arXiv preprint arXiv:1804.09081}, 2018.

\bibitem[Engstrom et~al.(2017)Engstrom, Tran, Tsipras, Schmidt, and
  Madry]{engstrom2017rotation}
Logan Engstrom, Brandon Tran, Dimitris Tsipras, Ludwig Schmidt, and Aleksander
  Madry.
\newblock A rotation and a translation suffice: Fooling {CNN}s with simple
  transformations.
\newblock \emph{arXiv preprint arXiv:1712.02779}, 2017.

\bibitem[Erlingsson et~al.(2014)Erlingsson, Pihur, and Korolova]{rappor:15}
\'{U}lfar Erlingsson, Vasyl Pihur, and Aleksandra Korolova.
\newblock {RAPPOR}: Randomized aggregatable privacy-preserving ordinal
  response.
\newblock In \emph{{ACM CCS}}, 2014.
\newblock ISBN 978-1-4503-2957-6.
\newblock \doi{10.1145/2660267.2660348}.
\newblock URL \url{http://doi.acm.org/10.1145/2660267.2660348}.

\bibitem[Erlingsson et~al.(2019)Erlingsson, Feldman, Mironov, Raghunathan,
  Talwar, and Thakurta]{erlingsson2019amplification}
{\'U}lfar Erlingsson, Vitaly Feldman, Ilya Mironov, Ananth Raghunathan, Kunal
  Talwar, and Abhradeep Thakurta.
\newblock Amplification by shuffling: From local to central differential
  privacy via anonymity.
\newblock In \emph{SODA}, pages 2468--2479, 2019.

\bibitem[{EU CORDIS}(2019)]{melloddy19pharma}
{EU CORDIS}.
\newblock Machine learning ledger orchestration for drug discovery, 2019.
\newblock URL
  \url{https://cordis.europa.eu/project/rcn/223634/factsheet/en?WT.mc_id=RSS-Feed&WT.rss_f=project&WT.rss_a=223634&WT.rss_ev=a}.
\newblock Retrieved Aug 2019.

\bibitem[Falkner et~al.(2018)Falkner, Klein, and Hutter]{falkner2018bohb}
Stefan Falkner, Aaron Klein, and Frank Hutter.
\newblock {BOHB}: Robust and efficient hyperparameter optimization at scale.
\newblock \emph{arXiv preprint arXiv:1807.01774}, 2018.

\bibitem[Fallah et~al.(2020)Fallah, Mokhtari, and
  Ozdaglar]{fallah2020personalized}
Alireza Fallah, Aryan Mokhtari, and Asuman Ozdaglar.
\newblock Personalized federated learning: A meta-learning approach.
\newblock \emph{arXiv preprint arXiv:2002.07948}, 2020.

\bibitem[Fan and Vercauteren(2012)]{BFV2}
Junfeng Fan and Frederik Vercauteren.
\newblock Somewhat practical fully homomorphic encryption.
\newblock \emph{{IACR} Cryptology ePrint Archive}, 2012:\penalty0 144, 2012.

\bibitem[Fang et~al.(2019)Fang, Cao, Jia, and Gong]{fang2019local}
Minghong Fang, Xiaoyu Cao, Jinyuan Jia, and Neil~Zhenqiang Gong.
\newblock Local model poisoning attacks to {B}yzantine-robust federated
  learning.
\newblock \emph{arXiv preprint arXiv:1911.11815}, 2019.

\bibitem[{FeatureCloud}(2019)]{featurecloud19ehr}
{FeatureCloud}.
\newblock Feature{C}loud: Our vision, 2019.
\newblock URL \url{https://featurecloud.eu/about/our-vision/}.
\newblock Retrieved Aug 2019.

\bibitem[Feldman et~al.(2018)Feldman, Mironov, Talwar, and
  Thakurta]{feldman2018privacy}
Vitaly Feldman, Ilya Mironov, Kunal Talwar, and Abhradeep Thakurta.
\newblock Privacy amplification by iteration.
\newblock In \emph{2018 IEEE 59th Annual Symposium on Foundations of Computer
  Science (FOCS)}, pages 521--532. IEEE, 2018.

\bibitem[Feutry et~al.(2018)Feutry, Piantanida, Bengio, and
  Duhamel]{DBLP:journals/corr/abs-1802-09386}
Cl{\'{e}}ment Feutry, Pablo Piantanida, Yoshua Bengio, and Pierre Duhamel.
\newblock Learning anonymized representations with adversarial neural networks.
\newblock \emph{CoRR}, abs/1802.09386, 2018.
\newblock URL \url{http://arxiv.org/abs/1802.09386}.

\bibitem[Finn et~al.(2017)Finn, Abbeel, and Levine]{finn17maml}
Chelsea Finn, Pieter Abbeel, and Sergey Levine.
\newblock Model-agnostic meta-learning for fast adaptation of deep networks.
\newblock In \emph{Proceedings of the 34th International Conference on Machine
  Learning}, 2017.

\bibitem[Francillon et~al.(2014)Francillon, Nguyen, Rasmussen, and
  Tsudik]{DBLP:conf/date/FrancillonNRT14}
Aur{\'{e}}lien Francillon, Quan Nguyen, Kasper~Bonne Rasmussen, and Gene
  Tsudik.
\newblock A minimalist approach to remote attestation.
\newblock In \emph{{DATE}}, pages 1--6. European Design and Automation
  Association, 2014.

\bibitem[Fredrikson et~al.(2015)Fredrikson, Jha, and
  Ristenpart]{fredrikson2015model}
Matt Fredrikson, Somesh Jha, and Thomas Ristenpart.
\newblock Model inversion attacks that exploit confidence information and basic
  countermeasures.
\newblock In \emph{Proceedings of the 22nd ACM SIGSAC Conference on Computer
  and Communications Security}, pages 1322--1333. ACM, 2015.

\bibitem[Fung et~al.(2018)Fung, Yoon, and Beschastnikh]{fung2018mitigating}
Clement Fung, Chris~JM Yoon, and Ivan Beschastnikh.
\newblock Mitigating sybils in federated learning poisoning.
\newblock \emph{arXiv preprint arXiv:1808.04866}, 2018.

\bibitem[Furukawa et~al.(2017)Furukawa, Lindell, Nof, and
  Weinstein]{DBLP:conf/eurocrypt/FurukawaLNW17}
Jun Furukawa, Yehuda Lindell, Ariel Nof, and Or~Weinstein.
\newblock High-throughput secure three-party computation for malicious
  adversaries and an honest majority.
\newblock In \emph{{EUROCRYPT} {(2)}}, volume 10211 of \emph{Lecture Notes in
  Computer Science}, pages 225--255, 2017.

\bibitem[Gaier and Ha(2019)]{gaier2019weight}
Adam Gaier and David Ha.
\newblock Weight agnostic neural networks.
\newblock \emph{arXiv preprint arXiv:1906.04358}, 2019.

\bibitem[Gandikota et~al.(2019)Gandikota, Maity, and
  Mazumdar]{gandikota2019vqsgd}
Venkata Gandikota, Raj~Kumar Maity, and Arya Mazumdar.
\newblock vq{SGD}: Vector quantized stochastic gradient descent.
\newblock \emph{arXiv preprint arXiv:1911.07971}, 2019.

\bibitem[Gasc{\'{o}}n et~al.(2017)Gasc{\'{o}}n, Schoppmann, Balle, Raykova,
  Doerner, Zahur, and Evans]{GasconSB0DZE17}
Adri{\`{a}} Gasc{\'{o}}n, Phillipp Schoppmann, Borja Balle, Mariana Raykova,
  Jack Doerner, Samee Zahur, and David Evans.
\newblock Privacy-preserving distributed linear regression on high-dimensional
  data.
\newblock \emph{PoPETs}, 2017\penalty0 (4):\penalty0 345--364, 2017.

\bibitem[Gennaro et~al.(2010)Gennaro, Gentry, and
  Parno]{DBLP:conf/crypto/GennaroGP10}
Rosario Gennaro, Craig Gentry, and Bryan Parno.
\newblock Non-interactive verifiable computing: Outsourcing computation to
  untrusted workers.
\newblock In \emph{{CRYPTO}}, volume 6223 of \emph{Lecture Notes in Computer
  Science}, pages 465--482. Springer, 2010.

\bibitem[Gennaro et~al.(2013)Gennaro, Gentry, Parno, and
  Raykova]{DBLP:conf/eurocrypt/GennaroGP013}
Rosario Gennaro, Craig Gentry, Bryan Parno, and Mariana Raykova.
\newblock Quadratic span programs and succinct {NIZK}s without {PCP}s.
\newblock In \emph{{EUROCRYPT}}, volume 7881 of \emph{Lecture Notes in Computer
  Science}, pages 626--645. Springer, 2013.

\bibitem[Gentry(2009)]{gentry2009fully}
Craig Gentry.
\newblock Fully homomorphic encryption using ideal lattices.
\newblock In \emph{Proceedings of the forty-first annual ACM symposium on
  Theory of computing}, pages 169--178, 2009.

\bibitem[Gentry and Halevi(2019)]{DBLP:conf/tcc/GentryH19}
Craig Gentry and Shai Halevi.
\newblock Compressible {FHE} with applications to {PIR}.
\newblock In \emph{{TCC} {(2)}}, volume 11892 of \emph{Lecture Notes in
  Computer Science}, pages 438--464. Springer, 2019.

\bibitem[Geyer et~al.(2017)Geyer, Klein, and Nabi]{geyer2017differentiallyyer}
Robin~C. Geyer, Tassilo Klein, and Moin Nabi.
\newblock Differentially private federated learning: {A} client level
  perspective.
\newblock \emph{CoRR}, abs/1712.07557, 2017.
\newblock URL \url{http://arxiv.org/abs/1712.07557}.

\bibitem[Ghazi et~al.(2019{\natexlab{a}})Ghazi, Golowich, Kumar, Pagh, and
  Velingker]{ghazi2019private}
Badih Ghazi, Noah Golowich, Ravi Kumar, Rasmus Pagh, and Ameya Velingker.
\newblock On the power of multiple anonymous messages.
\newblock \emph{arXiv:1908.11358}, 2019{\natexlab{a}}.

\bibitem[Ghazi et~al.(2019{\natexlab{b}})Ghazi, Pagh, and
  Velingker]{ghazi2019scalable_1}
Badih Ghazi, Rasmus Pagh, and Ameya Velingker.
\newblock Scalable and differentially private distributed aggregation in the
  shuffled model.
\newblock \emph{arXiv preprint arXiv:1906.08320}, 2019{\natexlab{b}}.

\bibitem[Ghazi et~al.(2020{\natexlab{a}})Ghazi, Golowich, Kumar, Manurangsi,
  Pagh, and Velingker]{pure-dp-shuffled}
Badih Ghazi, Noah Golowich, Ravi Kumar, Pasin Manurangsi, Rasmus Pagh, and
  Ameya Velingker.
\newblock Pure differentially private summation from anonymous messages.
\newblock In \emph{ITC}, pages 15:1--15:23, 2020{\natexlab{a}}.

\bibitem[Ghazi et~al.(2020{\natexlab{b}})Ghazi, Kumar, Manurangsi, and
  Pagh]{counting_shuffle_ICML}
Badih Ghazi, Ravi Kumar, Pasin Manurangsi, and Rasmus Pagh.
\newblock Private counting from anonymous messages: Near-optimal accuracy with
  vanishing communication overhead.
\newblock In \emph{ICML}, 2020{\natexlab{b}}.

\bibitem[Ghazi et~al.(2020{\natexlab{c}})Ghazi, Manurangsi, Pagh, and
  Velingker]{ghazi2019scalable}
Badih Ghazi, Pasin Manurangsi, Rasmus Pagh, and Ameya Velingker.
\newblock Private aggregation from fewer anonymous messages.
\newblock In \emph{EUROCRYPT}, pages 798--827, 2020{\natexlab{c}}.

\bibitem[Ghosh et~al.(2009)Ghosh, Roughgarden, and
  Sundararajan]{Ghosh:2009:UUP:1536414.1536464}
Arpita Ghosh, Tim Roughgarden, and Mukund Sundararajan.
\newblock Universally utility-maximizing privacy mechanisms.
\newblock In \emph{Proceedings of the Forty-first Annual ACM Symposium on
  Theory of Computing}, STOC '09, pages 351--360, New York, NY, USA, 2009. ACM.
\newblock ISBN 978-1-60558-506-2.
\newblock \doi{10.1145/1536414.1536464}.
\newblock URL \url{http://doi.acm.org/10.1145/1536414.1536464}.

\bibitem[Gilad{-}Bachrach et~al.(2016)Gilad{-}Bachrach, Dowlin, Laine, Lauter,
  Naehrig, and Wernsing]{gbdllnwCN}
Ran Gilad{-}Bachrach, Nathan Dowlin, Kim Laine, Kristin~E. Lauter, Michael
  Naehrig, and John Wernsing.
\newblock Crypto{N}ets: Applying neural networks to encrypted data with high
  throughput and accuracy.
\newblock In \emph{Proceedings of the 33nd International Conference on Machine
  Learning, {ICML} 2016, New York City, NY, USA, June 19-24, 2016}, pages
  201--210, 2016.
\newblock URL \url{http://proceedings.mlr.press/v48/gilad-bachrach16.html}.

\bibitem[Girgis et~al.(2020)Girgis, Data, Diggavi, Kairouz, and
  Suresh]{girgis2020shuffled}
Antonious~M Girgis, Deepesh Data, Suhas Diggavi, Peter Kairouz, and
  Ananda~Theertha Suresh.
\newblock Shuffled model of federated learning: Privacy, communication and
  accuracy trade-offs.
\newblock \emph{arXiv preprint arXiv:2008.07180}, 2020.

\bibitem[Goldreich et~al.(1987)Goldreich, Micali, and Wigderson]{goldreich87}
O.~Goldreich, S.~Micali, and A.~Wigderson.
\newblock How to play any mental game.
\newblock In \emph{Proceedings of the Nineteenth Annual ACM Symposium on Theory
  of Computing}, STOC '87, pages 218--229, New York, NY, USA, 1987. ACM.
\newblock ISBN 0-89791-221-7.
\newblock \doi{10.1145/28395.28420}.
\newblock URL \url{http://doi.acm.org/10.1145/28395.28420}.

\bibitem[Goldwasser et~al.(1989)Goldwasser, Micali, and
  Rackoff]{DBLP:journals/siamcomp/GoldwasserMR89}
Shafi Goldwasser, Silvio Micali, and Charles Rackoff.
\newblock The knowledge complexity of interactive proof systems.
\newblock \emph{{SIAM} J. Comput.}, 18\penalty0 (1):\penalty0 186--208, 1989.

\bibitem[Goldwasser et~al.(2008)Goldwasser, Kalai, and
  Rothblum]{DBLP:conf/stoc/GoldwasserKR08}
Shafi Goldwasser, Yael~Tauman Kalai, and Guy~N. Rothblum.
\newblock Delegating computation: interactive proofs for muggles.
\newblock In \emph{{STOC}}, pages 113--122. {ACM}, 2008.

\bibitem[Goodfellow et~al.(2015{\natexlab{a}})Goodfellow, Shlens, and
  Szegedy]{DBLP:journals/corr/GoodfellowSS14}
Ian~J. Goodfellow, Jonathon Shlens, and Christian Szegedy.
\newblock Explaining and harnessing adversarial examples.
\newblock In \emph{3rd International Conference on Learning Representations,
  {ICLR} 2015, San Diego, CA, USA, May 7-9, 2015, Conference Track
  Proceedings}, 2015{\natexlab{a}}.
\newblock URL \url{http://arxiv.org/abs/1412.6572}.

\bibitem[Goodfellow et~al.(2015{\natexlab{b}})Goodfellow, Shlens, and
  Szegedy]{goodfellow2014explaining}
Ian~J Goodfellow, Jonathon Shlens, and Christian Szegedy.
\newblock Explaining and harnessing adversarial examples.
\newblock \emph{ICLR}, 2015{\natexlab{b}}.

\bibitem[Goryczka and Xiong(2017)]{DBLP:journals/tdsc/GoryczkaX17}
Slawomir Goryczka and Li~Xiong.
\newblock A comprehensive comparison of multiparty secure additions with
  differential privacy.
\newblock \emph{{IEEE} Trans. Dependable Sec. Comput.}, 14\penalty0
  (5):\penalty0 463--477, 2017.
\newblock \doi{10.1109/TDSC.2015.2484326}.
\newblock URL \url{https://doi.org/10.1109/TDSC.2015.2484326}.

\bibitem[Gu et~al.(2017)Gu, Dolan-Gavitt, and Garg]{gu2017badnets}
Tianyu Gu, Brendan Dolan-Gavitt, and Siddharth Garg.
\newblock Bad{N}ets: Identifying vulnerabilities in the machine learning model
  supply chain.
\newblock \emph{arXiv preprint arXiv:1708.06733}, 2017.

\bibitem[Gupta and Raskar(2018)]{gupta2018distributed}
Otkrist Gupta and Ramesh Raskar.
\newblock Distributed learning of deep neural network over multiple agents.
\newblock \emph{Journal of Network and Computer Applications}, 116:\penalty0
  1--8, 2018.

\bibitem[Haddadpour et~al.(2019)Haddadpour, Kamani, Mahdavi, and
  Cadambe]{haddadpour2019local}
Farzin Haddadpour, Mohammad~Mahdi Kamani, Mehrdad Mahdavi, and Viveck~R
  Cadambe.
\newblock Local {SGD} with periodic averaging: Tighter analysis and adaptive
  synchronization.
\newblock \emph{arXiv preprint arXiv:1910.13598}, 2019.

\bibitem[Haeberlen et~al.(2011)Haeberlen, Pierce, and
  Narayan]{haeberlen2011differential}
Andreas Haeberlen, Benjamin~C Pierce, and Arjun Narayan.
\newblock Differential privacy under fire.
\newblock In \emph{USENIX Security Symposium}, 2011.

\bibitem[Halevi et~al.(2011)Halevi, Lindell, and Pinkas]{halevi2011secure}
Shai Halevi, Yehuda Lindell, and Benny Pinkas.
\newblock Secure computation on the web: Computing without simultaneous
  interaction.
\newblock In \emph{Annual Cryptology Conference}, pages 132--150. Springer,
  2011.

\bibitem[Hamer et~al.(2020)Hamer, Mohri, and Suresh]{hamer2020fedboost}
Jenny Hamer, Mehryar Mohri, and Ananda~Theertha Suresh.
\newblock Fedboost: A communication-efficient algorithm for federated learning.
\newblock In \emph{International Conference on Machine Learning}, pages
  3973--3983. PMLR, 2020.

\bibitem[Han et~al.(2015)Han, Mao, and Dally]{han2015deep}
Song Han, Huizi Mao, and William~J Dally.
\newblock Deep compression: Compressing deep neural networks with pruning,
  trained quantization and huffman coding.
\newblock \emph{arXiv preprint arXiv:1510.00149}, 2015.

\bibitem[Han et~al.(2018)Han, \"Ozg\"ur, and Weissman]{han2018}
Yanjun Han, Ayfer \"Ozg\"ur, and Tsachy Weissman.
\newblock Geometric lower bounds for distributed parameter estimation under
  communication constraints.
\newblock In \emph{Proceedings of Machine Learning Research}, pages 1--26, 75,
  2018.

\bibitem[Hard et~al.(2018)Hard, Rao, Mathews, Beaufays, Augenstein, Eichner,
  Kiddon, and Ramage]{hard18gboard}
Andrew Hard, Kanishka Rao, Rajiv Mathews, Fran{\c{c}}oise Beaufays, Sean
  Augenstein, Hubert Eichner, Chlo{\'{e}} Kiddon, and Daniel Ramage.
\newblock Federated learning for mobile keyboard prediction.
\newblock \emph{arXiv preprint 1811.03604}, 2018.

\bibitem[Hardt et~al.(2016)Hardt, Price, and Srebro]{hardt2016}
Moritz Hardt, Eric Price, and Nathan Srebro.
\newblock Equality of opportunity in supervised learning.
\newblock In \emph{Advances in Neural Information Processing Systems}, 2016.

\bibitem[Hardy et~al.(2017)Hardy, Henecka, Ivey-Law, Nock, Patrini, Smith, and
  Thorne]{Hardy2017-da}
Stephen Hardy, Wilko Henecka, Hamish Ivey-Law, Richard Nock, Giorgio Patrini,
  Guillaume Smith, and Brian Thorne.
\newblock Private federated learning on vertically partitioned data via entity
  resolution and additively homomorphic encryption.
\newblock \emph{arXiv preprint arXiv:1711.10677}, 2017.

\bibitem[Hashimoto et~al.(2018)Hashimoto, Srivastava, Namkoong, and
  Liang]{hashimoto2018fairness}
Tatsunori Hashimoto, Megha Srivastava, Hongseok Namkoong, and Percy Liang.
\newblock Fairness without demographics in repeated loss minimization.
\newblock In \emph{International Conference on Machine Learning}, pages
  1934--1943, 2018.

\bibitem[He et~al.(2019)He, Tan, Tang, Qiu, and Liu]{he2019central}
Chaoyang He, Conghui Tan, Hanlin Tang, Shuang Qiu, and Ji~Liu.
\newblock Central server free federated learning over single-sided trust social
  networks.
\newblock \emph{arXiv preprint arXiv:1910.04956}, 2019.

\bibitem[He et~al.(2020{\natexlab{a}})He, Annavaram, and
  Avestimehr]{FedGKT2020}
Chaoyang He, Murali Annavaram, and Salman Avestimehr.
\newblock Group knowledge transfer: Federated learning of large cnns at the
  edge.
\newblock In \emph{Advances in Neural Information Processing Systems 34},
  2020{\natexlab{a}}.

\bibitem[He et~al.(2020{\natexlab{b}})He, Annavaram, and
  Avestimehr]{he2020fednas}
Chaoyang He, Murali Annavaram, and Salman Avestimehr.
\newblock Fednas: Federated deep learning via neural architecture search.
\newblock 2020{\natexlab{b}}.

\bibitem[He et~al.(2020{\natexlab{c}})He, Li, So, Zeng, Zhang, Wang, Wang,
  Vepakomma, Singh, Qiu, Zhu, Wang, Shen, Zhao, Kang, Liu, Raskar, Yang,
  Annavaram, and Avestimehr]{he2020fedml}
Chaoyang He, Songze Li, Jinhyun So, Xiao Zeng, Mi~Zhang, Hongyi Wang, Xiaoyang
  Wang, Praneeth Vepakomma, Abhishek Singh, Hang Qiu, Xinghua Zhu, Jianzong
  Wang, Li~Shen, Peilin Zhao, Yan Kang, Yang Liu, Ramesh Raskar, Qiang Yang,
  Murali Annavaram, and Salman Avestimehr.
\newblock Fedml: A research library and benchmark for federated machine
  learning, 2020{\natexlab{c}}.

\bibitem[He et~al.(2020{\natexlab{d}})He, Ye, Shen, and Zhang]{MiLeNAS2020}
Chaoyang He, Haishan Ye, Li~Shen, and Tong Zhang.
\newblock Milenas: Efficient neural architecture search via mixed-level
  reformulation.
\newblock In \emph{Proceedings of IEEE Conference on Computer Vision and
  Pattern Recognition (CVPR)}, 2020{\natexlab{d}}.

\bibitem[He et~al.(2018)He, Bian, and Jaggi]{he2018cola}
Lie He, An~Bian, and Martin Jaggi.
\newblock {COLA}: Decentralized linear learning.
\newblock In \emph{NeurIPS 2018 - Advances in Neural Information Processing
  Systems 31}, 2018.

\bibitem[H{\'e}bert-Johnson et~al.(2018)H{\'e}bert-Johnson, Kim, Reingold, and
  Rothblum]{hebert2018multicalibration}
{\'U}rsula H{\'e}bert-Johnson, Michael Kim, Omer Reingold, and Guy Rothblum.
\newblock Multicalibration: Calibration for the (computationally-identifiable)
  masses.
\newblock In \emph{International Conference on Machine Learning}, pages
  1944--1953, 2018.

\bibitem[HElib()]{HElib}
HElib.
\newblock {HE}lib.
\newblock \url{https://github.com/homenc/HElib}, October 2019.

\bibitem[Hoffman et~al.(2018)Hoffman, Mohri, and Zhang]{hoffman2018algorithms}
Judy Hoffman, Mehryar Mohri, and Ningshan Zhang.
\newblock Algorithms and theory for multiple-source adaptation.
\newblock In \emph{Advances in Neural Information Processing Systems}, pages
  8246--8256, 2018.

\bibitem[Horvath et~al.(2019)Horvath, Ho, Horvath, Sahu, Canini, and
  Richtarik]{horvath2019natural}
Samuel Horvath, Chen-Yu Ho, Ludovit Horvath, Atal~Narayan Sahu, Marco Canini,
  and Peter Richtarik.
\newblock Natural compression for distributed deep learning.
\newblock \emph{arXiv preprint arXiv:1905.10988}, 2019.

\bibitem[Hsieh et~al.(2019)Hsieh, Phanishayee, Mutlu, and
  Gibbons]{hsieh2019noniid}
Kevin Hsieh, Amar Phanishayee, Onur Mutlu, and Phillip~B. Gibbons.
\newblock The non-{IID} data quagmire of decentralized machine learning, 2019.
\newblock URL \url{https://arxiv.org/abs/1910.00189}.

\bibitem[Hsu et~al.(2019)Hsu, Qi, and Brown]{hsu2019measuring}
Tzu-Ming~Harry Hsu, Hang Qi, and Matthew Brown.
\newblock Measuring the effects of non-identical data distribution for
  federated visual classification.
\newblock \emph{arXiv preprint arXiv:1909.06335}, 2019.

\bibitem[Hu et~al.(2019)Hu, Liu, Kong, and Niu]{hu2019learning}
Yaochen Hu, Peng Liu, Linglong Kong, and Di~Niu.
\newblock Learning privately over distributed features: An admm sharing
  approach, 2019.

\bibitem[Huang et~al.(2015)Huang, Mitra, and Vaidya]{Huang2015a}
Zhenqi Huang, Sayan Mitra, and Nitin Vaidya.
\newblock {Differentially Private Distributed Optimization}.
\newblock In \emph{ICDCN}, 2015.

\bibitem[Huo et~al.(2018)Huo, Gu, and Huang]{huo2018training}
Zhouyuan Huo, Bin Gu, and Heng Huang.
\newblock Training neural networks using features replay.
\newblock In \emph{Advances in Neural Information Processing Systems}, pages
  6659--6668, 2018.

\bibitem[Intel(2012)]{intel2012architecture}
R~Intel.
\newblock Architecture instruction set extensions programming reference.
\newblock \emph{Intel Corporation, Feb}, 2012.

\bibitem[Ion et~al.(2017)Ion, Kreuter, Nergiz, Patel, Saxena, Seth, Shanahan,
  and Yung]{DBLP:journals/iacr/IonKNPSSSY17}
Mihaela Ion, Ben Kreuter, Erhan Nergiz, Sarvar Patel, Shobhit Saxena, Karn
  Seth, David Shanahan, and Moti Yung.
\newblock Private intersection-sum protocol with applications to attributing
  aggregate ad conversions.
\newblock \emph{{IACR} Cryptology ePrint Archive}, 2017:\penalty0 738, 2017.

\bibitem[Ion et~al.(2019)Ion, Kreuter, Nergiz, Patel, Raykova, Saxena, Seth,
  Shanahan, and Yung]{DBLP:journals/iacr/IonKNPRSSSY19}
Mihaela Ion, Ben Kreuter, Ahmet~Erhan Nergiz, Sarvar Patel, Mariana Raykova,
  Shobhit Saxena, Karn Seth, David Shanahan, and Moti Yung.
\newblock On deploying secure computing commercially: Private intersection-sum
  protocols and their business applications.
\newblock \emph{{IACR} Cryptology ePrint Archive}, 2019:\penalty0 723, 2019.

\bibitem[Ishai et~al.(2003)Ishai, Kilian, Nissim, and Petrank]{IKNP}
Yuval Ishai, Joe Kilian, Kobbi Nissim, and Erez Petrank.
\newblock Extending oblivious transfers efficiently.
\newblock In \emph{{CRYPTO}}, volume 2729 of \emph{Lecture Notes in Computer
  Science}, pages 145--161. Springer, 2003.

\bibitem[Jaderberg et~al.(2017)Jaderberg, Czarnecki, Osindero, Vinyals, Graves,
  Silver, and Kavukcuoglu]{jaderberg2017decoupled}
Max Jaderberg, Wojciech~Marian Czarnecki, Simon Osindero, Oriol Vinyals, Alex
  Graves, David Silver, and Koray Kavukcuoglu.
\newblock Decoupled neural interfaces using synthetic gradients.
\newblock In \emph{Proceedings of the 34th International Conference on Machine
  Learning-Volume 70}, pages 1627--1635. JMLR. org, 2017.

\bibitem[Jagielski et~al.(2018)Jagielski, Kearns, Mao, Oprea, Roth,
  Sharifi{-}Malvajerdi, and Ullman]{jagielski2018privatefair}
Matthew Jagielski, Michael~J. Kearns, Jieming Mao, Alina Oprea, Aaron Roth,
  Saeed Sharifi{-}Malvajerdi, and Jonathan Ullman.
\newblock Differentially private fair learning.
\newblock \emph{CoRR}, abs/1812.02696, 2018.
\newblock URL \url{http://arxiv.org/abs/1812.02696}.

\bibitem[Jagielski et~al.(2020)Jagielski, Ullman, and
  Oprea]{jagielski2020auditing}
Matthew Jagielski, Jonathan Ullman, and Alina Oprea.
\newblock Auditing differentially private machine learning: How private is
  private sgd?
\newblock \emph{Advances in Neural Information Processing Systems}, 33, 2020.

\bibitem[Jeong et~al.(2018)Jeong, Oh, Kim, Park, Bennis, and Kim]{FD}
Eunjeong Jeong, Seungeun Oh, Hyesung Kim, Jihong Park, Mehdi Bennis, and
  Seong{-}Lyun Kim.
\newblock Communication-efficient on-device machine learning: Federated
  distillation and augmentation under non-{IID} private data.
\newblock \emph{CoRR}, abs/1811.11479, 2018.
\newblock URL \url{http://arxiv.org/abs/1811.11479}.

\bibitem[Jia and Jafar(2019)]{Jia2019OnTC}
Zhuqing Jia and Syed~Ali Jafar.
\newblock On the capacity of secure distributed matrix multiplication.
\newblock \emph{ArXiv}, abs/1908.06957, 2019.

\bibitem[Jiang et~al.(2019)Jiang, Kone{\v{c}}n{\'y}, Rush, and
  Kannan]{jiang2019improving}
Yihan Jiang, Jakub Kone{\v{c}}n{\'y}, Keith Rush, and Sreeram Kannan.
\newblock Improving federated learning personalization via model agnostic meta
  learning.
\newblock \emph{arXiv preprint arXiv:1909.12488}, 2019.

\bibitem[{Kadhe} et~al.(2017){Kadhe}, {Garcia}, {Heidarzadeh}, {Rouayheb}, and
  {Sprintson}]{pirsideinfo}
S.~{Kadhe}, B.~{Garcia}, A.~{Heidarzadeh}, S.~E. {Rouayheb}, and
  A.~{Sprintson}.
\newblock Private information retrieval with side information: The single
  server case.
\newblock In \emph{2017 55th Annual Allerton Conference on Communication,
  Control, and Computing (Allerton)}, pages 1099--1106, Oct 2017.
\newblock \doi{10.1109/ALLERTON.2017.8262860}.

\bibitem[Kairouz et~al.(2014)Kairouz, Oh, and Viswanath]{kairouz2014extremal}
Peter Kairouz, Sewoong Oh, and Pramod Viswanath.
\newblock Extremal mechanisms for local differential privacy.
\newblock In Z.~Ghahramani, M.~Welling, C.~Cortes, N.~D. Lawrence, and K.~Q.
  Weinberger, editors, \emph{Advances in Neural Information Processing Systems
  27}, pages 2879--2887. Curran Associates, Inc., 2014.

\bibitem[Kairouz et~al.(2016)Kairouz, Bonawitz, and
  Ramage]{kairouz2016discrete}
Peter Kairouz, K.~A. Bonawitz, and Daniel Ramage.
\newblock Discrete distribution estimation under local privacy.
\newblock In \emph{International Conference on Machine Learning}, pages
  2436--2444, 2016.

\bibitem[Kairouz et~al.(2017)Kairouz, Oh, and
  Viswanath]{kairouz2017composition}
Peter Kairouz, Sewoong Oh, and Pramod Viswanath.
\newblock The composition theorem for differential privacy.
\newblock \emph{IEEE Transactions on Information Theory}, 63\penalty0
  (6):\penalty0 4037--4049, 2017.

\bibitem[Kairouz et~al.(2020)Kairouz, Liao, Huang, and
  Sankar]{kairouz20learning}
Peter Kairouz, Jiachun Liao, Chong Huang, and Lalitha Sankar.
\newblock Censored and fair universal representations using generative
  adversarial models.
\newblock \emph{arXiv preprint arXiv:1910.00411}, 2020.

\bibitem[Kairouz et~al.(2021{\natexlab{a}})Kairouz, Liu, and
  Steinke]{kairouz2021distributed}
Peter Kairouz, Ziyu Liu, and Thomas Steinke.
\newblock The distributed discrete gaussian mechanism for federated learning
  with secure aggregation, 2021{\natexlab{a}}.

\bibitem[Kairouz et~al.(2021{\natexlab{b}})Kairouz, McMahan, Song, Thakkar,
  Thakurta, and Xu]{kairouz2021practical}
Peter Kairouz, Brendan McMahan, Shuang Song, Om~Thakkar, Abhradeep Thakurta,
  and Zheng Xu.
\newblock Practical and private (deep) learning without sampling or shuffling,
  2021{\natexlab{b}}.

\bibitem[Kamishima et~al.(2011)Kamishima, Akaho, and
  Sakuma]{kamishima2011fairness}
Toshihiro Kamishima, Shotaro Akaho, and Jun Sakuma.
\newblock Fairness-aware learning through regularization approach.
\newblock In \emph{2011 IEEE 11th International Conference on Data Mining
  Workshops}, pages 643--650. IEEE, 2011.

\bibitem[Kang et~al.(2019{\natexlab{a}})Kang, Sun, Hendrycks, Brown, and
  Steinhardt]{kang2019testing}
Daniel Kang, Yi~Sun, Dan Hendrycks, Tom Brown, and Jacob Steinhardt.
\newblock Testing robustness against unforeseen adversaries.
\newblock \emph{arXiv preprint arXiv:1908.08016}, 2019{\natexlab{a}}.

\bibitem[Kang et~al.(2019{\natexlab{b}})Kang, Xiong, Niyato, Xie, and
  Zhang]{kang2019incentiveB}
Jiawen Kang, Zehui Xiong, Dusit Niyato, Shengli Xie, and Junshan Zhang.
\newblock Incentive mechanism for reliable federated learning: A joint
  optimization approach to combining reputation and contract theory.
\newblock \emph{IEEE Internet of Things Journal}, 2019{\natexlab{b}}.

\bibitem[Kang et~al.(2019{\natexlab{c}})Kang, Xiong, Niyato, Yu, Liang, and
  Kim]{kang19incentive}
Jiawen Kang, Zehui Xiong, Dusit Niyato, Han Yu, Ying{-}Chang Liang, and Dong~In
  Kim.
\newblock Incentive design for efficient federated learning in mobile networks:
  {A} contract theory approach.
\newblock In \emph{{IEEE} {VTS} Asia Pacific Wireless Communications Symposium,
  {APWCS} 2019, Singapore, August 28-30, 2019}, pages 1--5, 2019{\natexlab{c}}.

\bibitem[Karimi et~al.(2016)Karimi, Nutini, and Schmidt]{karimi2016linear}
Hamed Karimi, Julie Nutini, and Mark Schmidt.
\newblock Linear convergence of gradient and proximal-gradient methods under
  the {P}olyak-{\l}ojasiewicz condition.
\newblock In \emph{Joint European Conference on Machine Learning and Knowledge
  Discovery in Databases}, pages 795--811. Springer, 2016.

\bibitem[Karimireddy et~al.(2019)Karimireddy, Rebjock, Stich, and
  Jaggi]{karimireddy2019ef}
Sai~Praneeth Karimireddy, Quentin Rebjock, Sebastian Stich, and Martin Jaggi.
\newblock Error feedback fixes {S}ign{SGD} and other gradient compression
  schemes.
\newblock In \emph{ICML}, 2019.

\bibitem[Karimireddy et~al.(2020{\natexlab{a}})Karimireddy, Jaggi, Kale, Mohri,
  Reddi, Stich, and Suresh]{karimireddy2020mime}
Sai~Praneeth Karimireddy, Martin Jaggi, Satyen Kale, Mehryar Mohri, Sashank~J
  Reddi, Sebastian~U Stich, and Ananda~Theertha Suresh.
\newblock Mime: Mimicking centralized stochastic algorithms in federated
  learning.
\newblock \emph{arXiv preprint arXiv:2008.03606}, 2020{\natexlab{a}}.

\bibitem[Karimireddy et~al.(2020{\natexlab{b}})Karimireddy, Kale, Mohri, Reddi,
  Stich, and Suresh]{karimireddy2019scaffold}
Sai~Praneeth Karimireddy, Satyen Kale, Mehryar Mohri, Sashank Reddi, Sebastian
  Stich, and Ananda~Theertha Suresh.
\newblock Scaffold: Stochastic controlled averaging for federated learning.
\newblock In \emph{International Conference on Machine Learning}, pages
  5132--5143. PMLR, 2020{\natexlab{b}}.

\bibitem[Kasiviswanathan et~al.(2011)Kasiviswanathan, Lee, Nissim,
  Raskhodnikova, and Smith]{KLNRS11}
Shiva~Prasad Kasiviswanathan, Homin~K. Lee, Kobbi Nissim, Sofya Raskhodnikova,
  and Adam~D. Smith.
\newblock What can we learn privately?
\newblock \emph{{SIAM} J. Comput.}, 40\penalty0 (3):\penalty0 793--826, 2011.
\newblock URL \url{https://doi.org/10.1137/090756090}.

\bibitem[Kearns et~al.(2015)Kearns, Roth, Wu, and
  Yaroslavtsev]{kearns2015protected}
Michael~J. Kearns, Aaron Roth, Zhiwei~Steven Wu, and Grigory Yaroslavtsev.
\newblock Privacy for the protected (only).
\newblock \emph{CoRR}, abs/1506.00242, 2015.
\newblock URL \url{http://arxiv.org/abs/1506.00242}.

\bibitem[Khaled et~al.(2019{\natexlab{a}})Khaled, Mishchenko, and
  Richtárik]{khaled2019analysis}
Ahmed Khaled, Konstantin Mishchenko, and Peter Richtárik.
\newblock First analysis of local {GD} on heterogeneous data,
  2019{\natexlab{a}}.
\newblock URL \url{https://arxiv.org/abs/1909.04715}.

\bibitem[Khaled et~al.(2019{\natexlab{b}})Khaled, Mishchenko, and
  Richtárik]{khaled2019better}
Ahmed Khaled, Konstantin Mishchenko, and Peter Richtárik.
\newblock Better communication complexity for local {SGD}, 2019{\natexlab{b}}.
\newblock URL \url{https://arxiv.org/abs/1909.04746}.

\bibitem[Khodak et~al.(2019)Khodak, Balcan, and Talwalkar]{khodak19adaptive}
Mikhail Khodak, Maria-Florina Balcan, and Ameet Talwalkar.
\newblock Adaptive gradient-based meta-learning methods.
\newblock In \emph{Advances in Neural Information Processing Systems}, 2019.

\bibitem[Kifer and Machanavajjhala(2014)]{pufferfish}
Daniel Kifer and Ashwin Machanavajjhala.
\newblock Pufferfish: A framework for mathematical privacy definitions.
\newblock \emph{ACM Transactions on Database Systems}, 39\penalty0
  (1):\penalty0 3:1--3:36, 2014.

\bibitem[Kim et~al.(2017)Kim, Sun, Yu, and Jiang]{DBLP:conf/kdd/KimSYJ17}
Yejin Kim, Jimeng Sun, Hwanjo Yu, and Xiaoqian Jiang.
\newblock Federated tensor factorization for computational phenotyping.
\newblock In \emph{Proceedings of the 23rd {ACM} {SIGKDD} International
  Conference on Knowledge Discovery and Data Mining, Halifax, NS, Canada,
  August 13 - 17, 2017}, pages 887--895, 2017.
\newblock \doi{10.1145/3097983.3098118}.
\newblock URL \url{https://doi.org/10.1145/3097983.3098118}.

\bibitem[King et~al.(1995)King, Feng, and Sutherland]{king1995statlog}
Ross~D. King, Cao Feng, and Alistair Sutherland.
\newblock Stat{L}og: comparison of classification algorithms on large
  real-world problems.
\newblock \emph{Applied Artificial Intelligence an International Journal},
  9\penalty0 (3):\penalty0 289--333, 1995.

\bibitem[Koeberl et~al.(2014)Koeberl, Schulz, Sadeghi, and
  Varadharajan]{DBLP:conf/eurosys/KoeberlSSV14}
Patrick Koeberl, Steffen Schulz, Ahmad{-}Reza Sadeghi, and Vijay Varadharajan.
\newblock Trust{L}ite: a security architecture for tiny embedded devices.
\newblock In \emph{EuroSys}, pages 10:1--10:14. {ACM}, 2014.

\bibitem[Koh and Liang(2017)]{koh2017understanding}
Pang~Wei Koh and Percy Liang.
\newblock Understanding black-box predictions via influence functions.
\newblock In \emph{Proceedings of the 34th International Conference on Machine
  Learning-Volume 70}, pages 1885--1894. JMLR. org, 2017.

\bibitem[Koh et~al.(2018)Koh, Steinhardt, and Liang]{koh2018stronger}
Pang~Wei Koh, Jacob Steinhardt, and Percy Liang.
\newblock Stronger data poisoning attacks break data sanitization defenses.
\newblock \emph{arXiv preprint arXiv:1811.00741}, 2018.

\bibitem[Kohavi and John(1995)]{kohavi1995automatic}
Ron Kohavi and George~H John.
\newblock Automatic parameter selection by minimizing estimated error.
\newblock In \emph{Machine Learning Proceedings 1995}, pages 304--312.
  Elsevier, 1995.

\bibitem[Koloskova et~al.(2019)Koloskova, Stich, and Jaggi]{Koloskova2019}
Anastasia Koloskova, Sebastian~U Stich, and Martin Jaggi.
\newblock {Decentralized Stochastic Optimization and Gossip Algorithms with
  Compressed Communication}.
\newblock In \emph{ICML}, 2019.

\bibitem[Koloskova et~al.(2020{\natexlab{a}})Koloskova, Lin, Stich, and
  Jaggi]{koloskova2019deep}
Anastasia Koloskova, Tao Lin, Sebastian~U Stich, and Martin Jaggi.
\newblock Decentralized deep learning with arbitrary communication compression.
\newblock \emph{International Conference on Learning Representations (ICLR)},
  2020{\natexlab{a}}.

\bibitem[Koloskova et~al.(2020{\natexlab{b}})Koloskova, Loizou, Boreiri, Jaggi,
  and Stich]{decentralized_sgd2}
Anastasia Koloskova, Nicolas Loizou, Sadra Boreiri, Martin Jaggi, and
  Sebastian~U. Stich.
\newblock {A Unified Theory of Decentralized SGD with Changing Topology and
  Local Updates}.
\newblock In \emph{ICML}, 2020{\natexlab{b}}.

\bibitem[Kone{\v{c}}n{\'y} and Richt{\'a}rik(2018)]{konevcny2018randomized}
Jakub Kone{\v{c}}n{\'y} and Peter Richt{\'a}rik.
\newblock Randomized distributed mean estimation: Accuracy vs communication.
\newblock \emph{Frontiers in Applied Mathematics and Statistics}, 4:\penalty0
  62, 2018.

\bibitem[Kone{\v{c}}n{\'y} et~al.(2016)Kone{\v{c}}n{\'y}, McMahan, Yu,
  Richt{\'a}rik, Suresh, and Bacon]{konevcny2016federated}
Jakub Kone{\v{c}}n{\'y}, H~Brendan McMahan, Felix~X. Yu, Peter Richt{\'a}rik,
  Ananda~Theertha Suresh, and Dave Bacon.
\newblock Federated learning: Strategies for improving communication
  efficiency.
\newblock \emph{arXiv preprint arXiv:1610.05492}, 2016.

\bibitem[Kuppam et~al.(2019)Kuppam, McKenna, Pujol, Hay, Machanavajjhala, and
  Miklau]{kuppam2019fair}
Satya Kuppam, Ryan McKenna, David Pujol, Michael Hay, Ashwin Machanavajjhala,
  and Gerome Miklau.
\newblock Fair decision making using privacy-protected data.
\newblock \emph{CoRR}, abs/1905.12744, 2019.
\newblock URL \url{http://arxiv.org/abs/1905.12744}.

\bibitem[Kurakin et~al.(2016)Kurakin, Goodfellow, and
  Bengio]{kurakin2016adversarial}
Alexey Kurakin, Ian Goodfellow, and Samy Bengio.
\newblock Adversarial machine learning at scale.
\newblock \emph{arXiv preprint arXiv:1611.01236}, 2016.

\bibitem[Kushilevitz and Nisan(1997)]{Kushilevitz_Nisan_cc}
Eyal Kushilevitz and Noam Nisan.
\newblock \emph{Communication Complexity}.
\newblock Cambridge University Press, New York, NY, USA, 1997.
\newblock ISBN 0-521-56067-5.

\bibitem[Kushilevitz and Ostrovsky(1997)]{Kushilevitz97replicationis}
Eyal Kushilevitz and Rafail Ostrovsky.
\newblock Replication is not needed: Single database, computationally-private
  information retrieval.
\newblock In \emph{In Proc. of the 38th Annu. IEEE Symp. on Foundations of
  Computer Science}, pages 364--373, 1997.

\bibitem[Kusner et~al.(2017)Kusner, Loftus, Russell, and
  Silva]{kusner2017counterfactual}
Matt~J Kusner, Joshua Loftus, Chris Russell, and Ricardo Silva.
\newblock Counterfactual fairness.
\newblock In \emph{Advances in Neural Information Processing Systems}, pages
  4066--4076, 2017.

\bibitem[Kwon et~al.(2016)Kwon, Lazar, Devadas, and Ford]{kwon2016riffle}
Albert Kwon, David Lazar, Srinivas Devadas, and Bryan Ford.
\newblock Riffle.
\newblock \emph{Proceedings on Privacy Enhancing Technologies}, 2016\penalty0
  (2):\penalty0 115--134, 2016.

\bibitem[Laguel et~al.(2020)Laguel, Pillutla, Malick, and
  Harchaoui]{fair_quantile}
Yassine Laguel, Krishna Pillutla, Jérôme Malick, and Zaid Harchaoui.
\newblock {Device Heterogeneity in Federated Learning: A Superquantile
  Approach}.
\newblock \emph{arXiv preprint arXiv:2002.11223}, 2020.

\bibitem[Lake et~al.(2017)Lake, Salakhutdinov, Gross, and
  Tenenbaum]{lake11omniglot}
Brenden~M. Lake, Ruslan Salakhutdinov, Jason Gross, and Joshua~B. Tenenbaum.
\newblock One shot learning of simple visual concepts.
\newblock In \emph{Proceedings of the Conference of the Cognitive Science
  Society (CogSci)}, 2017.

\bibitem[Lalitha et~al.(2019{\natexlab{a}})Lalitha, Kilinc, Javidi, and
  Koushanfar]{Lalitha2019}
Anusha Lalitha, Osman~Cihan Kilinc, Tara Javidi, and Farinaz Koushanfar.
\newblock {Peer-to-peer Federated Learning on Graphs}.
\newblock Technical report, arXiv:1901.11173, 2019{\natexlab{a}}.

\bibitem[Lalitha et~al.(2019{\natexlab{b}})Lalitha, Wang, Kilinc, Lu, Javidi,
  and Koushanfar]{BayesFL}
Anusha Lalitha, Xinghan Wang, Osman Kilinc, Yongxi Lu, Tara Javidi, and Farinaz
  Koushanfar.
\newblock Decentralized {B}ayesian learning over graphs.
\newblock \emph{arXiv preprint: 1905.10466}, 2019{\natexlab{b}}.

\bibitem[Lamport et~al.(1982)Lamport, Shostak, and Pease]{lamport1982byzantine}
Leslie Lamport, Robert Shostak, and Marshall Pease.
\newblock The {B}yzantine generals problem.
\newblock \emph{ACM Transactions on Programming Languages and Systems
  (TOPLAS)}, 4\penalty0 (3):\penalty0 382--401, 1982.

\bibitem[Lan(2012)]{Lan2012}
Guanghui Lan.
\newblock An optimal method for stochastic composite optimization.
\newblock \emph{Mathematical Programming}, 133\penalty0 (1):\penalty0 365--397,
  Jun 2012.
\newblock ISSN 1436-4646.
\newblock \doi{10.1007/s10107-010-0434-y}.
\newblock URL \url{https://doi.org/10.1007/s10107-010-0434-y}.

\bibitem[Lapets et~al.(2016)Lapets, Volgushev, Bestavros, Jansen, and
  Varia]{DBLP:conf/secdev/LapetsVBJV16}
Andrei Lapets, Nikolaj Volgushev, Azer Bestavros, Frederick Jansen, and Mayank
  Varia.
\newblock Secure {MPC} for analytics as a web application.
\newblock In \emph{SecDev}, pages 73--74. {IEEE} Computer Society, 2016.

\bibitem[L{\'{e}}cuyer et~al.(2019)L{\'{e}}cuyer, Atlidakis, Geambasu, Hsu, and
  Jana]{DBLP:conf/sp/LecuyerAG0J19}
Mathias L{\'{e}}cuyer, Vaggelis Atlidakis, Roxana Geambasu, Daniel Hsu, and
  Suman Jana.
\newblock Certified robustness to adversarial examples with differential
  privacy.
\newblock In \emph{2019 {IEEE} Symposium on Security and Privacy, {SP} 2019,
  San Francisco, CA, USA, May 19-23, 2019}, pages 656--672, 2019.
\newblock \doi{10.1109/SP.2019.00044}.
\newblock URL \url{https://doi.org/10.1109/SP.2019.00044}.

\bibitem[Lepoint et~al.(2020)Lepoint, Patel, Raykova, Seth, and
  Trieu]{DBLP:journals/iacr/LepointPRST20}
Tancr{\`{e}}de Lepoint, Sarvar Patel, Mariana Raykova, Karn Seth, and Ni~Trieu.
\newblock Private join and compute from {PIR} with default.
\newblock \emph{{IACR} Cryptol. ePrint Arch.}, 2020:\penalty0 1011, 2020.

\bibitem[Leroy et~al.(2018)Leroy, Coucke, Lavril, Gisselbrecht, and
  Dureau]{leroy2018federated}
David Leroy, Alice Coucke, Thibaut Lavril, Thibault Gisselbrecht, and Joseph
  Dureau.
\newblock Federated learning for keyword spotting.
\newblock \emph{arXiv preprint arXiv:{1810.05512}}, 2018.

\bibitem[Li et~al.(2019{\natexlab{a}})Li, Khodak, Caldas, and
  Talwalkar]{li19dpmeta}
Jeffrey Li, Mikhail Khodak, Sebastian Caldas, and Ameet Talwalkar.
\newblock Differentially private meta-learning.
\newblock \emph{arXiv preprint arXiv:1909.05830}, 2019{\natexlab{a}}.

\bibitem[Li et~al.(2018)Li, Sahu, Zaheer, Sanjabi, Talwalkar, and
  Smith]{li2018federated}
Tian Li, Anit~Kumar Sahu, Manzil Zaheer, Maziar Sanjabi, Ameet Talwalkar, and
  Virginia Smith.
\newblock Federated optimization in heterogeneous networks, 2018.
\newblock URL \url{https://arxiv.org/abs/1812.06127}.

\bibitem[Li et~al.(2019{\natexlab{b}})Li, Sahu, Talwalkar, and
  Smith]{li2019federated}
Tian Li, Anit~Kumar Sahu, Ameet Talwalkar, and Virginia Smith.
\newblock Federated learning: Challenges, methods, and future directions,
  2019{\natexlab{b}}.

\bibitem[Li et~al.(2019{\natexlab{c}})Li, Sanjabi, and Smith]{li2019fair}
Tian Li, Maziar Sanjabi, and Virginia Smith.
\newblock Fair resource allocation in federated learning.
\newblock \emph{arXiv preprint arXiv:1905.10497}, 2019{\natexlab{c}}.

\bibitem[Li et~al.(2019{\natexlab{d}})Li, Huang, Yang, Wang, and
  Zhang]{li2019convergence}
Xiang Li, Kaixuan Huang, Wenhao Yang, Shusen Wang, and Zhihua Zhang.
\newblock On the convergence of {F}ed{A}vg on non-{IID} data.
\newblock \emph{arXiv preprint arXiv:1907.02189}, 2019{\natexlab{d}}.

\bibitem[Li et~al.(2019{\natexlab{e}})Li, Yang, Wang, and
  Zhang]{li2019communication}
Xiang Li, Wenhao Yang, Shusen Wang, and Zhihua Zhang.
\newblock Communication efficient decentralized training with multiple local
  updates.
\newblock \emph{arXiv preprint arXiv:1910.09126}, 2019{\natexlab{e}}.

\bibitem[Lian et~al.(2017)Lian, Zhang, Zhang, Hsieh, Zhang, and Liu]{Lian2017b}
Xiangru Lian, Ce~Zhang, Huan Zhang, Cho-Jui Hsieh, Wei Zhang, and Ji~Liu.
\newblock {Can Decentralized Algorithms Outperform Centralized Algorithms? A
  Case Study for Decentralized Parallel Stochastic Gradient Descent}.
\newblock In \emph{NIPS}, 2017.

\bibitem[Lian et~al.(2018)Lian, Zhang, Zhang, and Liu]{Lian2018}
Xiangru Lian, Wei Zhang, Ce~Zhang, and Ji~Liu.
\newblock {Asynchronous Decentralized Parallel Stochastic Gradient Descent}.
\newblock In \emph{ICML}, 2018.

\bibitem[libsnark()]{libsnark}
libsnark.
\newblock {libsnark}: a c++ library for {zkSNARK} proofs.
\newblock \url{https://github.com/scipr-lab/libsnark}, December 2019.

\bibitem[Lie and Maniatis(2017)]{lie2017glimmers}
David Lie and Petros Maniatis.
\newblock Glimmers: Resolving the privacy/trust quagmire.
\newblock In \emph{Proceedings of the 16th Workshop on Hot Topics in Operating
  Systems}, pages 94--99. ACM, 2017.

\bibitem[Lin et~al.(2016)Lin, Talathi, and Annapureddy]{lin2016fixed}
Darryl Lin, Sachin Talathi, and Sreekanth Annapureddy.
\newblock Fixed point quantization of deep convolutional networks.
\newblock In \emph{International Conference on Machine Learning}, pages
  2849--2858, 2016.

\bibitem[Lin et~al.(2020)Lin, Stich, and Jaggi]{lin2018don}
Tao Lin, Sebastian~U Stich, and Martin Jaggi.
\newblock Don't use large mini-batches, use local {SGD}.
\newblock \emph{International Conference on Learning Representations (ICLR)},
  2020.

\bibitem[Lin et~al.(2017)Lin, Han, Mao, Wang, and Dally]{lin2017deep}
Yujun Lin, Song Han, Huizi Mao, Yu~Wang, and William~J Dally.
\newblock Deep gradient compression: Reducing the communication bandwidth for
  distributed training.
\newblock \emph{arXiv preprint arXiv:1712.01887}, 2017.

\bibitem[Little(1993)]{little1993poststratification}
R.~J.~A. Little.
\newblock Post-stratification: A modeler's perspective.
\newblock \emph{Journal of the American Statistical Association}, 88\penalty0
  (423):\penalty0 1001--1012, 1993.
\newblock ISSN 01621459.

\bibitem[Liu et~al.(2018{\natexlab{a}})Liu, Simonyan, and Yang]{liu2018darts}
Hanxiao Liu, Karen Simonyan, and Yiming Yang.
\newblock {DARTS}: Differentiable architecture search.
\newblock \emph{arXiv preprint arXiv:1806.09055}, 2018{\natexlab{a}}.

\bibitem[Liu et~al.(2018{\natexlab{b}})Liu, Dolan-Gavitt, and
  Garg]{liu2018fine}
Kang Liu, Brendan Dolan-Gavitt, and Siddharth Garg.
\newblock Fine-pruning: Defending against backdooring attacks on deep neural
  networks.
\newblock In \emph{International Symposium on Research in Attacks, Intrusions,
  and Defenses}, pages 273--294. Springer, 2018{\natexlab{b}}.

\bibitem[Liu and Oh(2019)]{liu2019minimax}
Xiyang Liu and Sewoong Oh.
\newblock Minimax rates of estimating approximate differential privacy.
\newblock \emph{arXiv preprint arXiv:1905.10335}, 2019.

\bibitem[Liu et~al.(2019)Liu, Kang, Zhang, Li, Cheng, Chen, Hong, and
  Yang]{liuVFL}
Yang Liu, Yan Kang, Xinwei Zhang, Liping Li, Yong Cheng, Tianjian Chen, Mingyi
  Hong, and Qiang Yang.
\newblock A communication efficient vertical federated learning framework.
\newblock \emph{CoRR}, abs/1912.11187, 2019.
\newblock URL \url{http://arxiv.org/abs/1912.11187}.

\bibitem[{Liu} et~al.(2020){Liu}, {Kang}, {Xing}, {Chen}, and
  {Yang}]{liu2018secure}
Yang {Liu}, Yan {Kang}, Chaoping {Xing}, Tianjian {Chen}, and Qiang {Yang}.
\newblock A secure federated transfer learning framework.
\newblock \emph{IEEE Intelligent Systems}, 35\penalty0 (4):\penalty0 70--82,
  2020.
\newblock \doi{10.1109/MIS.2020.2988525}.

\bibitem[Liu et~al.(2020{\natexlab{a}})Liu, Yi, and Chen]{liu2020backdoor}
Yang Liu, Zhihao Yi, and Tianjian Chen.
\newblock Backdoor attacks and defenses in feature-partitioned collaborative
  learning, 2020{\natexlab{a}}.

\bibitem[Liu et~al.(2018{\natexlab{c}})Liu, Ma, Aafer, Lee, Zhai, Wang, and
  Zhang]{DBLP:conf/ndss/LiuMALZW018}
Yingqi Liu, Shiqing Ma, Yousra Aafer, Wen{-}Chuan Lee, Juan Zhai, Weihang Wang,
  and Xiangyu Zhang.
\newblock Trojaning attack on neural networks.
\newblock In \emph{25th Annual Network and Distributed System Security
  Symposium, {NDSS} 2018, San Diego, California, USA, February 18-21, 2018},
  2018{\natexlab{c}}.
\newblock URL
  \url{http://wp.internetsociety.org/ndss/wp-content/uploads/sites/25/2018/02/ndss2018\_03A-5\_Liu\_paper.pdf}.

\bibitem[Liu et~al.(2020{\natexlab{b}})Liu, Suresh, Yu, Kumar, and
  Riley]{liu2020learning}
Yuhan Liu, Ananda~Theertha Suresh, Felix Xinnan~X Yu, Sanjiv Kumar, and Michael
  Riley.
\newblock Learning discrete distributions: user vs item-level privacy.
\newblock \emph{Advances in Neural Information Processing Systems}, 33,
  2020{\natexlab{b}}.

\bibitem[Ludwig et~al.(2020)Ludwig, Baracaldo, Thomas, Zhou, Anwar, Rajamoni,
  Ong, Radhakrishnan, Verma, Sinn, et~al.]{IBMFL}
Heiko Ludwig, Nathalie Baracaldo, Gegi Thomas, Yi~Zhou, Ali Anwar, Shashank
  Rajamoni, Yuya Ong, Jayaram Radhakrishnan, Ashish Verma, Mathieu Sinn, et~al.
\newblock {IBM} federated learning: An enterprise framework white paper {V}0.1.
\newblock \emph{arXiv preprint arXiv:2007.10987}, 2020.

\bibitem[Luo et~al.(2019)Luo, Wu, Luo, Huang, Huang, Liu, and
  Yang]{luo2019real}
Jiahuan Luo, Xueyang Wu, Yun Luo, Anbu Huang, Yunfeng Huang, Yang Liu, and
  Qiang Yang.
\newblock Real-world image datasets for federated learning.
\newblock \emph{arXiv preprint arXiv:1910.11089}, 2019.

\bibitem[Luo et~al.(2018)Luo, Tian, Qin, Chen, and Liu]{luo2018neural}
Renqian Luo, Fei Tian, Tao Qin, Enhong Chen, and Tie-Yan Liu.
\newblock Neural architecture optimization.
\newblock In \emph{Advances in neural information processing systems}, pages
  7816--7827, 2018.

\bibitem[Lyu et~al.(2020)Lyu, Yu, Nandakumar, Li, Ma, Jin, Yu, and
  Ng]{Lyu-et-al:2020TPDS}
Lingjuan Lyu, Jiangshan Yu, Karthik Nandakumar, Yitong Li, Xingjun Ma, Jiong
  Jin, Han Yu, and Kee~Siong Ng.
\newblock Towards fair and privacy-preserving federated deep models.
\newblock \emph{IEEE Transactions on Parallel and Distributed Systems},
  31\penalty0 (11):\penalty0 2524--2541, 2020.

\bibitem[Ma et~al.(2019{\natexlab{a}})Ma, Zhang, Lou, Ho, Xiong, and
  Jiang]{ma19cikm}
Jing Ma, Qiuchen Zhang, Jian Lou, Joyce Ho, Li~Xiong, and Xiaoqian Jiang.
\newblock Privacy-preserving tensor factorization for collaborative health data
  analysis.
\newblock In \emph{ACM CIKM}, volume~2, 2019{\natexlab{a}}.

\bibitem[Ma et~al.(2019{\natexlab{b}})Ma, Zhu, and Hsu]{MZH19}
Yuzhe Ma, Xiaojin Zhu, and Justin Hsu.
\newblock Data poisoning against differentially-private learners: Attacks and
  defenses.
\newblock In \emph{International Joint Conference on Artificial Intelligence
  (IJCAI), Macao, China}, 2019{\natexlab{b}}.
\newblock URL \url{https://arxiv.org/abs/1903.09860}.

\bibitem[Madras et~al.(2018)Madras, Creager, Pitassi, and Zemel]{Madras2018}
David Madras, Elliot Creager, Toniann Pitassi, and Richard Zemel.
\newblock Learning adversarially fair and transferable representations.
\newblock In \emph{ICML}, 2018.

\bibitem[Madry et~al.(2017)Madry, Makelov, Schmidt, Tsipras, and
  Vladu]{madry2017towards}
Aleksander Madry, Aleksandar Makelov, Ludwig Schmidt, Dimitris Tsipras, and
  Adrian Vladu.
\newblock Towards deep learning models resistant to adversarial attacks.
\newblock \emph{ICLR}, 2017.

\bibitem[Mansour et~al.(2009{\natexlab{a}})Mansour, Mohri, and
  Rostamizadeh]{mansour2009domain}
Yishay Mansour, Mehryar Mohri, and Afshin Rostamizadeh.
\newblock Domain adaptation: Learning bounds and algorithms.
\newblock \emph{arXiv preprint arXiv:0902.3430}, 2009{\natexlab{a}}.

\bibitem[Mansour et~al.(2009{\natexlab{b}})Mansour, Mohri, and
  Rostamizadeh]{mansour2009domainb}
Yishay Mansour, Mehryar Mohri, and Afshin Rostamizadeh.
\newblock Domain adaptation with multiple sources.
\newblock In \emph{Advances in neural information processing systems}, pages
  1041--1048, 2009{\natexlab{b}}.

\bibitem[Mansour et~al.(2020{\natexlab{a}})Mansour, Mohri, Ro, and
  Suresh]{mansour2020three}
Yishay Mansour, Mehryar Mohri, Jae Ro, and Ananda~Theertha Suresh.
\newblock Three approaches for personalization with applications to federated
  learning.
\newblock \emph{arXiv preprint arXiv:2002.10619}, 2020{\natexlab{a}}.

\bibitem[Mansour et~al.(2020{\natexlab{b}})Mansour, Mohri, Suresh, and
  Wu]{mansour2020theory}
Yishay Mansour, Mehryar Mohri, Ananda~Theertha Suresh, and Ke~Wu.
\newblock A theory of multiple-source adaptation with limited target labeled
  data.
\newblock \emph{arXiv preprint arXiv:2007.09762}, 2020{\natexlab{b}}.

\bibitem[Martin et~al.(2019)Martin, Kanai, Kamatani, Okada, Neale, and
  Daly]{martin2019current}
Alicia~R Martin, Masahiro Kanai, Yoichiro Kamatani, Yukinori Okada, Benjamin~M
  Neale, and Mark~J Daly.
\newblock Current clinical use of polygenic scores will risk exacerbating
  health disparities.
\newblock \emph{BioRxiv}, page 441261, 2019.

\bibitem[McMahan and Ramage(2017)]{flblog17}
H~Brendan McMahan and Daniel Ramage.
\newblock Federated learning: Collaborative machine learning without
  centralized training data, April 2017.
\newblock URL
  \url{https://ai.googleblog.com/2017/04/federated-learning-collaborative.html}.
\newblock Google AI Blog.

\bibitem[McMahan and Streeter(2010)]{mcmahan2010adaptive}
H~Brendan McMahan and Matthew Streeter.
\newblock Adaptive bound optimization for online convex optimization.
\newblock \emph{arXiv preprint arXiv:1002.4908}, 2010.

\bibitem[McMahan et~al.(1812)McMahan, Andrew, Erlingsson, Chien, Mironov,
  Papernot, and Kairouz]{mcmahan2018general}
H~Brendan McMahan, Galen Andrew, Ulfar Erlingsson, Steve Chien, Ilya Mironov,
  Nicolas Papernot, and Peter Kairouz.
\newblock A general approach to adding differential privacy to iterative
  training procedures. dec 2018.
\newblock \emph{URL https://arxiv. org/abs}, 1812.

\bibitem[McMahan et~al.(2017 (original version on arxiv Feb. 2016))McMahan,
  Moore, Ramage, Hampson, and y~Arcas]{mcmahan17fedavg}
H~Brendan McMahan, Eider Moore, Daniel Ramage, Seth Hampson, and Blaise~Aguera
  y~Arcas.
\newblock Communication-efficient learning of deep networks from decentralized
  data.
\newblock In \emph{Proceedings of the 20th International Conference on
  Artificial Intelligence and Statistics}, pages 1273--1282, 2017 (original
  version on arxiv Feb. 2016).

\bibitem[McMahan et~al.(2018)McMahan, Ramage, Talwar, and Zhang]{mcmahan18dplm}
H~Brendan McMahan, Daniel Ramage, Kunal Talwar, and Li~Zhang.
\newblock Learning differentially private recurrent language models.
\newblock In \emph{International Conference on Learning Representations
  (ICLR)}, 2018.

\bibitem[McSherry and Talwar(2007)]{mcsherry2007mechanism}
Frank McSherry and Kunal Talwar.
\newblock Mechanism design via differential privacy.
\newblock In \emph{FOCS}, pages 94--103, 2007.

\bibitem[Mei and Zhu(2015)]{mei2015using}
Shike Mei and Xiaojin Zhu.
\newblock Using machine teaching to identify optimal training-set attacks on
  machine learners.
\newblock In \emph{Twenty-Ninth AAAI Conference on Artificial Intelligence},
  2015.

\bibitem[Melis et~al.(2018)Melis, Song, De~Cristofaro, and
  Shmatikov]{melis2018exploiting}
Luca Melis, Congzheng Song, Emiliano De~Cristofaro, and Vitaly Shmatikov.
\newblock Exploiting unintended feature leakage in collaborative learning.
\newblock \emph{arXiv preprint arXiv:1805.04049}, 2018.

\bibitem[Mhamdi et~al.(2018)Mhamdi, Guerraoui, and Rouault]{mhamdi2018hidden}
El~Mahdi~El Mhamdi, Rachid Guerraoui, and S{\'e}bastien Rouault.
\newblock The hidden vulnerability of distributed learning in {B}yzantium.
\newblock In \emph{ICML}, 2018.

\bibitem[Micali(2000)]{DBLP:journals/siamcomp/Micali00}
Silvio Micali.
\newblock Computationally sound proofs.
\newblock \emph{{SIAM} J. Comput.}, 30\penalty0 (4):\penalty0 1253--1298, 2000.

\bibitem[Mireshghallah et~al.(2020)Mireshghallah, Taram, , Vepakomma, Singh,
  Raskar, and Hadi]{pdlSurvey}
Fatemehsadat Mireshghallah, Mohammadkazem Taram, , Praneeth Vepakomma, Abhishek
  Singh, Ramesh Raskar, and Esmaeilzadeh Hadi.
\newblock Privacy in deep learning: A survey.
\newblock \emph{arXiv preprint arXiv:2004.12254}, 2020.

\bibitem[Mironov(2012)]{mironov2012significance}
Ilya Mironov.
\newblock On significance of the least significant bits for differential
  privacy.
\newblock In \emph{Proceedings of the 2012 ACM conference on Computer and
  communications security}, pages 650--661. ACM, 2012.

\bibitem[Mironov(2017)]{mironov2017renyi}
Ilya Mironov.
\newblock R{\'e}nyi differential privacy.
\newblock In \emph{2017 IEEE 30th Computer Security Foundations Symposium
  (CSF)}, pages 263--275. IEEE, 2017.

\bibitem[Mironov et~al.(2009)Mironov, Pandey, Reingold, and
  Vadhan]{Mironov-CDP}
Ilya Mironov, Omkant Pandey, Omer Reingold, and Salil Vadhan.
\newblock Computational differential privacy.
\newblock In \emph{Advances in Cryptology---CRYPTO}, pages 126--142, 2009.

\bibitem[Mironov et~al.(2019)Mironov, Talwar, and Zhang]{mironov2019r}
Ilya Mironov, Kunal Talwar, and Li~Zhang.
\newblock R$\backslash$'enyi differential privacy of the sampled {G}aussian
  mechanism.
\newblock \emph{arXiv preprint arXiv:1908.10530}, 2019.

\bibitem[Mitchell et~al.(2018)Mitchell, Potash, and
  Barocas]{mitchell2018prediction}
Shira Mitchell, Eric Potash, and Solon Barocas.
\newblock Prediction-based decisions and fairness: A catalogue of choices,
  assumptions, and definitions.
\newblock \emph{arXiv preprint arXiv:1811.07867}, 2018.

\bibitem[Mnih and Hinton(2012)]{mnih2012learning}
Volodymyr Mnih and Geoffrey~E Hinton.
\newblock Learning to label aerial images from noisy data.
\newblock In \emph{Proceedings of the 29th International conference on machine
  learning (ICML-12)}, pages 567--574, 2012.

\bibitem[Mohassel and Zhang(2017)]{secureml}
Payman Mohassel and Yupeng Zhang.
\newblock Secure{ML}: {A} system for scalable privacy-preserving machine
  learning.
\newblock In \emph{{IEEE} Symposium on Security and Privacy}, pages 19--38.
  {IEEE} Computer Society, 2017.

\bibitem[Mohri et~al.(2019)Mohri, Sivek, and Suresh]{Mohri2019}
Mehryar Mohri, Gary Sivek, and Ananda~Theertha Suresh.
\newblock {Agnostic Federated Learning}.
\newblock In \emph{ICML}, 2019.

\bibitem[Moreno-Torres et~al.(2012)Moreno-Torres, Raeder, Alaiz-Rodr\'{\i}Guez,
  Chawla, and Herrera]{torres2012unifying}
Jose~G. Moreno-Torres, Troy Raeder, Roc\'{\i}O Alaiz-Rodr\'{\i}Guez, Nitesh~V.
  Chawla, and Francisco Herrera.
\newblock A unifying view on dataset shift in classification.
\newblock \emph{Pattern Recogn.}, 45\penalty0 (1), January 2012.

\bibitem[{Musketeer}(2019)]{musketeer19mfg}
{Musketeer}.
\newblock Musketeer: About, 2019.
\newblock URL \url{http://musketeer.eu/project/}.
\newblock Retrieved Aug 2019.

\bibitem[Naim et~al.(2019)Naim, Ye, and Rouayheb]{onoffisit}
Carolina Naim, Fangwei Ye, and Salim~El Rouayheb.
\newblock {ON-OFF} privacy with correlated requests.
\newblock In \emph{2019 IEEE International Symposium on Information Theory
  (ISIT)}, July 2019.

\bibitem[Natarajan et~al.(2013)Natarajan, Dhillon, Ravikumar, and
  Tewari]{natarajan2013learning}
Nagarajan Natarajan, Inderjit~S Dhillon, Pradeep~K Ravikumar, and Ambuj Tewari.
\newblock Learning with noisy labels.
\newblock In \emph{Advances in neural information processing systems}, pages
  1196--1204, 2013.

\bibitem[Neglia et~al.(2020)Neglia, Xu, Towsley, and Calbi]{neglia2020}
Giovanni Neglia, Chuan Xu, Don Towsley, and Gianmarco Calbi.
\newblock Decentralized gradient methods: does topology matter?
\newblock In \emph{AISTATS}, 2020.

\bibitem[Nichol et~al.(2018)Nichol, Achiam, and Schulman]{nichol18reptile}
Alex Nichol, Joshua Achiam, and John Schulman.
\newblock On first-order meta-learning algorithms.
\newblock \emph{arXiv preprint arXiv:1803.02999}, 2018.

\bibitem[Nikolaenko et~al.(2013)Nikolaenko, Weinsberg, Ioannidis, Joye, Boneh,
  and Taft]{DBLP:conf/sp/NikolaenkoWIJBT13}
Valeria Nikolaenko, Udi Weinsberg, Stratis Ioannidis, Marc Joye, Dan Boneh, and
  Nina Taft.
\newblock Privacy-preserving ridge regression on hundreds of millions of
  records.
\newblock In \emph{{IEEE} Symposium on Security and Privacy}, pages 334--348.
  {IEEE} Computer Society, 2013.

\bibitem[Niu et~al.(2019)Niu, Wu, Tang, Hua, Jia, Lv, Wu, and
  Chen]{niu2019secure}
Chaoyue Niu, Fan Wu, Shaojie Tang, Lifeng Hua, Rongfei Jia, Chengfei Lv, Zhihua
  Wu, and Guihai Chen.
\newblock Secure federated submodel learning.
\newblock \emph{arXiv preprint arXiv:1911.02254}, 2019.

\bibitem[NSA(2012)]{nsa2012defense}
NSA.
\newblock Defense in depth: A practical strategy for achieving {I}nformation
  {A}ssurance in today's highly networked environments.
\newblock Technical report, NSA, 2012.

\bibitem[Oktay et~al.(2019)Oktay, Ball{\'e}, Singh, and
  Shrivastava]{oktay2019model}
Deniz Oktay, Johannes Ball{\'e}, Saurabh Singh, and Abhinav Shrivastava.
\newblock Model compression by entropy penalized reparameterization.
\newblock \emph{arXiv preprint arXiv:1906.06624}, 2019.

\bibitem[Olumofin and Goldberg(2011)]{olumofin2011revisiting}
Femi Olumofin and Ian Goldberg.
\newblock Revisiting the computational practicality of private information
  retrieval.
\newblock In \emph{International Conference on Financial Cryptography and Data
  Security}, pages 158--172. Springer, 2011.

\bibitem[Palisade()]{Palisade}
Palisade.
\newblock {PALISADE} lattice cryptography library.
\newblock \url{https://gitlab.com/palisade/palisade-release }, October 2019.

\bibitem[Pan and Yang(2010)]{Pan:2010:STL:1850483.1850545}
Sinno~Jialin Pan and Qiang Yang.
\newblock A survey on transfer learning.
\newblock \emph{IEEE Transactions on Knowledge and Data Engineering},
  22\penalty0 (10):\penalty0 1345--1359, 2010.

\bibitem[Papernot et~al.(2017)Papernot, McDaniel, Goodfellow, Jha, Celik, and
  Swami]{papernot2017practical}
Nicolas Papernot, Patrick McDaniel, Ian Goodfellow, Somesh Jha, Z~Berkay Celik,
  and Ananthram Swami.
\newblock Practical black-box attacks against machine learning.
\newblock In \emph{Proceedings of the 2017 ACM on Asia conference on computer
  and communications security}, pages 506--519. ACM, 2017.

\bibitem[Papernot et~al.(2020)Papernot, Thakurta, Song, Chien, and
  Erlingsson]{papernot2020tempered}
Nicolas Papernot, Abhradeep Thakurta, Shuang Song, Steve Chien, and {\'U}lfar
  Erlingsson.
\newblock Tempered sigmoid activations for deep learning with differential
  privacy.
\newblock \emph{arXiv preprint arXiv:2007.14191}, 2020.

\bibitem[Park et~al.(2018)Park, Samarakoon, Bennis, and Debbah]{EdgeML}
Jihong Park, Sumudu Samarakoon, Mehdi Bennis, and M{\'{e}}rouane Debbah.
\newblock Wireless network intelligence at the edge.
\newblock \emph{CoRR}, abs/1812.02858, 2018.
\newblock URL \url{http://arxiv.org/abs/1812.02858}.

\bibitem[Parno et~al.(2016)Parno, Howell, Gentry, and
  Raykova]{DBLP:journals/cacm/ParnoHG016}
Bryan Parno, Jon Howell, Craig Gentry, and Mariana Raykova.
\newblock Pinocchio: nearly practical verifiable computation.
\newblock \emph{Commun. {ACM}}, 59\penalty0 (2):\penalty0 103--112, 2016.

\bibitem[Paszke et~al.(2019)Paszke, Gross, Massa, Lerer, Bradbury, Chanan,
  Killeen, Lin, Gimelshein, Antiga, Desmaison, Kopf, Yang, DeVito, Raison,
  Tejani, Chilamkurthy, Steiner, Fang, Bai, and
  Chintala]{pytorch_NEURIPS2019_9015}
Adam Paszke, Sam Gross, Francisco Massa, Adam Lerer, James Bradbury, Gregory
  Chanan, Trevor Killeen, Zeming Lin, Natalia Gimelshein, Luca Antiga, Alban
  Desmaison, Andreas Kopf, Edward Yang, Zachary DeVito, Martin Raison, Alykhan
  Tejani, Sasank Chilamkurthy, Benoit Steiner, Lu~Fang, Junjie Bai, and Soumith
  Chintala.
\newblock Pytorch: An imperative style, high-performance deep learning library.
\newblock In H.~Wallach, H.~Larochelle, A.~Beygelzimer, F.~d\textquotesingle
  Alch\'{e}-Buc, E.~Fox, and R.~Garnett, editors, \emph{Advances in Neural
  Information Processing Systems 32}, pages 8024--8035. Curran Associates,
  Inc., 2019.
\newblock URL
  \url{http://papers.neurips.cc/paper/9015-pytorch-an-imperative-style-high-performance-deep-learning-library.pdf}.

\bibitem[Patel and Dieuleveut(2019)]{patel19communication}
Kumar~Kshitij Patel and Aymeric Dieuleveut.
\newblock Communication trade-offs for synchronized distributed {SGD} with
  large step size.
\newblock \emph{NeurIPS}, 2019.

\bibitem[Patel et~al.(2018)Patel, Persiano, and Yeo]{Patel18googlePIR}
Sarvar Patel, Giuseppe Persiano, and Kevin Yeo.
\newblock Private stateful information retrieval.
\newblock In \emph{Proceedings of the 2018 ACM SIGSAC Conference on Computer
  and Communications Security}, CCS '18, pages 1002--1019, New York, NY, USA,
  2018. ACM.
\newblock ISBN 978-1-4503-5693-0.
\newblock \doi{10.1145/3243734.3243821}.
\newblock URL \url{http://doi.acm.org/10.1145/3243734.3243821}.

\bibitem[Patrini et~al.(2016)Patrini, Nock, Hardy, and Caetano]{pnhcFL}
Giorgio Patrini, Richard Nock, Stephen Hardy, and Tib{\'{e}}rio~S. Caetano.
\newblock Fast learning from distributed datasets without entity matching.
\newblock In \emph{Proceedings of the Twenty-Fifth International Joint
  Conference on Artificial Intelligence, {IJCAI} 2016, New York, NY, USA, 9-15
  July 2016}, pages 1909--1917, 2016.
\newblock URL \url{http://www.ijcai.org/Abstract/16/273}.

\bibitem[Pedregosa(2016)]{pedregosa2016hyperparameter}
Fabian Pedregosa.
\newblock Hyperparameter optimization with approximate gradient.
\newblock \emph{arXiv preprint arXiv:1602.02355}, 2016.

\bibitem[Pham et~al.(2018)Pham, Guan, Zoph, Le, and Dean]{pham2018efficient}
Hieu Pham, Melody Guan, Barret Zoph, Quoc Le, and Jeff Dean.
\newblock Efficient neural architecture search via parameter sharing.
\newblock In \emph{International Conference on Machine Learning}, pages
  4092--4101, 2018.

\bibitem[Pichai(May 7, 2019)]{sundar2019nyt}
Sundar Pichai.
\newblock {Google's {S}undar {P}ichai: Privacy Should Not Be a Luxury Good}.
\newblock \emph{New York Times}, May 7, 2019.

\bibitem[Pichapati et~al.(2019)Pichapati, Suresh, Yu, Reddi, and
  Kumar]{pichapati2019adaclip}
Venkatadheeraj Pichapati, Ananda~Theertha Suresh, Felix~X Yu, Sashank~J Reddi,
  and Sanjiv Kumar.
\newblock Ada{C}li{P}: Adaptive clipping for private {SGD}.
\newblock \emph{arXiv preprint arXiv:1908.07643}, 2019.

\bibitem[Pihur et~al.(2018)Pihur, Korolova, Liu, Sankuratripati, Yung, Huang,
  and Zeng]{snap}
Vasyl Pihur, Aleksandra Korolova, Frederick Liu, Subhash Sankuratripati, Moti
  Yung, Dachuan Huang, and Ruogu Zeng.
\newblock Differentially-private {``Draw and Discard"} machine learning.
\newblock \emph{CoRR}, abs/1807.04369, 2018.
\newblock URL \url{http://arxiv.org/abs/1807.04369}.

\bibitem[Pillutla et~al.(2019)Pillutla, Kakade, and
  Harchaoui]{pillutla2019robust}
Krishna Pillutla, Sham~M Kakade, and Zaid Harchaoui.
\newblock Robust aggregation for federated learning.
\newblock \emph{arXiv preprint arXiv:1912.13445}, 2019.

\bibitem[Quionero-Candela et~al.(2009)Quionero-Candela, Sugiyama, Schwaighofer,
  and Lawrence]{candela2009datasetshift}
Joaquin Quionero-Candela, Masashi Sugiyama, Anton Schwaighofer, and Neil~D.
  Lawrence.
\newblock \emph{Dataset Shift in Machine Learning}.
\newblock The MIT Press, 2009.
\newblock ISBN 0262170051, 9780262170055.

\bibitem[Rajput et~al.(2019)Rajput, Wang, Charles, and
  Papailiopoulos]{rajput2019detox}
Shashank Rajput, Hongyi Wang, Zachary Charles, and Dimitris Papailiopoulos.
\newblock {DETOX}: A redundancy-based framework for faster and more robust
  gradient aggregation.
\newblock \emph{arXiv preprint arXiv:1907.12205}, 2019.

\bibitem[Ramage and Mazzocchi(2020)]{fablog20}
Daniel Ramage and Stefano Mazzocchi.
\newblock Federated analytics: Collaborative data science without data
  collection, May 2020.
\newblock URL
  \url{https://ai.googleblog.com/2020/05/federated-analytics-collaborative-data.html}.
\newblock Google AI Blog.

\bibitem[Ramaswamy et~al.(2019)Ramaswamy, Mathews, Rao, and
  Beaufays]{ramaswamy19emoji}
Swaroop Ramaswamy, Rajiv Mathews, Kanishka Rao, and Fran{\c{c}}oise Beaufays.
\newblock Federated learning for emoji prediction in a mobile keyboard.
\newblock \emph{arXiv preprint 1906.04329}, 2019.

\bibitem[Ramaswamy et~al.(2020)Ramaswamy, Thakkar, Mathews, Andrew, McMahan,
  and Beaufays]{ramaswamy2020training}
Swaroop Ramaswamy, Om~Thakkar, Rajiv Mathews, Galen Andrew, H~Brendan McMahan,
  and Fran{\c{c}}oise Beaufays.
\newblock Training production language models without memorizing user data.
\newblock \emph{arXiv preprint arXiv:2009.10031}, 2020.

\bibitem[Rastogi and Nath(2010)]{Rastogi:2010:DPA:1807167.1807247}
Vibhor Rastogi and Suman Nath.
\newblock Differentially private aggregation of distributed time-series with
  transformation and encryption.
\newblock In \emph{Proceedings of the 2010 ACM SIGMOD International Conference
  on Management of Data}, SIGMOD '10, pages 735--746, New York, NY, USA, 2010.
  ACM.
\newblock ISBN 978-1-4503-0032-2.
\newblock \doi{10.1145/1807167.1807247}.
\newblock URL \url{http://doi.acm.org/10.1145/1807167.1807247}.

\bibitem[Ravi and Larochelle(2017)]{ravi17miniimagenet}
Sachin Ravi and Hugo Larochelle.
\newblock Optimization as a model for few-shot learning.
\newblock In \emph{Proceedings of the 5th International Conference on Learning
  Representations}, 2017.

\bibitem[Real et~al.(2017)Real, Moore, Selle, Saxena, Suematsu, Tan, Le, and
  Kurakin]{real2017large}
Esteban Real, Sherry Moore, Andrew Selle, Saurabh Saxena, Yutaka~Leon Suematsu,
  Jie Tan, Quoc~V Le, and Alexey Kurakin.
\newblock Large-scale evolution of image classifiers.
\newblock In \emph{Proceedings of the 34th International Conference on Machine
  Learning-Volume 70}, pages 2902--2911. JMLR. org, 2017.

\bibitem[Real et~al.(2019)Real, Aggarwal, Huang, and Le]{real2019regularized}
Esteban Real, Alok Aggarwal, Yanping Huang, and Quoc~V Le.
\newblock Regularized evolution for image classifier architecture search.
\newblock In \emph{Proceedings of the AAAI Conference on Artificial
  Intelligence}, volume~33, pages 4780--4789, 2019.

\bibitem[Reddi et~al.(2020)Reddi, Charles, Zaheer, Garrett, Rush,
  Kone{\v{c}}n{\`y}, Kumar, and McMahan]{reddi2020adaptive}
Sashank Reddi, Zachary Charles, Manzil Zaheer, Zachary Garrett, Keith Rush,
  Jakub Kone{\v{c}}n{\`y}, Sanjiv Kumar, and H~Brendan McMahan.
\newblock Adaptive federated optimization.
\newblock \emph{arXiv preprint arXiv:2003.00295}, 2020.

\bibitem[Reisizadeh et~al.(2019{\natexlab{a}})Reisizadeh, Mokhtari, Hassani,
  Jadbabaie, and Pedarsani]{reisizadeh2019fedpaq}
Amirhossein Reisizadeh, Aryan Mokhtari, Hamed Hassani, Ali Jadbabaie, and
  Ramtin Pedarsani.
\newblock Fedpaq: A communication-efficient federated learning method with
  periodic averaging and quantization.
\newblock \emph{arXiv preprint arXiv:1909.13014}, 2019{\natexlab{a}}.

\bibitem[Reisizadeh et~al.(2019{\natexlab{b}})Reisizadeh, Taheri, Mokhtari,
  Hassani, and Pedarsani]{reisizadeh2019robust}
Amirhossein Reisizadeh, Hossein Taheri, Aryan Mokhtari, Hamed Hassani, and
  Ramtin Pedarsani.
\newblock Robust and communication-efficient collaborative learning.
\newblock \emph{arXiv:1907.10595}, 2019{\natexlab{b}}.

\bibitem[Reyzin et~al.(2018)Reyzin, Smith, and Yakoubov]{hatelove}
Leonid Reyzin, Adam~D. Smith, and Sophia Yakoubov.
\newblock Turning {HATE} into {LOVE:} homomorphic ad hoc threshold encryption
  for scalable {MPC}.
\newblock \emph{{IACR} Cryptology ePrint Archive}, 2018:\penalty0 997, 2018.

\bibitem[Riazi et~al.(2019)Riazi, Laine, Pelton, and Dai]{riazi2019heax}
M~Sadegh Riazi, Kim Laine, Blake Pelton, and Wei Dai.
\newblock {HEAX}: High-performance architecture for computation on
  homomorphically encrypted data in the cloud.
\newblock \emph{arXiv preprint arXiv:1909.09731}, 2019.

\bibitem[Richardson et~al.(2019)Richardson, Schultz, and
  Crawford]{richardson2019dirty}
Rashida Richardson, Jason Schultz, and Kate Crawford.
\newblock Dirty data, bad predictions: How civil rights violations impact
  police data, predictive policing systems, and justice.
\newblock \emph{New York University Law Review Online, Forthcoming}, 2019.

\bibitem[Ripley(1993)]{ripley1993statistical}
Brian~D Ripley.
\newblock Statistical aspects of neural networks.
\newblock \emph{Networks and chaos—statistical and probabilistic aspects},
  50:\penalty0 40--123, 1993.

\bibitem[Rivest et~al.(1978)Rivest, Adleman, and Dertouzos]{Rivest1978}
Ronald~L Rivest, Len Adleman, and Michael~L Dertouzos.
\newblock On data banks and privacy homomorphisms.
\newblock \emph{Foundations of Secure Computation, Academia Press}, pages
  169--179, 1978.

\bibitem[Rodr{\'\i}guez-Barroso et~al.(2020)Rodr{\'\i}guez-Barroso, Stipcich,
  Jim{\'e}nez-L{\'o}pez, Ruiz-Mill{\'a}n, Mart{\'\i}nez-C{\'a}mara,
  Gonz{\'a}lez-Seco, Luz{\'o}n, Veganzones, and
  Herrera]{rodriguez2020federated}
Nuria Rodr{\'\i}guez-Barroso, Goran Stipcich, Daniel Jim{\'e}nez-L{\'o}pez,
  Jos{\'e}~Antonio Ruiz-Mill{\'a}n, Eugenio Mart{\'\i}nez-C{\'a}mara, Gerardo
  Gonz{\'a}lez-Seco, M~Victoria Luz{\'o}n, Miguel~Angel Veganzones, and
  Francisco Herrera.
\newblock Federated learning and differential privacy: Software tools analysis,
  the sherpa. ai fl framework and methodological guidelines for preserving data
  privacy.
\newblock \emph{Information Fusion}, 64:\penalty0 270--292, 2020.

\bibitem[Roth et~al.(2019)Roth, Noble, Falk, and
  Haeberlen]{DBLP:conf/sosp/RothNFH19}
Edo Roth, Daniel Noble, Brett~Hemenway Falk, and Andreas Haeberlen.
\newblock Honeycrisp: large-scale differentially private aggregation without a
  trusted core.
\newblock In \emph{{SOSP}}, pages 196--210. {ACM}, 2019.

\bibitem[Ryffel et~al.(2018)Ryffel, Trask, Dahl, Wagner, Mancuso, Rueckert, and
  Passerat-Palmbach]{PySyft}
Theo Ryffel, Andrew Trask, Morten Dahl, Bobby Wagner, Jason Mancuso, Daniel
  Rueckert, and Jonathan Passerat-Palmbach.
\newblock A generic framework for privacy preserving deep learning, 2018.

\bibitem[{Sabater} et~al.(2020){Sabater}, {Bellet}, and {Ramon}]{sabater2020}
C{\'e}sar {Sabater}, Aur{\'e}lien {Bellet}, and Jan {Ramon}.
\newblock {Distributed Differentially Private Averaging with Improved Utility
  and Robustness to Malicious Parties}.
\newblock \emph{arXiv preprint arXiv:2006.07218}, 2020.

\bibitem[Salmon et~al.(2011)Salmon, Moraes, Dror, and Shaw]{salmon2011parallel}
John~K Salmon, Mark~A Moraes, Ron~O Dror, and David~E Shaw.
\newblock Parallel random numbers: {A}s easy as 1, 2, 3.
\newblock In \emph{Proceedings of 2011 International Conference for High
  Performance Computing, Networking, Storage and Analysis}, page~16. ACM, 2011.

\bibitem[Samarakoon et~al.(2018)Samarakoon, Bennis, Saad, and Debbah]{FL5G}
Sumudu Samarakoon, Mehdi Bennis, Walid Saad, and M{\'{e}}rouane Debbah.
\newblock Federated learning for ultra-reliable low-latency {V2V}
  communications.
\newblock \emph{CoRR}, abs/1805.09253, 2018.
\newblock URL \url{http://arxiv.org/abs/1805.09253}.

\bibitem[Sambasivan et~al.(2018)Sambasivan, Checkley, Batool, Ahmed, Nemer,
  Gayt{\'a}n-Lugo, Matthews, Consolvo, and Churchill]{sambasivan2018privacy}
Nithya Sambasivan, Garen Checkley, Amna Batool, Nova Ahmed, David Nemer,
  Laura~Sanely Gayt{\'a}n-Lugo, Tara Matthews, Sunny Consolvo, and Elizabeth
  Churchill.
\newblock " privacy is not for me, it's for those rich women": Performative
  privacy practices on mobile phones by women in south asia.
\newblock In \emph{Fourteenth Symposium on Usable Privacy and Security
  ($\{$SOUPS$\}$ 2018)}, pages 127--142, 2018.

\bibitem[Sathya et~al.(2018)Sathya, Vepakomma, Raskar, Ramachandra, and
  Bhattacharya]{sathya2018review}
Sai~Sri Sathya, Praneeth Vepakomma, Ramesh Raskar, Ranjan Ramachandra, and
  Santanu Bhattacharya.
\newblock A review of homomorphic encryption libraries for secure computation.
\newblock \emph{arXiv preprint arXiv:1812.02428}, 2018.

\bibitem[Sattler et~al.(2019)Sattler, Wiedemann, M{\"u}ller, and
  Samek]{sattler2019robust}
Felix Sattler, Simon Wiedemann, Klaus-Robert M{\"u}ller, and Wojciech Samek.
\newblock Robust and communication-efficient federated learning from non-{IID}
  data.
\newblock \emph{arXiv preprint arXiv:1903.02891}, 2019.

\bibitem[Schnell(2013)]{sEP}
R.~Schnell.
\newblock Efficient private record linkage of very large datasets.
\newblock In \emph{59$^{th}$ World Statistics Congress}, 2013.

\bibitem[Schnell et~al.(2011)Schnell, Bachteler, and Reiher]{schnell11}
R.~Schnell, T.~Bachteler, and J.~Reiher.
\newblock A novel error-tolerant anonymous linking code.
\newblock Technical report, Paper No. WP-GRLC-2011-02, German Record Linkage
  Center Working Paper Series, 2011.

\bibitem[Schnorr(1990)]{Schnorr:1990:EIS:111563.111628}
Claus~P. Schnorr.
\newblock Efficient identification and signatures for smart cards.
\newblock In \emph{Proceedings of the Workshop on the Theory and Application of
  Cryptographic Techniques on Advances in Cryptology}, EUROCRYPT '89, 1990.

\bibitem[SEAL()]{SEAL}
SEAL.
\newblock {M}icrosoft {SEAL} (release 3.6).
\newblock \url{https://github.com/Microsoft/SEAL}, November 2020.
\newblock Microsoft Research, Redmond, WA.

\bibitem[Seide and Agarwal(2016)]{cntk}
Frank Seide and Amit Agarwal.
\newblock Cntk: Microsoft's open-source deep-learning toolkit.
\newblock In \emph{Proceedings of the 22nd ACM SIGKDD International Conference
  on Knowledge Discovery and Data Mining}, KDD '16, page 2135, New York, NY,
  USA, 2016. Association for Computing Machinery.
\newblock ISBN 9781450342322.
\newblock \doi{10.1145/2939672.2945397}.
\newblock URL \url{https://doi.org/10.1145/2939672.2945397}.

\bibitem[Seshadri et~al.(2007)Seshadri, Luk, Perrig, van Doom, and
  Khosla]{DBLP:series/ais/SeshadriLPDK07}
Arvind Seshadri, Mark Luk, Adrian Perrig, Leendert van Doom, and Pradeep~K.
  Khosla.
\newblock Pioneer: Verifying code integrity and enforcing untampered code
  execution on legacy systems.
\newblock In \emph{Malware Detection}, volume~27 of \emph{Advances in
  Information Security}, pages 253--289. Springer, 2007.

\bibitem[Shafahi et~al.(2019)Shafahi, Najibi, Ghiasi, Xu, Dickerson, Studer,
  Davis, Taylor, and Goldstein]{shafahi2018free}
Ali Shafahi, Mahyar Najibi, Amin Ghiasi, Zheng Xu, John Dickerson, Christoph
  Studer, Larry~S Davis, Gavin Taylor, and Tom Goldstein.
\newblock Adversarial training for free.
\newblock \emph{NeurIPS}, 2019.

\bibitem[Sharma et~al.(2019)Sharma, Vepakomma, Swedish, Chang, Kalpathy-Cramer,
  and Raskar]{sharma2019expertmatcher}
Vivek Sharma, Praneeth Vepakomma, Tristan Swedish, Ken Chang, Jayashree
  Kalpathy-Cramer, and Ramesh Raskar.
\newblock Expert{M}atcher: Automating {ML} model selection for clients using
  hidden representations.
\newblock \emph{arXiv preprint arXiv:1910.03731}, 2019.

\bibitem[Sharma and Chen(2017)]{sharma2017attacking}
Yash Sharma and Pin-Yu Chen.
\newblock Attacking the {M}adry defense model with $ l\_1 $-based adversarial
  examples.
\newblock \emph{arXiv preprint arXiv:1710.10733}, 2017.

\bibitem[SHELL()]{SHELL}
SHELL.
\newblock \url{https://github.com/google/shell-encryption}, December 2020.
\newblock Google.

\bibitem[Shen and Sanghavi(2019)]{pmlr-v97-shen19e}
Yanyao Shen and Sujay Sanghavi.
\newblock Learning with bad training data via iterative trimmed loss
  minimization.
\newblock In Kamalika Chaudhuri and Ruslan Salakhutdinov, editors,
  \emph{Proceedings of the 36th International Conference on Machine Learning},
  volume~97 of \emph{Proceedings of Machine Learning Research}, pages
  5739--5748, Long Beach, California, USA, 09--15 Jun 2019. PMLR.
\newblock URL \url{http://proceedings.mlr.press/v97/shen19e.html}.

\bibitem[Shi et~al.(2011)Shi, Chan, Rieffel, Chow, and Song]{shi2011privacy}
Elaine Shi, HTH Chan, Eleanor Rieffel, Richard Chow, and Dawn Song.
\newblock Privacy-preserving aggregation of time-series data.
\newblock In \emph{Annual Network \& Distributed System Security Symposium
  (NDSS)}, 2011.

\bibitem[Shokri et~al.(2017)Shokri, Stronati, Song, and
  Shmatikov]{shokri2017membership}
Reza Shokri, Marco Stronati, Congzheng Song, and Vitaly Shmatikov.
\newblock Membership inference attacks against machine learning models.
\newblock In \emph{2017 IEEE Symposium on Security and Privacy (SP)}, pages
  3--18. IEEE, 2017.

\bibitem[Shridhar et~al.(2019)Shridhar, Laumann, and Liwicki]{BL-overview}
Kumar Shridhar, Felix Laumann, and Marcus Liwicki.
\newblock A comprehensive guide to {B}ayesian convolutional neural network with
  variational inference.
\newblock \emph{arXiv preprint: 1901.02731}, 2019.

\bibitem[Silver et~al.(2013)Silver, Yang, and Li]{silver13lifelong}
Daniel~L. Silver, Qiang Yang, and Lianghao Li.
\newblock Lifelong machine learning systems: Beyond learning algorithms.
\newblock In \emph{AAAI Spring Symposium Series}, 2013.

\bibitem[Singh et~al.(2019)Singh, Vepakomma, Gupta, and
  Raskar]{vepakomma2019splitComm}
Abhishek Singh, Praneeth Vepakomma, Otkrist Gupta, and Ramesh Raskar.
\newblock Detailed comparison of communication efficiency of split learning and
  federated learning.
\newblock \emph{arXiv preprint arXiv:1909.09145}, 2019.

\bibitem[Singh et~al.(2020)Singh, Chopra, Sharma, Garza, Zhang, Vepakomma, and
  Raskar]{channelPruning}
Abhishek Singh, Ayush Chopra, Vivek Sharma, Ethan Garza, Emily Zhang, Praneeth
  Vepakomma, and Ramesh Raskar.
\newblock {DISCO}: Dynamic and invariant sensitive channel obfuscation for deep
  neural networks.
\newblock 2020.

\bibitem[Sion and Carbunar(2007)]{sion2007computational}
Radu Sion and Bogdan Carbunar.
\newblock On the computational practicality of private information retrieval.
\newblock In \emph{Proceedings of the Network and Distributed Systems Security
  Symposium}, pages 2006--06. Internet Society, 2007.

\bibitem[Smith et~al.(2017)Smith, Chiang, Sanjabi, and Talwalkar]{Smith2017}
Virginia Smith, Chao-Kai Chiang, Maziar Sanjabi, and Ameet~S. Talwalkar.
\newblock {Federated Multi-Task Learning}.
\newblock In \emph{NIPS}, 2017.

\bibitem[Snell et~al.(2017)Snell, Swersky, and Zemel]{snell17protonets}
Jake Snell, Kevin Swersky, and Richard~S. Zemel.
\newblock Prototypical networks for few-shot learning.
\newblock In \emph{Advances in Neural Information Processing Systems}, 2017.

\bibitem[Snoek et~al.(2015)Snoek, Rippel, Swersky, Kiros, Satish, Sundaram,
  Patwary, Prabhat, and Adams]{snoek2015scalable}
Jasper Snoek, Oren Rippel, Kevin Swersky, Ryan Kiros, Nadathur Satish,
  Narayanan Sundaram, Mostofa Patwary, Mr~Prabhat, and Ryan Adams.
\newblock Scalable {B}ayesian optimization using deep neural networks.
\newblock In \emph{International conference on machine learning}, pages
  2171--2180, 2015.

\bibitem[So et~al.(2020{\natexlab{a}})So, Guler, and Avestimehr]{BREA2020}
Jinhyun So, Basak Guler, and A.~Salman Avestimehr.
\newblock Byzantine-resilient secure federated learning.
\newblock \emph{IEEE Journal on Selected Areas in Communication, Series on
  Machine Learning for Communications and Networks}, 2020{\natexlab{a}}.

\bibitem[So et~al.(2020{\natexlab{b}})So, Guler, and Avestimehr]{so2020turbo}
Jinhyun So, Basak Guler, and A~Salman Avestimehr.
\newblock Turbo-aggregate: Breaking the quadratic aggregation barrier in secure
  federated learning.
\newblock \emph{arXiv preprint arXiv:2002.04156}, 2020{\natexlab{b}}.

\bibitem[Song et~al.(2019)Song, Shokri, and Mittal]{song:ccs19}
Liwei Song, Reza Shokri, and Prateek Mittal.
\newblock Privacy risks of securing machine learning models against adversarial
  examples.
\newblock In \emph{In Proceedings of the {ACM} Conference on Computer and
  Communication Security (CCS)}, 2019.

\bibitem[Srinathan and Rangan(2000)]{srinathan2000efficient}
K~Srinathan and C~Pandu Rangan.
\newblock Efficient asynchronous secure multiparty distributed computation.
\newblock In \emph{International Conference on Cryptology in India}, pages
  117--129. Springer, 2000.

\bibitem[Srivastava et~al.(2019)Srivastava, Bellet, Tommasi, and
  Vincent]{Srivastava2019a}
Brij Mohan~Lal Srivastava, Aur\'elien Bellet, Marc Tommasi, and Emmanuel
  Vincent.
\newblock {P}rivacy-{P}reserving {A}dversarial {R}epresentation {L}earning in
  {ASR}: {R}eality or {I}llusion?
\newblock In \emph{Annual Conference of the International Speech Communication
  Association (Interspeech)}, 2019.

\bibitem[Steinhardt et~al.(2017)Steinhardt, Koh, and
  Liang]{steinhardt2017certified}
Jacob Steinhardt, Pang Wei~W Koh, and Percy~S Liang.
\newblock Certified defenses for data poisoning attacks.
\newblock In \emph{Advances in neural information processing systems}, pages
  3517--3529, 2017.

\bibitem[Steinke and Ullman(2017)]{steinke2017tight}
Thomas Steinke and Jonathan Ullman.
\newblock Tight lower bounds for differentially private selection.
\newblock In \emph{FOCS}, pages 552--563, 2017.

\bibitem[Stich(2019)]{stich2018local}
Sebastian~U Stich.
\newblock Local {SGD} converges fast and communicates little.
\newblock In \emph{International Conference on Learning Representations
  (ICLR)}, 2019.

\bibitem[Stich and Karimireddy(2019)]{stich2019error}
Sebastian~U Stich and Sai~Praneeth Karimireddy.
\newblock The error-feedback framework: Better rates for {SGD} with delayed
  gradients and compressed communication.
\newblock \emph{arXiv:1909.05350}, 2019.

\bibitem[Su and Vaidya(2016)]{Su2016FaultTolerantMO}
Lili Su and Nitin~H. Vaidya.
\newblock {Fault-Tolerant Multi-Agent Optimization: Optimal Iterative
  Distributed Algorithms}.
\newblock In \emph{PODC}, 2016.

\bibitem[Subramanyan et~al.(2017)Subramanyan, Sinha, Lebedev, Devadas, and
  Seshia]{subramanyan2017formal}
Pramod Subramanyan, Rohit Sinha, Ilia Lebedev, Srinivas Devadas, and Sanjit~A
  Seshia.
\newblock A formal foundation for secure remote execution of enclaves.
\newblock In \emph{Proceedings of the 2017 ACM SIGSAC Conference on Computer
  and Communications Security}, pages 2435--2450. ACM, 2017.

\bibitem[Sun et~al.(2019)Sun, Kairouz, Suresh, and McMahan]{sun2019backdoor}
Ziteng Sun, Peter Kairouz, Ananda~Theertha Suresh, and H~Brendan McMahan.
\newblock Can you really backdoor federated learning?
\newblock \emph{arXiv preprint arXiv:1911.07963}, 2019.

\bibitem[{support.google}(2019)]{messages19privacy}
{support.google}.
\newblock Your chats stay private while {M}essages improves suggestions, 2019.
\newblock URL \url{https://support.google.com/messages/answer/9327902}.
\newblock Retrieved Aug 2019.

\bibitem[Suresh et~al.(2017)Suresh, Yu, Kumar, and
  McMahan]{suresh2017distributed}
Ananda~Theertha Suresh, Felix~X. Yu, Sanjiv Kumar, and H~Brendan McMahan.
\newblock Distributed mean estimation with limited communication.
\newblock In \emph{Proceedings of the 34th International Conference on Machine
  Learning-Volume 70}, pages 3329--3337. JMLR. org, 2017.

\bibitem[Szegedy et~al.(2013)Szegedy, Zaremba, Sutskever, Bruna, Erhan,
  Goodfellow, and Fergus]{szegedy2013intriguing}
Christian Szegedy, Wojciech Zaremba, Ilya Sutskever, Joan Bruna, Dumitru Erhan,
  Ian Goodfellow, and Rob Fergus.
\newblock Intriguing properties of neural networks.
\newblock \emph{ICLR}, 2013.

\bibitem[Sz{\'e}kely et~al.(2007)Sz{\'e}kely, Rizzo, Bakirov,
  et~al.]{szekely2007measuring}
G{\'a}bor~J Sz{\'e}kely, Maria~L Rizzo, Nail~K Bakirov, et~al.
\newblock Measuring and testing dependence by correlation of distances.
\newblock \emph{The annals of statistics}, 35\penalty0 (6):\penalty0
  2769--2794, 2007.

\bibitem[Tang et~al.(2018)Tang, Lian, Yan, Zhang, and Liu]{Tang2018}
Hanlin Tang, Xiangru Lian, Ming Yan, Ce~Zhang, and Ji~Liu.
\newblock D2: Decentralized training over decentralized data.
\newblock In \emph{ICML}, 2018.

\bibitem[Tang et~al.(2019)Tang, Lian, Qiu, Yuan, Zhang, Zhang, and
  Liu]{tang2019texttt}
Hanlin Tang, Xiangru Lian, Shuang Qiu, Lei Yuan, Ce~Zhang, Tong Zhang, and
  Ji~Liu.
\newblock Deep{S}queeze: Parallel stochastic gradient descent with double-pass
  error-compensated compression.
\newblock \emph{arXiv preprint arXiv:1907.07346}, 2019.

\bibitem[Tang et~al.(2017)Tang, Korolova, Bai, Wang, and Wang]{appleepsilon}
Jun Tang, Aleksandra Korolova, Xiaolong Bai, Xueqiang Wang, and XiaoFeng Wang.
\newblock Privacy loss in {Apple}'s implementation of differential privacy on
  {MacOS} 10.12.
\newblock \emph{CoRR}, abs/1709.02753, 2017.
\newblock URL \url{http://arxiv.org/abs/1709.02753}.

\bibitem[Thakkar et~al.(2019)Thakkar, Andrew, and
  McMahan]{thakkar2019differentially}
Om~Thakkar, Galen Andrew, and H~Brendan McMahan.
\newblock Differentially private learning with adaptive clipping.
\newblock \emph{arXiv preprint arXiv:1905.03871}, 2019.

\bibitem[Tram{\`{e}}r and Boneh(2019)]{tramer2018slalom}
Florian Tram{\`{e}}r and Dan Boneh.
\newblock Slalom: Fast, verifiable and private execution of neural networks in
  trusted hardware.
\newblock In \emph{International Conference on Learning Representations}, 2019.
\newblock URL \url{https://openreview.net/forum?id=rJVorjCcKQ}.

\bibitem[Tram{\`e}r and Boneh(2019)]{tramer2019adversarial}
Florian Tram{\`e}r and Dan Boneh.
\newblock Adversarial training and robustness for multiple perturbations.
\newblock \emph{arXiv preprint arXiv:1904.13000}, 2019.

\bibitem[Tram{\`e}r and Boneh(2020)]{tramer2020differentially}
Florian Tram{\`e}r and Dan Boneh.
\newblock Differentially private learning needs better features (or much more
  data).
\newblock \emph{arXiv preprint arXiv:2011.11660}, 2020.

\bibitem[Tram{\`{e}}r et~al.(2016)Tram{\`{e}}r, Zhang, Juels, Reiter, and
  Ristenpart]{DBLP:conf/uss/TramerZJRR16}
Florian Tram{\`{e}}r, Fan Zhang, Ari Juels, Michael~K. Reiter, and Thomas
  Ristenpart.
\newblock Stealing machine learning models via prediction {API}s.
\newblock In \emph{25th {USENIX} Security Symposium, {USENIX} Security 16,
  Austin, TX, USA, August 10-12, 2016.}, pages 601--618, 2016.
\newblock URL
  \url{https://www.usenix.org/conference/usenixsecurity16/technical-sessions/presentation/tramer}.

\bibitem[Tram{\`{e}}r et~al.(2017)Tram{\`{e}}r, Zhang, Lin, Hubaux, Juels, and
  Shi]{DBLP:conf/eurosp/TramerZLHJS17}
Florian Tram{\`{e}}r, Fan Zhang, Huang Lin, Jean{-}Pierre Hubaux, Ari Juels,
  and Elaine Shi.
\newblock Sealed-glass proofs: Using transparent enclaves to prove and sell
  knowledge.
\newblock In \emph{2017 {IEEE} European Symposium on Security and Privacy,
  EuroS{\&}P 2017, Paris, France, April 26-28, 2017}, pages 19--34, 2017.

\bibitem[Tram{\`{e}}r et~al.(2018)Tram{\`{e}}r, Kurakin, Papernot, Goodfellow,
  Boneh, and McDaniel]{DBLP:conf/iclr/TramerKPGBM18}
Florian Tram{\`{e}}r, Alexey Kurakin, Nicolas Papernot, Ian~J. Goodfellow, Dan
  Boneh, and Patrick~D. McDaniel.
\newblock Ensemble adversarial training: Attacks and defenses.
\newblock In \emph{6th International Conference on Learning Representations,
  {ICLR} 2018, Vancouver, BC, Canada, April 30 - May 3, 2018, Conference Track
  Proceedings}, 2018.

\bibitem[Tram{\`{e}}r et~al.(2020)Tram{\`{e}}r, Behrmann, Carlini, Papernot,
  and Jacobsen]{DBLP:conf/icml/TramerBCPJ20}
Florian Tram{\`{e}}r, Jens Behrmann, Nicholas Carlini, Nicolas Papernot, and
  J{\"{o}}rn{-}Henrik Jacobsen.
\newblock Fundamental tradeoffs between invariance and sensitivity to
  adversarial perturbations.
\newblock In \emph{Proceedings of the 37th International Conference on Machine
  Learning, {ICML} 2020, 13-18 July 2020, Virtual Event}, volume 119 of
  \emph{Proceedings of Machine Learning Research}, pages 9561--9571. {PMLR},
  2020.
\newblock URL \url{http://proceedings.mlr.press/v119/tramer20a.html}.

\bibitem[Tran et~al.(2018)Tran, Li, and Madry]{tran2018spectral}
Brandon Tran, Jerry Li, and Aleksander Madry.
\newblock Spectral signatures in backdoor attacks.
\newblock In \emph{Advances in Neural Information Processing Systems}, pages
  8000--8010, 2018.

\bibitem[Ullman(2018)]{Ullman18}
Jonathan Ullman.
\newblock Tight lower bounds for locally differentially private selection.
\newblock Technical Report abs/1802.02638, arXiv, 2018.
\newblock URL \url{http://arxiv.org/abs/1802.02638}.

\bibitem[v2~Authors(2019)]{GDLv2}
The Google-Landmark v2~Authors.
\newblock Google landmark dataset v2, 2019.
\newblock URL \url{https://github.com/cvdfoundation/google-landmark}.

\bibitem[Vaidya et~al.(2008)Vaidya, Yu, and Jiang]{vaidya2008ppsvm}
Jaideep Vaidya, Hwanjo Yu, and Xiaoqian Jiang.
\newblock Privacy-preserving {SVM} classification.
\newblock \emph{Knowl. Inf. Syst.}, 14\penalty0 (2), January 2008.

\bibitem[Van~Bulck et~al.(2018)Van~Bulck, Minkin, Weisse, Genkin, Kasikci,
  Piessens, Silberstein, Wenisch, Yarom, and Strackx]{van2018foreshadow}
Jo~Van~Bulck, Marina Minkin, Ofir Weisse, Daniel Genkin, Baris Kasikci, Frank
  Piessens, Mark Silberstein, Thomas~F Wenisch, Yuval Yarom, and Raoul Strackx.
\newblock Foreshadow: Extracting the keys to the intel $\{$SGX$\}$ kingdom with
  transient out-of-order execution.
\newblock In \emph{27th $\{$USENIX$\}$ Security Symposium ($\{$USENIX$\}$
  Security 18)}, pages 991--1008, 2018.

\bibitem[Vanhaesebrouck et~al.(2017)Vanhaesebrouck, Bellet, and
  Tommasi]{Vanhaesebrouck2017}
Paul Vanhaesebrouck, Aur{\'e}lien Bellet, and Marc Tommasi.
\newblock Decentralized collaborative learning of personalized models over
  networks.
\newblock In \emph{AISTATS}, 2017.

\bibitem[Vepakomma et~al.(2018{\natexlab{a}})Vepakomma, Gupta, Swedish, and
  Raskar]{vepakomma2018split}
Praneeth Vepakomma, Otkrist Gupta, Tristan Swedish, and Ramesh Raskar.
\newblock Split learning for health: Distributed deep learning without sharing
  raw patient data.
\newblock \emph{arXiv preprint arXiv:1812.00564}, 2018{\natexlab{a}}.

\bibitem[Vepakomma et~al.(2018{\natexlab{b}})Vepakomma, Tonde, Elgammal,
  et~al.]{vepakomma2018supervised}
Praneeth Vepakomma, Chetan Tonde, Ahmed Elgammal, et~al.
\newblock Supervised dimensionality reduction via distance correlation
  maximization.
\newblock \emph{Electronic Journal of Statistics}, 12\penalty0 (1):\penalty0
  960--984, 2018{\natexlab{b}}.

\bibitem[Vepakomma et~al.(2020)Vepakomma, Singh, and
  Raskar]{vepakomma2019reducing}
Praneeth Vepakomma, Otkrist Singh, Abhishek~Gupta, and Ramesh Raskar.
\newblock Nopeek: Information leakage reduction to share activations in
  distributed deep learning.
\newblock \emph{arXiv preprint arXiv:2008.09161}, 2020.

\bibitem[Vogels et~al.(2019)Vogels, Karimireddy, and Jaggi]{vogels2019powersgd}
Thijs Vogels, Sai~Praneeth Karimireddy, and Martin Jaggi.
\newblock Power{SGD}: Practical low-rank gradient compression for distributed
  optimization.
\newblock In \emph{NeurIPS 2019 - Advances in Neural Information Processing
  Systems 32}, 2019.

\bibitem[Wahby et~al.(2018)Wahby, Tzialla, Shelat, Thaler, and
  Walfish]{DBLP:conf/sp/WahbyTSTW18}
Riad~S. Wahby, Ioanna Tzialla, Abhi Shelat, Justin Thaler, and Michael Walfish.
\newblock Doubly-efficient zksnarks without trusted setup.
\newblock In \emph{2018 {IEEE} Symposium on Security and Privacy, {SP} 2018,
  Proceedings, 21-23 May 2018, San Francisco, California, {USA}}, 2018.

\bibitem[Wang et~al.(2019{\natexlab{a}})Wang, Yao, Shan, Li, Viswanath, Zheng,
  and Zhao]{wang2019neural}
Bolun Wang, Yuanshun Yao, Shawn Shan, Huiying Li, Bimal Viswanath, Haitao
  Zheng, and Ben~Y Zhao.
\newblock Neural cleanse: Identifying and mitigating backdoor attacks in neural
  networks.
\newblock In \emph{2019 IEEE Symposium on Security and Privacy}. IEEE,
  2019{\natexlab{a}}.

\bibitem[Wang et~al.(2020{\natexlab{a}})Wang, Sreenivasan, Rajput, Vishwakarma,
  Agarwal, yong Sohn, Lee, and Papailiopoulos]{wang2020attack}
Hongyi Wang, Kartik Sreenivasan, Shashank Rajput, Harit Vishwakarma, Saurabh
  Agarwal, Jy~yong Sohn, Kangwook Lee, and Dimitris Papailiopoulos.
\newblock Attack of the tails: Yes, you really can backdoor federated learning,
  2020{\natexlab{a}}.

\bibitem[Wang and Joshi(2018)]{wang2018cooperative}
Jianyu Wang and Gauri Joshi.
\newblock {Cooperative {SGD}: A unified framework for the design and analysis
  of communication-efficient {SGD} algorithms}.
\newblock \emph{preprint}, August 2018.
\newblock URL \url{https://arxiv.org/abs/1808.07576}.

\bibitem[Wang and Joshi(2019)]{wang2018adaptive}
Jianyu Wang and Gauri Joshi.
\newblock {Adaptive Communication Strategies for Best Error-Runtime Trade-offs
  in Communication-Efficient Distributed {SGD}}.
\newblock In \emph{Proceedings of the SysML Conference}, April 2019.
\newblock URL \url{https://arxiv.org/abs/1810.08313}.

\bibitem[Wang et~al.(2019{\natexlab{b}})Wang, Sahu, Joshi, and
  Kar]{wang2019matcha}
Jianyu Wang, Anit Sahu, Gauri Joshi, and Soummya Kar.
\newblock {{MATCHA}: Speeding Up Decentralized {SGD} via Matching Decomposition
  Sampling}.
\newblock \emph{preprint}, May 2019{\natexlab{b}}.
\newblock URL \url{https://arxiv.org/abs/1905.09435}.

\bibitem[Wang et~al.(2019{\natexlab{c}})Wang, Tantia, Ballas, and
  Rabbat]{wang2019slowmo}
Jianyu Wang, Vinayak Tantia, Nicolas Ballas, and Michael Rabbat.
\newblock Slow{M}o: Improving communication-efficient distributed {SGD} with
  slow momentum.
\newblock \emph{arXiv preprint arXiv:1910.00643}, 2019{\natexlab{c}}.

\bibitem[Wang et~al.(2020{\natexlab{b}})Wang, Liu, Liang, Joshi, and
  Poor]{wang2020tackling}
Jianyu Wang, Qinghua Liu, Hao Liang, Gauri Joshi, and H~Vincent Poor.
\newblock Tackling the objective inconsistency problem in heterogeneous
  federated optimization.
\newblock \emph{Advances in Neural Information Processing Systems}, 33,
  2020{\natexlab{b}}.

\bibitem[Wang et~al.(2019{\natexlab{d}})Wang, Mathews, Kiddon, Eichner,
  Beaufays, and Ramage]{wang2019federated}
Kangkang Wang, Rajiv Mathews, Chloé Kiddon, Hubert Eichner, Françoise
  Beaufays, and Daniel Ramage.
\newblock Federated evaluation of on-device personalization.
\newblock \emph{arXiv preprint arXiv:1910.10252}, 2019{\natexlab{d}}.

\bibitem[Wang et~al.(2018{\natexlab{a}})Wang, Zhu, Torralba, and
  Efros]{wang2018dataset}
Tongzhou Wang, Jun-Yan Zhu, Antonio Torralba, and Alexei~A Efros.
\newblock Dataset distillation.
\newblock \emph{arXiv preprint arXiv:1811.10959}, 2018{\natexlab{a}}.

\bibitem[Wang et~al.(2018{\natexlab{b}})Wang, Balle, and
  Kasiviswanathan]{wang2018subsampled}
Yu-Xiang Wang, Borja Balle, and Shiva Kasiviswanathan.
\newblock Subsampled {R}$\backslash$'enyi differential privacy and analytical
  moments accountant.
\newblock \emph{arXiv preprint arXiv:1808.00087}, 2018{\natexlab{b}}.

\bibitem[Warner(1965)]{Warner65}
Stanley~L. Warner.
\newblock Randomized response: A survey technique for eliminating evasive
  answer bias.
\newblock \emph{Journal of the American Statistical Association}, 60\penalty0
  (309):\penalty0 63--69, 1965.

\bibitem[{WeBank}(2019)]{webankswissre19reinsurance}
{WeBank}.
\newblock We{B}ank and {S}wiss re signed cooperation {MOU}, 2019.
\newblock URL
  \url{https://finance.yahoo.com/news/webank-swiss-signed-cooperation-mou-112300218.html}.
\newblock Retrieved Aug 2019.

\bibitem[Wong et~al.(2019)Wong, Schmidt, and Kolter]{wong2019wasserstein}
Eric Wong, Frank~R Schmidt, and J~Zico Kolter.
\newblock Wasserstein adversarial examples via projected sinkhorn iterations.
\newblock \emph{ICML}, 2019.

\bibitem[Wood et~al.(2014)]{wood2014ethereum}
Gavin Wood et~al.
\newblock Ethereum: A secure decentralised generalised transaction ledger.
\newblock \emph{Ethereum project yellow paper}, 151\penalty0 (2014):\penalty0
  1--32, 2014.

\bibitem[{Woodruff} and {Yekhanin}(2005)]{yekhaninpir}
D.~{Woodruff} and S.~{Yekhanin}.
\newblock A geometric approach to information-theoretic private information
  retrieval.
\newblock In \emph{20th Annual IEEE Conference on Computational Complexity
  (CCC'05)}, pages 275--284, June 2005.
\newblock \doi{10.1109/CCC.2005.2}.

\bibitem[Woodworth et~al.(2018)Woodworth, Wang, McMahan, and
  Srebro]{woodworth18graphoracle}
Blake Woodworth, Jialei Wang, H.~Brendan McMahan, and Nathan Srebro.
\newblock Graph oracle models, lower bounds, and gaps for parallel stochastic
  optimization.
\newblock In \emph{Advances in Neural Information Processing Systems (NIPS)},
  2018.
\newblock URL \url{https://arxiv.org/abs/1805.10222}.

\bibitem[Woodworth et~al.(2020)Woodworth, Patel, Stich, Dai, Bullins, McMahan,
  Shamir, and Srebro]{woodworth2020local}
Blake Woodworth, Kumar~Kshitij Patel, Sebastian~U Stich, Zhen Dai, Brian
  Bullins, H~Brendan McMahan, Ohad Shamir, and Nathan Srebro.
\newblock Is local sgd better than minibatch sgd?
\newblock \emph{arXiv preprint arXiv:2002.07839}, 2020.

\bibitem[Wu et~al.(2017)Wu, Guo, Suresh, Kumar, Holtmann-Rice, Simcha, and
  Yu]{wu2017multiscale}
Xiang Wu, Ruiqi Guo, Ananda~Theertha Suresh, Sanjiv Kumar, Daniel~N
  Holtmann-Rice, David Simcha, and Felix~X. Yu.
\newblock Multiscale quantization for fast similarity search.
\newblock In \emph{Advances in Neural Information Processing Systems}, pages
  5745--5755, 2017.

\bibitem[Xie et~al.(2019{\natexlab{a}})Xie, Wu, van~der Maaten, Yuille, and
  He]{xie2018feature}
Cihang Xie, Yuxin Wu, Laurens van~der Maaten, Alan Yuille, and Kaiming He.
\newblock Feature denoising for improving adversarial robustness.
\newblock \emph{CVPR}, 2019{\natexlab{a}}.

\bibitem[Xie(2019)]{xie2019zeno++}
Cong Xie.
\newblock Zeno++: robust asynchronous {SGD} with arbitrary number of
  {B}yzantine workers.
\newblock \emph{arXiv preprint arXiv:1903.07020}, 2019.

\bibitem[Xie et~al.(2019{\natexlab{b}})Xie, Koyejo, Gupta, and
  Lin]{xie2019local}
Cong Xie, Oluwasanmi Koyejo, Indranil Gupta, and Haibin Lin.
\newblock Local adaalter: Communication-efficient stochastic gradient descent
  with adaptive learning rates.
\newblock \emph{arXiv preprint arXiv:1911.09030}, 2019{\natexlab{b}}.

\bibitem[Xie et~al.(2019{\natexlab{c}})Xie, Koyejo, and
  Gupta]{xie2019practicalsecure}
Cong Xie, Sanmi Koyejo, and Indranil Gupta.
\newblock Practical distributed learning: Secure machine learning with
  communication-efficient local updates.
\newblock In \emph{European Conference on Machine Learning and Principles and
  Practice of Knowledge Discovery in Databases (ECML PKDD)},
  2019{\natexlab{c}}.

\bibitem[Xie et~al.(2019{\natexlab{d}})Xie, Koyejo, and Gupta]{xie2019zeno}
Cong Xie, Sanmi Koyejo, and Indranil Gupta.
\newblock Zeno: Distributed stochastic gradient descent with suspicion-based
  fault-tolerance.
\newblock In \emph{International Conference on Machine Learning}, pages
  6893--6901, 2019{\natexlab{d}}.

\bibitem[Xie et~al.(2018)Xie, Zheng, Liu, and Lin]{xie2018snas}
Sirui Xie, Hehui Zheng, Chunxiao Liu, and Liang Lin.
\newblock {SNAS}: stochastic neural architecture search.
\newblock \emph{arXiv preprint arXiv:1812.09926}, 2018.

\bibitem[Xie et~al.(2019{\natexlab{e}})Xie, Zhang, Zhang, Papamanthou, and
  Song]{DBLP:conf/crypto/XieZZPS19}
Tiancheng Xie, Jiaheng Zhang, Yupeng Zhang, Charalampos Papamanthou, and Dawn
  Song.
\newblock Libra: Succinct zero-knowledge proofs with optimal prover
  computation.
\newblock In \emph{{CRYPTO} {(3)}}, volume 11694 of \emph{Lecture Notes in
  Computer Science}, pages 733--764. Springer, 2019{\natexlab{e}}.

\bibitem[Yang et~al.(2019)Yang, Liu, Chen, and
  Tong]{DBLP:journals/corr/abs-1902-04885}
Qiang Yang, Yang Liu, Tianjian Chen, and Yongxin Tong.
\newblock Federated machine learning: Concept and applications.
\newblock \emph{CoRR}, abs/1902.04885, 2019.
\newblock URL \url{http://arxiv.org/abs/1902.04885}.

\bibitem[Yang et~al.(2018)Yang, Andrew, Eichner, Sun, Li, Kong, Ramage, and
  Beaufays]{yang18gboardquery}
Timothy Yang, Galen Andrew, Hubert Eichner, Haicheng Sun, Wei Li, Nicholas
  Kong, Daniel Ramage, and Fran{\c{c}}oise Beaufays.
\newblock Applied federated learning: Improving {G}oogle keyboard query
  suggestions.
\newblock \emph{arXiv preprint 1812.02903}, 2018.

\bibitem[Yao(1982)]{yao1982protocols}
Andrew~C Yao.
\newblock Protocols for secure computations.
\newblock In \emph{{Symposium on Foundations of Computer Science}}, 1982.

\bibitem[Yao(1986)]{DBLP:conf/focs/Yao86}
Andrew~Chi{-}Chih Yao.
\newblock How to generate and exchange secrets (extended abstract).
\newblock In \emph{{FOCS}}, pages 162--167. {IEEE} Computer Society, 1986.

\bibitem[Ye et~al.(2019)Ye, Naim, and Rouayheb]{onoffitw}
Fangwei Ye, Carolina Naim, and Salim~El Rouayheb.
\newblock Preserving {ON-OFF} privacy for past and future requests.
\newblock In \emph{2019 IEEE Information Theory Workshop (ITW)}, August 2019.

\bibitem[Ye and Barg(2018)]{ye2018optimal}
Min Ye and Alexander Barg.
\newblock Optimal schemes for discrete distribution estimation under locally
  differential privacy.
\newblock \emph{IEEE Transactions on Information Theory}, 2018.

\bibitem[Yeom et~al.(2018)Yeom, Giacomelli, Fredrikson, and
  Jha]{yeom2018privacy}
Samuel Yeom, Irene Giacomelli, Matt Fredrikson, and Somesh Jha.
\newblock Privacy risk in machine learning: Analyzing the connection to
  overfitting.
\newblock In \emph{2018 IEEE 31st Computer Security Foundations Symposium
  (CSF)}, pages 268--282. IEEE, 2018.

\bibitem[Yin et~al.(2019)Yin, Chen, Ramchandran, and
  Bartlett]{yin2018byzantine}
Dong Yin, Yudong Chen, Kannan Ramchandran, and Peter Bartlett.
\newblock Byzantine-robust distributed learning: Towards optimal statistical
  rates.
\newblock In \emph{ICML}, 2019.

\bibitem[Yu et~al.(2018{\natexlab{a}})Yu, Tang, Renggli, Kassing, Singla,
  Alistarh, Zhang, and Liu]{yu2018distributed}
Chen Yu, Hanlin Tang, Cedric Renggli, Simon Kassing, Ankit Singla, Dan
  Alistarh, Ce~Zhang, and Ji~Liu.
\newblock Distributed learning over unreliable networks.
\newblock \emph{arXiv preprint arXiv:1810.07766}, 2018{\natexlab{a}}.

\bibitem[Yu et~al.(2020{\natexlab{a}})Yu, Liu, Liu, Chen, Cong, Weng, Niyato,
  and Yang]{Yu-et-al:2020IS}
Han Yu, Zelei Liu, Yang Liu, Tianjian Chen, Mingshu Cong, Xi~Weng, Dusit
  Niyato, and Qiang Yang.
\newblock A sustainable incentive scheme for federated learning.
\newblock \emph{IEEE Intelligent Systems}, 35\penalty0 (4):\penalty0 58--69,
  2020{\natexlab{a}}.

\bibitem[Yu et~al.(2018{\natexlab{b}})Yu, Yang, and Zhu]{yu2018parallel}
Hao Yu, Sen Yang, and Shenghuo Zhu.
\newblock Parallel restarted {SGD} for non-convex optimization with faster
  convergence and less communication.
\newblock \emph{arXiv preprint arXiv:1807.06629}, 2018{\natexlab{b}}.

\bibitem[Yu et~al.(2019)Yu, Jin, and Yang]{yu2019linear}
Hao Yu, Rong Jin, and Sen Yang.
\newblock On the linear speedup analysis of communication efficient momentum
  {SGD} for distributed non-convex optimization.
\newblock \emph{arXiv preprint arXiv:1905.03817}, 2019.

\bibitem[Yu et~al.(2020{\natexlab{b}})Yu, Bagdasaryan, and
  Shmatikov]{yu2020salvaging}
Tao Yu, Eugene Bagdasaryan, and Vitaly Shmatikov.
\newblock Salvaging federated learning by local adaptation.
\newblock \emph{arXiv preprint arXiv:2002.04758}, 2020{\natexlab{b}}.

\bibitem[Zafar et~al.(2017)Zafar, Valera, Rodriguez, and Gummadi]{zafar2017}
Muhammad~Bila Zafar, Isabel Valera, Manuel~Gomez Rodriguez, and Krishna~P.
  Gummadi.
\newblock Fairness constraints: Mechanisms for fair classification.
\newblock In \emph{Proceedings of the 20th International Conference on
  Artificial Intelligence and Statistics}, 2017.

\bibitem[Zantedeschi et~al.(2019)Zantedeschi, Bellet, and
  Tommasi]{Zantedeschi2019}
Valentina Zantedeschi, Aurélien Bellet, and Marc Tommasi.
\newblock {Fully Decentralized Joint Learning of Personalized Models and
  Collaboration Graphs}.
\newblock Technical report, arXiv:1901.08460, 2019.

\bibitem[Zhang et~al.(2015)Zhang, Choromanska, and LeCun]{zhang2015deep}
Sixin Zhang, Anna~E Choromanska, and Yann LeCun.
\newblock Deep learning with elastic averaging {SGD}.
\newblock In \emph{Advances in Neural Information Processing Systems}, pages
  685--693, 2015.

\bibitem[Zhang and Yang(2017)]{DBLP:journals/corr/ZhangY17aa}
Yu~Zhang and Qiang Yang.
\newblock A survey on multi-task learning.
\newblock \emph{CoRR}, abs/1707.08114, 2017.
\newblock URL \url{http://arxiv.org/abs/1707.08114}.

\bibitem[Zhang et~al.(2013)Zhang, Duchi, Jordan, and Wainwright]{duchi2013}
Yuchen Zhang, John Duchi, Micheal~I. Jordan, and Martin~J. Wainwright.
\newblock Information-theoretic lower bounds for distributed statistical
  estimation with communication constraints.
\newblock In \emph{Advances in Neural Information Processing Systems}, pages
  2328--2336, 2013.

\bibitem[Zhao et~al.(2019)Zhao, Yu, Zhao, and Liu]{zhao2019decentralized}
Yawei Zhao, Chen Yu, Peilin Zhao, and Ji~Liu.
\newblock Decentralized online learning: Take benefits from others' data
  without sharing your own to track global trend.
\newblock \emph{arXiv preprint arXiv:1901.10593}, 2019.

\bibitem[Zhu and Gupta(2017)]{zhu2017prune}
Michael Zhu and Suyog Gupta.
\newblock To prune, or not to prune: exploring the efficacy of pruning for
  model compression.
\newblock \emph{arXiv preprint arXiv:1710.01878}, 2017.

\bibitem[Zhu et~al.(2019)Zhu, Kairouz, Sun, McMahan, and Li]{zhu2019federated}
Wennan Zhu, Peter Kairouz, Haicheng Sun, Brendan McMahan, and Wei Li.
\newblock Federated heavy hitters discovery with differential privacy.
\newblock \emph{arXiv preprint arXiv:1902.08534}, 2019.

\bibitem[Zhu(2015)]{zhu2015machine}
Xiaojin Zhu.
\newblock Machine teaching: An inverse problem to machine learning and an
  approach toward optimal education.
\newblock In \emph{Twenty-Ninth AAAI Conference on Artificial Intelligence},
  2015.

\end{thebibliography}
\end{small}

\appendix

\section{Software and Datasets for Federated Learning}
\label{sec:datasets-and-software}

\paragraph{Software for simulation} Simulations of federated learning require dealing with multiple issues that do not arise in datacenter ML research, for example, efficiently processing partitioned datasets, with computations running on different simulated devices, each with a variable amount of data. FL research also requires different metrics such as the number of bytes upload or downloaded by device, as well as the ability to simulate issues like time-varying arrival of different clients or client drop-out that is potentially correlated with the nature of the local dataset. With this in mind, the development of open software frameworks for federated learning research (simulation) has the potential to greatly accelerate research progress. Several platforms are available or in development, including \cite{sathya2018review}:
\begin{itemize}
    \item TensorFlow Federated~\citep{tff} specifically targets research use cases, providing large-scale simulation capabilities as well as flexible orchestration for the control of sampling.  
    \item FedML \citep{he2020fedml} is a research-oriented library. It supports three platforms: on-device training for IoT and mobile devices, distributed computing, and single-machine simulation. For research diversity, FedML also supports various algorithms (e.g., decentralized learning, vertical FL, and split learning), models, and datasets.
    \item PySyft \citep{PySyft}  is a Python library for secure, private Deep Learning. PySyft decouples private data from model training, using federated learning, differential privacy, and multi-party computation (MPC) within PyTorch.  
    \item Leaf \citep{Leaf} provides multiple datasets (see below), as well as simulation and evaluation capabilities.
    \item Sherpa.ai Federated Learning and Differential Privacy Framework \citep{rodriguez2020federated} is an open source federated learning and differential privacy framework which provides methodologies, pipelines, and evaluation techniques for federated learning.
    \item PyVertical \citep{PyVertical} is a project focusing on federated learning with data partitioned by features (also referred to as vertical partitioning) in the cross-silo setting; see \cref{ssec:cross-silo}.
\end{itemize}
\paragraph{Production-oriented software} In addition to the above simulation platforms, several production-oriented federated learning platforms are being developed:
\begin{itemize}
    \item FATE (Federated AI Technology Enabler) \citep{FATE} is an open-source project intended to provide a secure computing framework to support the federated AI ecosystem.
    \item PaddleFL \citep{PaddleFL} is an open source federated learning framework based on PaddlePaddle \citep{PaddlePaddle}. In PaddleFL, several federated learning strategies and training strategies are provided with application demonstrations.
    \item Clara Training Framework \citep{ClaraTraining} includes the support of cross-silo federated learning based on a server-client approach with data privacy protection.
    \item IBM Federated Learning \citep{IBMFL} is a Python-based federated learning framework for enterprise environments, which provides a basic fabric for adding advanced features.
    \item Flower framework \citep{beutel2020flower} supports implementation and experimentation of federated learning algorithms on mobile and embedded devices with a real-world system conditions simulation.
    \item Fedlearner \citep{Fedlearner} is an open source federated learning framework that enables joint modeling of data distributed between institutions.
\end{itemize}
Such production-oriented federated learning platforms must address problems that do not exist in simulation such as authentication, communication protocols, encryption and deployment to physical devices or silos. Note that while TensorFlow Federated is listed under ``Software for simulation'', its design includes abstractions for aggregation and broadcast, and serialization of all TensorFlow computations for execution in non-Python environments, making it suitable for use as a component in a production system.

\paragraph{Datasets} Federated learning is adopted when the data is decentralized and typically unbalanced (different clients have different numbers of examples) and not identically distributed (each client's data is drawn from a different distribution). The open source package TensorFlow Federated~\citep{tff} supports loading decentralized dataset in a simulated environment with each client id corresponding to a TensorFlow Dataset Object. These datasets can easily be converted to numpy arrays for use in other frameworks.\footnote{\url{https://www.tensorflow.org/datasets/api_docs/python/tfds/as_numpy}.} At the time of writing, three datasets are supported and we recommend researchers to benchmark on them.

\begin{itemize}
    \item \textit{EMNIST} dataset \cite{cohen2017emnist} consists of 671,585 images of digits and upper and lower case English characters (62 classes).
    The federated version splits the dataset into 3,400 unbalanced clients indexed by the original writer of the digits/characters.
    The non-IID distribution comes from the unique writing style of each person.
    \item \textit{Stackoverflow\footnote{\url{https://www.kaggle.com/stackoverflow/stackoverflow}}} dataset consists of question and answer from Stack Overflow with metadata like timestamps, scores, etc.
    The training dataset has more than 342,477 unique users with 135,818,730 examples.
    Note that the timestamp information can be helpful to simulate the pattern of incoming data.
    \item \textit{Shakespeare} is a language modeling dataset derived from \textit{The Complete Works of William Shakespeare}. 
    It consists of 715 characters whose contiguous lines are examples in the client dataset. The train set has 16,068 examples and test set has 2,356 examples.
\end{itemize}
The preprocessing for \textit{EMNIST} and \textit{Shakespeare} are provided by the Leaf project \cite{caldas2018leaf}, which also provides federated versions of the sentiment140 and celebA datasets. These datasets have enough clients that they can be used to simulate cross-device FL scenarios, but for questions where scale is particularly important, they may be too small. In this respect \textit{Stackoverflow} provides the most realistic example of a cross-device FL problem.

\paragraph{Cross-silo datasets}
One example is the iNaturalist dataset\footnote{\url{https://www.inaturalist.org/}} which consists of large numbers of observations of various organisms all over the world. One can partition it by the geolocation or the author of an observation. If we partition it by the group an organism belongs to, like kingdom, phylum, etc., then the clients have totally different labels and biological closeness between two clients is already known. This makes it a very suitable dataset to study federated transfer learning and multi-task learning in cross-silo settings. 

Another example is the Google-Landmark-v2 \citep{GDLv2} that includes over 5 million images of more than 200 thousand different types of landmark. Similar to the iNaturalist dataset, one can split the dataset by authors, but due to the difference in scale with iNaturalist dataset, Google Landmark Dataset provides much more diversity and creates even greater challenges to large-scale federated learning.


\citet{luo2019real} has recently published a federated dataset for computer vision. The dataset contains more than $900$ annotated street images generated from $26$ street cameras and $7$ object categories annotated with detailed bounding box. Due to the relatively small number of examples in the dataset, it may not adequately reflect a challenging realistic scenario.

\paragraph{The need for more datasets}
Developing new federated learning datasets that are representative of real-world problems is an important question for the community to address. Platforms like TensorFlow Federated~\citep{tff} welcome the contribution of new datasets and may be able to provide hosting support.

While completely new datasets are always interesting, in many cases it is possible to partition existing open datasets, treating each split as a client. Different partitioning strategies may be appropriate for different research questions, but often unbalanced and non-IID partitions will be most relevant. It is also interesting to maintain as much additional meta information (timestamp, geolocation, etc.) as possible.

In particular, there is a need for feature-partitioned datasets, as will be discussed in \cref{ssec:cross-silo}. For example, a patient may go to one medical institute for a pathology test and go to another for radiology picture archiving, in which case the features of one sample are partitioned over two institutes regulated by HIPAA. \citep{annas2003hipaa}.

\end{document}